%% file: main.tex
\documentclass{article}

\usepackage{settings/arxiv}

\usepackage[utf8]{inputenc} %
\usepackage[T1]{fontenc}    %
\usepackage{hyperref}       %
\usepackage{url}            %
\usepackage{booktabs}       %
\usepackage{amsfonts}       %
\usepackage{nicefrac}       %
\usepackage{microtype}      %
\usepackage{graphicx}
\usepackage{natbib}
\usepackage{doi}
\usepackage{xcolor}
\usepackage{amsmath}
\usepackage{float}
\usepackage{rotating}

\usepackage{subcaption}
\usepackage{algorithm}
\usepackage{algpseudocode}
\usepackage{wrapfig}

\usepackage{pdflscape}

\usepackage{lineno}

\title{MOON-RL: \textbf{M}ulti-\textbf{O}bjective \textbf{O}ptimization via Objective Space \textbf{N}avigation with  \textbf{R}einforcement \textbf{L}earning }

\date{} 					%

\author{
 \href{https://orcid.org/0009-0003-6546-720X}{\includegraphics[scale=0.06]{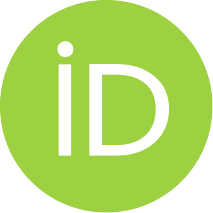}\hspace{1mm} Andrii Artomov}~\thanks{Corresponding author}\\
	Laboratory for Applied Mechanical Design\\
	Ecole Polytechnique F\'ed\'erale de Lausanne (EPFL)\\
	CH-1015 Lausanne, Switzerland \\
	\texttt{\href{mailto:andrii.artomov@epfl.ch}{andrii.artomov@epfl.ch}} \\
	\And
	\href{https://orcid.org/0000-0002-6763-3625}{\includegraphics[scale=0.06]{assets/orcid.pdf}\hspace{1mm} Soheyl Massoudi} \\
	Department of Mechanical and Process Engineering\\
	ETH Z\"urich\\
    8092 Zurich, Switzerland \\
	\texttt{\href{mailto:smassoudi@ethz.ch}{smassoudi@ethz.ch}} \\
	\AND
	\href{https://orcid.org/0000-0001-8822-9974}{\includegraphics[scale=0.06]{assets/orcid.pdf}\hspace{1mm} J\"urg Schiffmann}\\
	Laboratory for Applied Mechanical Design\\
	Ecole Polytechnique F\'ed\'erale de Lausanne (EPFL)\\
	CH-1015 Lausanne, Switzerland \\
	\texttt{\href{mailto:jurg.schiffmann@epfl.ch}{jurg.schiffmann@epfl.ch}} \\
}

\renewcommand{\headeright}{}
\renewcommand{\shorttitle}{MOON-RL}

\begin{document}

\maketitle %

\input{sections/Abstract}


\input{sections/1_Introduction}

\input{sections/2_Related_Work}
\input{sections/3_Methods}
\input{sections/4_Experiments} %

\input{sections/5_Discussion_and_Conclusion}

\input{sections/Paper_Statements}

\newpage

\newpage
\bibliographystyle{plain}
\bibliography{references}

\newpage
\input{sections/Appendix}

\end{document}

%% file: sections/Abstract.tex
\begin{abstract}

Many engineering design problems involve multiple conflicting objectives and strict feasibility constraints. They are typically solved with population-based optimizers that require many evaluations and offer limited online control over where the search is focused.
We present MOON-RL, a reinforcement-learning-based optimization framework that formulates constrained multi-objective search as preference-directed navigation in the objective space.
In MOON-RL, a lightweight PPO policy receives objective-space and constraint-related observations and acts as an instance-specific adaptive search process. It learns how to update the design so that the current solution advances along a user-specified trade-off direction, while a simple meta-algorithm periodically resets the policy, resamples starting designs from a continuously updated archive of non-dominated solutions, and adjusts step-size and penalty schedules. We evaluate MOON-RL on real-world actuator-design problems and additional convergence-hard constrained benchmarks, comparing against established evolutionary algorithms under matched evaluation budgets. In focused search scenarios, MOON-RL achieves competitive  performance in the region of interest, and ablation studies show that both the RL-driven adaptive sampling process and the meta-algorithm contribute to this behaviour, suggesting that repurposing RL training as objective-space navigation is a viable building block for interactive, preference-guided design tools.Additional experiments show that the framework also accommodates adaptive non-RL samplers, while PPO remains a strong choice for fixed-target, preference-directed objective-space navigation.

\end{abstract}

%% file: sections/1_Introduction.tex
\newpage

\section{Introduction}
\label{sec:intro}

Many real-world design problems involve multiple, often conflicting objectives and strict feasibility constraints. Unlike single-objective optimization, the outcome of multi-objective optimization (MOO) is a set of trade-off solutions known as the Pareto front. Population-based methods like NSGA-II \citep{deb2002fast} and MOEA/D \citep{zhang_moead_2007} are robust and parallelizable, but in their standard formulations they commonly distribute evaluations across broad regions of the objective space to approximate the Pareto front. From a practical engineering perspective, however, it can be more beneficial to guide the search toward a user-specified region of interest instead of spreading computational resources over the whole front, as the designer might be interested only in a specific region of the objective space. That region may not be known in advance: as feasible solutions reveal which trade-offs are attainable, the designer may wish to redirect the search toward newly identified areas of interest. Preference-based and interactive variants of evolutionary algorithms can address this by incorporating decision-maker preferences during the search and concentrating evaluations in regions of interest \citep{lopez2014including}. The practical challenge is therefore not only to specify a preferred direction in objective space, but also to determine which changes to the design variables will produce progress in that direction.

Reinforcement learning (RL) has emerged as a tool for adaptive optimization. Existing approaches include (i) multi-objective RL (MORL), where agents learn to balance vector-valued rewards across tasks or preferences \cite{liu_efficient_2024}, (ii) scalarization-based MOO, which trains a separate policy for each weight vector \citep{lin_pareto_2022}, and (iii) direct RL-based multi-objective design optimization, where the agent itself generates design updates during the optimization process \citep{wang2023multi,de2025multi}.
These methods use RL directly as the optimizer but, to our knowledge,do not formulate the search around a user-specified reference direction in objective space.

Recent works have also used RL to assist or configure evolutionary multi-objective algorithms, for example by selecting variation operators~\cite{tian2022deep}, configuring algorithmic components for constrained multi-objective optimization~\cite{ming2025automated}, performing algorithm selection~\cite{guo2024deep}, or adapting environmental selection and constraint handling~\cite{si2025efficient}. Other approaches use RL for adapting reference vectors, selecting guiding vectors, or allocating computational resources during the search~\cite{liang2024information,shao2024deep,dang2022dynamic}. In these approaches, the evolutionary algorithm remains the core optimizer, while RL controls or configures its components. MOON-RL differs from this setting: PPO is not used to assist or configure an evolutionary optimizer. Instead, the policy-learning loop directly generates design-variable updates guided by a user-specified direction in objective space.

Here, we use RL to instantiate preference-directed objective-space navigation as the search process itself. Navigation is defined in objective space by a reference direction from the objective vector of a starting design toward a user-specified target, but it is executed through changes to the design variables. At each step, a lightweight navigation agent uses objective-space and constraint-related observations to generate a continuous design-variable perturbation, the resulting design is evaluated, and its displacement relative to the reference direction informs the policy updates that follow. MOON-RL uses the target as a directional reference without requiring it to be a final solution, so the target may be infeasible, unreachable, or non-Pareto-optimal.

\begin{figure} [h]
    \centering
    \begin{subfigure}[b]{0.3\textwidth}
        \includegraphics[width=\textwidth]{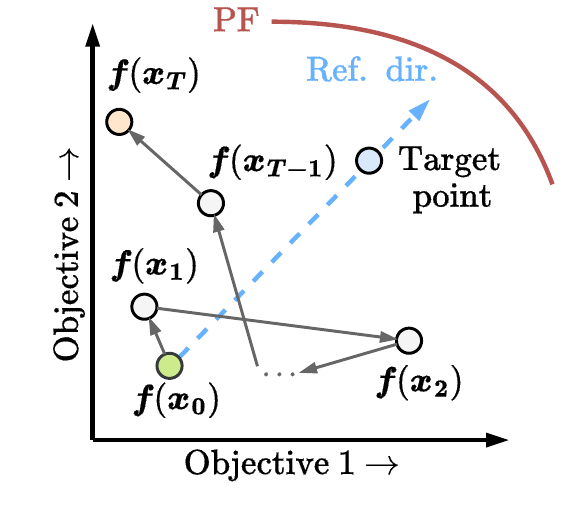}
        \begin{picture}(0,0)
            \put(-2,25){a.} %
        \end{picture}
        \label{fig:episode-rollout-1}
    \end{subfigure}
    \hspace{0.1\textwidth}
    \begin{subfigure}[b]{0.3\textwidth}
        \includegraphics[width=\textwidth]{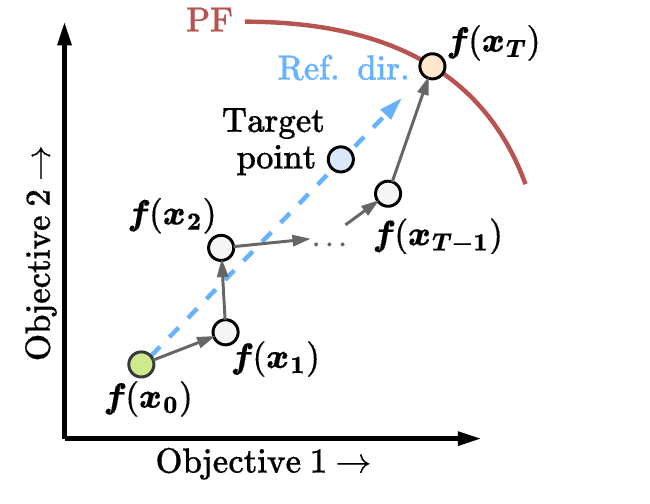}
        \begin{picture}(0,0)
            \put(-2,25){b.} %
        \end{picture}
        \label{fig:episode-rollout-2}
    \end{subfigure}
    \caption{ $T$-step rollouts for an untrained (a) and trained (b) agent in a 2-objective max-max problem. The trained agent drives the solution efficiently towards the Pareto Front along a reference direction defined by a target point in the objective space}
    \label{fig:episode-rollout}
\end{figure}

In contrast to classical RL, we do not seek a reusable policy. Instead, we repurpose the RL training loop itself as an adaptive sampling process that reacts to recent evaluations. The policy’s continual updates serve to bias exploration toward the desired direction, effectively turning the RL agent into a self-adjusting sampler for the current navigation task. This avoids spending evaluations on preference‑generalization whose cost is not amortized in static design tasks, where each optimization instance and preference direction is solved only once. Figure \ref{fig:episode-rollout} illustrates this idea: as the reinforcement learning agent learns, the sampling becomes more aligned with the chosen reference direction, and the solution advances toward the Pareto front (PF).

We instantiate this idea with Proximal Policy Optimization (PPO) and a reward that combines (i) directional progress, (ii) Gaussian fall-off orthogonal to the reference direction, and (iii) soft penalties for constraint violation. A simple meta-algorithm coordinates the training, periodically re-initializing both the navigation agent and its starting design point.
The new start is sampled from the continuously updated archive of non-dominated solutions, so each restart can resume the search from a different already discovered region of the objective space. This encourages exploration of under-represented regions without maintaining a full population. Linear schedules for scaling and penalty parameters modulate exploration and feasibility pressure over time. This mechanism mitigates the tendency of the single-point RL search to remain stuck in a local optima region while remaining conceptually straightforward.

Through this approach, MOON-RL becomes an objective space navigation framework that couples a user-specified direction directly to candidate generation: each navigation step produces an evaluated design update, and its displacement relative to the specified direction informs the updates that follow. In the experiments presented here, the target point is fixed within each optimization run. As feasible trade-offs are uncovered, a designer may wish to refine the region of interest and update the preference direction. Quantitative evaluation of such within-run retargeting remains future work.

We evaluate MOON-RL on actuator design problems from the MODAct benchmark suite \citep{picard2020realistic}, which represent real-world engineering optimization tasks with nonlinear objectives and constraints. These problems were chosen for their relevance to practical design workflows, where conflicting objectives and feasibility requirements mirror those encountered in real applications. ``CS1'' and ``CT1'' involve two objectives each, while ``CTS1'' includes three objectives. For each MODAct problem, we perform runs toward five distinct target points in the objective space, representing different user preference directions. To complement these experiments, we include the DAS-CMOP benchmark suite as an independent constrained test setting. The main comparison uses DAS-CMOP3 with difficulty level 11, while additional experiments cover DAS-CMOP1 and DAS-CMOP3 under diversity-, feasibility-, convergence-, and combined-difficulty settings. For statistical significance, each experiment is repeated with 30 random seeds, and we report the median and 5th--95th percentile results. All experiments share identical evaluation budgets of 300,000 function evaluations.

Our contributions can be summarized as follows:

\begin{enumerate}
    \item \textcolor{black}{We propose \textcolor{black}{MOON-RL, a framework for} preference-directed objective-space navigation \textcolor{black}{in} constrained multi-objective optimization. \textcolor{black}{Each navigation step applies a perturbation to the design variables and evaluates the result; the objective-space displacement this produces relative to a user-specified reference direction informs the updates that follow. We instantiate this process with a PPO agent} trained online within the optimization instance\textcolor{black}{: rather than producing a reusable policy, the learning process itself becomes the adaptive search mechanism.}}

    \item We design a soft directional reward that promotes exploration along the reference direction, naturally forming a localized Pareto trade-off region, while a soft constraint penalty encourages feasibility without hindering exploration.

    \item We introduce a simple meta-algorithm that coordinates the online search with periodic sampler and archive-based starting-point resets, along with evolving parameter schedules, mitigating local stagnation and reducing dependence on any single starting design.

    \item We validate the proposed framework on realistic actuator-design problems from the MODAct benchmark suite, representative of complex, constrained engineering optimization tasks, and on DAS-CMOP benchmark problems with multiple difficulty settings to assess behavior beyond the actuator-design setting. Further, we compare MOON-RL against established and recent constrained MOEAs, including APSEA, LCMEA, and CMODRL, demonstrating competitive preference-guided behavior in focused regions, while identifying regimes where specialized constrained MOEAs remain stronger. Additional results show that the proposed framework also accommodates adaptive non-RL samplers, including the cross-entropy method (CEM) and CMA-ES. PPO is level with or slightly behind CMA-ES on the MODAct problems and clearly ahead on ``DAS-CMOP3-11''.
\end{enumerate}

\newpage

%% file: sections/2_Related_Work.tex
\section{Related Work}

We consider standard MOO settings involving constrained vector-valued objectives and standard RL settings based on Markov Decision Processes (MDPs). Formal definitions for MOO and RL are provided in Appendix \ref{app:formalism}.

Evolutionary Algorithms (EAs)~\citep{deb_multi-objective_2011} such as NSGA-II~\citep{deb2002fast}, NSGA-III~\citep{deb2013evolutionary, jain2013evolutionary} or
MOEA/D~\citep{zhang_moead_2007}, for instance, are gradient-free, highly parallel, and robust to non-convex landscapes. In their standard formulations, these methods commonly aim to approximate broad regions of the Pareto front, which can require many function evaluations. Preference-based and interactive variants can instead incorporate decision-maker preferences during the search and concentrate evaluations in a specified region of interest; for example, López-Jaimes and Coello Coello~\citep{lopez2014including} introduce a reference-point-based preference relation that guides the evolving population toward a preferred region. Reference-point-based EAs such as R-NSGA-II~\citep{deb2006reference} and R-NSGA-III~\citep{vesikar2018reference} follow the same idea, biasing selection toward solutions close to one or more user-supplied reference points; we use R-NSGA-II and R-NSGA-III as preference-guided baselines with the same target points as MOON-RL.

Early RL-based optimizers converted MOO into a collection of single-objective tasks by drawing scalarization weights and training one policy per weight vector \citep{parisi_policy_2014, xu_prediction-guided_2020}. This decomposition is straightforward but scales linearly with the number of desired trade-offs and offers no mechanism for adapting the search once preferences change.
To reduce the number of trained policies, preference-conditioned approaches learn a single network that accepts a weight vector as an additional input \citep{yang_generalized_2019, zhang_meta-learning-based_2023}. Although elegant, conditioning significantly enlarges the state–action space, slows learning, and requires careful weight sampling to cover the Pareto front.
A complementary line (unified under the MORL/D taxonomy \citep{felten_multi-objective_2024}) maintains multiple specialist policies but couples them through shared critics or experience replay \citep{zhao_deep_2024, zhang_multi-objective_2022}. These algorithms typically maximize hyper-volume or epsilon-optimality, yet they inherit the high data demand of deep RL and are designed for dynamic environments. In static design optimization, the learned policy is discarded after one run, so the cost of generalization is not amortized.

Another related direction uses RL to control or configure components within evolutionary multi-objective algorithms. RL has been used to adapt reference vectors in many-objective optimization, select guiding vectors in large-scale sparse optimization, and allocate computational resources among subspaces in multimodal optimization~\cite{liang2024information,shao2024deep,dang2022dynamic}. Other approaches use RL for adaptive variation-operator selection~\cite{tian2022deep}, dynamic selection among differential evolution algorithms~\cite{guo2024deep}, automated configuration of environmental selection and offspring-generation operators in constrained MOO~\cite{ming2025automated}, and selection of constraint-handling techniques~\cite{si2025efficient}. In these methods, the evolutionary algorithm remains the core optimizer, while RL controls components of the search. This differs from MOON-RL, where PPO is not used to control an evolutionary algorithm, but is itself used as an instance-specific adaptive search process for objective-space navigation.

RL has also been used directly as a multi-objective engineering optimizer. Wang et al.~\citep{wang2023multi} formulate the current design vector as the RL state and use actions to define sequential design-space search steps, with a vector reward based on changes in the objectives. De Santanna et al.~\citep{de2025multi} similarly use DDPG to generate designs directly, assigning feasible designs a dominance-based scalar reward and infeasible designs a separate negative reward. These works establish direct RL-based multi-objective design optimization. In contrast to MOON-RL, neither formulation uses a user-specified reference or aspiration direction to define a preference-directed objective-space navigation task.

Interactive objective-space navigation is likewise established in the MCDM literature. O-NAUTILUS~\citep{saini2022optimistic} starts from a pre-generated set of evaluated solutions, which is used to train surrogate models from which an optimistic front is computed. The method allows the decision maker to navigate a step point in objective space between predefined rungs using reference-point information and to request additional preference-informed exact evaluations through a separate surrogate-assisted infill procedure. The resulting evaluation is then used to update the known set, surrogate models, and optimistic front, after which navigation is restarted with the step point returned to the combined-set nadir point. MOON-RL differs in how navigation is coupled to candidate generation: within each rollout, the policy directly generates a design-variable update, the resulting design is evaluated, and its objective-space displacement provides feedback for the following navigation step. The two methods therefore also differ in what they require to begin. O-NAUTILUS needs a pre-generated set of evaluated solutions from which the surrogate models and the fronts are constructed, whereas MOON-RL requires a single feasible starting design and produces its evaluated solutions as navigation proceeds.

Recent work increasingly treats the \emph{objective space itself} as the search
domain.  Distributional methods train ensembles that approximate the Pareto
front directly in objective space \citep{cai_distributional_2023,reymond_pareto_2022}, while iterative solvers decompose the front into a sequence of constrained sub-problems \citep{ropke_divide_2025}.  These studies highlight the value of objective-space reasoning, yet they still rely on batch training or computationally heavy constrained policy-optimization steps, leaving \emph{online, instance-specific navigation} largely unexplored in this line of work.

In deep RL, feasibility is typically enforced via Lagrangian critics or learned operator selection \citep{ming_constrained_2024}; both improve constraint satisfaction but introduce additional training complexity.  For \emph{static} tasks, RL has proved effective in combinatorial routing and circuit design by sequentially constructing solutions \citep{bello_neural_2017,kool_attention_2019,brown_deep_2022}. However, these
studies are single-objective or hybridize RL with evolutionary search, and they do not consider preference-directed search toward a user-specified region of the objective space.

Evolutionary optimizers remain robust but are \emph{evaluation-hungry} in many engineering settings, while preference-based variants can focus the search toward decision-maker-specified regions of interest~\citep{lopez2014including}. Prior work also establishes both interactive objective-space navigation~\citep{saini2022optimistic} and direct RL-based multi-objective design optimization~\citep{wang2023multi,de2025multi}. Taken together, these studies clarify the specific contribution of MOON-RL: we use online RL to realize preference-directed objective-space navigation as the search process itself, translating objective-space directional feedback into evaluated design-variable updates while a non-dominated archive and restart mechanism support continued search from previously discovered trade-off solutions. To our knowledge, no other work uses online, instance-specific reinforcement learning as the adaptive sampler within a preference-directed objective-space navigation framework.

%% file: sections/3_Methods.tex
\section{Methods}
\label{sec:algo}

\subsection{From Multi-Objective Optimization to RL: Navigating the Objective Space}

MOO in engineering typically involves finding design vectors  $\boldsymbol{x} \in \mathbb{R}^n$ whose objective values $\boldsymbol{f}(\boldsymbol{x}) \in \mathbb{R}^m$ lie on the Pareto front. This process is conventionally treated as a static mapping: designs are evaluated independently, and non-dominated solutions are accumulated through iterative sampling. While robust, these methods are sample-inefficient, as each evaluation is effectively a one-shot attempt. In MOON-RL, we formulate the search as a \emph{sequential decision-making problem in the objective space}. Starting from an initial design, the agent applies a sequence of modifications $\boldsymbol{m_t}$ that move the corresponding point in the objective space \textcolor{teal}{$\boldsymbol{f}(\boldsymbol{x})$} along a desired direction. This allows us to leverage reinforcement learning as an efficient sampling strategy, where the agent learns how to make targeted improvements along the reference direction using feedback from previous steps.

Navigation is defined in objective space through the positions and displacements of evaluated designs, while the agent acts by modifying the design variables. Figure \ref{fig:episode-rollout} shows how a design’s position in objective space evolves across an episode, contrasting freshly initialized agent (Figure \ref{fig:episode-rollout}a) versus the one after training (Figure \ref{fig:episode-rollout}b). Each episode starts from an initial design $\boldsymbol{x_0}$ (user-defined or randomly sampled) and runs for a fixed number of steps $T$. At each step $t$, the agent receives the observation $\boldsymbol{s_t}$ defined in Section~\ref{subsec:observation_and_reward}, samples an action $\boldsymbol{a_t}$ from its policy, and modifies the design, producing $\boldsymbol{f}(\boldsymbol{x_{t+1}})$. The agent is rewarded for advancing along a reference direction defined by a user-specified target point.The target point is used only to define this direction, it is not interpreted as a prescribed final solution and does not need to be reachable, feasible, or Pareto-optimal. For a given starting point, only the direction from the starting point to the target point matters; their distance has no effect. Note, that in the current implementation, the initial design $\boldsymbol{x_0}$ is assumed to be feasible. If no feasible design is available, the feasible archive and archive-based reset mechanism cannot operate directly, and an additional feasibility-search or repair stage would be required.

\subsection{Policy Architecture \& Action Modeling}

We use the clipped version of Proximal Policy Optimization (PPO) \cite{schulman2017proximal} algorithm, a stochastic policy gradient method that models action selection via probability distributions. As training progresses, the action distributions typically become narrower, shifting from exploration toward exploitation. We leverage this behavior in MOO to encourage diverse sampling early on, followed by more focused refinement near the reference direction.

We use separate networks for the actor and the critic. In our implementation, the actor consists of two independent networks: one predicts the means ($\mu$) and the other predicts the standard deviations ($\sigma$) for a normal distribution used to sample actions ($\boldsymbol{a_t} \sim  \mathcal{N}(\boldsymbol{\mu}, \boldsymbol{\sigma}^2)$). Mean outputs are passed through a $tanh$ activation to stay within (–1, 1), and standard deviations are passed through $softplus$ to ensure they are strictly positive.

When the agent is initialized or reset, actions are sampled from a near-zero mean, unit-variance distribution $\boldsymbol{a}_t^{\text{untrained}} \sim \mathcal{N}(\boldsymbol{\mu} \approx 0, {\boldsymbol{\sigma}}^2 \approx 1)$. Since design variables vary in scale, applying the same sampling distribution across all dimensions can result in overly large changes for some variables and negligible ones for others. To correct for this, we scale actions using a vector of scaling factors $\boldsymbol{\kappa} \in \mathbb{R}^n$, where each component is proportional to the range of the corresponding design variable. The resulting modification is computed as

\begin{equation}
    \boldsymbol{m}_t = \boldsymbol{\kappa} \odot \boldsymbol{a}_t ,
\end{equation}

where $\odot$ denotes element-wise multiplication. The new candidate design is then obtained as

\begin{equation}
    \boldsymbol{x}_{t+1} =\mathrm{clip} \left(\boldsymbol{x}_t + \boldsymbol{m}_t,\boldsymbol{\ell},\boldsymbol{u}\right),
\end{equation}

where $\mathrm{clip}(\cdot)$ enforces the lower and upper design-variable bounds component-wise. For a trained agent, this implies modifications $\boldsymbol{m}_t \sim \mathcal{N}(\boldsymbol{\mu} \odot \boldsymbol{\kappa}, \boldsymbol{\sigma}^2 \odot \boldsymbol{\kappa}^2)$. The additive update allows both increases and decreases in design variables relative to their current values. Initially, $\boldsymbol{\kappa}$ is set to a fraction of each variable’s range and gradually reduced during the optimization run. As the agent trains, $\boldsymbol{\mu}$ tends to shift directionally while $\boldsymbol{\sigma}$ decreases, reflecting a transition from exploration to confident design updates.

\subsection{Observation \& Reward}
\label{subsec:observation_and_reward}

\begin{figure}
    \centering
    \begin{subfigure}[b]{0.3\textwidth}
        \includegraphics[width=\textwidth]{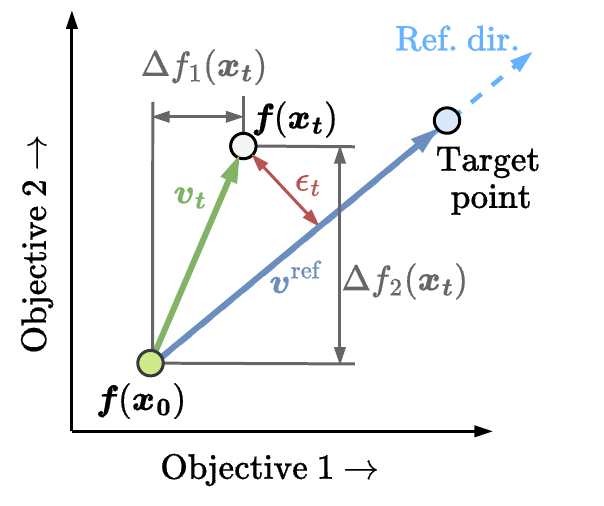}
        \begin{picture}(0,0)
            \put(0,30){a.} %
        \end{picture}

    \end{subfigure}
    \hspace{0.1\textwidth}
    \begin{subfigure}[b]{0.3\textwidth}
        \includegraphics[width=\textwidth]{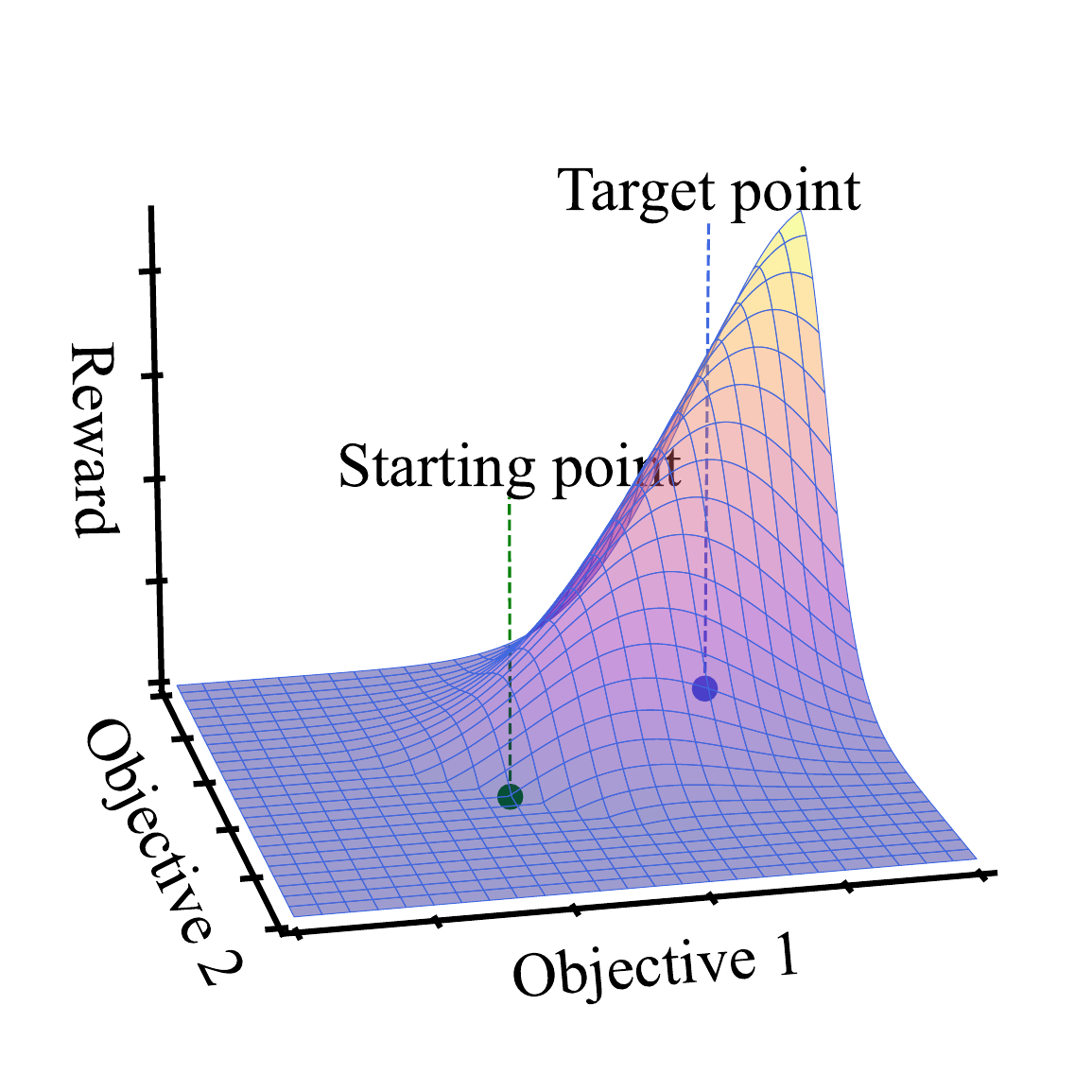}
        \begin{picture}(0,0)
            \put(0,30){b.} %
        \end{picture}
        \label{fig:reward_surf}
    \end{subfigure}
    \caption{ (a) monitored geometric metrics in the objective space, and (b) visualization of the reward function (without constraint penalty) for a two-objective optimization.}
    \label{fig:obj_space_metrics_and_reward}
\end{figure}

We define the observation vector $\boldsymbol{s_t}$ based on geometric features in the objective space and constraint-related information. The observation at step $t$ is:

\begin{equation}
    \boldsymbol{s_t} =
    [\Delta f_1(\boldsymbol{x_t}),\
    \Delta f_2(\boldsymbol{x_t}),\dots,\
    \Delta f_m(\boldsymbol{x_t}),\
    {
    \left\langle
    \frac{\boldsymbol{v^{\text{ref}}}}
    {\lVert \boldsymbol{v^{\text{ref}}} \rVert_2},
    \boldsymbol{v_t}
    \right\rangle},\
    CV_1(\boldsymbol{x_t}),\
    CV_2(\boldsymbol{x_t}),\dots,\
    CV_{p}(\boldsymbol{x_t}),\
    \frac{T-t}{T}]^T
\end{equation}

where $\Delta f_i(\boldsymbol{x_t}) = f_i(\boldsymbol{x_t}) - f_i(\boldsymbol{x_0})$ is the change in objective $i$ relative to the initial design. The vectors $\boldsymbol{v_t}$ and $\boldsymbol{v^{\text{ref}}}$ point from the initial objective value to the current one and to the user-defined target, respectively.
The scalar $\left\langle\frac{\boldsymbol{v^{\text{ref}}}}{\lVert \boldsymbol{v^{\text{ref}}} \rVert_2},\boldsymbol{v_t}\right\rangle$ represents the progress along the reference direction (i.e., the scalar projection of $\boldsymbol{v_t}$ onto the reference direction) in objective space between the initial objective vector $\boldsymbol{f}(\boldsymbol{x_0})$ and the current objective vector $\boldsymbol{f}(\boldsymbol{x_t})$; $\boldsymbol{v^{\text{ref}}}/\lVert\boldsymbol{v^{\text{ref}}}\rVert_2$ defines the unit reference direction (see Figure \ref{fig:obj_space_metrics_and_reward}).The term $(T-t)/T$ represents the fraction of the episode remaining at step $t$.
Finally, $CV$ terms represent the nonnegative violations of the $p$ inequality constraints handled by the present implementation. Specifically, $CV_i(\boldsymbol{x_t})=\max(g_i(\boldsymbol{x_t}),0)$, so each term remains zero whenever the corresponding constraint is satisfied. These terms are kept in their current form because they indicate the feasibility status of the candidate design at step \(t\). This differs from the objective terms, which are expressed as changes from the episode starting point because the reward measures objective-space progress relative to this starting point and along the reference direction.

The design vector $\boldsymbol{x_t}$ is excluded from the observation. This choice is empirically motivated: adding $\boldsymbol{x_t}$ increases the observation dimension without providing a systematic benefit in our ablation study reported in Appendix~\ref{app:observation_ablation}. We therefore retain the lower-dimensional objective-space observation. Because different designs can produce the same reduced observation while responding differently to the same action, $\boldsymbol{s_t}$ is not guaranteed to satisfy the Markov property.

The reward $r_t$ is computed from the position of the resulting point $\boldsymbol{f}(\boldsymbol{x_{t+1}})$ in the objective space, where $\boldsymbol{x_{t+1}}$ is obtained by applying action $\boldsymbol{a_t}$ to $\boldsymbol{x_t}$. The reward is computed after applying the objective-space scaling described in the experimental setup. The reward is designed to encourage progress along the reference direction in the objective space (see Figure \ref{fig:obj_space_metrics_and_reward}b).
It increases with progress along the reference direction and decays with perpendicular deviation, according to the exponential term $e^{-\epsilon_{t+1}^2}$, where $\epsilon_{t+1}$ is the perpendicular distance from $\boldsymbol{f}(\boldsymbol{x_{t+1}})$ to the reference direction vector $\boldsymbol{v^{\text{ref}}}$. We first compute the pure directional reward as

\begin{equation}
    R^{\text{dir}}_t = \max\left(
    \textcolor{black}{
    \left\langle
    \frac{\boldsymbol{v^{\text{ref}}}}
    {\lVert \boldsymbol{v^{\text{ref}}} \rVert_2},
    \boldsymbol{v_{t+1}}
    \right\rangle}
    e^{-\epsilon_{t+1}^2},0\right).
    \label{eq:pure_reward}
\end{equation}

Constraint violations are aggregated as the sum of positive violations and are incorporated through a clipped penalty term. The implemented reward is
\begin{equation}
    r_t = R^{\text{dir}}_t - \lambda \min\left(R^{\text{dir}}_t \textcolor{black}{\sum_{i=1}^{p} CV_i(\boldsymbol{x_{t+1}})}, R^{\text{dir}}_t\right).
    \label{eq:reward}
\end{equation}

Constraint violations are incorporated into the reward via the penalty coefficient $\lambda$, which controls the strength of the penalty. The clipping is applied only to the constraint-penalty contribution, not directly to the final reward. Since the clipped term is then multiplied by $\lambda$, the weighted penalty is not bounded by $R_t^{\mathrm{dir}}$; for $\lambda > 1$, sufficiently infeasible designs may receive negative rewards. This approach keeps the unweighted penalty on the same scale as the directional reward, preventing largely infeasible outliers from dominating the PPO update only because of the numerical magnitude of the constraint violation. %
Further, the penalty is scaled with the directional reward, ensuring that the penalty remains proportionally significant even when the agent makes large steps in the objective space. This discourages the agent from compensating for infeasibility by simply maximizing progress along the reference direction. The aggregation of constraint violations is used only in the scalar reward penalty because feasibility requires all constraints to be satisfied simultaneously, while individual constraint violations are still included separately in the observation vector.

\subsection{Implementation Remarks and Algorithm Summary}

We maintain an archive $\mathcal{P}$ of feasible non-dominated solutions, which is updated at every step during optimization. This archive provides an evolving approximation of the Pareto front, capturing the best trade-offs identified by the agent over time.

In engineering problems, the mapping from design space to objective space is often nonlinear and non-convex, making it impractical to train a single policy that generalizes across the entire domain. Instead, our agent focuses on one local task at a time and continues learning the local mapping between design variables and objective space throughout the optimization process. This contrasts with standard RL usage, which typically separates training from deployment. When the task changes—setting a new starting point—the agent is reset and learning begins again, continuing within the new local context.

\begin{figure}
    \centering
    \includegraphics[width=1\linewidth]{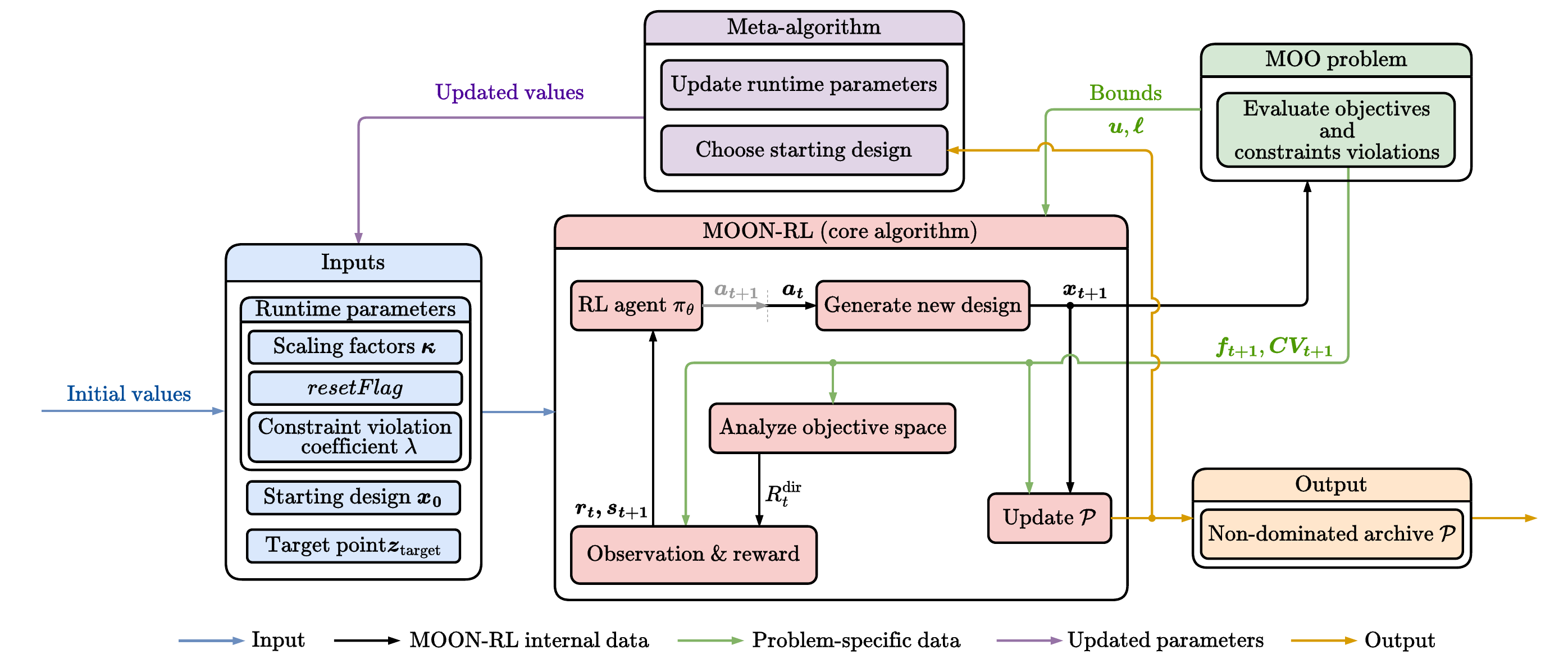}
\caption{Overview of the MOON-RL framework and data flow. The framework starts from the user-defined inputs, including the runtime parameters, initial design $\boldsymbol{x}_0$, and target point $\boldsymbol{z}_{\text{target}}$. The policy $\pi_\theta$ produces an action $a_t$, which is used to generate a new candidate design $\boldsymbol{x}_{t+1}$. The candidate design is evaluated through the multi-objective evaluation functions, producing the objective values $\boldsymbol{f}_{t+1}$ and constraint violations $\boldsymbol{CV}_{t+1}$. These evaluated quantities are used to update the non-dominated archive $\mathcal{P}$, construct the observation and reward, and analyze the position of the candidate design in objective space relative to the target direction. The meta-algorithm updates runtime parameters, such as scaling factors and constraint violation coefficients, and selects new starting designs during restarts. Arrow colors distinguish initial inputs, MOON-RL internal data, problem-specific data, updated parameters, and archive output.}
\label{fig:framewrk-components}
\end{figure}

Figure \ref{fig:framewrk-components} illustrates the framework and component interactions. While the evaluation function (used to compute $\boldsymbol{f}(\boldsymbol{x_t})$, $\boldsymbol{CV}(\boldsymbol{x_t})$, and bounds $\boldsymbol{u}$ and $\boldsymbol{\ell}$) depends on the specific design problem, the rest of the algorithm operates solely in the objective space. Except for constraint penalties, both observations and rewards rely only on geometric position in the objective space, enabling broad applicability across domains (only the evaluation logic must be adapted for new problems).

The algorithm runs until a fixed evaluation budget is reached, progressively moving the design from $\boldsymbol{f}(\boldsymbol{x_0})$ toward a target point $\boldsymbol{z}_\text{targ}$. Runtime parameters (e.g., penalties, scaling, target) can be adjusted externally to steer the search. In our experiments, the target point remains fixed throughout each run, although the framework allows it to be changed at runtime if needed. A detailed pseudocode for the algorithm is provided in Appendix \ref{app:core_pseudo}.

\subsection{Guiding meta-algorithm}
\label{subsec:meta-algo}

The core algorithm navigates the objective space from a fixed starting point toward a target, but it relies on external logic to set the starting point, target, and runtime parameters such as $\lambda$ and $\boldsymbol{\kappa}$. This guidance can be
user-defined (interactive setting) or provided by a meta-algorithm (automated setting).

In our experiments, we use a simple, fixed meta-algorithm to enable
repeatable, fully automated runs. The meta-algorithm is not a separate optimizer and does not optimize an objective function of its own. Instead, it coordinates the PPO-based search by updating runtime parameters and selecting restart points from the current archive. Developing more sophisticated strategies is beyond the scope of this
work, but the modular structure of our framework allows such components to be replaced without altering the core algorithm.

Policy-gradient methods like PPO typically converge to a locally optimal policy rather than a globally optimal one. In the context of MOO, this challenge is often intensified by the presence of many local optima in the objective space, which can cause the agent to converge to one of these local optima.
To mitigate this, our meta-algorithm resets the agent at fixed intervals and selects a new starting point in the objective space, sampled from the current Pareto archive. This enables the search to stride across the objective space and to escape local optima: the agent advances along a reference direction, and after each reset, the reference direction is recomputed from the new starting point to the same target point, and the agent is reinitialized for navigation from the new starting point. Thanks to the agent’s lightweight design, resets remain computationally efficient and allow rapid re-learning. The new starting point is drawn using a softmax distribution with a temperature parameter that depends on each solution’s proximity to the target point. %
More explicitly, each archive design $\boldsymbol{x}^{(i)} \in \mathcal{P}$ is first ranked according to its Euclidean distance to the target point in objective space,

\begin{equation}
    d^{(i)} = \left\|   \boldsymbol{f}(\boldsymbol{x}^{(i)})  - \boldsymbol{z}_{\mathrm{target}}\right\|_2 .
\end{equation}

The distance-based rank is converted into a score $s^{(i)}$, with larger scores assigned to archive points closer to the target. The probability of selecting archive design $i$ as the next starting point is then

\begin{equation}
    p^{(i)}  =  \frac{\exp(s^{(i)} / \tau)}  {\sum_j \exp(s^{(j)} / \tau)} ,
\end{equation}

where $\tau$ is the softmax temperature. This biases restarts toward the target point while preserving stochasticity in the selected starting design.

To support exploration early in training, we initialize the scaling factors $\boldsymbol{\kappa} $ with relatively large values, allowing action distributions to span broader regions of the design space. As optimization progresses, $\boldsymbol{\kappa} $ is gradually reduced to focus the search on local neighborhoods. The constraint penalty coefficient $\lambda$ follows an inverse schedule: it begins small to tolerate constraint violations during early exploration, then increases over time to promote feasibility. Both parameters are updated using simple linear schedules at fixed intervals:

\textcolor{black}{
\begin{equation}
    \boldsymbol{\kappa}
    \leftarrow
    \max\left(
        \boldsymbol{\kappa}_{\mathrm{limit}},
        \boldsymbol{\kappa} - \Delta \boldsymbol{\kappa}
    \right),
    \qquad
    \lambda
    \leftarrow
    \min\left(
        \lambda_{\mathrm{limit}},
        \lambda + \Delta \lambda
    \right).
\end{equation}
}

The maximum and minimum operations enforce the prescribed lower and upper schedule limits, respectively. The corresponding update logic is provided in Appendix  \ref{app:meta_pseudo} , and an additional sensitivity analysis of the constraint-penalty schedule is reported in Appendix~\ref{app:lambda_magnitude_ablation}..

In an interactive setting, the choice of target point, the selection and resetting of starting points, and the scheduling of  $\boldsymbol{\kappa} $ and $\lambda$ could be made directly by a human designer. While our main results focus on this fully automated setting, we also provide an illustrative demonstration of an
interactive scenario: a short screen-capture of a prototype interface built on top of this implementation, in which the user changes the target during a run and observes the agent redirect its navigation without being reset. The user can also select a new starting design, which resets the agent and starts a new local navigation task (supplementary Movie 1\footnote{Supplementary Movie 1 is available at \url{https://doi.org/10.5281/zenodo.24119189}.}). The movie provides a qualitative demonstration of these capabilities. Quantitative evaluation of adaptation speed, post-retargeting performance, and human interaction is left for future work.

%% file: sections/4_Experiments.tex
\newpage
\section{Experiments}

\subsection{Benchmark problems}

Our primary benchmark is the MODAct suite \cite{picard2020realistic}, a real-world multi-objective optimization task that involves the design of electromechanical actuators composed of an electric motor and a multi-stage gearbox. In this work we use three problems from MODAct: CS1, CT1, and CTS1. They share the same design space and constraint set but use different combinations of cost, torque excess, and safety factor as objectives, which leads to different optimisation dynamics and trade-offs. CS1 and CT1 each have 2 objectives, and CTS1 has 3 objectives. All three problems are highly constrained, multi-modal (with multiple local optima), and have relatively low feasibility ratios: $FsR_{CS1} = 0.0022$ for CS1, for CT1 $FsR_{CT1} = 0.0342$ and for CTS1 $FsR_{CTS1} = 0.0343$. This makes them representative of realistic but challenging design tasks.

In the original implementation of the MODAct suite \footnote{We used implementation from the following GitHub repository mentioned in \cite{picard2020realistic}: \href{https://github.com/epfl-lamd/modact}{https://github.com/epfl-lamd/modact}, last accessed in 2024. The repository also contains the best known Pareto fronts for proposed problems} problems include 20 design variables describing motor type, coil fill, gearbox configuration, and gear positions. Some variables encode both discrete (e.g. number of teeth) or categorical (motor type) variables and continuous ones (e.g. motor fill factor) in a single float with the integer part representing the discrete value and the fractional part encoding the continuous parameter. We separate these composite variables to allow our method to operate on individual design variables, bringing the total number of design variables to 27. For fairness, benchmark algorithms apply the same separation scheme. In both MOON-RL and the baseline optimization algorithms, discrete and categorical variables are rounded to the nearest integer after sampling.

To complement MODAct with additional problems, we include the DAS-CMOP3 problem with difficulty setting 11. DAS-CMOP problems \cite{fan2020difficulty} are scalable, two-objective constrained benchmarks with 30 decision variables and tunable difficulty controlled by a triplet $(\eta, \zeta, \gamma)$ that adjusts diversity, feasibility, and convergence hardness. The configuration $(\eta, \zeta, \gamma) = (0, 0, 0.75)$ yields a convergence-hard problem: the feasible region remains sizable, but the Pareto front is difficult to approach. This allows us to evaluate our algorithm under convergence-hard conditions in a benchmark that is independent of the actuator-design setting. This setting is also relevant for assessing whether MOON-RL can guide progress along a prescribed direction in objective space when approaching the Pareto front is difficult. We used the Pymoo \cite{pymoo} implementation for this problem. For completeness, additional DAS-CMOP experiments on DAS-CMOP1 and DAS-CMOP3 under diversity-, feasibility-, convergence-, and combined-difficulty settings are provided in Appendix~\ref{app:additional_dascmop}.

\subsection{Benchmark optimization algorithms}

We compare MOON-RL first to a set of established and recent constrained multi-objective evolutionary algorithms implemented in the Pymoo~\cite{pymoo} and PlatEMO~\cite{PlatEMO} libraries. The established baselines include NSGA-II, NSGA-III, MOEA/D, R-NSGA-II, R-NSGA-III, and C-TAEA. R-NSGA-II \cite{deb2006reference} and R-NSGA-III \cite{vesikar2018reference} are conceptually closest to our method, as they perform focused search toward user-defined target points. In our experiments, we used the same target points for R-NSGA variants as for MOON-RL to ensure aligned evaluation goals. MOEA/D \cite{zhang2007moea} decomposes the problem into scalar subproblems using fixed reference directions. NSGA-III \cite{deb2013evolutionary, jain2013evolutionary} promotes diversity by associating individuals with reference directions. C-TAEA \cite{li2018two} uses two archives to separately manage feasible and infeasible solutions. We tested multiple configurations per method . The full settings are provided in Appendix \ref{app:GA-config}.

We also include the more recent algorithms APSEA~\cite{tian2025adaptive}, LCMEA~\cite{si2025efficient}, and CMODRL~\cite{ming2025automated}, using their PlatEMO implementations~\cite{PlatEMO}. APSEA is an adaptive multi-population constrained MOEA, LCMEA is a large-scale constrained MOEA with directed offspring sampling and reinforcement-learning-assisted constraint handling, and CMODRL is a deep-reinforcement-learning-assisted constrained MOEA. The latter is therefore the most conceptually related to MOON-RL among these additional evolutionary baselines, although reinforcement learning is used there to configure an evolutionary optimizer rather than to act as the optimizer itself.

\subsection{Experimental protocol }

All algorithms are given a fixed budget of 300,000 function evaluations, matching the evaluation budget used in the original MODAct paper~\citep{picard2020realistic}, and runs were repeated 30 times with different random seeds. To avoid any possible correlation between runs, the seeds themselves were randomly sampled. They are summarized in Appendix \ref{app:seeds}.

For each experiment we specify a reference direction in the objective space through a target point. This target point defines the region of the Pareto front toward which MOON-RL and the reference-direction baselines are steered. In the main results we report one target point per problem; additional runs with alternative target points on the MODAct problems are reported in the appendix \ref{app:cs1-additional-results}.

Each run is initialised from a randomly sampled feasible design. To identify a feasible starting solution, we sample candidate designs uniformly at random until a feasible one is found, and count these evaluations against the overall budget. On CS1, which has the lowest feasibility ratio among the tested benchmarks, this requires on average $\sim 454$ evaluations ($1/\mathrm{FsR}$); the other problems require fewer samples on average. For the tested problems, this overhead is small relative to the total budget of 300,000 function evaluations. However, the current implementation assumes that a feasible initial design can be found by random sampling; for problems where this is not practical, an additional feasibility-search or repair stage would be required.

Each episode comprised $10$ parallel trajectories, each consisting of $8$ optimization steps, to gather enough data between learning iterations. Here and throughout the manuscript, the term episode denotes one such batch of parallel trajectories. Based on the evaluation budget, this results in $3750$ episodes.

The objectives in our benchmarks live on different numerical scales, which would otherwise distort the reward and the hypervolume calculations. We rescale the objectives for MOON-RL by positive constants estimated from the distribution of 100 randomly sampled feasible designs. These constants are used only for numerical normalization of the objective space; they do not encode preferences or objective trade-off weights. This crude calibration is intentional: it mirrors the information available to a designer at the beginning of a project, where only a small set of feasible solutions is known, and no detailed prior knowledge of the numerical scales is available. The same scaling factors are used for all algorithms and for hypervolume computation, ensuring fair comparison. Concretely, we divide each objective by its scaling factor before passing it to the RL agent and when computing hypervolume. For the MODAct problems, cost is divided by 0.1, torque excess by 0.2, and the safety factor by 20 whenever they appear as objectives. For DAS-CMOP3, both objectives are divided by 0.1.

\subsection{Used performance metrics and analysis approach}

\begin{wrapfigure}{r}{0.3\textwidth}
\centering
   \includegraphics[width=0.33\textwidth]{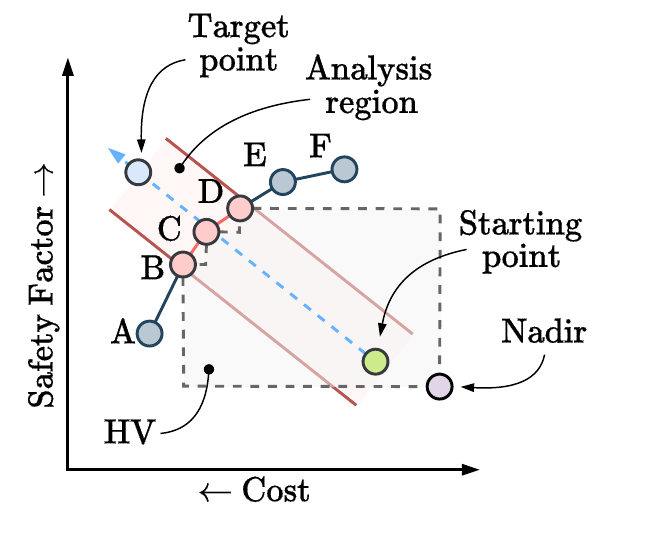}
   \caption{Analysis region (shaded red) example for CS1: only points B, C, and D are chosen for analysis.}
   \label{fig:focused-analysis-diagram}
\end{wrapfigure}

Hypervolume (HV) \cite{zitzler1998multiobjective} is the primary metric we use to evaluate performance. It is calculated as an area/volume that is dominated by the solutions in the Pareto Front with respect to a chosen reference point. Higher values indicate better obtained fronts. For each problem, we use the nadir of the best-known Pareto front in the scaled objective space as the reference point for all HV calculations. We track the HV as a function of the number of evaluation calls to assess the convergence effectiveness of an algorithm.

MOON-RL and the R-NSGA variants are configured to focus search along a user-defined reference direction, whereas algorithms such as NSGA-II and NSGA-III are designed to explore the entire front. To ensure a fair comparison in the focused setting, we restrict the HV analysis to an ``analysis region'' around the reference direction, defined by the fixed threshold $\epsilon < 1$, where $\epsilon$ is the perpendicular distance from the reference direction used in the Gaussian falloff term of Eq.\ref{eq:reward}. The analysis region also helps isolate intentional progress and avoid crediting incidental discoveries (see Figure\ref{fig:focused-analysis-diagram}). The same region is applied to all algorithms on the corresponding problem.

To place the analysis region consistently, we fix one reference starting point per problem and define the reference direction as the vector from this point to the chosen target point. Since MOON-RL uses a different randomly sampled feasible point in each run, we uniformly sample one million designs, filter for feasibility, and select the median objective-space location of the feasible subset as the shared reference starting point. The reference starting points in the objective space are as follows: CS1: $(0.85, 32.3)$, CT1: $(0.6564, -0.4538)$, CTS1: $(0.6624, -0.4509, 40.5)$, and DAS-CMOP3-11: $(5.70, 5.97)$.

To capture convergence speed of algorithms, we report the number of evaluations they required to reach $50\%$ and $75\%$ of the best-known hypervolume in the relevant region (analysis band or full front). The best-known hypervolume is computed from the best-known Pareto front for each problem.

To complement HV, we report the \emph{Generational Distance} (GD) \citep{van1999multiobjective}, which quantifies the proximity of the obtained solutions to the best-known Pareto front at the end of a run. GD is calculated as the root-mean-square of the Euclidean distances from each solution to its closest point on the best-known Pareto front in the scaled objective space, so smaller values indicate that the given run lies closer to the best-known front. We use GD rather than the \emph{Inverted Generational Distance} (IGD), because IGD measures distances from the reference Pareto front to the obtained solution set and therefore rewards coverage of the full front. This is not aligned with our focused-search setting, where the goal is to evaluate convergence in a preference-relevant region rather than coverage of regions the method was not designed to search.

We also compute the \emph{Empirical Attainment Function} (EAF) \cite{da2010attainment} to characterize variability across runs. It provides a probabilistic description of the outcomes across multiple independent runs. When computing the EAF, the objective space is discretized into a regular grid and, for each grid cell, we count how many runs contain at least one solution that dominates that cell. This allows us to report performance statistically in terms of a median attainment surface and corresponding lower and upper percentile surfaces, which summarize both typical performance and its variability across runs.

For visual clarity, the figures in the main text show MOON-RL performance together with a compact subset of representative classical baselines and the most relevant algorithms for each benchmark case. Full numerical results for all compared algorithms are reported in the appendices \ref{app:cs1_full_table} - \ref{app:cs1_global_full_table}.

\subsection{Focused search scenario}
\label{subsec:main_results}

In this scenario, we consider one user-specified target point per problem, defining a single reference direction and corresponding analysis band. MOON-RL dedicates its evaluation budget to navigating along this direction, while R-NSGA-II and R-NSGA-III are configured with the same target point. Other baseline algorithms (NSGA-II, NSGA-III, MOEA/D, C-TAEA, APSEA, LCMEA, and CMODRL) remain unchanged but are evaluated within the same analysis band.

\begin{figure}[H]
        \centering
        \begin{subfigure}[b]{0.49\textwidth}
            \includegraphics[width=\textwidth]{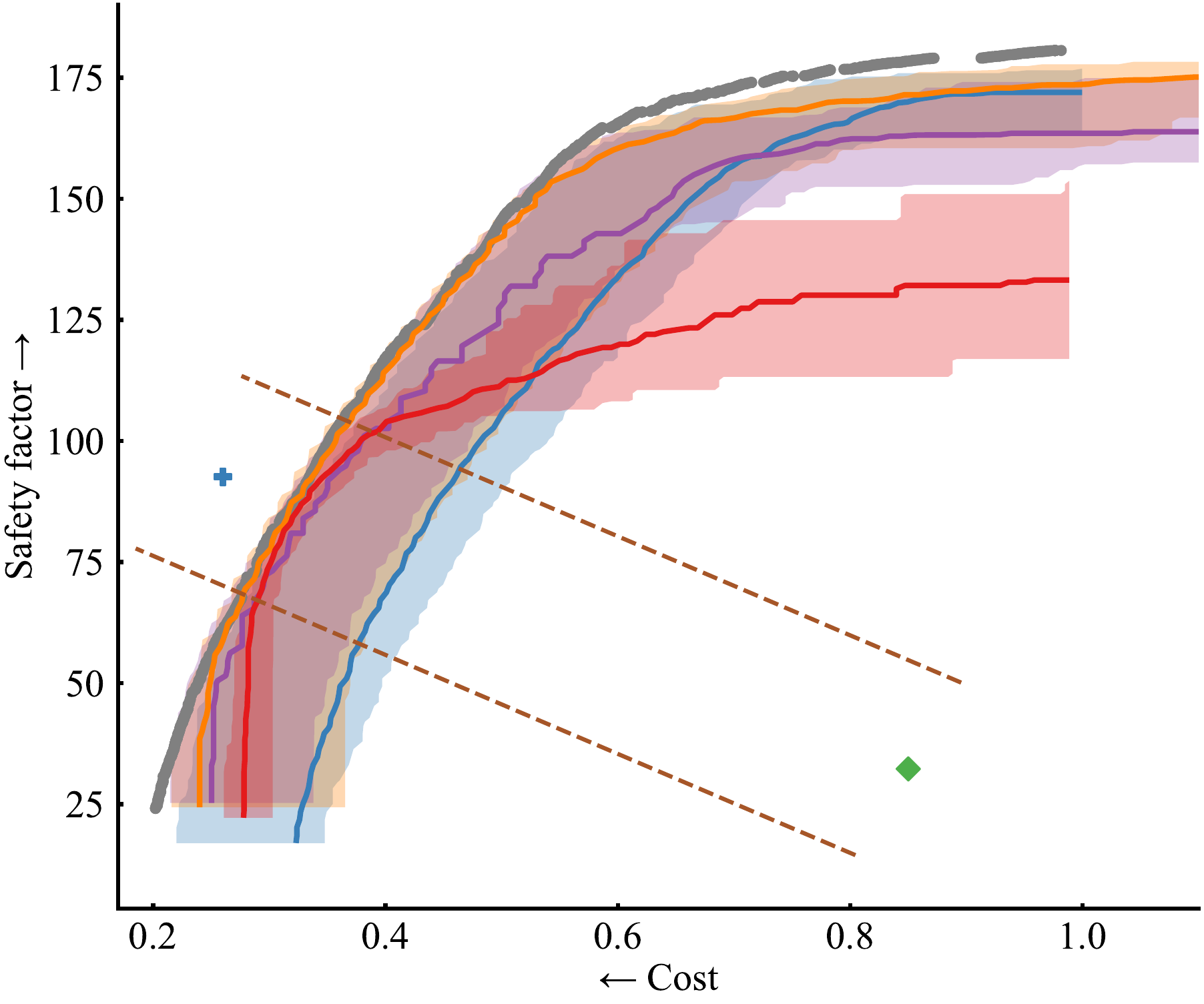}
            \begin{picture}(0,0)
                \put(0,195){a.} %
            \end{picture}

        \end{subfigure}
        \hfill
        \begin{subfigure}[b]{0.49\textwidth}
            \includegraphics[width=\textwidth]{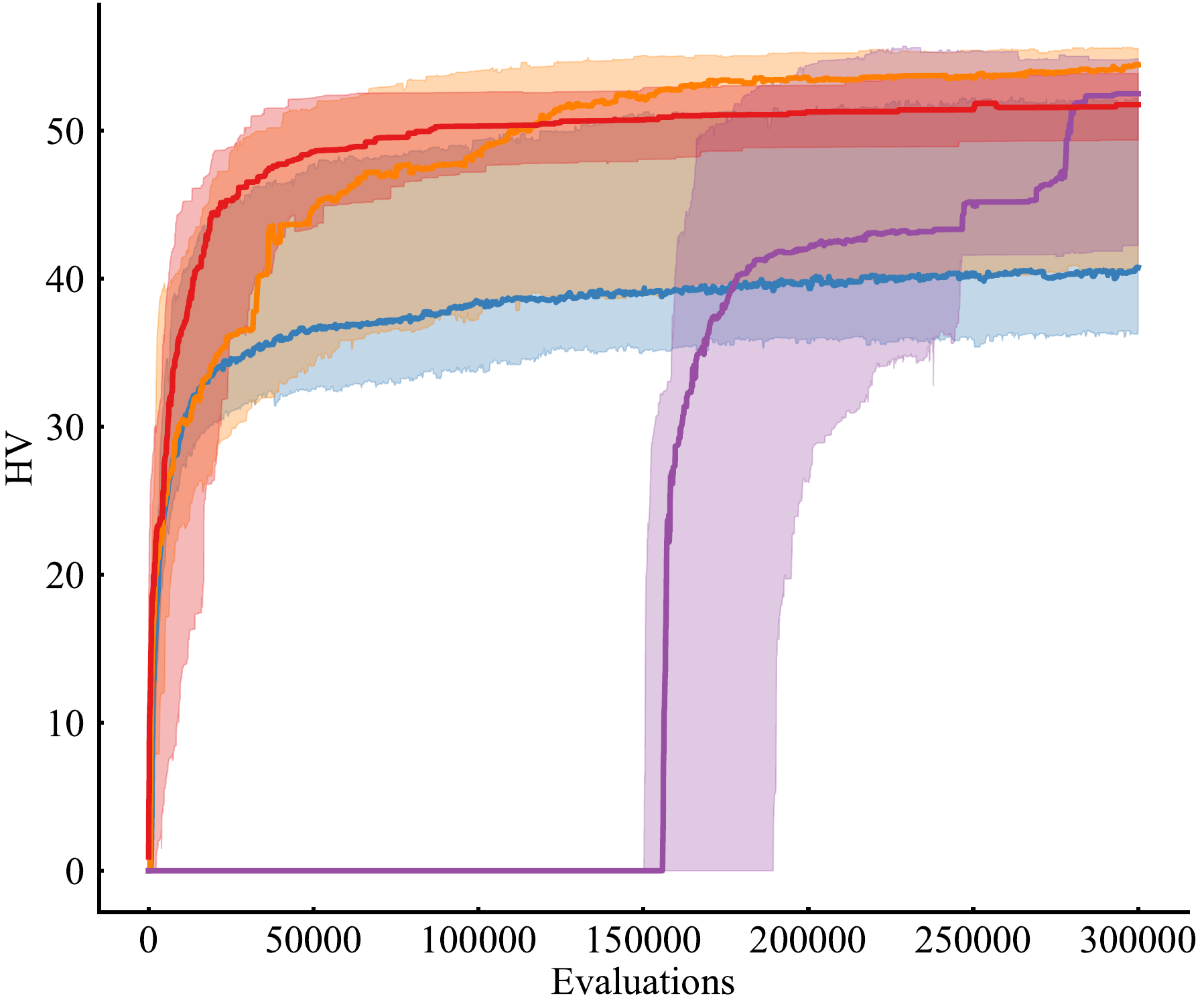}
            \begin{picture}(0,0)
                \put(0,195){b.} %
            \end{picture}
        \end{subfigure}

        \centering
        \vspace*{-1em} %
        {
        \begin{subfigure}[b]{0.6\textwidth}
            \includegraphics[width=\textwidth]{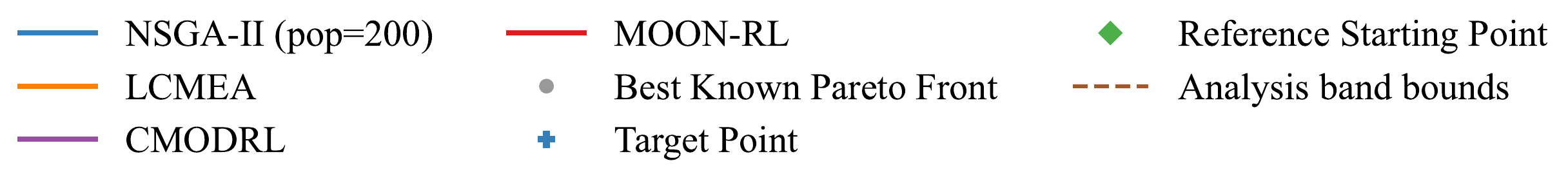}
        \end{subfigure}
        }

        \caption{``CS1'' problem: EAF (a) and HV progression (b) of different algorithms, including $5^{th}$ and $95^{th}$ percentiles.}
        \label{fig:cs1_best_p1}
    \end{figure}

For CS1 we focus on a target point at $(0.26, 92.6)$ in the objective space. Figure \ref{fig:cs1_best_p1} shows the median empirical attainment functions and hypervolume progression within the analysis band around the corresponding reference direction. In this scenario, MOON-RL remains competitive and performs particularly well in terms of reliability and early progress. Its median attainment surface approaches the best-known Pareto front closely along the direction of interest, while several baseline algorithms often stall in local optima. Table 1 confirms this observation quantitatively: LCMEA achieves the highest final hypervolume and the lowest median generational distance, while MOON-RL remains close to the best-performing methods and reaches $50\%$ of its final hypervolume in the analysis band faster than the recent PlatEMO baselines. For the $75\%$ hypervolume threshold, the focused R-NSGA variants show slightly smaller median evaluation counts, but only 7 out of 30 runs reach this level, whereas MOON-RL does so in 30 out of 30 runs. Results of the other baselines as well as the percentiles of already mentioned metrics are summarized in the full table in Appendix \ref{app:cs1_full_table}. Additional experiments toward alternative target points on CS1 are reported in Appendix \ref{app:cs1-additional-results}.

\begin{figure}[H]
        \centering
        \begin{subfigure}[b]{0.49\textwidth}
            \includegraphics[width=\textwidth]{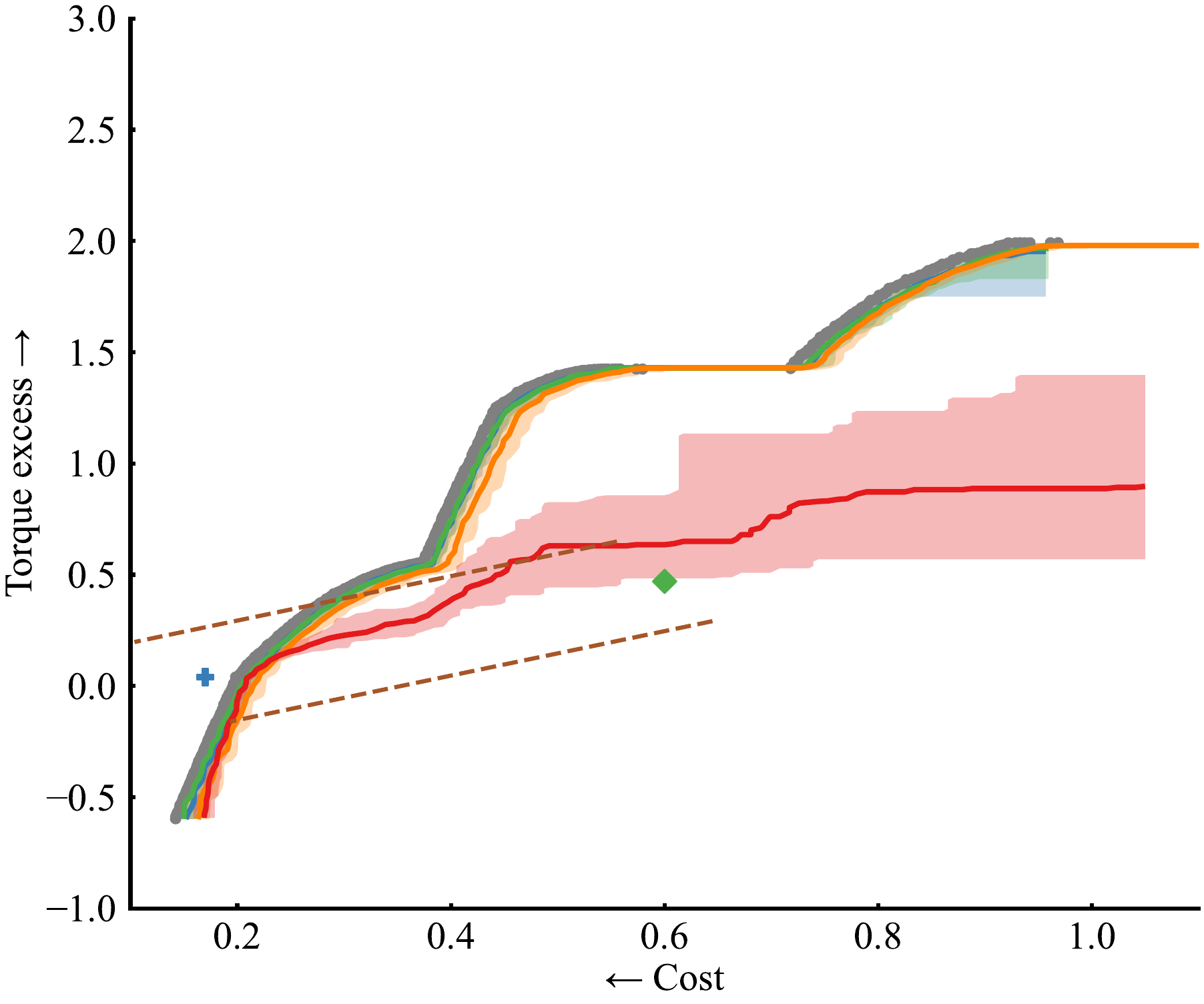}
            \begin{picture}(0,0)
                \put(0,195){a.} %
            \end{picture}

        \end{subfigure}
        \hfill
        \begin{subfigure}[b]{0.49\textwidth}
            \includegraphics[width=\textwidth]{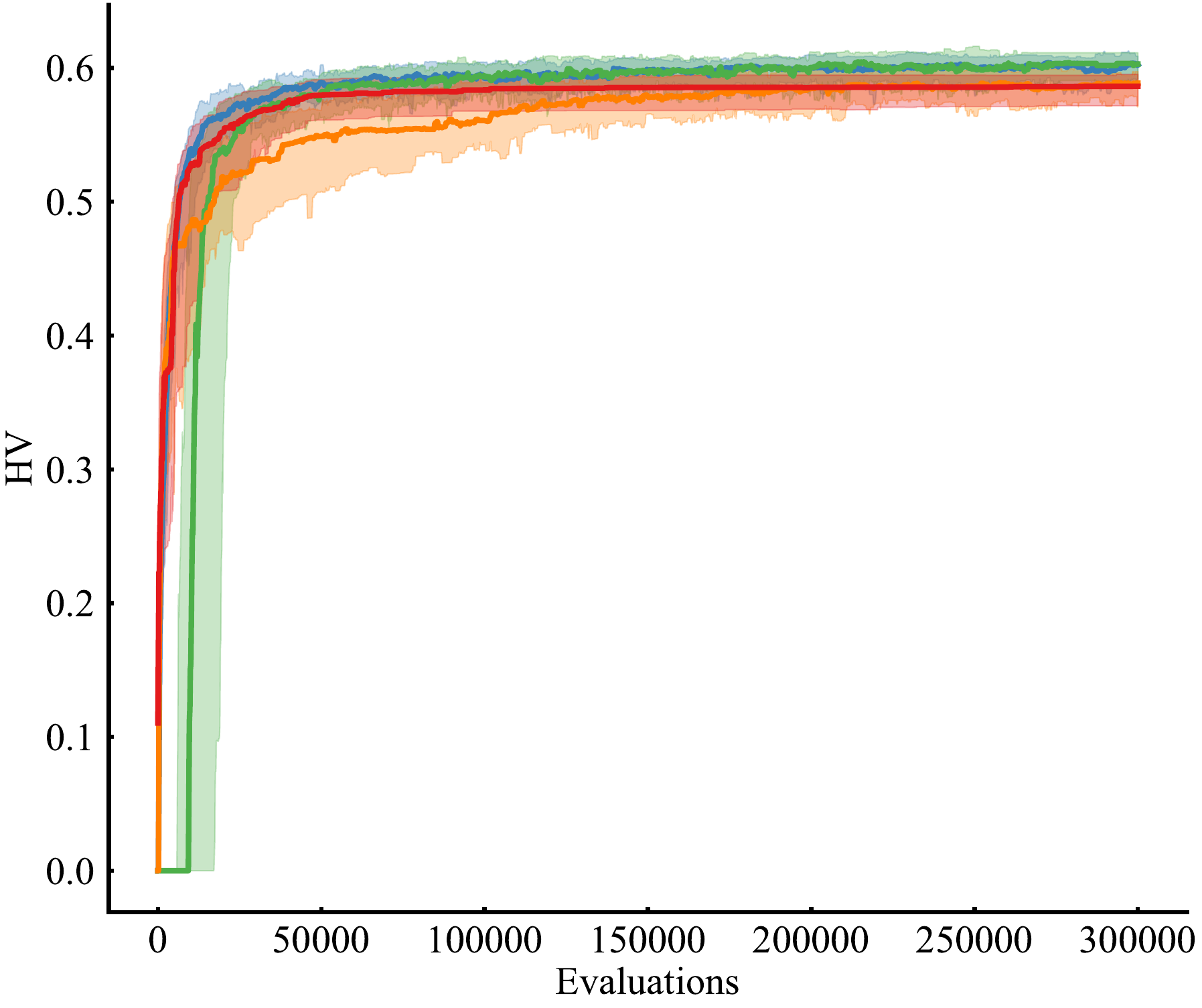}
            \begin{picture}(0,0)
                \put(0,195){b.} %
            \end{picture}
        \end{subfigure}

        \centering
        \vspace*{-1em} %
        {
        \begin{subfigure}[b]{0.6\textwidth}
            \includegraphics[width=\textwidth]{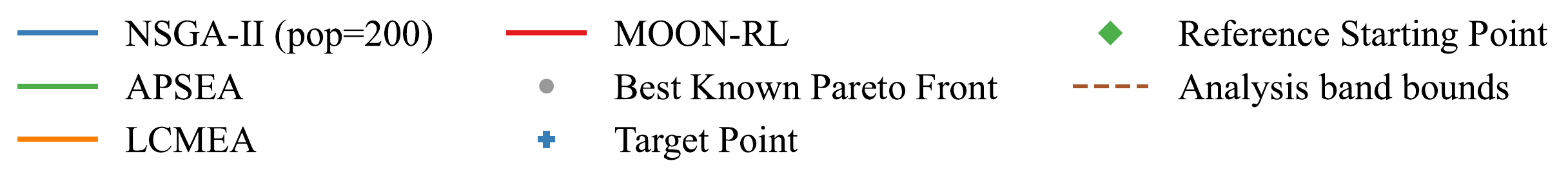}
        \end{subfigure}
        }

        \caption{``CT1'' problem: EAF (a) and HV progression (b) of different algorithms, including $5^{th}$ and $95^{th}$ percentiles.}
        \label{fig:ct1_best_p2}

    \end{figure}

For CT1, we focus on a target point at $(0.17, 0.04)$ in the objective space. The median attainment surfaces and hypervolume curves in Figure \ref{fig:ct1_best_p2} indicate that the best baseline algorithms solve this problem slightly better overall, while MOON-RL still finds solutions close to the best-known Pareto front in the region of interest. The data in Table \ref{tab:main_results_best} suggests that the best baselines (APSEA, NSGA-II, NSGA-III, and R-NSGA-II) achieve marginally higher final hypervolume and lower generational distance than MOON-RL. However, MOON-RL reaches $50\%$ of the best-known hypervolume in the analysis band earlier than most baselines. For the $75\%$ threshold, R-NSGA-II and NSGA-III reach the target in fewer evaluations. A more detailed table with additional metrics is included in Appendix \ref{app:ct1_full_table}, and the runs toward alternative target points on CT1 are provided in Appendix \ref{app:ct1-additional-results}.

\begin{figure}[H]
        \centering
        \begin{subfigure}[b]{0.49\textwidth}
            \includegraphics[width=\textwidth]{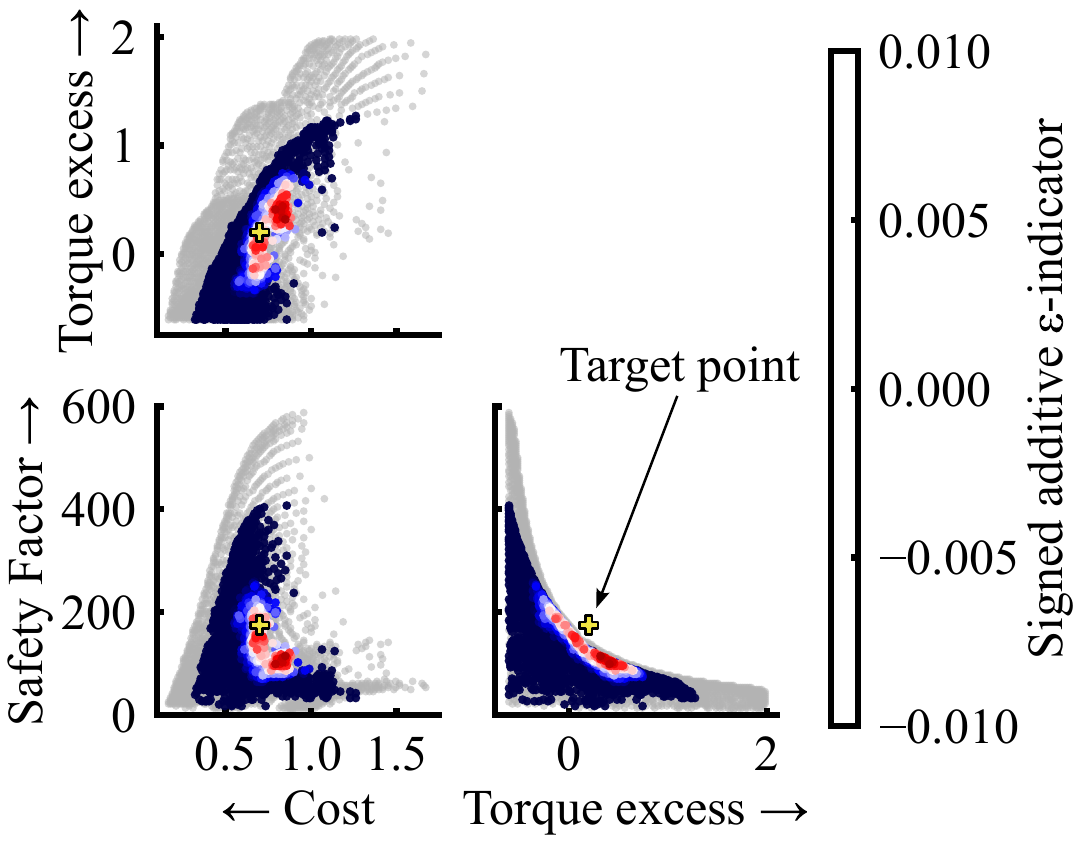}
            \begin{picture}(0,0)
                \put(0,180){a.} %
            \end{picture}

        \end{subfigure}
        \hfill
        \begin{subfigure}[b]{0.49\textwidth}
            \includegraphics[width=\textwidth]{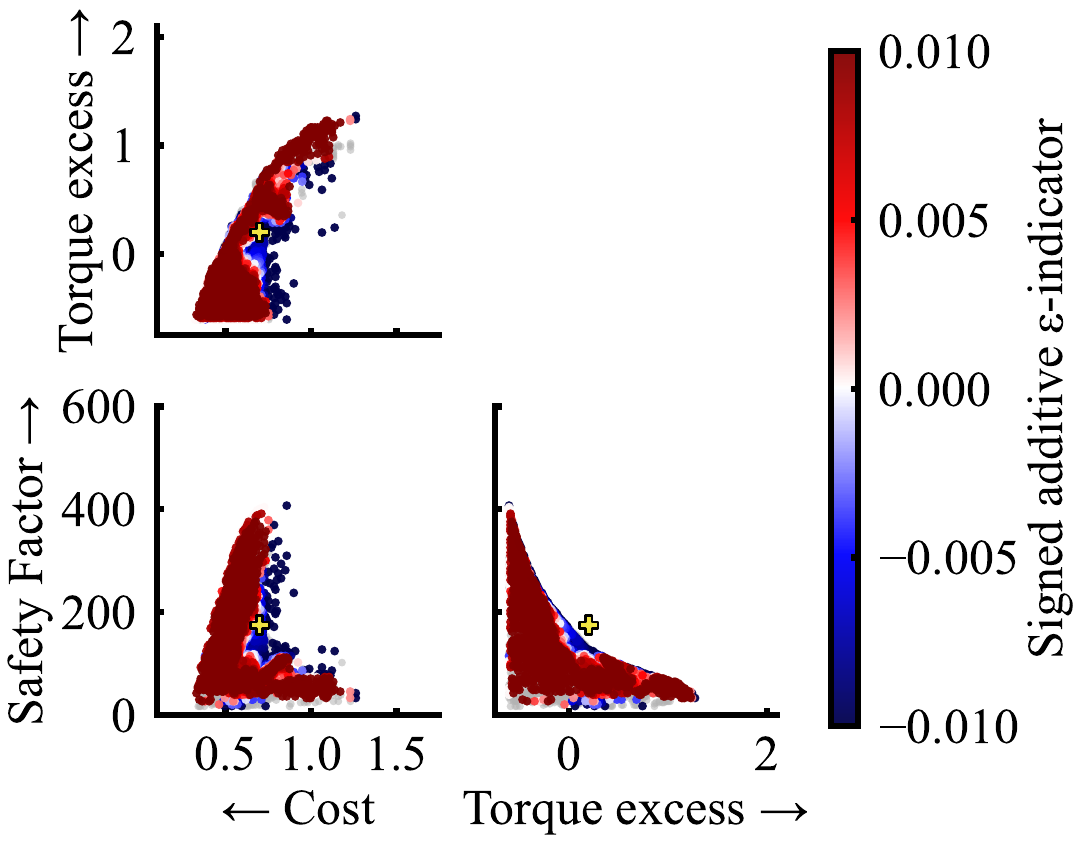}
            \begin{picture}(0,0)
                \put(0,180){b.} %
            \end{picture}
        \end{subfigure}

        \centering
        \vspace*{-1em} %

        \centering
        \begin{subfigure}[b]{0.49\textwidth}
            \includegraphics[width=\textwidth]{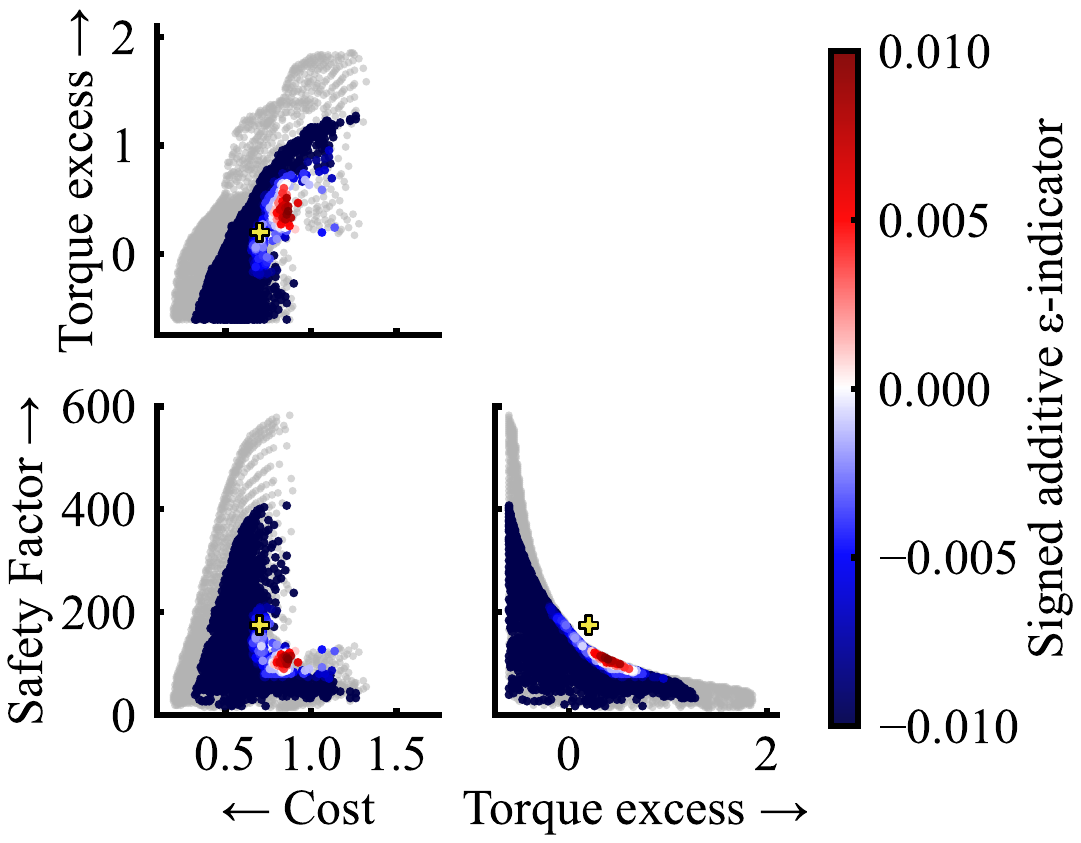}
            \begin{picture}(0,0)
                \put(0,180){c.} %
            \end{picture}

        \end{subfigure}
        \hfill
        \begin{subfigure}[b]{0.49\textwidth}
            \includegraphics[width=\textwidth]{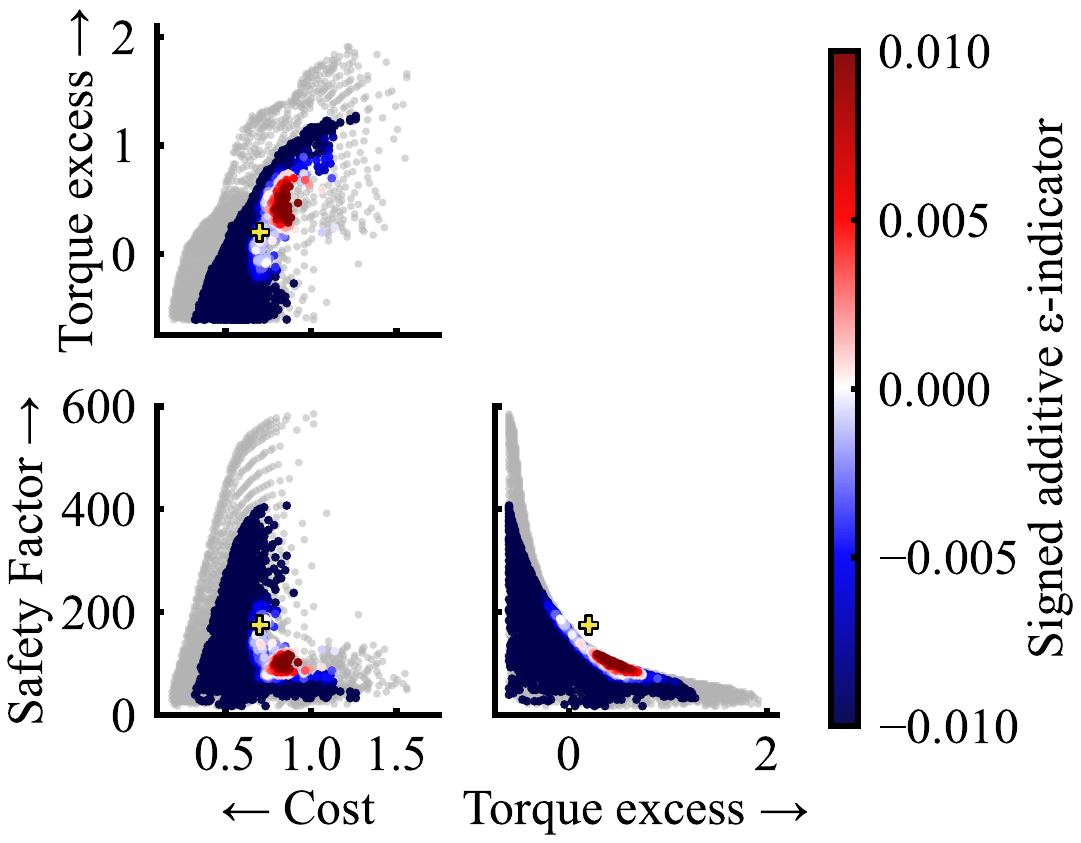}
            \begin{picture}(0,0)
                \put(0,180){d.} %
            \end{picture}
        \end{subfigure}

        \centering
        \vspace*{-1em} %

        \centering
        \begin{subfigure}[b]{0.49\textwidth}
            \includegraphics[width=\textwidth]{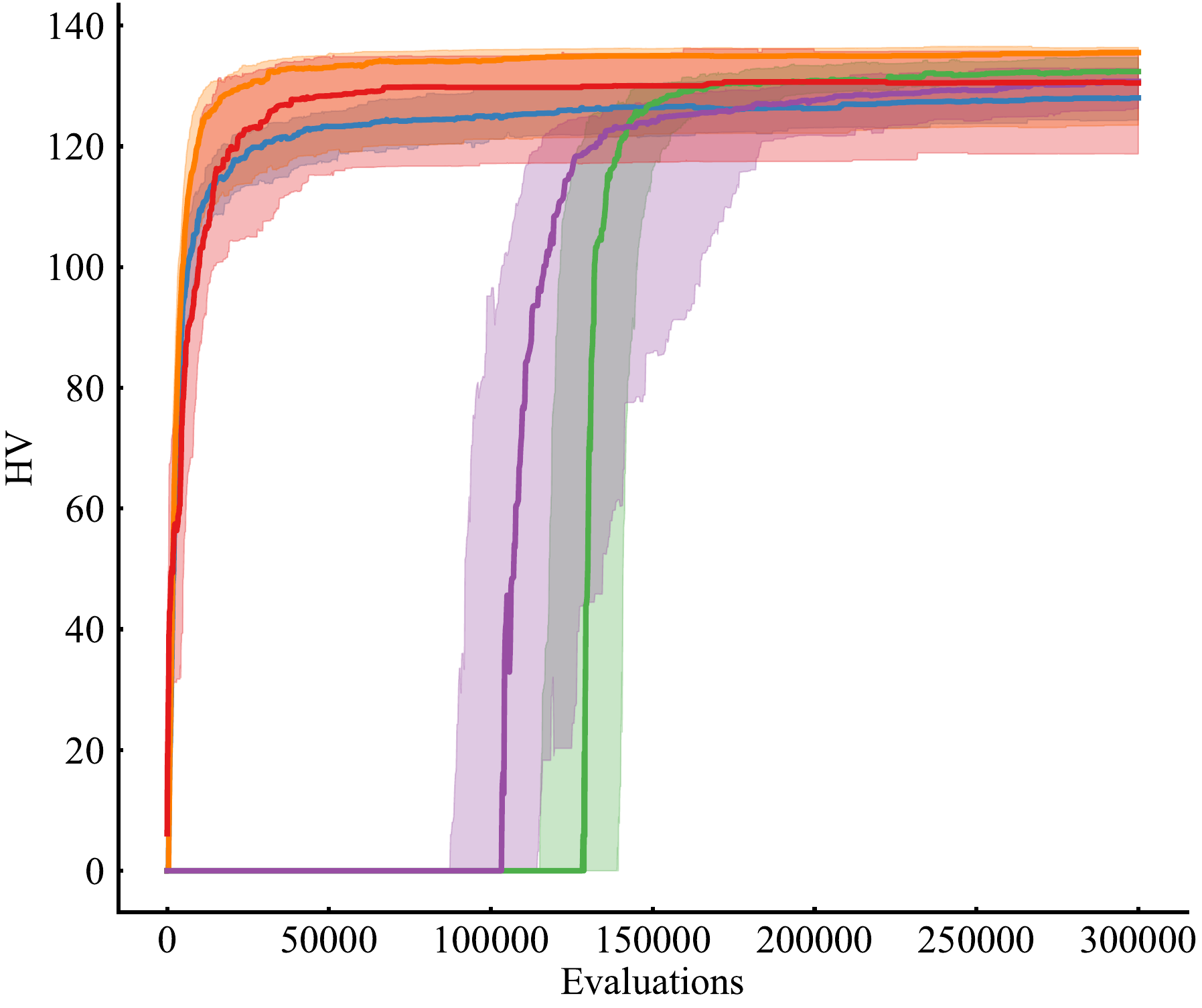}
            \begin{picture}(0,0)
                \put(-0.5,180){e.} %
            \end{picture}

        \end{subfigure}
        \centering
        \vspace*{-1em} %

        \centering
        \begin{subfigure}[b]{0.49\textwidth}
            \includegraphics[width=\textwidth]{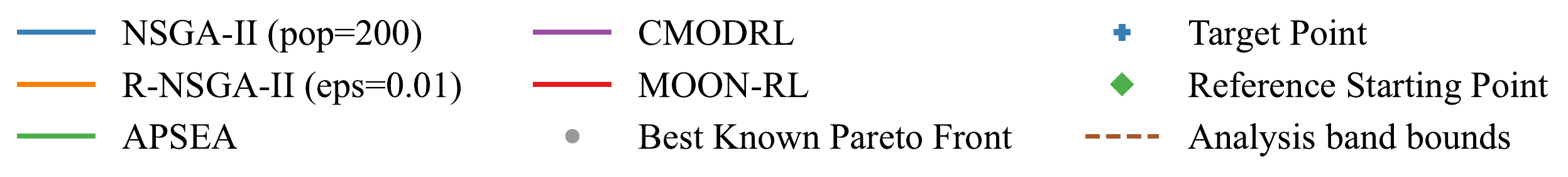}
            \label{fig:cts1-legend}
        \end{subfigure}

        \centering

        \caption{``CTS1'' problem: Comparison of median EAFs for MOON-RL against (a) NSGA-II (pop.\ 200), (b) R-NSGA-II ($\varepsilon = 0.01$), (c) APSEA, and (d) CMODRL. The target point is shown as a yellow cross. Grey points depict the baseline algorithm EAF, while colored points correspond to the MOON-RL EAF. The color encodes the signed additive $\varepsilon$-indicator: red regions indicate that MOON-RL outperforms the baseline, blue regions that the baseline outperforms MOON-RL, and white denotes parity. (e) HV progression including $5^{th}$ and $95^{th}$ percentiles.}
        \label{fig:cts1_p1}
    \end{figure}

For CTS1, a three-objective problem, we focus the algorithm on a target point at $(1.45, 1.58, 65.0)$ in the objective space. Visualising the full three-dimensional empirical attainment function is difficult, so Figure \ref{fig:cts1_p1} shows pairwise projections of the median attainment surfaces in each objective plane. Grey points correspond to the baseline median EAF, and colored points to the MOON-RL median EAF. To assess where MOON-RL improves over or falls behind each baseline, we compute a signed additive $\epsilon$-indicator \cite{zitzler2003performance} for each individual MOON-RL point with respect to the baseline Pareto front, i.e., every point produced by MOON-RL is compared directly to the baseline front. More details on how the $\epsilon$-indicator was calculated are given in Appendix \ref{app:additive-epsilon-indicator}. We then take the inverse of this value so that positive numbers correspond to regions where MOON-RL dominates the baseline and negative numbers to regions where it is dominated; values close to zero indicate parity. These scores are mapped to a fixed red--white--blue color scale: red marks regions where MOON-RL performs better, blue regions where it performs worse, and white means parity. The same color scale and value range are used for all plots, including those in Appendix \ref{app:cts1-additional-results}, to make comparisons across baselines easier. These visualizations are mainly qualitative and highlight objective-space regions where MOON-RL gains or loses relative to each baseline. On CTS1, MOON-RL remains competitive in the region of interest, while the strongest evolutionary baselines achieve better results in either final accuracy or convergence speed. Table \ref{tab:main_results_best} shows that R-NSGA-II attains the highest median hypervolume among the methods shown in the main table, while APSEA and CMODRL achieve the lowest median generational distance. In terms of convergence, R-NSGA-II and NSGA-II reach the $50\%$ and $75\%$ hypervolume thresholds faster than MOON-RL, whereas APSEA and CMODRL reach better final generational distance but require substantially more evaluations. Overall, CTS1 demonstrates that MOON-RL remains competitive in three dimensions, but the comparison also shows a clear trade-off between final accuracy and convergence speed across the evaluated methods. The full table with results is available in Appendix \ref{app:cts1_full_table}.

\begin{figure}[H]
        \centering
        \begin{subfigure}[b]{0.49\textwidth}
            \includegraphics[width=\textwidth]{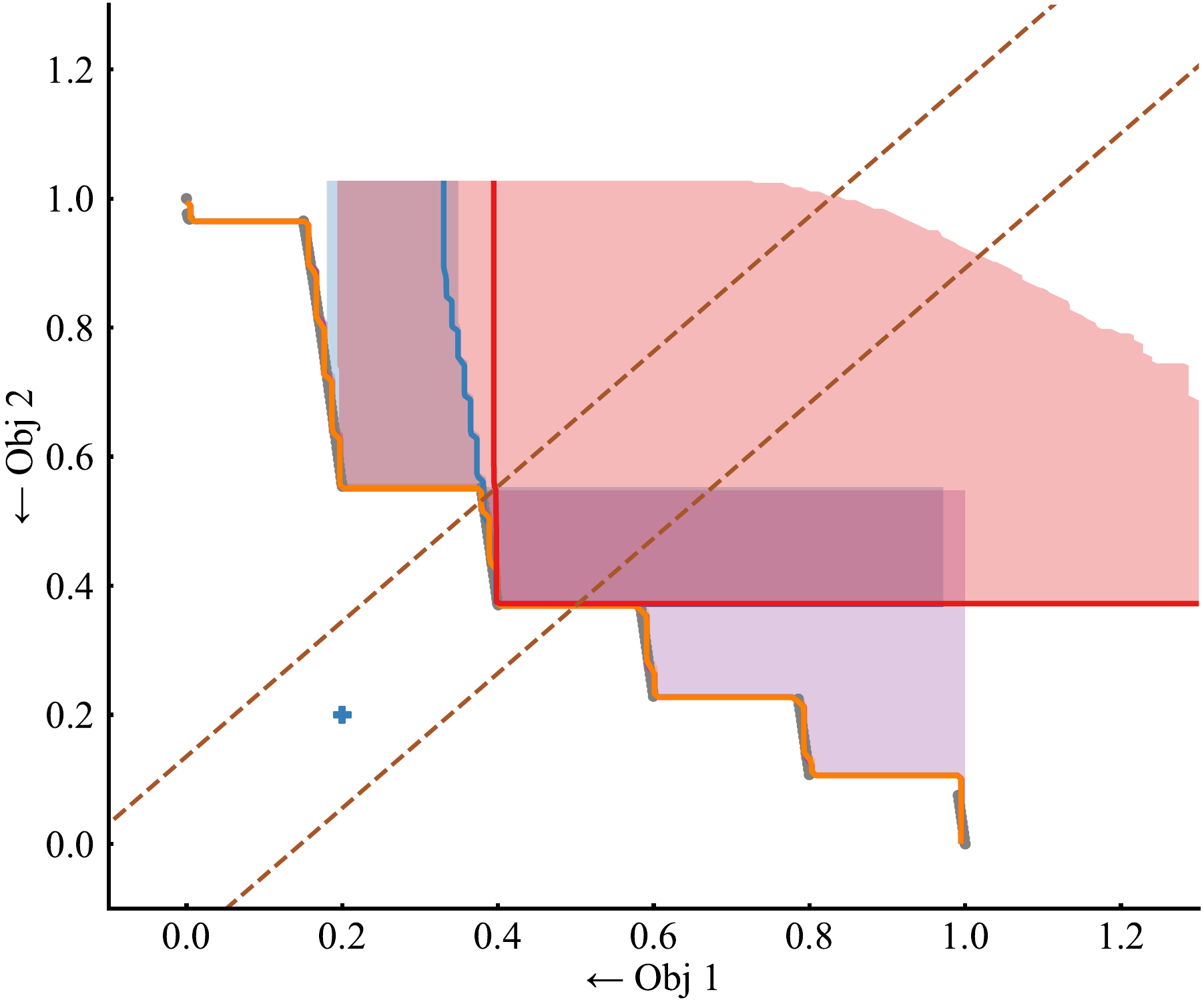}
            \begin{picture}(0,0)
                \put(0,195){a.} %
            \end{picture}

        \end{subfigure}
        \hfill
        \begin{subfigure}[b]{0.49\textwidth}
            \includegraphics[width=\textwidth]{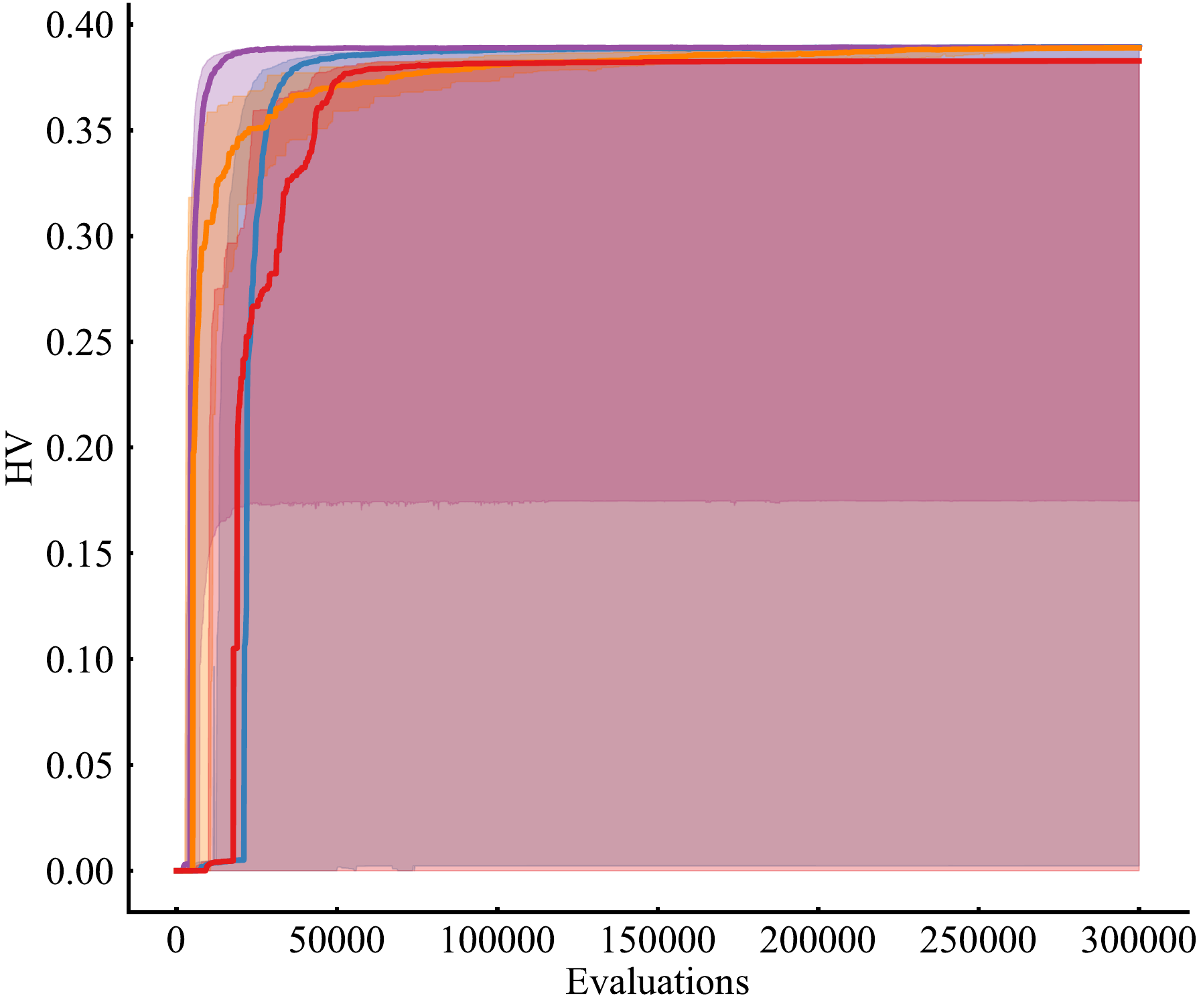}
            \begin{picture}(0,0)
                \put(0,195){b.} %
            \end{picture}
        \end{subfigure}

        \centering
        \vspace*{-1em} %
        {
        \begin{subfigure}[b]{0.6\textwidth}
            \includegraphics[width=\textwidth]{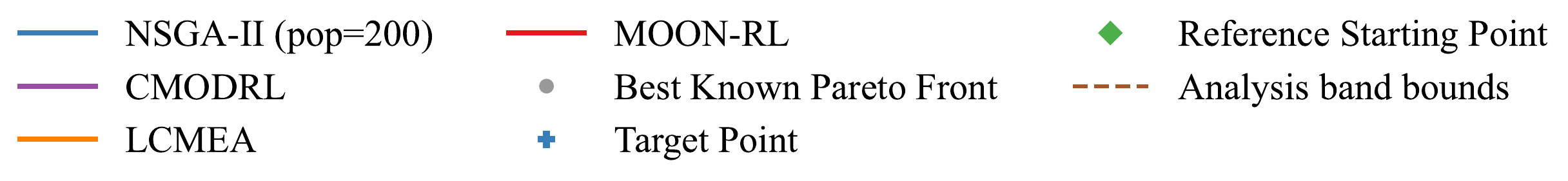}
        \end{subfigure}
        }

        \caption{``DAS-CMOP 3-11'' problem: EAF (a) and HV progression (b) of different algorithms, including $5^{th}$ and $95^{th}$ percentiles.}
        \label{fig:dascmop_best}
    \end{figure}

For DAS-CMOP3-11, we use a target point at $(0.2, 0.2)$ in the objective space. On this convergence-hard synthetic problem, Figure \ref{fig:dascmop_best} shows that NSGA-II, LCMEA, and CMODRL reliably approach the best-known Pareto front in the region of interest, while MOON-RL behaves less consistently. Some MOON-RL runs also get stuck, but the median attainment surface remains close to the best-known front. The summary in Table \ref{tab:main_results_best} reflects this picture and the full table with additional baselines is available in Appendix \ref{app:dascmop3_11_full_table}. The final hypervolume and generational distance of MOON-RL are close to those of the strongest compared methods, but the evolutionary baselines reach the hypervolume thresholds more consistently and, in several cases, faster. On DAS-CMOP3-11, MOON-RL therefore remains competitive in some runs, but the comparison also highlights a limitation of the current implementation on convergence-hard constrained problems.

Additional DAS-CMOP experiments on DAS-CMOP1 and DAS-CMOP3 are reported in Appendix~\ref{app:additional_dascmop}. These experiments cover diversity-, feasibility-, convergence-, and combined-difficulty settings, while keeping the feasibility difficulty compatible with the current feasible-initialization procedure. Overall, the results support the same interpretation as the DAS-CMOP3-11 case: MOON-RL reaches the prescribed analysis band and remains relevant as a reference-directed navigation method, but the strongest constrained MOEAs, especially LCMEA and often NSGA-II, generally achieve better convergence and broader Pareto-front coverage. The gap is most visible on convergence-difficulty and several DAS-CMOP3 settings, whereas MOON-RL is closer to the compared baselines on the diversity-difficulty cases.

\subsection{Global search scenario (MODAct ``CS1'')}

Previous results reflect the intended use of the proposed algorithm, namely a focused search along the user-specified reference direction. To enable a fair comparison with algorithms that are designed to explore the entire Pareto front, we also conducted a complementary global sweep, running the same algorithm sequentially on five distinct target points: $(0.22, 51.2), (0.26, 92.6), (0.39, 138.0), (0.58, 172.0), \text{and } (0.85, 182.8)$. The evaluation budget was divided evenly across these runs and each run initialized from scratch.

\begin{figure}[H]
        \centering
        \begin{subfigure}[b]{0.49\textwidth}
            \includegraphics[width=\textwidth]{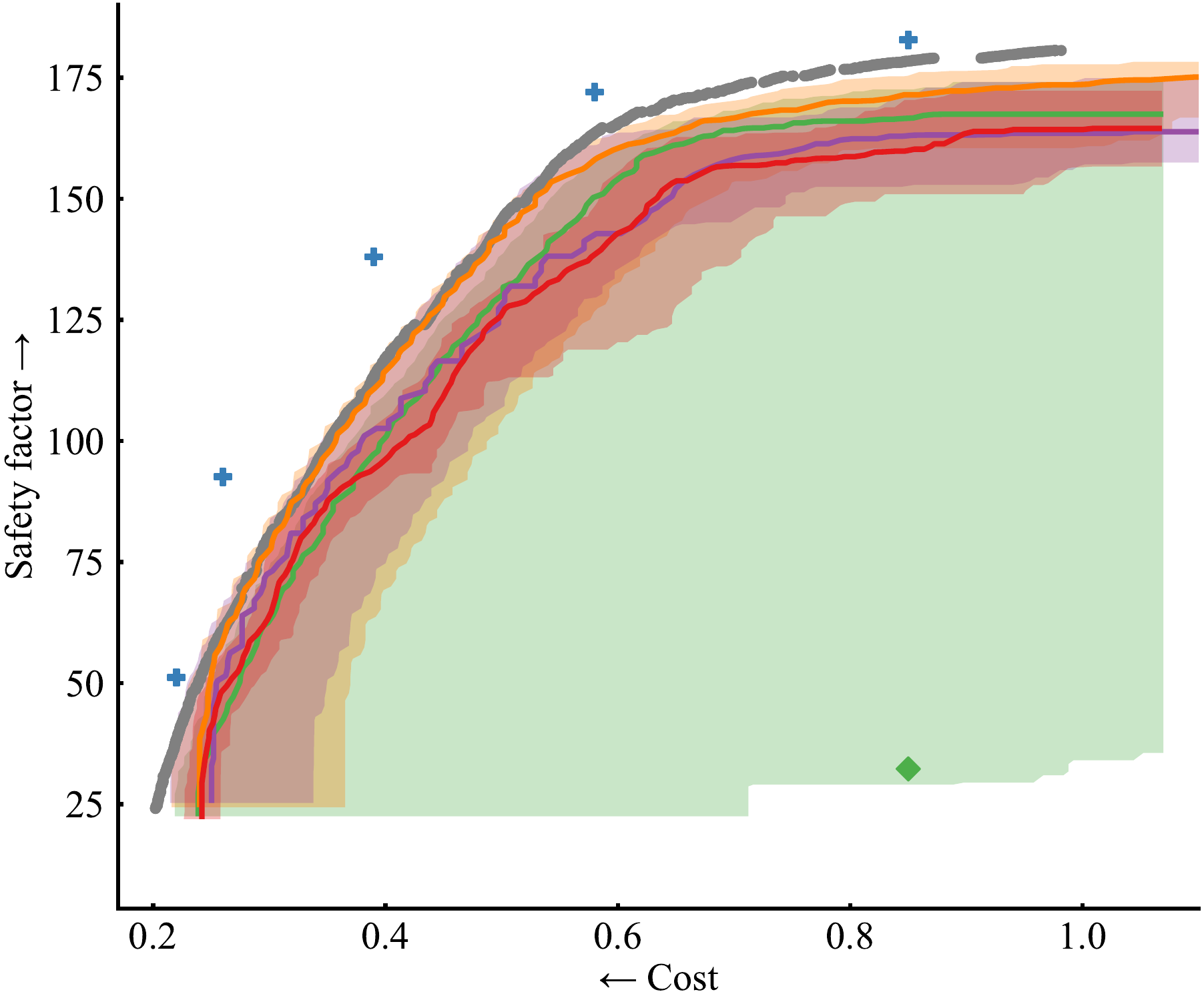}
            \begin{picture}(0,0)
                \put(0,195){a.} %
            \end{picture}

        \end{subfigure}
        \hfill
        \begin{subfigure}[b]{0.49\textwidth}
            \includegraphics[width=\textwidth]{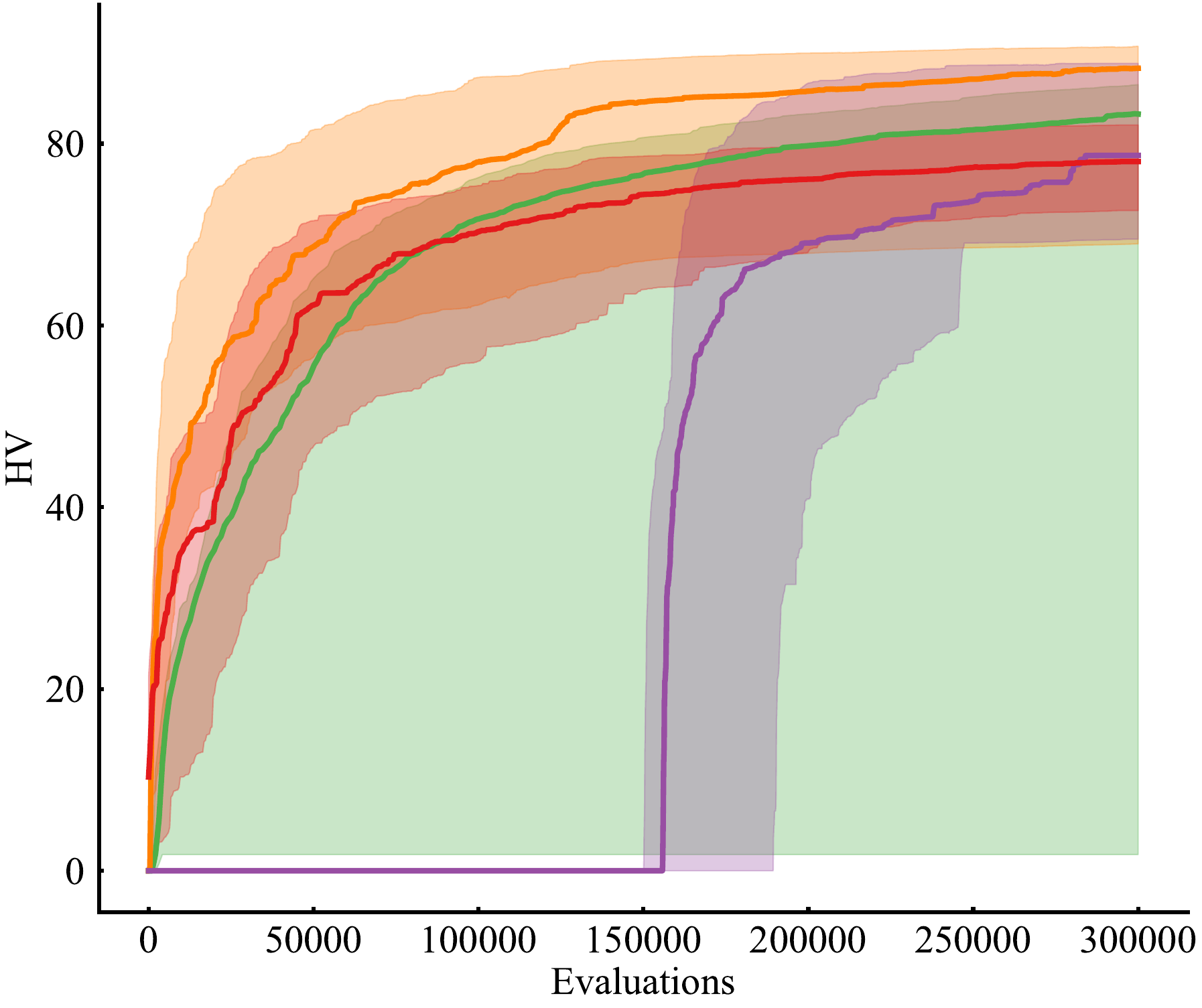}
            \begin{picture}(0,0)
                \put(0,195){b.} %
            \end{picture}
        \end{subfigure}

        \centering
        \vspace*{-1em} %
        {
        \begin{subfigure}[b]{0.6\textwidth}
            \includegraphics[width=\textwidth]{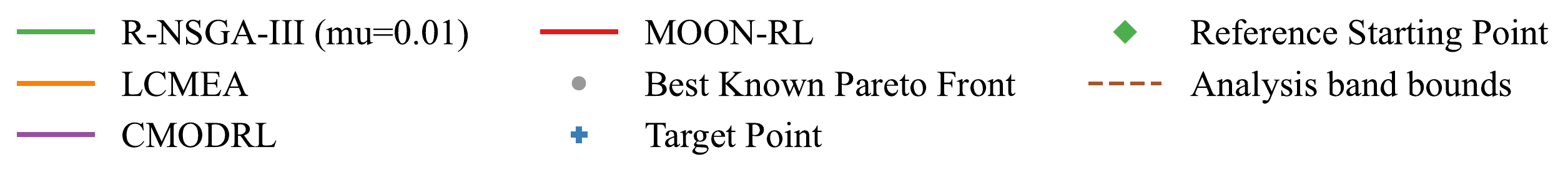}
        \end{subfigure}
        }

        \caption{``CS1'' problem (global search): EAF (a) and HV progression (b) of different algorithms, including $5^{th}$ and $95^{th}$ percentiles.}
        \label{fig:cs1_best_global}
    \end{figure}

In this scenario, LCMEA achieves the highest overall hypervolume and the lowest generational distance among the methods shown in Table \ref{tab:main_results_best}. CMODRL and R-NSGA-III also obtain high final hypervolume, while MOON-RL remains competitive despite being run sequentially with a divided budget. This is also reflected in the attainment surfaces in Figure \ref{fig:cs1_best_global}.

According to the EAF plot, MOON-RL remains close to the strongest compared methods over several target regions, although LCMEA and CMODRL provide better overall coverage of the front. This supports the interpretation that MOON-RL is better suited to focused objective-space navigation than to full-front approximation.

The slower HV progression of MOON-RL (Figure \ref{fig:cs1_best_global}) is due to the experimental setup, where each target was optimized sequentially and the algorithm was restarted from scratch for every direction.

\begin{table}
    \centering
    \caption{Results summary for main results. Success rate in parentheses shows how many runs reached the corresponding HV target. $\uparrow$ means higher is better, $\downarrow$ means lower is better. Only \emph{median} results for the \emph{best performing} algorithms are shown. Full tables with more algorithms and statistical results are available in Appendices~\ref{app:cs1_full_table}, \ref{app:ct1_full_table}, \ref{app:cts1_full_table}, \ref{app:dascmop3_11_full_table}, and \ref{app:cs1_global_full_table}.}
    \label{tab:main_results_best}
    
\begin{tabular}{lcccc}
        \toprule
        Algorithm & HV $\uparrow$ & GD $\downarrow$ & \shortstack{Eval @ 50\% HV $\downarrow$} & \shortstack{Eval @ 75\% HV $\downarrow$} \\
        \midrule
        \multicolumn{5}{c}{\emph{CS1}} \\
        \midrule
MOON-RL & $ 52.0 $ & $ 0.1 $ & $ \boldsymbol{4880}\ (\boldsymbol{30}/30) $ & $ 16440\ (\boldsymbol{30}/30) $ \\
APSEA & $ 40.2 $ & $ 1.1 $ & $ 123194\ (\boldsymbol{30}/30) $ & $ 135346\ (12/30) $ \\
LCMEA & $ \boldsymbol{54.6} $ & $ \boldsymbol{0.05} $ & $ 7800\ (29/30) $ & $ 30800\ (27/30) $ \\
CMODRL & $ 52.5 $ & $ 0.16 $ & $ 158325\ (29/30) $ & $ 183600\ (29/30) $ \\
NSGA-II (pop. 200) & $ 41.0 $ & $ 1.1 $ & $ 7600\ (\boldsymbol{30}/30) $ & $ 159800\ (13/30) $ \\
R-NSGA-II ($\varepsilon = 0.01$) & $ 37.9 $ & $ 1.3 $ & $ 6300\ (\boldsymbol{30}/30) $ & $ \boldsymbol{14400}\ (7/30) $ \\
NSGA-III (200 ref. dir.) & $ 39.1 $ & $ 1.3 $ & $ 8500\ (\boldsymbol{30}/30) $ & $ 83000\ (7/30) $ \\
R-NSGA-III ($\mu = 0.01$) & $ 38.4 $ & $ 1.3 $ & $ 7420\ (28/30) $ & $ 14840\ (7/30) $ \\
        \midrule
        \multicolumn{5}{c}{\emph{CT1}} \\
        \midrule
MOON-RL & $ 0.59 $ & $ 0.12 $ & $ \boldsymbol{1240}\ (\boldsymbol{30}/30) $ & $ 5120\ (\boldsymbol{30}/30) $ \\
APSEA & $ \boldsymbol{0.61} $ & $ \boldsymbol{0.07} $ & $ 11169\ (\boldsymbol{30}/30) $ & $ 13694\ (\boldsymbol{30}/30) $ \\
LCMEA & $ 0.59 $ & $ 0.17 $ & $ 1400\ (\boldsymbol{30}/30) $ & $ 5500\ (\boldsymbol{30}/30) $ \\
CMODRL & $ 0.60 $ & $ 0.10 $ & $ 148538\ (\boldsymbol{30}/30) $ & $ 149100\ (\boldsymbol{30}/30) $ \\
NSGA-II (pop. 200) & $ \boldsymbol{0.61} $ & $ \boldsymbol{0.07} $ & $ 2200\ (\boldsymbol{30}/30) $ & $ 5200\ (\boldsymbol{30}/30) $ \\
R-NSGA-II ($\varepsilon = 0.01$) & $ \boldsymbol{0.61} $ & $ \boldsymbol{0.07} $ & $ 2100\ (\boldsymbol{30}/30) $ & $ \boldsymbol{4000}\ (\boldsymbol{30}/30) $ \\
NSGA-III (200 ref. dir.) & $ \boldsymbol{0.61} $ & $ 0.08 $ & $ 2200\ (\boldsymbol{30}/30) $ & $ 4800\ (\boldsymbol{30}/30) $ \\
R-NSGA-III ($\mu = 0.01$) & $ 0.60 $ & $ 0.12 $ & $ 2756\ (24/30) $ & $ 5724\ (24/30) $ \\
        \midrule
        \multicolumn{5}{c}{\emph{CTS1}} \\
        \midrule
MOON-RL & $ 130.7 $ & $ 0.3 $ & $ 4240\ (\boldsymbol{30}/30) $ & $ 8400\ (\boldsymbol{30}/30) $ \\
APSEA & $ 132.8 $ & $ \boldsymbol{0.15} $ & $ 130540\ (\boldsymbol{30}/30) $ & $ 132309\ (\boldsymbol{30}/30) $ \\
LCMEA & $ 130.8 $ & $ 0.18 $ & $ 4600\ (\boldsymbol{30}/30) $ & $ 12800\ (\boldsymbol{30}/30) $ \\
CMODRL & $ 131.4 $ & $ \boldsymbol{0.15} $ & $ 134063\ (\boldsymbol{30}/30) $ & $ 138975\ (\boldsymbol{30}/30) $ \\
NSGA-II (pop. 200) & $ 128.1 $ & $ 0.2 $ & $ 3000\ (\boldsymbol{30}/30) $ & $ 5800\ (\boldsymbol{30}/30) $ \\
R-NSGA-II ($\varepsilon = 0.01$) & $ \boldsymbol{135.6} $ & $ 0.3 $ & $ \boldsymbol{2500}\ (\boldsymbol{30}/30) $ & $ \boldsymbol{4600}\ (\boldsymbol{30}/30) $ \\
NSGA-III (200 ref. dir.) & $ 111.1 $ & $ 0.4 $ & $ 3100\ (\boldsymbol{30}/30) $ & $ 6800\ (29/30) $ \\
R-NSGA-III ($\mu = 0.01$) & $ 134.4 $ & $ 0.3 $ & $ 3408\ (\boldsymbol{30}/30) $ & $ 6496\ (\boldsymbol{30}/30) $ \\
        \midrule
        \multicolumn{5}{c}{\emph{DAS-CMOP3-11}} \\
        \midrule
MOON-RL & $ 0.38 $ & $ 0.05 $ & $ 15360\ (24/30) $ & $ 28360\ (24/30) $ \\
APSEA & $ \boldsymbol{0.39} $ & $ \boldsymbol{0.02} $ & $ 8272\ (\boldsymbol{30}/30) $ & $ 10528\ (\boldsymbol{30}/30) $ \\
LCMEA & $ \boldsymbol{0.39} $ & $ \boldsymbol{0.02} $ & $ 5600\ (\boldsymbol{30}/30) $ & $ 8500\ (\boldsymbol{30}/30) $ \\
CMODRL & $ \boldsymbol{0.39} $ & $ \boldsymbol{0.02} $ & $ \boldsymbol{4275}\ (28/30) $ & $ \boldsymbol{5550}\ (28/30) $ \\
NSGA-II (pop. 200) & $ \boldsymbol{0.39} $ & $ \boldsymbol{0.02} $ & $ 19400\ (27/30) $ & $ 22400\ (27/30) $ \\
R-NSGA-II ($\varepsilon = 0.01$) & $ 0.01 $ & $ 6.83 $ & $ 48000\ (9/30) $ & $ 50600\ (9/30) $ \\
NSGA-III (200 ref. dir.) & $ \boldsymbol{0.39} $ & $ 0.03 $ & $ 26500\ (\boldsymbol{30}/30) $ & $ 31800\ (\boldsymbol{30}/30) $ \\
R-NSGA-III ($\mu = 0.01$) & $ 0.00 $ & $ 24.63 $ & $ 58580\ (1/30) $ & $ 61206\ (1/30) $ \\
        \midrule
        \multicolumn{5}{c}{\emph{CS1 (global search)}} \\
        \midrule
MOON-RL & $ 78.0 $ & $ 0.5 $ & $ 24800\ (\boldsymbol{30}/30) $ & $ 90200\ (\boldsymbol{30}/30) $ \\
APSEA & $ 69.3 $ & $ 1.1 $ & $ 123690\ (\boldsymbol{30}/30) $ & $ 174716\ (17/30) $ \\
LCMEA & $ \boldsymbol{88.3} $ & $ \boldsymbol{0.3} $ & $ 9600\ (29/30) $ & $ 43700\ (28/30) $ \\
CMODRL & $ 78.7 $ & $ 0.5 $ & $ 160500\ (29/30) $ & $ 202575\ (29/30) $ \\
NSGA-II (pop. 200) & $ 70.5 $ & $ 1.0 $ & $ \boldsymbol{8600}\ (\boldsymbol{30}/30) $ & $ 138100\ (20/30) $ \\
R-NSGA-II ($\varepsilon = 0.01$) & $ 68.9 $ & $ 1.1 $ & $ 8700\ (\boldsymbol{30}/30) $ & $ \boldsymbol{16800}\ (14/30) $ \\
NSGA-III (200 ref. dir.) & $ 67.7 $ & $ 1.2 $ & $ 10400\ (\boldsymbol{30}/30) $ & $ 133700\ (10/30) $ \\
R-NSGA-III ($\mu = 0.01$) & $ 83.3 $ & $ 0.4 $ & $ 33344\ (23/30) $ & $ 79192\ (19/30) $ \\
        \bottomrule
    \end{tabular}

\end{table}

\subsection{Ablation studies}
\label{subsec:ablations}

\begin{figure}[h]
        \centering
        \begin{subfigure}[b]{0.49\textwidth}
            \includegraphics[width=\textwidth]{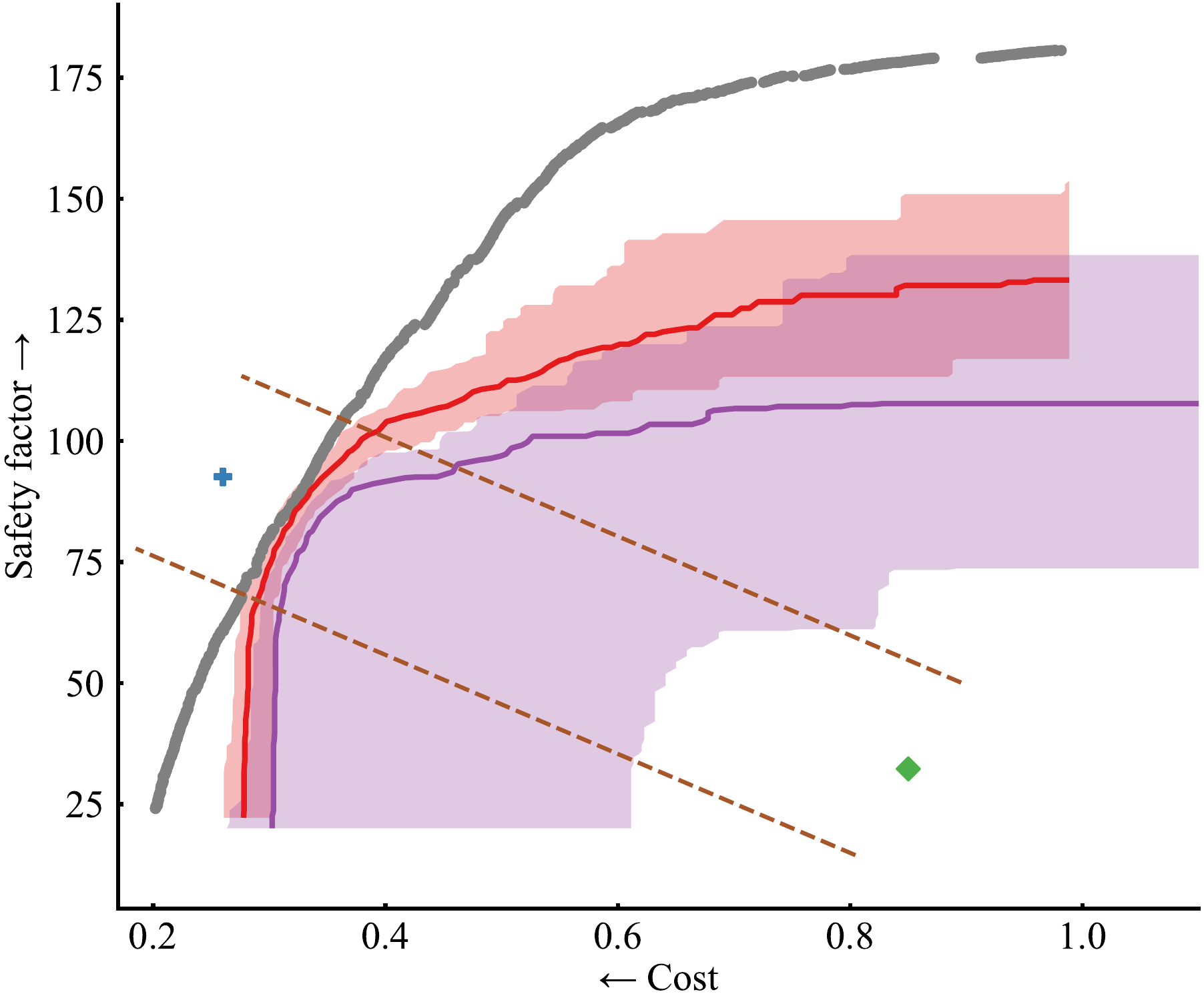}
            \begin{picture}(0,0)
                \put(0,195){a.} %
            \end{picture}

        \end{subfigure}
        \hfill
        \begin{subfigure}[b]{0.49\textwidth}
            \includegraphics[width=\textwidth]{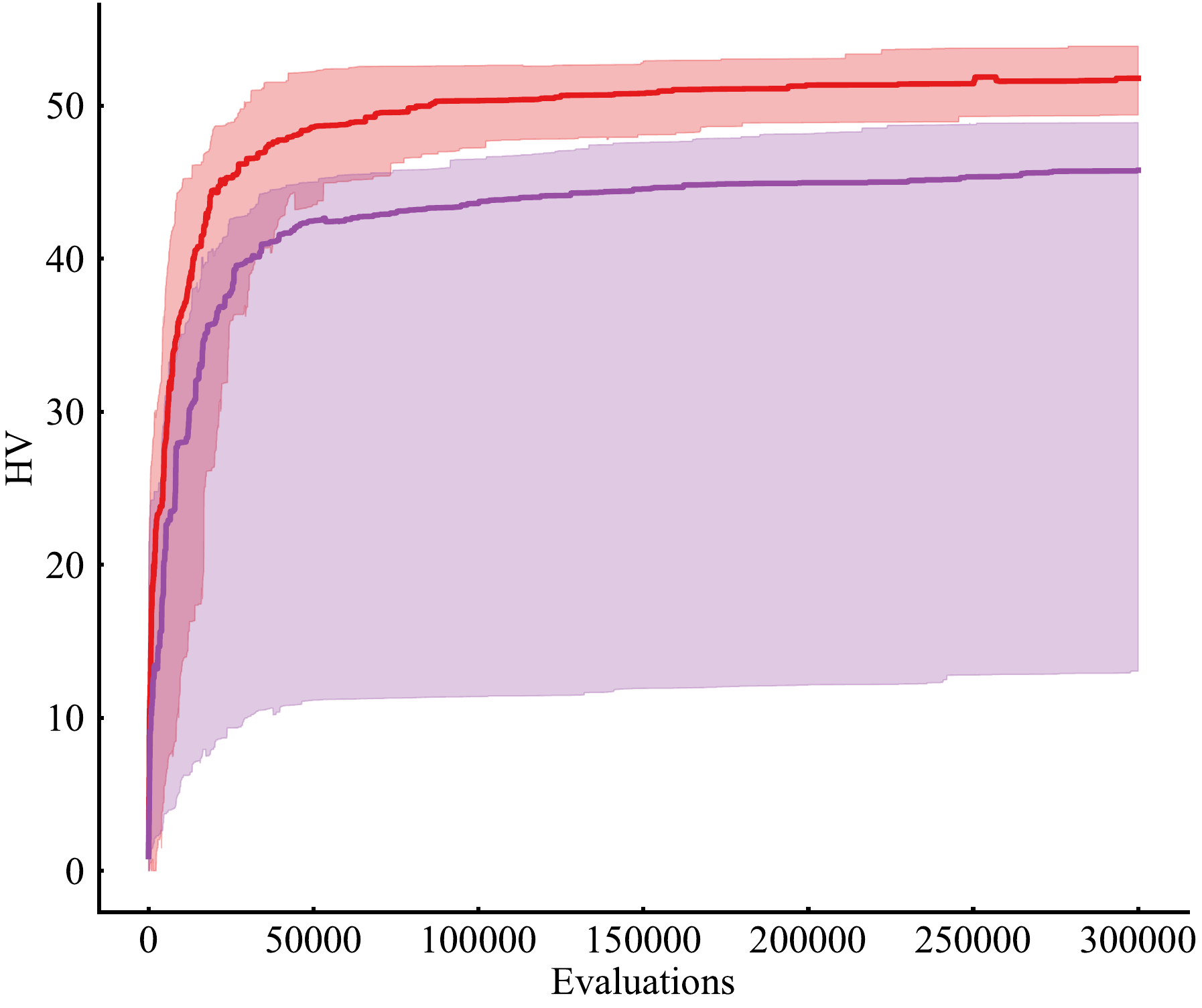}
            \begin{picture}(0,0)
                \put(0,195){b.} %
            \end{picture}
        \end{subfigure}
        \centering

        \vspace*{-1em} %

        \centering
        \begin{subfigure}[b]{0.49\textwidth}
            \includegraphics[width=\textwidth]{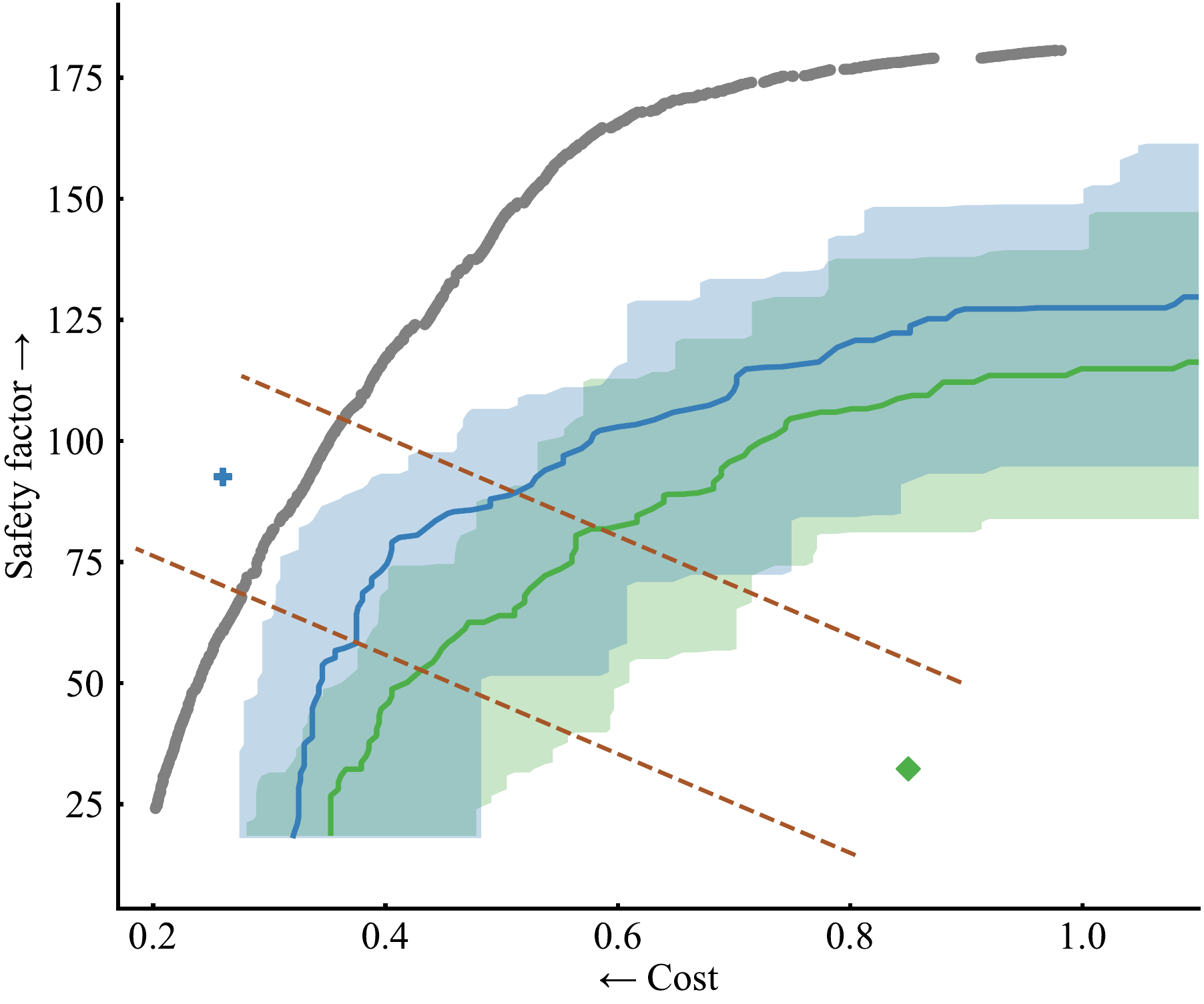}
            \begin{picture}(0,0)
                \put(0,195){c.} %
            \end{picture}

        \end{subfigure}
        \hfill
        \begin{subfigure}[b]{0.49\textwidth}
            \includegraphics[width=\textwidth]{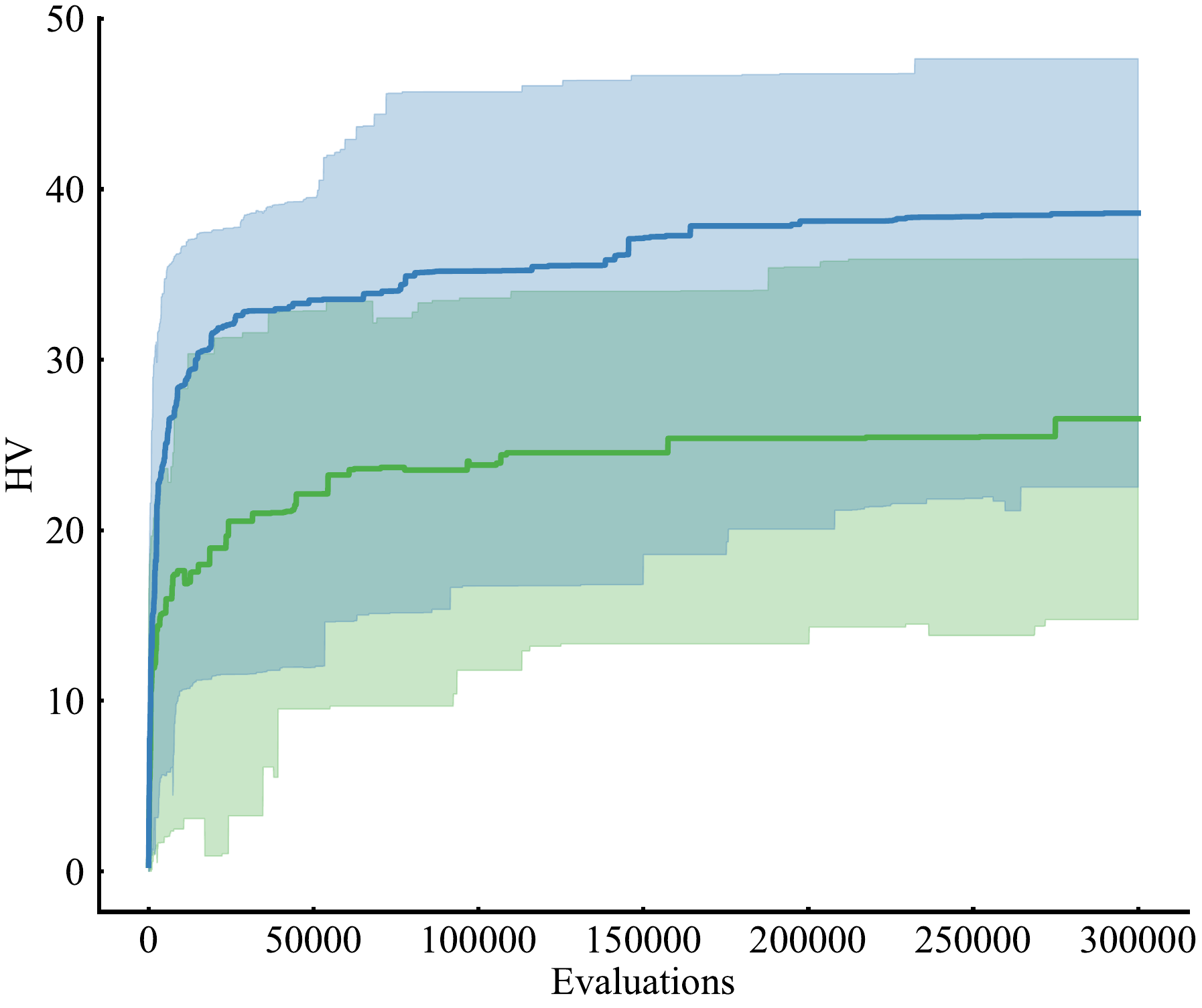}
            \begin{picture}(0,0)
                \put(0,195){d.} %
            \end{picture}
        \end{subfigure}
        \centering

        \vspace*{-1em} %

        {
        \begin{subfigure}[b]{0.8\textwidth}
            \includegraphics[width=\textwidth]{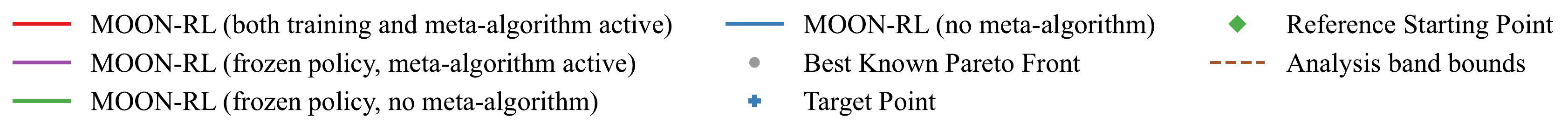}
        \end{subfigure}
        }

        \caption{ EAF and HV progression of MOON-RL run with and without meta-algorithm and with and without training the policy network. Panels (a, b) show runs with the meta-algorithm active, panels (c, d) runs without it. Focused search scenario.}
        \label{fig:ablation_metaNoMeta}
    \end{figure}

To assess the contribution of the meta-algorithm and the policy-training process, we perform ablation experiments on CS1 in the same focused setting as above. Additional ablations on CT1, CTS1, and DAS-CMOP are reported in Appendix \ref{app:additional_meta_no_meta} to verify that the observed effect is not specific to CS1. Figure \ref{fig:ablation_metaNoMeta} compares: the full MOON-RL algorithm with MOON-RL without the meta-algorithm (Figure \ref{fig:ablation_metaNoMeta} a \& b), and MOON-RL with a frozen policy (no training) with a variant with both the meta-algorithm disabled and the policy frozen (Figure \ref{fig:ablation_metaNoMeta} c \& d). In the frozen-policy variants, the PPO weights are fixed, so the policy induces a fixed, state-independent Gaussian sampling scheme over the design variables: at each step it generates a new design by adding zero-mean Gaussian noise to the current design, with the noise scale set by $\kappa$ and the design-variable ranges (see Section \ref{sec:algo}). Because the parameters never change, this sampler is non-adaptive and does not react to rewards or constraint violations, but it still provides a structured random search in the vicinity of the current design rather than uniform sampling. Together with the meta-algorithm, which periodically resets the policy and resamples the starting points from the continuously updated archive of non-dominated solutions, this non-adaptive sampler already produces reasonably strong performance.

The results highlight the importance of both components. Removing the meta-algorithm causes the algorithm to get stuck in local optima more often and leads to a clear degradation in both attainment and hypervolume. With the meta-algorithm active, both the frozen and the training-enabled variants improve relative to their counterparts without meta algorithm, showing that the restart and archive-based starting-point logic is an effective mechanism for escaping local optima and maintaining progress. In both settings, enabling RL training improves performance relative to the corresponding frozen-policy variant, and the full MOON-RL configuration achieves the highest median hypervolume and best empirical attainment among all four variants.

\begin{figure}[h]
        \centering
        \begin{subfigure}[b]{0.49\textwidth}
            \includegraphics[width=\textwidth]{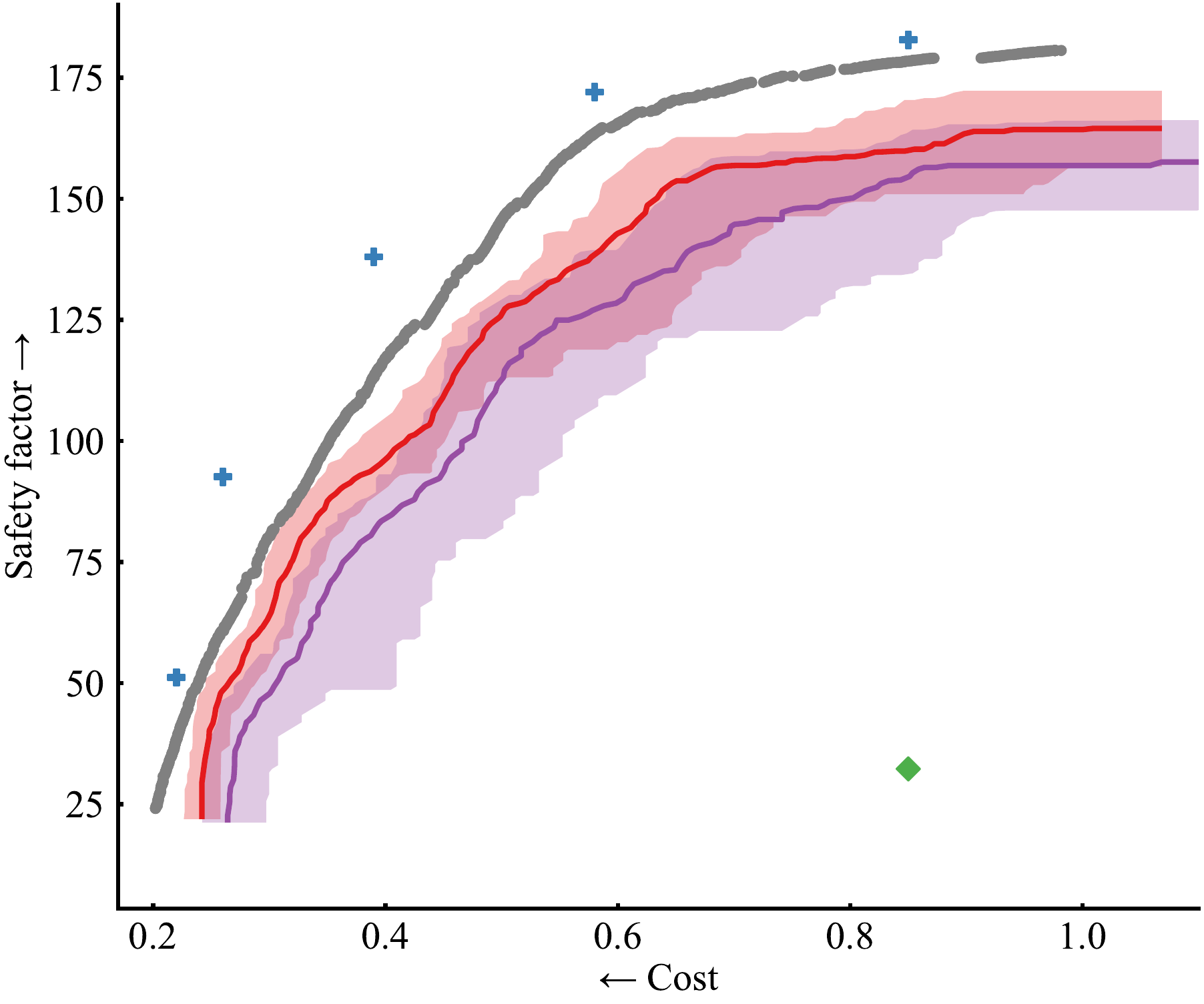}
            \begin{picture}(0,0)
                \put(0,195){a.} %
            \end{picture}

        \end{subfigure}
        \hfill
        \begin{subfigure}[b]{0.49\textwidth}
            \includegraphics[width=\textwidth]{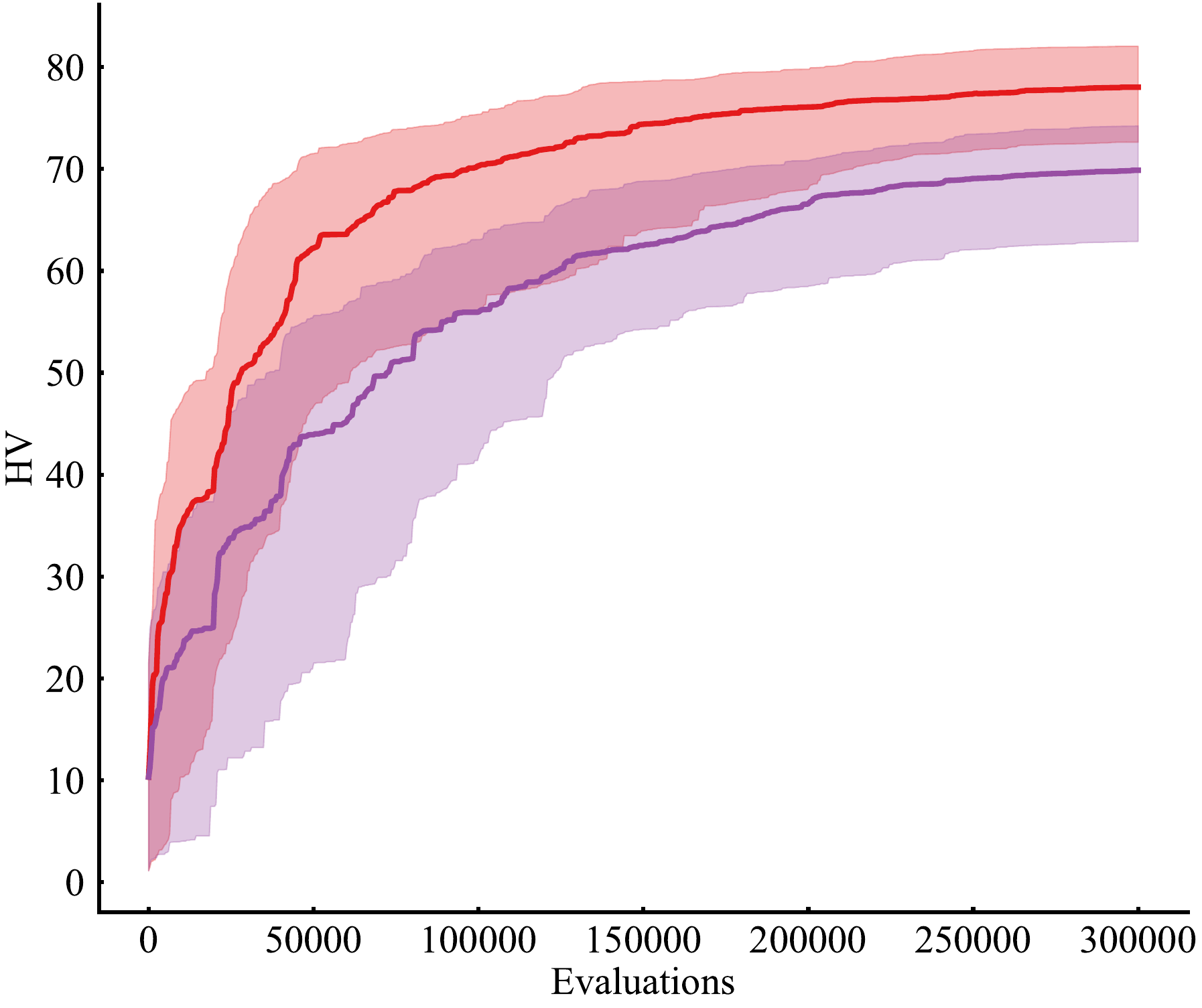}
            \begin{picture}(0,0)
                \put(0,195){b.} %
            \end{picture}
        \end{subfigure}
        \centering

        \vspace*{-1em} %

        \centering
        \begin{subfigure}[b]{0.49\textwidth}
            \includegraphics[width=\textwidth]{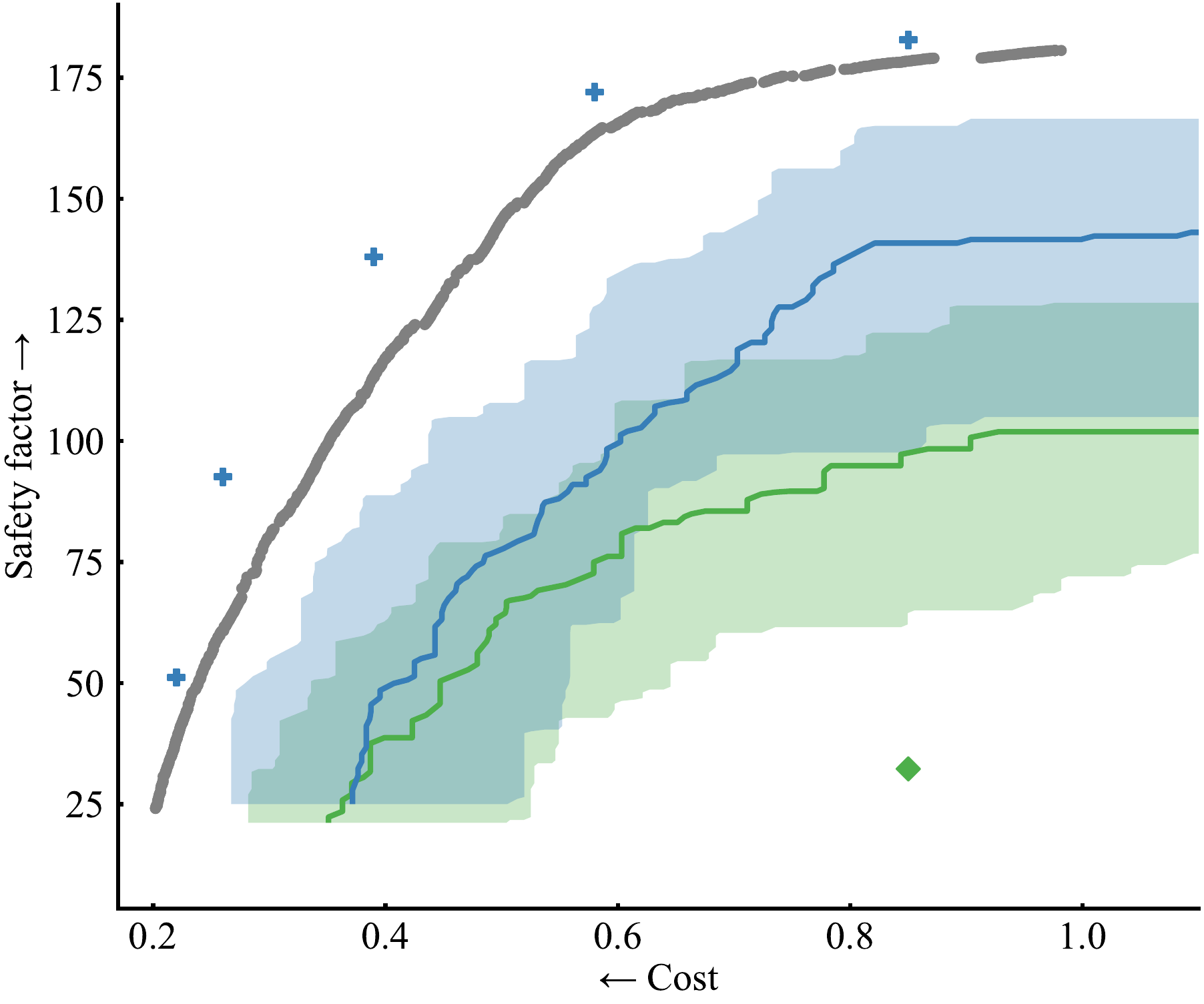}
            \begin{picture}(0,0)
                \put(0,195){c.} %
            \end{picture}

        \end{subfigure}
        \hfill
        \begin{subfigure}[b]{0.49\textwidth}
            \includegraphics[width=\textwidth]{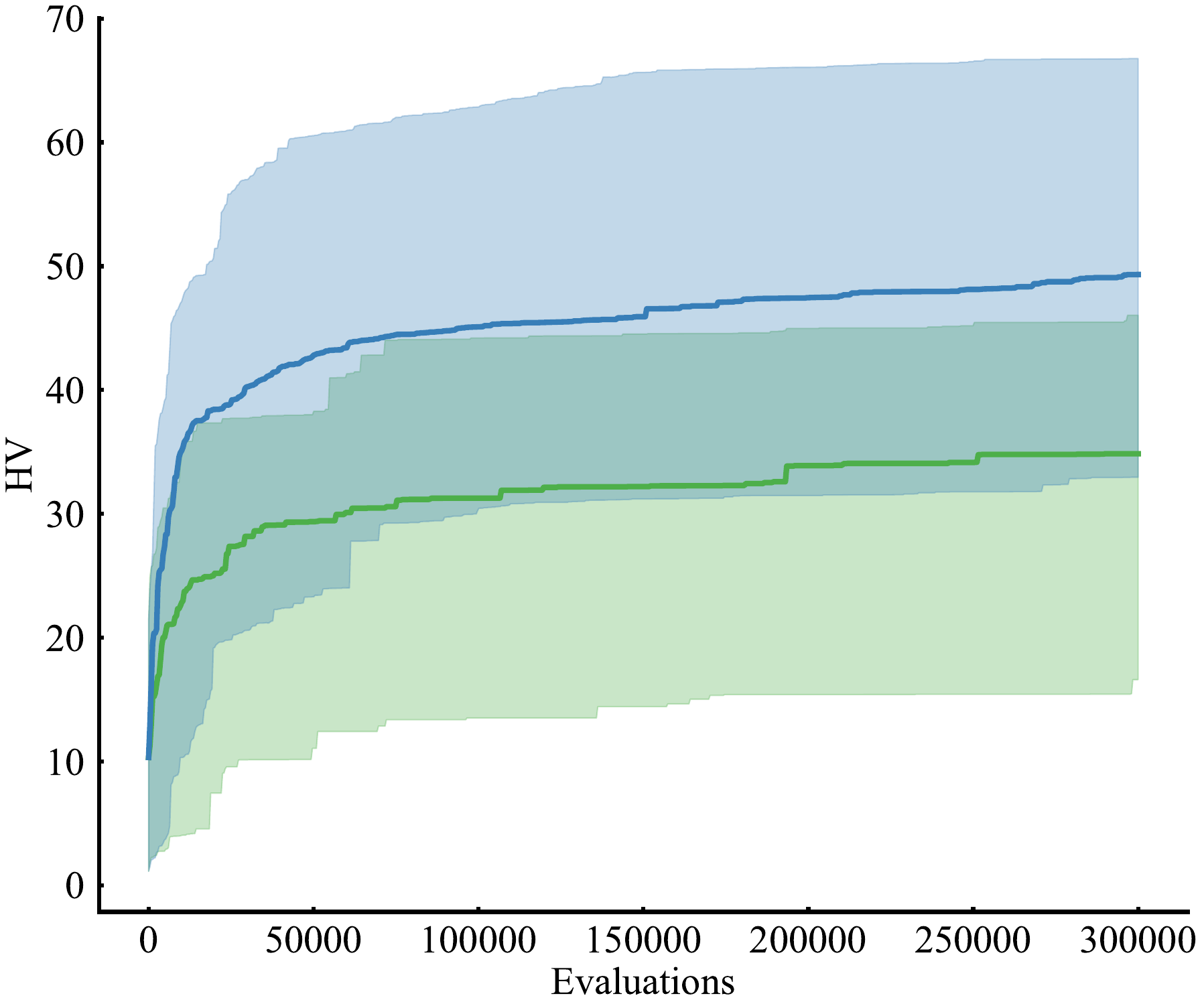}
            \begin{picture}(0,0)
                \put(0,195){d.} %
            \end{picture}
        \end{subfigure}
        \centering

        \vspace*{-1em} %

        {
        \begin{subfigure}[b]{0.8\textwidth}
            \includegraphics[width=\textwidth]{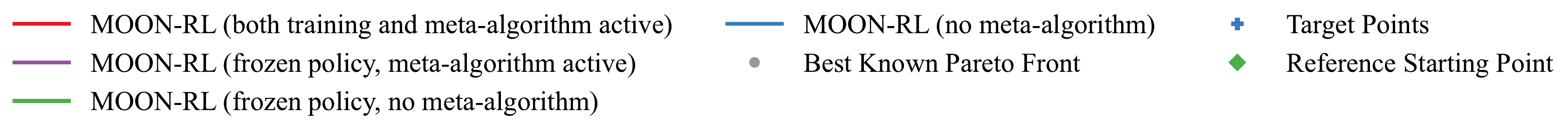}
        \end{subfigure}
        }

        \caption{ EAF and HV progression of MOON-RL run with and without meta-algorithm and with and without training the policy network. Panels (a, b) show runs with the meta-algorithm active, panels (c, d) runs without it. Global search scenario.}
        \label{fig:ablation_metaNoMeta_global}
    \end{figure}

Repeating the ablation in the global-search setting on CS1 (Figure \ref{fig:ablation_metaNoMeta_global}) yields qualitatively similar behaviour. The additional ablations on CT1, CTS1, and DAS-CMOP3-11 reported in Appendix \ref{app:additional_meta_no_meta} further support the same conclusion: in all tested problems, the frozen-policy variants are outperformed by the corresponding variants with RL training enabled. The difference is especially visible on ``DAS-CMOP3-11''. Overall, these ablations support a division of labour in MOON-RL: the meta-algorithm provides global exploration through restarts and archive updates, while the RL-driven adaptive sampling process refines navigation between resets, and both components contribute meaningfully to the observed performance. Complementary PPO diagnostics in Appendix~\ref{app:ppo-diagnostics} support this interpretation from the learning side: in isolated PPO runs without the MOON-RL meta-algorithm active, the episodic return increases before approaching a plateau, indicating that the actor learns to obtain higher reward under the proposed objective-space navigation reward. In an interactive setting, the meta-algorithm can be replaced with the user, who can at any moment reset the agent, the starting point, and the reference direction, if the Pareto Front appears to be stuck.

We also ran MOON-RL with different values of the meta-algorithm parameters, as summarized in Appendix \ref{app:ablation-additional}. On CS1, the additional experiments indicate that the starting-point reset interval has the strongest impact on performance: very rare resets clearly worsen both the median EAF and the width of the inter-percentile band, whereas resetting the starting point at least once every 100 episodes is sufficient, and more frequent resets bring only marginal gains. The decay interval for the scaling factors $\boldsymbol{\kappa}$ mainly controls how long the algorithm keeps using larger steps. If the decay is too fast, the step sizes become small early in the run, and the search is more likely to become trapped in local optima. By contrast, within the tested range, the constraint-penalty $\lambda$ growth interval produces almost indistinguishable EAFs, suggesting that MOON-RL is relatively insensitive to this parameter on ``CS1''.

\subsection{Alternative Adaptive Samplers within the Navigation Framework}
\label{subsec:alternative-samplers}

We developed MOON-RL to use a lightweight RL agent as the sampler that updates the design so that the current solution advances along the reference direction. To test whether reinforcement learning is the only effective choice, we also evaluate two adaptive non-RL samplers in place of PPO: the cross-entropy method (CEM) \cite{de2005tutorial}, which updates a diagonal Gaussian distribution from high-reward perturbations, and CMA-ES \cite{hansen2016cma}, which adapts the step size and full covariance of its sampling distribution. Neither conditions its proposal distribution directly on the objective-space observation. In each case, we retain the same target, reward, non-dominated archive, meta-algorithm, reset and parameter schedules, constraint handling, design-variable transformations, evaluation budget, and random seeds.

The comparison includes MOON-RL (PPO), MOON-RL (CEM), MOON-RL (CEM, independent trajectories), and MOON-RL (CMA-ES). PPO maintains ten trajectories over eight steps, each receiving one perturbation per step and advancing independently. CEM and CMA-ES update a sampling distribution from ranked candidate populations. For these configurations, we therefore reorganize the rollout as one sequential trajectory: at each of eight steps, ten candidates are sampled around the current design, and the highest-reward candidate becomes the design for the next step. The total budget remains $10 \times 8 = 80$ candidate evaluations per rollout block, but ten parallel trajectories are replaced by ten candidate samples at each sequential step. Budget and infrastructure are matched, while the rollout structure is not. MOON-RL (CEM, independent trajectories) restores the PPO rollout structure: each trajectory keeps the candidate it produced instead of all ten collapsing onto the highest-reward one. The CEM distribution remains shared, and is updated at each step from the perturbations pooled across all ten trajectories. CMA-ES uses the reference pycma implementation \cite{hansen2019pycma}. Full implementations and parameter settings are provided in Appendices~\ref{app:cem_implementation} and \ref{app:cma_implementation}.

    \begin{figure}[H]
        \centering
        \begin{subfigure}[b]{0.49\textwidth}
            \includegraphics[width=\textwidth]{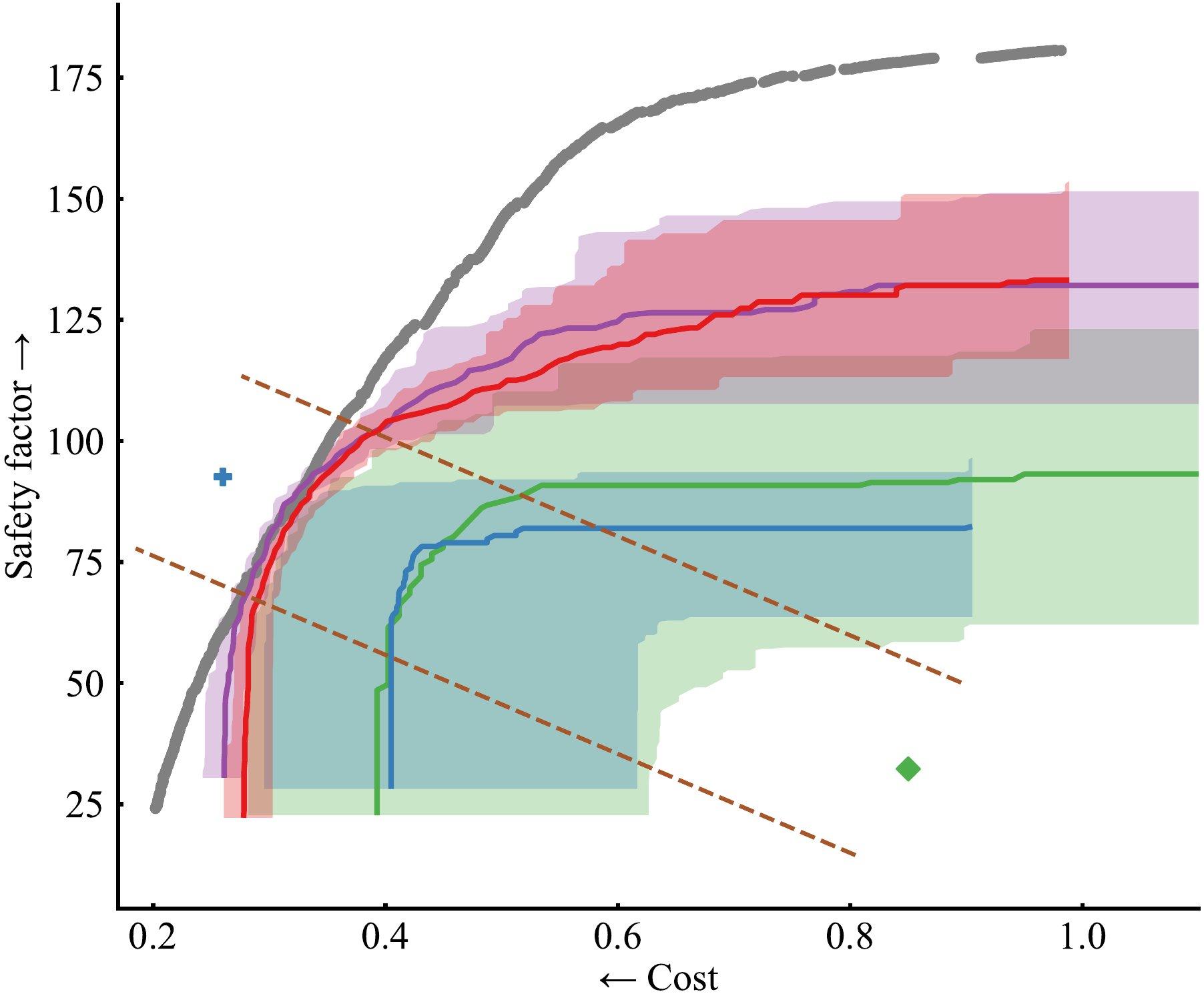}
            \begin{picture}(0,0)
                \put(0,195){a.} %
            \end{picture}

        \end{subfigure}
        \hfill
        \begin{subfigure}[b]{0.49\textwidth}
            \includegraphics[width=\textwidth]{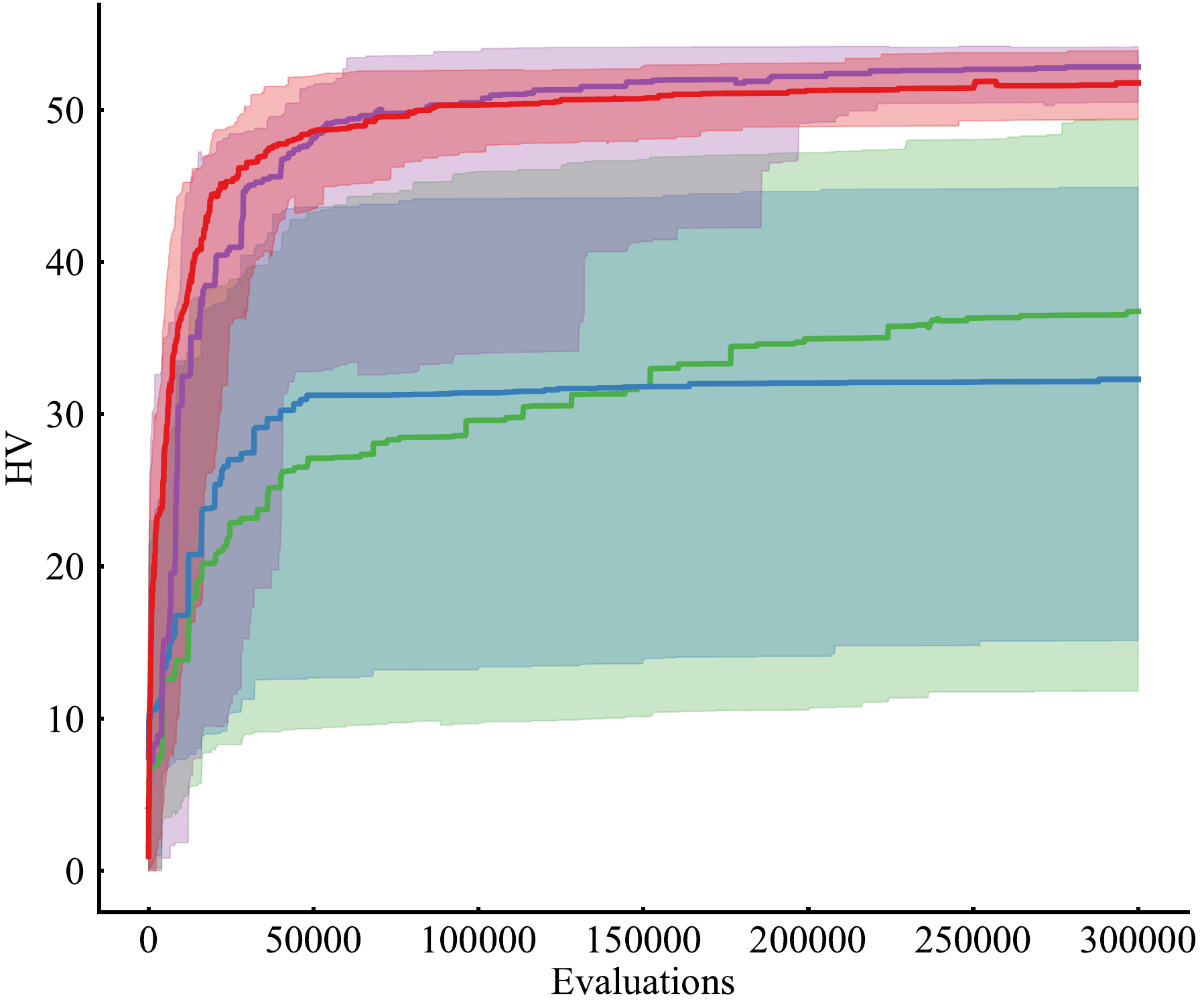}
            \begin{picture}(0,0)
                \put(0,195){b.} %
            \end{picture}
        \end{subfigure}

        \centering
        \vspace*{-1em} %
        {
        \begin{subfigure}[b]{0.6\textwidth}
            \includegraphics[width=\textwidth]{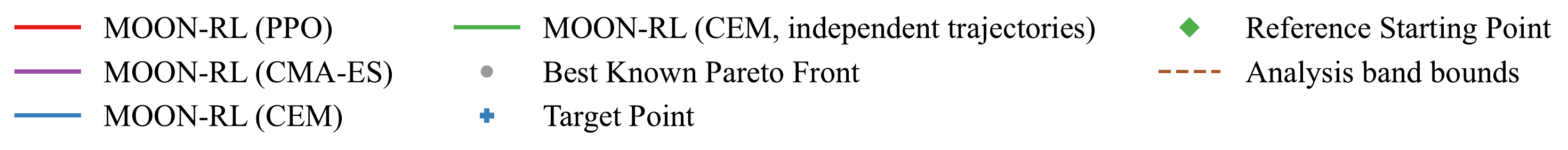}
        \end{subfigure}
        }

        \caption{``CS1'' problem: EAF (a) and HV progression (b) of MOON-RL with different adaptive samplers, including $5^{th}$ and $95^{th}$ percentiles.}
        \label{fig:cs1_non_rl}
    \end{figure}

    \begin{figure}[H]
        \centering
        \begin{subfigure}[b]{0.49\textwidth}
            \includegraphics[width=\textwidth]{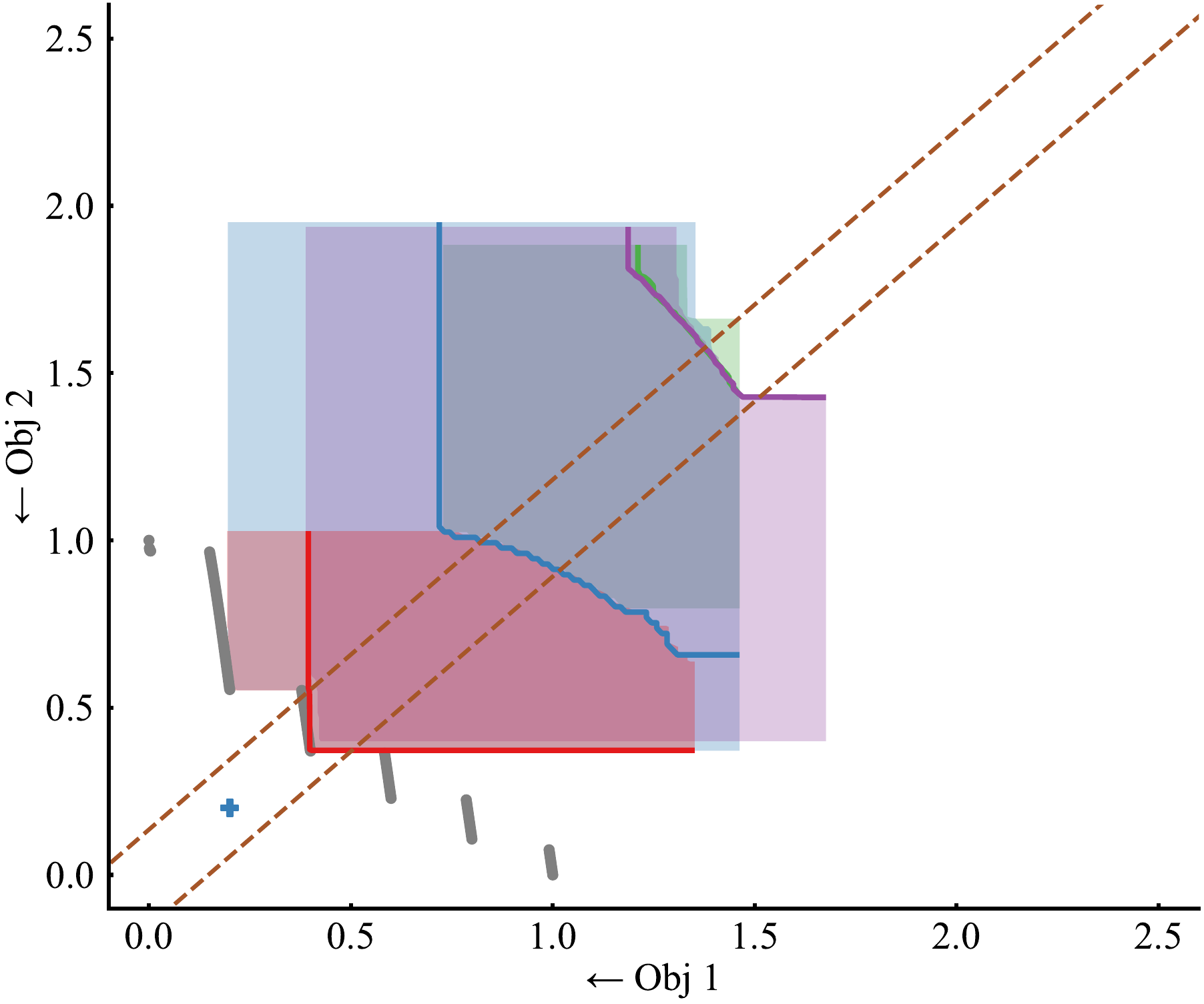}
            \begin{picture}(0,0)
                \put(0,195){a.} %
            \end{picture}

        \end{subfigure}
        \hfill
        \begin{subfigure}[b]{0.49\textwidth}
            \includegraphics[width=\textwidth]{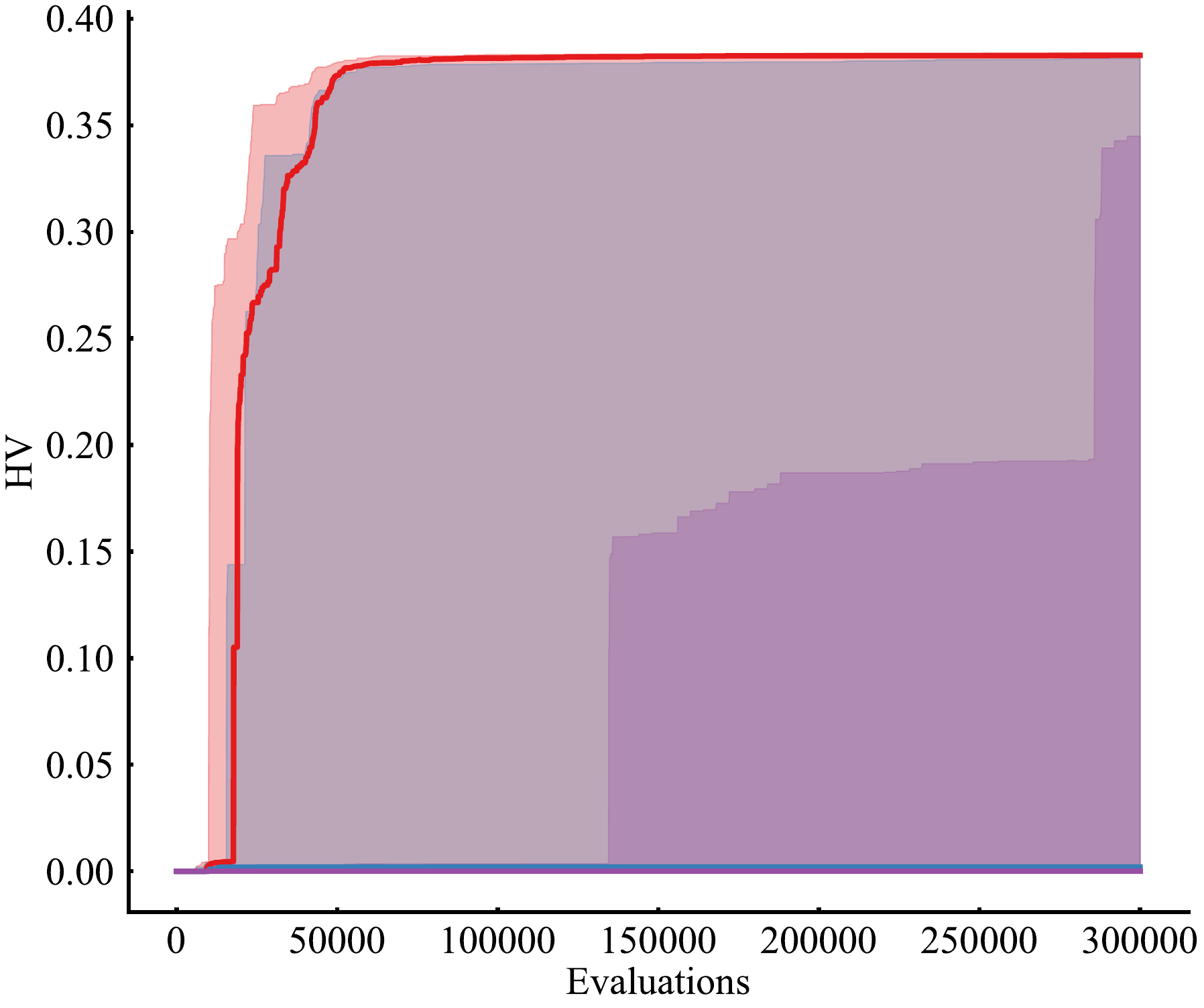}
            \begin{picture}(0,0)
                \put(0,195){b.} %
            \end{picture}
        \end{subfigure}

        \centering
        \vspace*{-1em} %
        {
        \begin{subfigure}[b]{0.6\textwidth}
            \includegraphics[width=\textwidth]{figures/adaptive_samplers/legend.pdf}
        \end{subfigure}
        }

        \caption{``DAS-CMOP 3-11'' problem: EAF (a) and HV progression (b) of MOON-RL with different adaptive samplers, including $5^{th}$ and $95^{th}$ percentiles.}
        \label{fig:dascmop3-11_non_rl}
    \end{figure}

Figure~\ref{fig:cs1_non_rl} shows the results on ``CS1'' and Figure~\ref{fig:dascmop3-11_non_rl} those on ``DAS-CMOP3-11''. The corresponding results for ``CT1'' and ``CTS1'' are reported in Appendix~\ref{app:adaptive_sampler_results}. On ``CS1'', MOON-RL (CMA-ES) and MOON-RL (PPO) attain closely comparable objective-space coverage and hypervolume progression, with CMA-ES reaching a slightly higher median toward the end of the budget and the uncertainty regions overlapping throughout. Both CEM configurations converge to lower hypervolume levels with considerably wider inter-percentile bands, and the independent-trajectory configuration closes only part of this gap. On ``CT1'' all four configurations attain similar results, with PPO and CMA-ES slightly ahead. On ``CTS1'', CMA-ES performs better than PPO, while both CEM configurations remain below either. On ``DAS-CMOP3-11'', PPO clearly outperforms every other configuration and advances farthest along the reference direction, followed by the sequential CEM configuration, with CMA-ES and the independent-trajectory CEM configuration further behind.

One possible explanation for the weaker CEM performance is premature contraction of its sampling distribution, since its variance is estimated directly from a small elite subset. CMA-ES instead adapts its step size from the evolution path, independently of the spread of the selected samples, and combines this with a full covariance model. Both samplers adapt, but CEM keeps pace only on ``CT1'' and falls behind on the other three problems. Between PPO and CMA-ES, neither is uniformly preferable across the examined problems.

Taken together, these results show that the navigation formulation is not intrinsically restricted to reinforcement learning. The framework accommodates adaptive samplers with substantially different internal mechanisms and rollout structures, and their relative effectiveness is problem-dependent. CMA-ES is slightly ahead of PPO on ``CS1'' and ``CTS1'' and level with it on ``CT1'', whereas PPO is clearly ahead on ``DAS-CMOP3-11''. No CEM configuration leads on any problem. This supports repurposing reinforcement learning as the sampler while showing that the formulation does not depend on it.

%% file: sections/5_Discussion_and_Conclusion.tex
\section{Discussion \& Conclusion}
\label{sec:conclusion}

The MOON-RL formulation of multi-objective search as an objective-space navigation task enables a reinforcement learning agent to make local, targeted improvements without approximating the entire Pareto front. In this work, we implement this idea in MOON-RL, where a lightweight policy steers design exploration along a reference direction in the objective space. Across constrained engineering benchmarks from the MODAct suite and additional DAS-CMOP problems with different difficulty settings, MOON-RL delivers competitive optimization results, with particularly strong performance on the CS1 problem.
At the same time, the comparison shows that specialized constrained MOEAs can be stronger on specific metrics or problem instances, especially when final front quality, convergence speed, or full-front coverage is prioritized.

Rather than aiming to compete with or to replace well-established multi-objective evolutionary algorithms or full-front coverage methods, MOON-RL offers a different lens on multi-objective optimization: it treats design as navigation in the objective space along a chosen preference direction. Unlike MORL or scalarization-based approaches that train weight-conditioned policies to generalize across a continuum of trade-off preferences, we deliberately focus on a simple policy tailored to a fixed preference direction. The results obtained on the design-related and synthetic benchmark problems considered here suggest that reinforcement learning can be used as an adaptive search operator for objective-space navigation in preference-directed engineering design. Under this view, MOON-RL can form the core of an interactive design companion that suggests local design updates toward user-specified trade-offs.

Although in this study the meta-algorithm is fixed, starting points are automatically chosen from the archive, and the target point remains fixed within each run, the same mechanism could provide a basis for future interactive use. MOON-RL only requires a starting design and a reference direction. A designer could in principle select or update the target point during runtime without resetting the agent, or treat the current state of an evolving design as a moving starting point, in which case the agent is reinitialized to navigate from whatever design is currently on the table. Supplementary Movie 1 demonstrates such interactive retargeting qualitatively, while its quantitative evaluation remains future work. Under this perspective, MOON-RL acts as a suggestive engine that continuously proposes directional updates to the design variables, indicating how to modify an existing design to progress towards preferred trade-offs rather than rediscovering the entire Pareto front from scratch.

Several limitations remain in the current implementation. MOON-RL assumes that a feasible initial design is available and does not include a cold-start feasibility-search or repair mechanism for cases where random sampling cannot provide one. The meta-algorithm is also deliberately simple, with fixed restart rules, archive usage, parameter schedules, design-variable scaling, and objective normalization. In addition, the present experiments focus on two- and three-objective problems, leaving the sample complexity and policy behavior in higher-dimensional or many-objective settings unexplored. The comparison of alternative samplers is also limited in scope. It covers three samplers on four problems, and identifying which sampler is best for objective-space navigation was not among the aims of this study. Finally, the results on the DAS-CMOP experiments and the CS1 global-search setting suggest that MOON-RL is better suited to focused objective-space navigation than to difficult full-front approximation or convergence-hard constrained search.

Future work should address these limitations. Restart schedules, archive usage, search directions, design-variable scaling, and objective normalization could be adjusted from feedback collected during optimization rather than fixed in advance. Larger objective spaces may require richer observation representations, deeper networks, and more structured exploration along multiple preference directions. The choice of sampler also deserves broader study: other evolution strategies, surrogate-based methods, and further reinforcement-learning algorithms, including off-policy ones, could all be evaluated within the same framework. Quantitative evaluation of within-run retargeting will require a dedicated protocol that defines how the analysis region changes with the preference direction and how comparator methods respond to a new target. Finally, integrating MOON-RL into an interactive environment would allow designers to query, update, and redirect the optimization process in real time, enabling direct study of how users interact with the navigation paradigm.

Despite these limitations, our results demonstrate that navigating the objective space with reinforcement learning can produce competitive outcomes in focused preference-guided regions of constrained engineering design problems. MOON-RL shows that even a lightweight policy and a simple meta-algorithm are sufficient to concentrate evaluations in regions of interest and to remain competitive with strong constrained MOEAs in several focused-search scenarios. Beyond showing that reinforcement learning can serve this purpose, this work specifies how to construct it: an objective-space observation, a reward combining directional progress with an orthogonal fall-off and a soft constraint penalty, a scaled continuous action on the design variables, and a meta-algorithm that resets the agent, resamples starting designs from the feasible archive, and adjusts the step-size and penalty schedules. The same construction also accommodates samplers that do not learn a policy, so the navigation formulation extends beyond the reinforcement-learning instantiation developed here. With further development of the meta-level components and integration into interactive interfaces where users can steer starting points and targets, MOON-RL has the potential to function as a practical design companion that complements classical population-based methods by concentrating computational effort on regions of the trade-off space that are most relevant to the design task.

%% file: sections/Paper_Statements.tex
\newpage

\section*{Funding}

The authors declare no relevant funding.

\section*{Author contributions:}

Andrii Artomov conceived the study, developed the methodology, implemented the software, performed the experiments, analyzed the results, and wrote the original manuscript draft. Soheyl Massoudi contributed to the literature review and assisted with manuscript editing and critical discussion. Jürg Schiffmann supervised the project and provided feedback on the methodology, results interpretation, and manuscript.

\section*{Competing interests}
The authors declare no competing interests.

\section*{Code availability statement}

The code accompanying this study is available at \url{https://github.com/artomov/MOON-RL} under the GNU General Public License v3.0. Version 1.0.0 is archived on Zenodo at \url{https://doi.org/10.5281/zenodo.24119851}.

\section*{Data availability statement}

The data generated and analysed during this study are available via Zenodo at \url{https://doi.org/10.5281/zenodo.21993233}.

%% file: sections/Appendix.tex
\appendix

\section{Formal definitions for MOO and RL}
\label{app:formalism}

\paragraph{Multi objective optimization  }
In order to provide the formalism behind the multi-objective optimization (MOO) let us consider, without loss of generality, minimization problem:

\begin{align*}
    \min_{\boldsymbol{x} \in \mathcal{X}} & \quad \boldsymbol{f}(\boldsymbol{x}) = \begin{bmatrix}
                                                        f_1(\boldsymbol{x}), f_2(\boldsymbol{x}), \dots, f_m(\boldsymbol{x})
                                                        \end{bmatrix}^T, \\
    \text{subject to } & \quad g_j(\boldsymbol{x}) \leq 0, \quad j = 1, \dots, p, \\
                       & \quad h_k(\boldsymbol{x}) = 0, \quad k = 1, \dots, q,\\
                       & \quad \boldsymbol{\ell} \le \boldsymbol{x} \le \boldsymbol{u}.
\end{align*}

Where $\boldsymbol{f} :  \mathcal{X}  \rightarrow \mathbb{R}^m$ is a vector of $m$ real-valued objective functions $f$, each mapping the design space $\mathcal{X} = \{ \boldsymbol{x} \in \mathbb{R}^n \mid \boldsymbol{\ell} \le \boldsymbol{x} \le \boldsymbol{u}\ \}$ to a real number. The image of $\boldsymbol{f}$ is called the objective space $ \mathcal{Z} = \{\boldsymbol{f}(\boldsymbol{x}) \in \mathbb{R}^m\}$. Each design $\boldsymbol{x} \in \mathbb{R}^n$ is a vector of $n$ design variables, bounded component-wise by lower bounds $\boldsymbol{\ell} = [l_1, \dots, l_n]^T$ and upper bounds $\boldsymbol{u} = [u_1, \dots, u_n]^T$. The set of all feasible solutions (i.e. designs that satisfy the constraints) is called feasible set $\mathcal{F} = \{ \boldsymbol{x} \in \mathbb{R}^n \mid \boldsymbol{\ell} \le \boldsymbol{x} \le \boldsymbol{u},\ g_j(\boldsymbol{x}) \le 0,\ h_k(\boldsymbol{x}) = 0 \} \subseteq \mathcal{X}$.

Design $\boldsymbol{x_1}$ is said to Pareto dominate design $\boldsymbol{x_2}$ (denoted as $\boldsymbol{x_1} \prec \boldsymbol{x_2}$) iff $\boldsymbol{f_i}(\boldsymbol{x_1}) \le \boldsymbol{f_i}(\boldsymbol{x_2}), \forall i \in [1,m]$, and $\exists j \in [1,m]$ for which $\boldsymbol{f_j}(\boldsymbol{x_1}) < \boldsymbol{f_j}(\boldsymbol{x_2})$.
A design $\boldsymbol{x^*} \in \mathcal{F}$ is called Pareto optimal if there is no other design that dominates it. Together, a set of non-dominated Pareto optimal designs form a  Pareto set $\mathcal{P^*} = \{\boldsymbol{x^*} \in \mathcal{F} \mid \not \exists \boldsymbol{x} \in \mathcal{F},\text{s.t. } \boldsymbol{x^*} \prec \boldsymbol{x}  \}$, which, in the objective space maps to the Pareto front $PF = \boldsymbol{f}(\mathcal{P^*}) \subseteq \mathcal{Z}$.

Naturally, the MOO aims to find a set of designs which maps to the Pareto front in the objective space

\paragraph{Reinforcement learning}

Reinforcement learning (RL) can be formalized as a Markov Decision Process (MDP) defined by the tuple $\langle \mathcal{S}, \mathcal{A}, P, R, \gamma \rangle$, where $\mathcal{S}$ is the state space, $\mathcal{A}$ is the action space, $P(s_{t+1} \mid s_t, a_t)$ is the transition probability, $R(s_t, a_t)$ is the reward function, and $\gamma \in [0,1]$ is the discount factor. At each time step $t$, an agent observes state $s_t$, executes action $a_t$, receives reward $r_t$, and transitions to the next state $s_{t+1}$. The objective of RL is to learn a policy $\pi(a \mid s)$ that maximizes the expected cumulative discounted reward, often referred to as return.

In deep reinforcement learning (DRL), the policy $\pi_\theta(a \mid s)$ is typically parameterized by a neural network with parameters $\theta$. Policy gradient methods directly optimize $\theta$ by estimating the gradient of the performance measure (e.g., the expected return) with respect to $\theta$, thereby iteratively improving the policy. Such methods allow for continuous action spaces and straightforward incorporation of function approximators like deep neural networks.

\newpage
\section{MOON-RL pseudocode}
\label{app:core_pseudo}

\begin{algorithm} [H]
\caption{MOON-RL}
\label{algo:core-algo}

\begin{algorithmic}
    \State \textbf{Input:}
    Initial starting design $\boldsymbol{x_0}^\text{init}$,
    target point $\boldsymbol{z}_{\text{targ}}$,
    upper $\boldsymbol{u}$ and lower $\boldsymbol{\ell}$ bounds for design variables,
    initial scaling factors $\boldsymbol{\kappa}^{\text{init}}$,
    initial constraints penalty  $\boldsymbol{\lambda}^{\text{init}}$,
    episode length $T$,
    max number of episodes $N^{\text{max}}$,
    policy uptdate interval $N^{\text{train}}$,
    PPO hyperparameters $\boldsymbol{\Theta}_{\textbf{PPO}}$,
    guiding meta-algorithm $MetaAlgorithm$, meta-algorithm parameters $\boldsymbol{\Theta_{\text{meta}}}$

    \State \textbf{Output:} List of non-dominated designs $\mathcal{P}$
\end{algorithmic}

\begin{algorithmic}[1]  %
    \State Initialize PPO agent $\pi_\theta$ with $\boldsymbol{\Theta}_{\textbf{PPO}}$
    \State Initialize $MetaAlgorithm$ with $\boldsymbol{\kappa}^{\text{init}}$, ${\lambda}^{\text{initial}}$ and $\Theta_{\text{meta}}$
    \Comment{Meta-algorithm is discussed in the section \ref{subsec:meta-algo}}
    \State Initialize non-dominated archive: $\mathcal{P} \gets \emptyset$
    \State Initialize starting design: $\boldsymbol{x_0} \gets \boldsymbol{x_0}^\text{init}$

    \For{$episode$ = 1 to $N^{\text{max}}$} %
        \State$(\boldsymbol{\kappa}, {\lambda}, resetFlag) \gets $ Get updated runtime parameters for $episode$ according to $MetaAlgorithm$
        \If{$resetFlag$ }
            \State Reset agent $\pi_\theta$
            \State Chose new starting point $\boldsymbol{x_0}$ from $\mathcal{P}$ according to $MetaAlgorithm$
        \EndIf

        \State$(\boldsymbol{f} (\boldsymbol{x_0}), \boldsymbol{CV}(\boldsymbol{x_0})) \gets Evaluate(\boldsymbol{x_0})$ \Comment{Evaluate starting design}
        \State $\boldsymbol{v^\text{ref}} \gets \boldsymbol{z}_{\text{targ}} - \boldsymbol{f}(\boldsymbol{x_0})$
        \State $\left(\left\langle
        \frac{\boldsymbol{v^\text{ref}}}{\lVert \boldsymbol{v^\text{ref}} \rVert_2},
        \boldsymbol{v_0}
        \right\rangle,\epsilon_0\right) \gets$ Calculate objective space metrics for a starting design

        \For{step $t$ = 0 to $T-1$} %
            \State $s_t \gets$ Construct the current observation vector
            \State Sample action from an agent: $\boldsymbol{a_t} \sim \pi_\theta$
            \State $\boldsymbol{m_t} \gets \boldsymbol{a_t} \ \boldsymbol{\kappa}$
            \State Create new design: $\boldsymbol{x_{t+1}} \gets \boldsymbol{x_{t}} + \boldsymbol{m_t}$
            \State Make sure the new design is within bounds: $\boldsymbol{x_{t+1}} \gets clip(\boldsymbol{x_{t+1}}, \boldsymbol{u}, \boldsymbol{\ell})$
            \State $(\boldsymbol{f} (\boldsymbol{x_{t+1}}), \boldsymbol{CV}(\boldsymbol{x_{t+1}})) \gets Evaluate(\boldsymbol{x_{t+1}})$ \Comment{Evaluate new design}
            \State $\left(\left\langle
            \frac{\boldsymbol{v^\text{ref}}}{\lVert \boldsymbol{v^\text{ref}} \rVert_2},
            \boldsymbol{v_{t+1}}
            \right\rangle,\epsilon_{t+1}\right) \gets$ Calculate objective space metrics for the new design
            \State $r_t \gets$ Calculate reward according to equation in section \ref{subsec:observation_and_reward}
            \State Store relevant transition metrics in PPO buffer \Comment{Standard for PPO algorithm}
            \State Update non-dominated archive $\mathcal{P}$
        \EndFor
        \If{$episode$ is a multiple of $N^\text{train}$}
            \State Train PPO agent $\pi_\theta$ on a collected buffer \Comment{See \cite{schulman2017proximal} for more details}
            \State Clear PPO buffer
        \EndIf
    \EndFor
    \State \Return{$\mathcal{P}$}
\end{algorithmic}
\end{algorithm}

\textbf{Equality constraints.} All benchmark problems considered in this study contain inequality constraints only. The present implementation does not handle equality constraints directly. An equality constraint $h_j(\boldsymbol{x})=0$ would first need to be converted by the problem interface into a tolerance-based nonnegative violation, for example $CV_j(\boldsymbol{x})=\max!\left(\lvert h_j(\boldsymbol{x})\rvert-\delta_j,0\right)$, where $\delta_j>0$ is a problem-specific feasibility tolerance. This conversion was not implemented or evaluated in the present study.

\newpage
\section{Meta-algorithm functions}
\label{app:meta_pseudo}

In our implementation, we reset the agent every fixed amount of episodes, meaning that $resetFlag$ in algorithm \ref{algo:core-algo} is set to $true$ after fixed periods. New starting point is sampled from the actual list of non-dominated solutions.  The sampling is done according softmax with temperature, where the scores are assigned according to proximity of designs in the non-dominated archive to the target point. This way designs that are closer to the target point have more probability to be chosen. The algorithm \ref{algo:softmax-sampler} shows the sampling approach.

\begin{algorithm}[H]
\caption{Meta-algorithm: starting-point sampling function}
\label{algo:softmax-sampler}

\begin{algorithmic}
    \State \textbf{Input:} Most–recent Pareto archive $\mathcal P=\{\boldsymbol{x}^{(i)}\}$ and corresponding objective values
        $\mathcal Z=\{\boldsymbol{f}(\boldsymbol{x}^{(i)}),\,\boldsymbol{x}^{(i)}\!\in\!\mathcal P\}$,
        temperature parameter schedule $\tau$, target point $\boldsymbol{z}_{\text{targ}}$
    \State \textbf{Output:} New starting design $\boldsymbol{x}_0$
\end{algorithmic}

\begin{algorithmic}[1]
    \Function{StartingDesignSampler}{$\mathcal P,\mathcal Z,\tau,\boldsymbol{z}_{\text{targ}}$}
        \ForAll{$(\boldsymbol{x}^{(i)},\boldsymbol{f}(\boldsymbol{x}^{(i)})) \in (\mathcal P,\mathcal Z)$}
            \State $d^{(i)} \gets \bigl\lVert \,\boldsymbol{f}(\boldsymbol{x}^{(i)})-\boldsymbol{z}_{\text{targ}} \bigr\rVert_2$
                  \Comment{$d^{(i)}$ is the Euclidean distance from $\boldsymbol{x}^{(i)}$ to the target}
        \EndFor
        \State $\text{Compute ranks } r^{(i)} \text{ of } d^{(i)}\quad$
               \Comment{Closer to the target $\Rightarrow$ higher rank}
        \State $s^{(i)} \gets -\,r^{(i)}$
               \Comment{Negate so that higher rank $\Rightarrow$ larger $s^{(i)}$}
        \State $\displaystyle
               p^{(i)} \gets
               \frac{\exp\!\bigl(s^{(i)}/\tau\bigr)}
                    {\sum\nolimits_{j}\exp\!\bigl(s^{(j)}/\tau\bigr)}$
        \State $\text{sample index } i_0 \sim \text{Categorical}\bigl(p^{(1)},p^{(2)},\dots\bigr)$
        \State $\boldsymbol{x}_0 \gets \boldsymbol{x}^{(i_0)}$ \Comment{Map index to the actual design}
        \State \Return $\boldsymbol{x}_0$
    \EndFunction
\end{algorithmic}
\end{algorithm}

In our implementation, when the optimization run starts we keep the scaling factors $\boldsymbol{\kappa}$ relatively large so that the action distributions cover relatively big portion of the design variable ranges which promotes larger steps in the design space and consequently helps to better explore the objective space. As the algorithm progresses, we gradually decrese the scaling factors to narrow down the search to the local neighborhoods of the discovered solutions.

Similarly, we keep the constraint penalty parameter $\lambda$ relatively small at the beginning, to better investigate the objective space even if it means finding unfeasible solutions and we increase it as the algorithm approaches later stages of the optimization run to favor feasible solutions.

We use simple linear schedules with fixed rate to decrease the values of $\boldsymbol{\kappa} $ and increase the value of $\lambda$ while keeping some limit values so that the schedulers stop when these values are reached. The functions used for scheduling $\boldsymbol{\kappa}$ and $\lambda$ are summarized in the algorithm \ref{algo:scheduler}.

\begin{algorithm} [H]
\caption{Meta-algorithm: step-wise scheduling functions for $\boldsymbol{\kappa}$ and $\lambda$}
\label{algo:scheduler}

\begin{algorithmic}
    \State \textbf{Input:} Episode index $episode$, current values of $\boldsymbol{\kappa}$ and $\lambda$, limit values $\boldsymbol{\kappa}^\text{limit}$ and $\lambda^\text{limit}$, scheduler rates $\Delta_\kappa$ and $\Delta_\lambda$, scheduler interval $S_\kappa$ and $S_\lambda$
    \State \textbf{Outputs:} Updated values of  $\boldsymbol{\kappa}$ and $\lambda$

\end{algorithmic}

\begin{algorithmic}[1]  %
    \Function{ScalingFactorsScheduler}{$episode$, $\boldsymbol{\kappa}$, $\boldsymbol{\kappa}^\text{limit}$, $S_\kappa$, $\Delta_\kappa$}
        \If{$episode$ is multiple of $S_\kappa$}
            \State $\boldsymbol{\kappa} \gets max(\boldsymbol{\kappa}^\text{limit},\boldsymbol{\kappa} - \Delta_\kappa)$
        \EndIf
        \State \Return{$\boldsymbol{\kappa}$}
    \EndFunction
        \Function{ConstraintsPenaltyScheduler}{$episode$, $\lambda$, $\lambda^\text{limit}$, $S_\lambda$, $\Delta_\lambda$}
        \If{$episode$ is multiple of $S_\lambda$}
            \State $\lambda \gets min(\lambda^\text{limit}, \lambda + \Delta_\lambda)$
        \EndIf
        \State \Return{$\lambda$}
    \EndFunction
\end{algorithmic}
\end{algorithm}

\newpage
\section{Random seeds}
\label{app:seeds}

Table~\ref{tab:seeds} summarizes the random seeds used for all of our experiments.

\begin{table}[h]
\centering
\caption{Random seeds used in all experiments}

\begin{tabular}{rrrrrr}
\toprule
0    & 9    & 55   & 159  & 248  & 921 \\
1118 & 217  & 2674 & 2957 & 2968 & 3008 \\
3376 & 3750 & 4352 & 4779 & 4844 & 5319 \\
5520 & 5806 & 5859 & 6215 & 7106 & 8523 \\
8642 & 8694 & 9213 & 9256 & 9348 & 9766 \\
\bottomrule
\end{tabular}
\label{tab:seeds}
\end{table}

These seeds were randomly generated to avoid potential correlations that may arise from sequential values.

\section{Additive $\epsilon$-indicator}
\label{app:additive-epsilon-indicator}

To visually compare the Pareto front $\mathcal{Q}$ obtained by MOON-RL, to a baseline front $\mathcal{B}$, we use the additive $\epsilon$-indicator $I_{\epsilon^+}$~\cite{zitzler2003performance} evaluated for each point  $q \in \mathcal{Q} \subset \mathbb{R}^m$. Assuming a maximization problem, for a point $q$ the indicator is computed as

$$ I_{\epsilon^+}(q \mid \mathcal{B}) = \min_{b \in \mathcal{B}} \; \max_{k} \bigl( b_k - q_k \bigr). $$

For each baseline design $b$, the inner maximum measures the worst-case objective difference between $q$ and $b$, and the outer minimum selects the baseline design that is closest to $q$ in this additive sense.

For visualization we plot the negated quantity $-I_{\epsilon^+}(q \mid \mathcal{B})$, so that, in this sign convention $  -I_{\varepsilon^+}(q \mid \mathcal{B}) > 0 $ indicates that $q$ dominates at least one design in $\mathcal{B}$ (our algorithm locally outperforms the baseline), values near zero correspond to approximate parity, and negative values indicate that $q$ is dominated by its closest baseline design.

\section{Axes scalings}
\label{app:axes_scaling}

To prevent any single objective from dominating due to its numerical scale, we apply fixed per-objective scaling factors for both MOON-RL and the hypervolume computation. The values used for each problem are listed in Table \ref{tab:axes-scaling}. They were chosen by inspecting the ranges of the objectives on a small set of feasible designs and selecting simple coefficients that bring the objectives to comparable magnitudes.

\begin{table}[ht]
    \centering
    \caption{Objective scaling factors used for MOON-RL and for hypervolume computation. A dash indicates that no third objective is defined for the problem.}
    \label{tab:axes-scaling}
    
\begin{tabular}{lccc}
        \toprule
        Problem      & Objective 1 & Objective 2 & Objective 3 \\
        \midrule
        CS1          & 0.1         & 20          & --          \\
        CT1          & 0.1         & 0.2         & --          \\
        CTS1         & 0.1         & 0.2         & 20          \\
        DAS-CMOP3-11 & 0.1         & 0.1         & --          \\
        \bottomrule
    \end{tabular}

\end{table}

\section{Agent hyperparameters}
\label{app:agent_hyperparameters}

Hyperparameters of the PPO agent presented in Table~\ref{tab:ppo_hyperparams} were empirically chosen on a different problem set. They showed good generalization capability across different optimization problems and were hence retained.

\begin{table}[ht]
\centering
\caption{PPO agent hyperparameters used in all experiments.}

\begin{tabular}{ll}
\toprule
\textbf{Parameter} & \textbf{Value / Description} \\
\midrule
Policy architecture & Separate networks for actor and critic \\
Actor architecture & Separate networks for means and standard deviation \\
Action distribution & Independent Gaussian per action dimension \\
Hidden layers & $2$ per network (all networks)\\
Hidden layer sizes & $64$ units \\
Hidden activation & ReLU \\
Mean output activation & $\tanh$ \\
Standard-deviation output & $\operatorname{softplus}(z+1)+10^{-2}$ \\
Optimizer & Adam, PyTorch defaults ($\beta_1=0.9$, $\beta_2=0.999$, $\epsilon=10^{-8}$) \\
Actor learning rate & $3 \times 10^{-4}$ \\
Critic learning rate & $3 \times 10^{-4}$ \\
Batch size & $32$ \\
Epochs per update & $30$ \\
Policy clip parameter & $0.2$ \\
Discount factor $\gamma$ & $0.99$ \\
GAE $\lambda$ & $0.9$ \\
Advantage normalization & Enabled \\
Probability-ratio calculation & Computed and clipped per action dimension \\
Critic loss & Mean-squared error to discounted returns \\
Entropy bonus & Not used \\
Value-function clipping & Not used \\
Gradient-norm clipping & Not used \\
Observation normalization & Disabled \\
Reward normalization & Not used \\
Weight initialization & PyTorch defaults; actor output-layer weights divided by $100$ \\
\bottomrule
\end{tabular}
\label{tab:ppo_hyperparams}
\end{table}

Table~\ref{tab:ppo_rollout_settings} summarizes the rollout structure and PPO update frequency. Here, an episode denotes one batch of parallel trajectories.

\begin{table}[ht]
\centering

\caption{PPO rollout and update settings used in the experiments.}

\begin{tabular}{ll}
\toprule
\textbf{Parameter} & \textbf{Value / Description} \\
\midrule
Trajectory length $T$ & $8$ steps \\
Parallel trajectories per episode $N$ & $10$ \\
Candidate evaluations per episode & $N \times T = 80$ \\
Policy update interval & After every episode \\
Number of episodes & $3750$ \\
Candidate evaluations from rollouts & $300{,}000$ \\
\bottomrule
\end{tabular}
\label{tab:ppo_rollout_settings}
\end{table}

\section{Meta-algorithm hyperparameters}
\label{app:meta-algo_hyperparameters}

Table \ref{tab:scheduler_params} summarizes the hyperparameters of the meta-algorithm used in our experiment. They were empirically chosen.

\begin{table}[ht]
\centering
\caption{Runtime scheduler parameters used to adjust scaling factors $\kappa$ and constraint penalties $\lambda$.}

\begin{tabular}{lll}
\toprule
\textbf{Parameter} & \textbf{Schedule Type} & \textbf{Value} \\
\midrule
Initial scaling factor $\kappa_0$ & Linear decay & $7\%$ of design-variable ranges \\
Target scaling factor $\kappa_{\text{min}}$ &  & $1\%$ \\
Decay interval $S_\kappa$ &  & Every $100$ episodes \\
Decay rate $\delta_\kappa$ &  & $1\%$ of design-variable ranges (subtracted per interval) \\
Minimum categorical-variable scaling & Fixed floor & $0.6$ for MODAct variables \\
\midrule
Initial constraint penalty $\lambda_0$ & Linear increase & $2$ \\
Target penalty $\lambda_{\text{max}}$ &  & $6$ \\
Growth interval $S_\lambda$ &  & Every $200$ episodes \\
Growth rate $\delta_\lambda$ &  & $2$ (added per interval) \\
\midrule
Reset interval $N_{\text{reset}}$ & Fixed & Every $50$ episodes \\
\midrule
Softmax temperature $\tau$ & Fixed & $2$  \\
\bottomrule
\end{tabular}
\label{tab:scheduler_params}
\end{table}

\newpage
\section{Benchmark algorithms parameters}
\label{app:GA-config}

Table~\ref{tab:ga_parameters} summarizes the key parameters explicitly specified in our experiments. Unless listed otherwise, the Pymoo \cite{pymoo} and PlatEMO \cite{PlatEMO} implementations were run with their respective default parameter values. For example, the Pymoo algorithms used Simulated Binary Crossover (SBX) with $\eta = 15$ and $p = 0.9$.

\begin{table}[htbp]
\centering
\caption{Benchmark Algorithm Parameters Used in Experiments}
\label{tab:ga_parameters}
\small

\begin{tabular}{@{}ll@{}}
\toprule
\textbf{Algorithm} & \textbf{Parameters} \\
\midrule
\textbf{NSGA-II} &

\begin{tabular}[t]{@{}l@{}}
Population size: 200 or 100\\
\end{tabular}
 \\
\addlinespace
\textbf{R-NSGA-II} &

\begin{tabular}[t]{@{}l@{}}
Population size: 200 \\
Reference point(s): Same as for MOON-RL, depending global or local search\\
$\varepsilon$: 0.01 or 0.001 \\
\end{tabular}
 \\
\addlinespace
\textbf{MOEA/D} &

\begin{tabular}[t]{@{}l@{}}
Reference directions: ``Das-Dennis'', 2 objectives, 199 partitions \\
Number of neighbors: 200 \\
Probability of neighbor mating: 0.7 \\
\end{tabular}
 \\
\addlinespace
\textbf{NSGA-III} &

\begin{tabular}[t]{@{}l@{}}
Population size: 200 \\
Reference directions: ``Das-Dennis'', 2 objectives, 14  or 199 partitions \\
\end{tabular}
 \\
\addlinespace
\textbf{R-NSGA-III} &

\begin{tabular}[t]{@{}l@{}}
Population size: 200 \\
Reference point(s): Same as for MOON-RL, depending global or local search\\
$\mu$: 0.1 or 0.01 \\
\end{tabular}
 \\
\addlinespace
\textbf{C-TAEA} &

\begin{tabular}[t]{@{}l@{}}
Reference directions: ``Das-Dennis'', 2 objectives, 14  or 199 partitions \\
\end{tabular}
 \\
\addlinespace
\textbf{APSEA} &

\begin{tabular}[t]{@{}l@{}}
Population size: 200 \\
$\alpha$: 0.05; $\beta$: 0.05; $c_p$: 5 \\
\end{tabular}
 \\
\addlinespace
\textbf{LCMEA} &

\begin{tabular}[t]{@{}l@{}}
Population size: 200 \\
\end{tabular}
 \\
\addlinespace
\textbf{CMODRL} &

\begin{tabular}[t]{@{}l@{}}
Population size: 100 \\
Reward step: 5 \\
\end{tabular}
 \\
\bottomrule
\end{tabular}

\end{table}

\newpage

\clearpage
\newpage
\section{CS1 main results full table}

Table~\ref{tab:results-cs1-full} reports the full results for the CS1 focused-search setting.

\label{app:cs1_full_table}

\begin{sidewaystable}

    \centering

    \caption{CS1 results, reported as median [$5^{th}$ percentile, $95^{th}$ percentile].}

    \label{tab:results-cs1-full}

\begin{tabular}{lcccc}

        \toprule

        Algorithm & HV $\uparrow$ & GD $\downarrow$ & \shortstack{Eval @ 50\% HV $\downarrow$} & \shortstack{Eval @ 75\% HV $\downarrow$} \\

        \midrule

        MOON-RL & $ 52.03 [\boldsymbol{49.39}, 53.88] $ & $ 0.11 [0.04, \boldsymbol{0.24}] $ & $ 4880 [\boldsymbol{1316}, 18964] (\boldsymbol{30}/30) $ & $ 16440 [7068, 36676] (\boldsymbol{30}/30) $ \\

        APSEA & $ 40.21 [36.42, 51.53] $ & $ 1.14 [0.25, 1.50] $ & $ 123194 [105610.8, 136288.4] (\boldsymbol{30}/30) $ & $ 135346 [118730, 269997.6] (12/30) $ \\

        LCMEA & $ \boldsymbol{54.57} [41.00, 55.68] $ & $ \boldsymbol{0.05} [\boldsymbol{0.03}, 1.08] $ & $ 7800 [2080, 17760] (29/30) $ & $ 30800 [10380, 213920] (27/30) $ \\

        CMODRL & $ 52.50 [42.25, \boldsymbol{55.74}] $ & $ 0.16 [\boldsymbol{0.03}, 0.96] $ & $ 158325 [152355, 197430] (29/30) $ & $ 183600 [160905, 233745] (29/30) $ \\

        NSGA-II (pop. 200) & $ 40.96 [36.81, 52.70] $ & $ 1.10 [0.19, 1.46] $ & $ 7600 [3690, 10620] (\boldsymbol{30}/30) $ & $ 159800 [10640, 274160] (13/30) $ \\

        NSGA-II (pop. 100) & $ 39.03 [23.15, 52.59] $ & $ 1.27 [0.18, 2.86] $ & $ \boldsymbol{4050} [1880, \boldsymbol{8365}] (28/30) $ & $ 153300 [\boldsymbol{5400}, 288760] (9/30) $ \\

        R-NSGA-II ($\varepsilon = 0.01$) & $ 37.86 [34.13, 50.16] $ & $ 1.29 [0.29, 1.58] $ & $ 6300 [3270, 9460] (\boldsymbol{30}/30) $ & $ 14400 [13300, 51880] (7/30) $ \\

        R-NSGA-II ($\varepsilon = 0.001$) & $ 35.22 [31.79, 47.92] $ & $ 1.46 [0.35, 1.78] $ & $ 6100 [3270, 10620] (\boldsymbol{30}/30) $ & $ \boldsymbol{10000} [8900, 18650] (6/30) $ \\

        NSGA-III (15 ref. dir.) & $ 39.06 [33.40, 48.46] $ & $ 1.27 [0.50, 1.78] $ & $ 8500 [4320, 13410] (\boldsymbol{30}/30) $ & $ 83000 [11280, 283080] (7/30) $ \\

        NSGA-III (200 ref. dir.) & $ 39.06 [33.40, 48.46] $ & $ 1.27 [0.50, 1.78] $ & $ 8500 [4320, 13410] (\boldsymbol{30}/30) $ & $ 83000 [11280, 283080] (7/30) $ \\

        R-NSGA-III ($\mu = 0.1$) & $ 38.67 [34.92, 50.57] $ & $ 1.29 [0.21, 1.64] $ & $ 7526 [3911, 11819] (\boldsymbol{30}/30) $ & $ 13250 [10472, 26012] (8/30) $ \\

        R-NSGA-III ($\mu = 0.01$) & $ 38.41 [16.31, 51.11] $ & $ 1.27 [0.25, 3.84] $ & $ 7420 [4240, 12476] (28/30) $ & $ 14840 [9412, \boldsymbol{16557}] (7/30) $ \\

        MOEA/D (200 ref. dir.) & $ 5.94 [0.16, 27.02] $ & $ 4.58 [2.29, 5.81] $ & $ 96000 [10680, 181320] (2/30) $ & $ - [-, -] (0/30) $ \\

        C-TAEA (15 ref. dir.) & $ 0.00 [0.00, 7.47] $ & $ 5.60 [3.06, 7.45] $ & $ 278415 [278415, 278415] (1/30) $ & $ - [-, -] (0/30) $ \\

        C-TAEA (200 ref. dir.) & $ 3.19 [0.00, 15.68] $ & $ 4.98 [2.70, 6.73] $ & $ - [-, -] (0/30) $ & $ - [-, -] (0/30) $ \\

        \bottomrule

    \end{tabular}

\end{sidewaystable}

\clearpage

\clearpage
\newpage
\section{CT1 main results full table}
\label{app:ct1_full_table}

Table~\ref{tab:results-ct1-full} reports the full results for the CT1 focused-search setting.

\begin{sidewaystable}

    \centering

    \caption{CT1 results, reported as median [$5^{th}$ percentile, $95^{th}$ percentile].}
    \label{tab:results-ct1-full}
    
\begin{tabular}{lcccc}
        \toprule
        Algorithm & HV $\uparrow$ & GD $\downarrow$ & \shortstack{Eval @ 50\% HV $\downarrow$} & \shortstack{Eval @ 75\% HV $\downarrow$} \\
        \midrule
        MOON-RL & $ 0.59 [0.57, 0.60] $ & $ 0.12 [0.08, 0.16] $ & $ \boldsymbol{1240.00} [\boldsymbol{436.00}, 4740.00] (\boldsymbol{30}/30) $ & $ 5120.00 [2836.00, 12340.00] (\boldsymbol{30}/30) $ \\
        APSEA & $ \boldsymbol{0.61} [\boldsymbol{0.61}, \boldsymbol{0.62}] $ & $ \boldsymbol{0.07} [\boldsymbol{0.03}, \boldsymbol{0.10}] $ & $ 11169.00 [7481.70, 20456.10] (\boldsymbol{30}/30) $ & $ 13693.50 [9095.85, 21595.95] (\boldsymbol{30}/30) $ \\
        LCMEA & $ 0.59 [0.59, 0.61] $ & $ 0.17 [0.09, 0.22] $ & $ 1400.00 [800.00, 2800.00] (\boldsymbol{30}/30) $ & $ 5500.00 [2690.00, 17300.00] (\boldsymbol{30}/30) $ \\
        CMODRL & $ 0.60 [0.59, 0.61] $ & $ 0.10 [0.06, 0.15] $ & $ 148537.50 [141945.00, 172181.25] (\boldsymbol{30}/30) $ & $ 149100.00 [142762.50, 173006.25] (\boldsymbol{30}/30) $ \\
        NSGA-II (pop. 200) & $ \boldsymbol{0.61} [0.60, \boldsymbol{0.62}] $ & $ \boldsymbol{0.07} [\boldsymbol{0.03}, \boldsymbol{0.10}] $ & $ 2200.00 [1600.00, 3220.00] (\boldsymbol{30}/30) $ & $ 5200.00 [3800.00, 7730.00] (\boldsymbol{30}/30) $ \\
        NSGA-II (pop. 100) & $ \boldsymbol{0.61} [0.60, \boldsymbol{0.62}] $ & $ 0.08 [\boldsymbol{0.03}, 0.12] $ & $ 1450.00 [700.00, \boldsymbol{2155.00}] (\boldsymbol{30}/30) $ & $ \boldsymbol{3500.00} [\boldsymbol{2200.00}, 7745.00] (\boldsymbol{30}/30) $ \\
        R-NSGA-II ($\varepsilon = 0.01$) & $ \boldsymbol{0.61} [0.60, 0.61] $ & $ \boldsymbol{0.07} [0.05, 0.11] $ & $ 2100.00 [1600.00, 2600.00] (\boldsymbol{30}/30) $ & $ 4000.00 [2890.00, \boldsymbol{6550.00}] (\boldsymbol{30}/30) $ \\
        R-NSGA-II ($\varepsilon = 0.001$) & $ 0.60 [0.60, 0.61] $ & $ 0.08 [0.06, 0.11] $ & $ 2000.00 [1600.00, 2600.00] (\boldsymbol{30}/30) $ & $ 4100.00 [3290.00, 7630.00] (\boldsymbol{30}/30) $ \\
        NSGA-III (15 ref. dir.) & $ \boldsymbol{0.61} [0.60, 0.61] $ & $ 0.08 [0.04, 0.11] $ & $ 2200.00 [1490.00, 3220.00] (\boldsymbol{30}/30) $ & $ 4800.00 [4090.00, 8150.00] (\boldsymbol{30}/30) $ \\
        NSGA-III (200 ref. dir.) & $ \boldsymbol{0.61} [0.60, 0.61] $ & $ 0.08 [0.04, 0.12] $ & $ 2200.00 [1490.00, 3220.00] (\boldsymbol{30}/30) $ & $ 4800.00 [4090.00, 8150.00] (\boldsymbol{30}/30) $ \\
        R-NSGA-III ($\mu = 0.1$) & $ 0.60 [0.19, 0.61] $ & $ 0.08 [0.05, 3.00] $ & $ 2544.00 [1547.60, 3180.00] (27/30) $ & $ 5300.00 [4430.80, 7144.40] (27/30) $ \\
        R-NSGA-III ($\mu = 0.01$) & $ 0.60 [0.09, 0.61] $ & $ 0.12 [0.06, 3.85] $ & $ 2756.00 [1515.80, 3572.20] (24/30) $ & $ 5724.00 [4303.60, 8204.40] (24/30) $ \\
        MOEA/D (200 ref. dir.) & $ 0.60 [0.51, 0.61] $ & $ 0.15 [0.08, 0.87] $ & $ 1600.00 [1000.00, 11030.00] (\boldsymbol{30}/30) $ & $ 15600.00 [2750.00, 152240.00] (28/30) $ \\
        C-TAEA (15 ref. dir.) & $ 0.12 [0.00, 0.33] $ & $ 3.60 [1.43, 5.33] $ & $ 241182.00 [132755.40, 295792.20] (3/30) $ & $ - [-, -] (0/30) $ \\
        C-TAEA (200 ref. dir.) & $ 0.31 [0.23, 0.38] $ & $ 2.02 [0.86, 2.71] $ & $ 164182.00 [115318.80, 268830.00] (13/30) $ & $ - [-, -] (0/30) $ \\
        \bottomrule
    \end{tabular}

\end{sidewaystable}

\clearpage

\clearpage
\newpage
\section{CTS1 main results full table}
\label{app:cts1_full_table}

Table~\ref{tab:results-cts1-full} reports the full results for the CTS1 focused-search setting.

\begin{sidewaystable}

    \centering

    \caption{CTS1 results, reported as median [$5^{th}$ percentile, $95^{th}$ percentile].}
    \label{tab:results-cts1-full}
    
\begin{tabular}{lcccc}
        \toprule
        Algorithm & HV $\uparrow$ & GD $\downarrow$ & \shortstack{Eval @ 50\% HV $\downarrow$} & \shortstack{Eval @ 75\% HV $\downarrow$} \\
        \midrule
        MOON-RL & $ 130.71 [118.96, 136.19] $ & $ 0.32 [0.22, 0.56] $ & $ 4240.00 [\boldsymbol{676.00}, 6600.00] (\boldsymbol{30}/30) $ & $ 8400.00 [5112.00, 12612.00] (\boldsymbol{30}/30) $ \\
        APSEA & $ 132.76 [131.02, 134.77] $ & $ \boldsymbol{0.15} [\boldsymbol{0.10}, \boldsymbol{0.22}] $ & $ 130540.0 [118425.4, 142160.5] (\boldsymbol{30}/30) $ & $ 132309.0 [121615.7, 144710.3] (\boldsymbol{30}/30) $ \\
        LCMEA & $ 130.76 [127.31, 133.18] $ & $ 0.18 [0.11, 0.26] $ & $ 4600.00 [1690.00, 8650.00] (\boldsymbol{30}/30) $ & $ 12800.00 [5890.00, 29280.00] (\boldsymbol{30}/30) $ \\
        CMODRL & $ 131.42 [127.98, 133.9] $ & $ \boldsymbol{0.15} [\boldsymbol{0.10}, \boldsymbol{0.22}] $ & $ 134062.5 [118211.25, 148702.5] (\boldsymbol{30}/30) $ & $ 138975.0 [122557.5, 154005.0] (\boldsymbol{30}/30) $ \\
        NSGA-II (pop. 200) & $ 135.09 [131.26, 136.30] $ & $ 0.18 [0.14, 0.27] $ & $ 3200.00 [2490.00, 4220.00] (\boldsymbol{30}/30) $ & $ 7400.00 [5780.00, 9200.00] (\boldsymbol{30}/30) $ \\
        NSGA-II (pop. 100) & $ 127.71 [123.11, 131.14] $ & $ 0.23 [0.16, 0.33] $ & $ \boldsymbol{1700.00} [1245.00, \boldsymbol{2210.00}] (\boldsymbol{30}/30) $ & $ \boldsymbol{4500.00} [\boldsymbol{3100.00}, 23665.00] (\boldsymbol{30}/30) $ \\
        R-NSGA-II ($\varepsilon = 0.01$) & $ 135.62 [123.52, 136.60] $ & $ 0.32 [0.26, 0.58] $ & $ 2500.00 [2000.00, 2910.00] (\boldsymbol{30}/30) $ & $ 4600.00 [3400.00, 5820.00] (\boldsymbol{30}/30) $ \\
        R-NSGA-II ($\varepsilon = 0.001$) & $ 130.36 [123.68, 135.44] $ & $ 0.38 [0.30, 0.58] $ & $ 2600.00 [2000.00, 3000.00] (\boldsymbol{30}/30) $ & $ 4600.00 [3600.00, \boldsymbol{5600.00}] (\boldsymbol{30}/30) $ \\
        NSGA-III (15 ref. dir.) & $ 129.93 [125.72, 132.66] $ & $ 0.16 [0.11, 0.25] $ & $ 3200.00 [2400.00, 3910.00] (\boldsymbol{30}/30) $ & $ 6200.00 [4870.00, 9240.00] (\boldsymbol{30}/30) $ \\
        NSGA-III (200 ref. dir.) & $ 82.80 [66.51, 94.16] $ & $ 0.43 [0.29, 0.74] $ & $ 4000.00 [3090.00, 8060.00] (\boldsymbol{30}/30) $ & $ 6400.00 [6400.00, 6400.00] (1/30) $ \\
        R-NSGA-III ($\mu = 0.1$) & $ \boldsymbol{136.77} [\boldsymbol{134.38}, \boldsymbol{137.52}] $ & $ 0.27 [0.23, 0.34] $ & $ 3195.00 [2651.85, 3621.00] (\boldsymbol{30}/30) $ & $ 6496.50 [4973.55, 8445.45] (\boldsymbol{30}/30) $ \\
        R-NSGA-III ($\mu = 0.01$) & $ 134.40 [131.24, 136.17] $ & $ 0.32 [0.24, 0.44] $ & $ 3408.00 [2651.85, 4164.15] (\boldsymbol{30}/30) $ & $ 6496.50 [5207.85, 9372.00] (\boldsymbol{30}/30) $ \\
        MOEA/D (200 ref. dir.) & $ 8.55 [0.00, 41.54] $ & $ 3.26 [1.56, 3.91] $ & $ 600.00 [600.00, 600.00] (1/30) $ & $ - [-, -] (0/30) $ \\
        C-TAEA (15 ref. dir.) & $ 126.84 [114.62, 131.30] $ & $ 0.21 [0.14, 0.52] $ & $ 3600.00 [1950.00, 5490.00] (\boldsymbol{30}/30) $ & $ 9240.00 [6228.00, 64026.00] (\boldsymbol{30}/30) $ \\
        C-TAEA (200 ref. dir.) & $ 47.63 [34.67, 59.37] $ & $ 2.31 [2.11, 2.56] $ & $ 300000.00 [300000.00, 300000.00] (1/30) $ & $ - [-, -] (0/30) $ \\
        \bottomrule
    \end{tabular}

\end{sidewaystable}

\clearpage

\clearpage
\newpage
\section{DAS-CMOP3-11 main results full table}
\label{app:dascmop3_11_full_table}

Table~\ref{tab:results-dascmop3-11-full} reports the full results for the DAS-CMOP3-11 focused-search setting.

\begin{sidewaystable}

    \centering

    \caption{DASCMOP3-11 results, reported as median [$5^{th}$ percentile, $95^{th}$ percentile].}

    \label{tab:results-dascmop3-11-full}

\begin{tabular}{lcccc}
        \toprule
        Algorithm & HV $\uparrow$ & GD $\downarrow$ & \shortstack{Eval @ 50\% HV $\downarrow$} & \shortstack{Eval @ 75\% HV $\downarrow$} \\
        \midrule
        MOON-RL & $ 0.38 [0.00, 0.38] $ & $ 0.05 [0.04, 5.11] $ & $ 15360.00 [10116.00, 31036.00] (24/30) $ & $ 28360.00 [15560.00, 38568.00] (24/30) $ \\
        APSEA & $ 0.39 [0.39, 0.39] $ & $ \boldsymbol{0.02} [\boldsymbol{0.02}, 0.03] $ & $ 8272.00 [6768.00, 11486.80] (\boldsymbol{30}/30) $ & $ 10528.00 [8732.60, 16290.20] (\boldsymbol{30}/30) $ \\
        LCMEA & $ 0.39 [0.39, 0.39] $ & $ \boldsymbol{0.02} [\boldsymbol{0.02}, 0.03] $ & $ 5600.00 [3090.00, 11330.00] (\boldsymbol{30}/30) $ & $ 8500.00 [3670.00, 18110.00] (\boldsymbol{30}/30) $ \\
        CMODRL & $ 0.39 [0.18, 0.39] $ & $ \boldsymbol{0.02} [\boldsymbol{0.02}, 0.03] $ & $ 4275.00 [3258.75, 6247.50] (28/30) $ & $ 5550.00 [4278.75, 8358.75] (28/30) $ \\
        NSGA-II (pop. 200) & $ \boldsymbol{0.39} [0.00, 0.39] $ & $ \boldsymbol{0.02} [\boldsymbol{0.02}, 0.03] $ & $ 19400.00 [13040.00, 54600.00] (27/30) $ & $ 22400.00 [15920.00, 57540.00] (27/30) $ \\
        NSGA-II (pop. 100) & $ \boldsymbol{0.40} [0.38, 0.41] $ & $ 0.03 [\boldsymbol{0.02}, 0.03] $ & $ 21500.00 [8780.00, 184800.00] (29/30) $ & $ 22600.00 [10280.00, 185700.00] (29/30) $ \\
        R-NSGA-II ($\varepsilon = 0.01$) & $ 0.01 [0.01, 0.38] $ & $ 6.83 [0.09, 6.88] $ & $ 48000.00 [20880.00, 177560.00] (9/30) $ & $ 50600.00 [23080.00, 180480.00] (9/30) $ \\
        R-NSGA-II ($\varepsilon = 0.001$) & $ 0.01 [0.01, 0.38] $ & $ 6.83 [0.13, 6.90] $ & $ 13400.00 [11670.00, 40850.00] (8/30) $ & $ 16100.00 [13670.00, 42880.00] (8/30) $ \\
        NSGA-III (15 ref. dir.) & $ \boldsymbol{0.40} [\boldsymbol{0.40}, 0.40] $ & $ 0.03 [\boldsymbol{0.02}, 0.03] $ & $ 26200.00 [19680.00, 86000.00] (29/30) $ & $ 29200.00 [21480.00, 88680.00] (29/30) $ \\
        NSGA-III (200 ref. dir.) & $ \boldsymbol{0.39} [0.39, 0.39] $ & $ 0.03 [\boldsymbol{0.02}, 0.03] $ & $ 26500.00 [15070.00, 92810.00] (\boldsymbol{30}/30) $ & $ 31800.00 [16800.00, 100050.00] (\boldsymbol{30}/30) $ \\
        R-NSGA-III ($\mu = 0.1$) & $ \boldsymbol{0.39} [0.00, 0.40] $ & $ 0.06 [0.03, 1.03] $ & $ 26462.00 [19462.70, 76618.60] (20/30) $ & $ 28785.00 [21775.60, 79466.80] (20/30) $ \\
        R-NSGA-III ($\mu = 0.01$) & $ 0.00 [0.00, 0.00] $ & $ 24.63 [16.89, 28.09] $ & $ 58580.00 [58580.00, 58580.00] (1/30) $ & $ 61206.00 [61206.00, 61206.00] (1/30) $ \\
        C-TAEA (15 ref. dir.) & $ \boldsymbol{0.39} [0.36, 0.40] $ & $ \boldsymbol{0.02} [\boldsymbol{0.02}, \boldsymbol{0.02}] $ & $ \boldsymbol{2332.50} [\boldsymbol{1518.75}, \boldsymbol{4148.25}] (\boldsymbol{30}/30) $ & $ \boldsymbol{2865.00} [\boldsymbol{1788.75}, \boldsymbol{4854.00}] (\boldsymbol{30}/30) $ \\
        C-TAEA (200 ref. dir.) & $ \boldsymbol{0.39} [0.39, \boldsymbol{0.47}] $ & $ \boldsymbol{0.02} [\boldsymbol{0.02}, 0.03] $ & $ 10600.00 [8690.00, 17600.00] (\boldsymbol{30}/30) $ & $ 14800.00 [11690.00, 22190.00] (\boldsymbol{30}/30) $ \\
        \bottomrule
    \end{tabular}

\end{sidewaystable}

\clearpage

\clearpage
\newpage
\section{CS1 (global search) main results full table}
\label{app:cs1_global_full_table}

Table~\ref{tab:results-cs1-global-full} reports the full results for the CS1 global-search setting.

\begin{sidewaystable}
    \centering
    \caption{CS1 global-search results, reported as median [$5^{th}$ percentile, $95^{th}$ percentile].}
    \label{tab:results-cs1-global-full}
    
\begin{tabular}{lcccc}
        \toprule
        Algorithm & HV $\uparrow$ & GD $\downarrow$ & \shortstack{Eval @ 50\% HV $\downarrow$} & \shortstack{Eval @ 75\% HV $\downarrow$} \\
        \midrule
        MOON-RL & $ 78.03 [\boldsymbol{72.64}, 82.05] $ & $ 0.54 [0.37, \boldsymbol{0.79}] $ & $ 24800.00 [8360.00, 48680.00] (\boldsymbol{30}/30) $ & $ 90200.00 [43460.00, 206740.00] (\boldsymbol{30}/30) $ \\
        APSEA & $ 69.32 [64.10, 84.65] $ & $ 1.08 [0.25, 1.31] $ & $ 123690.00 [107098.80, 139345.00] (\boldsymbol{30}/30) $ & $ 174716.00 [123107.20, 273444.80] (17/30) $ \\
        LCMEA & $ \boldsymbol{88.31} [68.96, \boldsymbol{90.71}] $ & $ \boldsymbol{0.25} [\boldsymbol{0.06}, 1.29] $ & $ 9600.00 [\boldsymbol{3160.00}, 24960.00] (29/30) $ & $ 43700.00 [13680.00, \boldsymbol{181730.00}] (28/30) $ \\
        CMODRL & $ 78.70 [69.49, 88.84] $ & $ 0.54 [0.07, 0.97] $ & $ 160500.00 [154290.00, 199710.00] (29/30) $ & $ 202575.00 [160815.00, 244920.00] (29/30) $ \\
        NSGA-II (pop. 200) & $ 70.53 [64.40, 87.34] $ & $ 1.02 [0.19, 1.35] $ & $ 8600.00 [4690.00, 12110.00] (\boldsymbol{30}/30) $ & $ 138100.00 [15730.00, 251640.00] (20/30) $ \\
        NSGA-II (pop. 100) & $ 68.54 [41.39, 86.19] $ & $ 1.10 [0.19, 2.22] $ & $ \boldsymbol{5300.00} [3235.00, \boldsymbol{9700.00}] (28/30) $ & $ 106100.00 [\boldsymbol{10580.00}, 192740.00] (12/30) $ \\
        R-NSGA-II ($\varepsilon = 0.01$) & $ 68.91 [59.47, 84.45] $ & $ 1.09 [0.28, 1.58] $ & $ 8700.00 [4090.00, 12000.00] (\boldsymbol{30}/30) $ & $ \boldsymbol{16800.00} [11830.00, 270450.00] (14/30) $ \\
        R-NSGA-II ($\varepsilon = 0.001$) & $ 67.09 [60.65, 85.74] $ & $ 1.18 [0.27, 1.48] $ & $ 8300.00 [4560.00, 12440.00] (\boldsymbol{30}/30) $ & $ 19900.00 [14490.00, 221300.00] (10/30) $ \\
        NSGA-III (15 ref. dir.) & $ 67.70 [59.40, 80.53] $ & $ 1.20 [0.48, 1.64] $ & $ 10400.00 [6470.00, 16820.00] (\boldsymbol{30}/30) $ & $ 133700.00 [24760.00, 259220.00] (10/30) $ \\
        NSGA-III (200 ref. dir.) & $ 67.70 [59.40, 80.53] $ & $ 1.20 [0.48, 1.64] $ & $ 10400.00 [6470.00, 16820.00] (\boldsymbol{30}/30) $ & $ 133700.00 [24760.00, 259220.00] (10/30) $ \\
        R-NSGA-III ($\mu = 0.1$) & $ 77.01 [2.29, 86.87] $ & $ 0.66 [0.16, 5.38] $ & $ 36470.00 [27092.00, 42617.80] (23/30) $ & $ 80234.00 [65958.60, 270294.80] (19/30) $ \\
        R-NSGA-III ($\mu = 0.01$) & $ 83.29 [1.79, 86.49] $ & $ 0.36 [0.19, 5.61] $ & $ 33344.00 [24070.20, 44806.00] (23/30) $ & $ 79192.00 [60331.80, 247475.00] (19/30) $ \\
        MOEA/D (200 ref. dir.) & $ 8.51 [2.99, 51.65] $ & $ 3.58 [0.45, 4.64] $ & $ 73000.00 [7950.00, 223900.00] (6/30) $ & $ - [-, -] (0/30) $ \\
        C-TAEA (15 ref. dir.) & $ 0.00 [0.00, 7.93] $ & $ 6.06 [3.68, 7.39] $ & $ 277680.00 [277680.00, 277680.00] (1/30) $ & $ - [-, -] (0/30) $ \\
        C-TAEA (200 ref. dir.) & $ 7.02 [0.10, 29.11] $ & $ 4.62 [2.95, 5.86] $ & $ - [-, -] (0/30) $ & $ - [-, -] (0/30) $ \\
        \bottomrule
    \end{tabular}

\end{sidewaystable}

\clearpage

\clearpage
\newpage
\section{MODact ``CS1'' additional results}
\label{app:cs1-additional-results}

\begin{figure}[H]
    \centering
    \begin{subfigure}[b]{0.49\textwidth}
        \includegraphics[width=\textwidth]{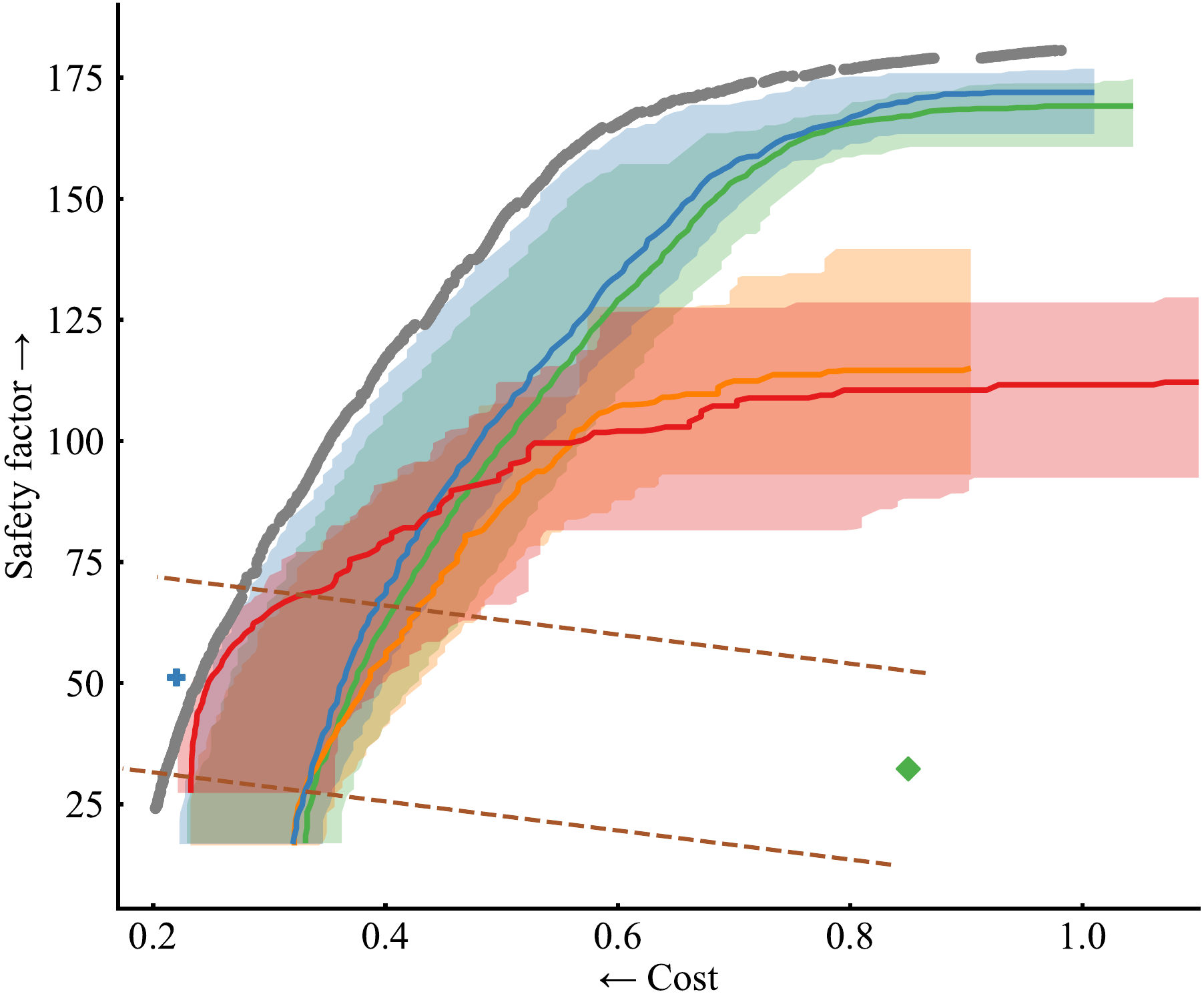}
        \begin{picture}(0,0)
            \put(0,195){a.} %
        \end{picture}

    \end{subfigure}
    \hfill
    \begin{subfigure}[b]{0.49\textwidth}
        \includegraphics[width=\textwidth]{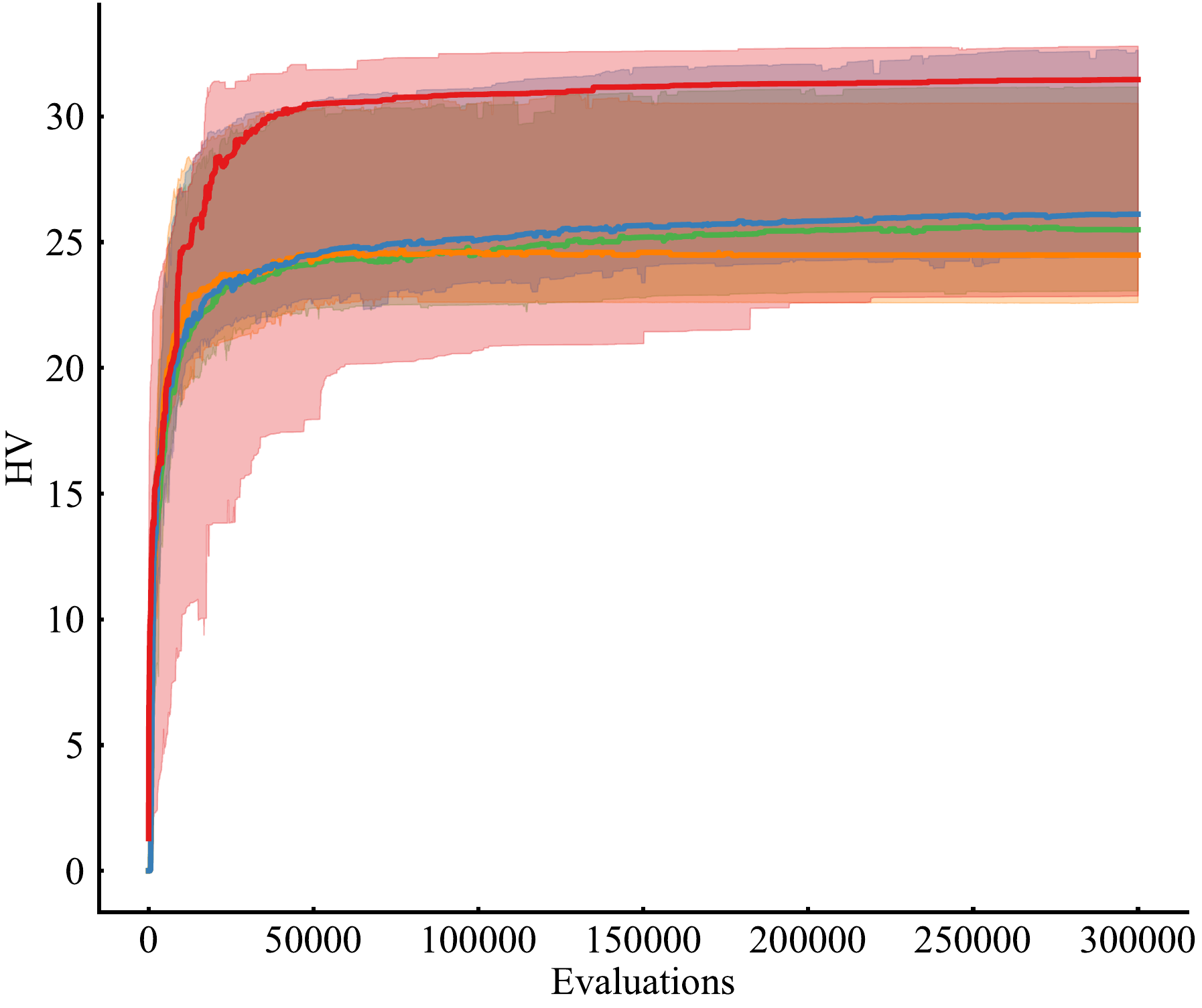}
        \begin{picture}(0,0)
            \put(0,195){b.} %
        \end{picture}
    \end{subfigure}

    \centering
    \vspace*{-1em} %
    {
    \begin{subfigure}[b]{0.6\textwidth}
        \includegraphics[width=\textwidth]{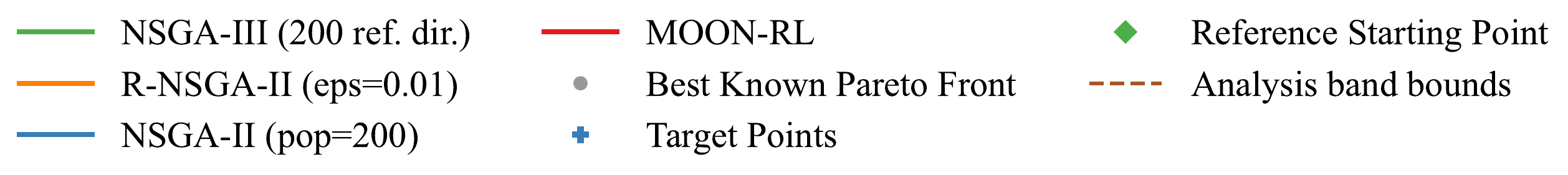}
    \end{subfigure}
    }

    \caption{ EAF (a) and HV progression (b) of different algorithms, including $5^{th}$ and $95^{th}$ percentiles. Target point: $(0.22, 51.2)$.}
    \label{fig:cs1_best_p2}
\end{figure}

\begin{figure}[H]
    \centering
    \begin{subfigure}[b]{0.49\textwidth}
        \includegraphics[width=\textwidth]{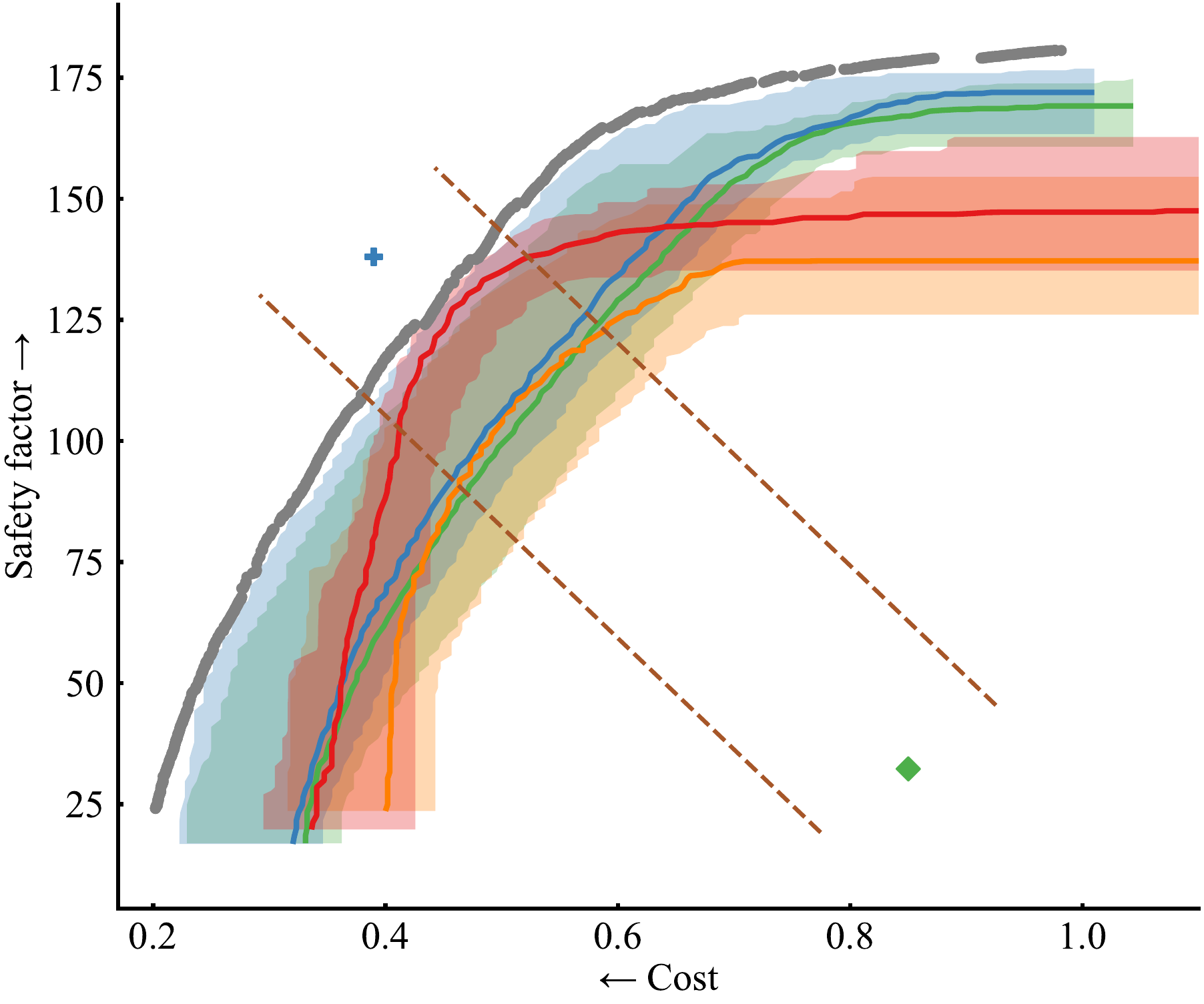}
        \begin{picture}(0,0)
            \put(0,195){a.} %
        \end{picture}

    \end{subfigure}
    \hfill
    \begin{subfigure}[b]{0.49\textwidth}
        \includegraphics[width=\textwidth]{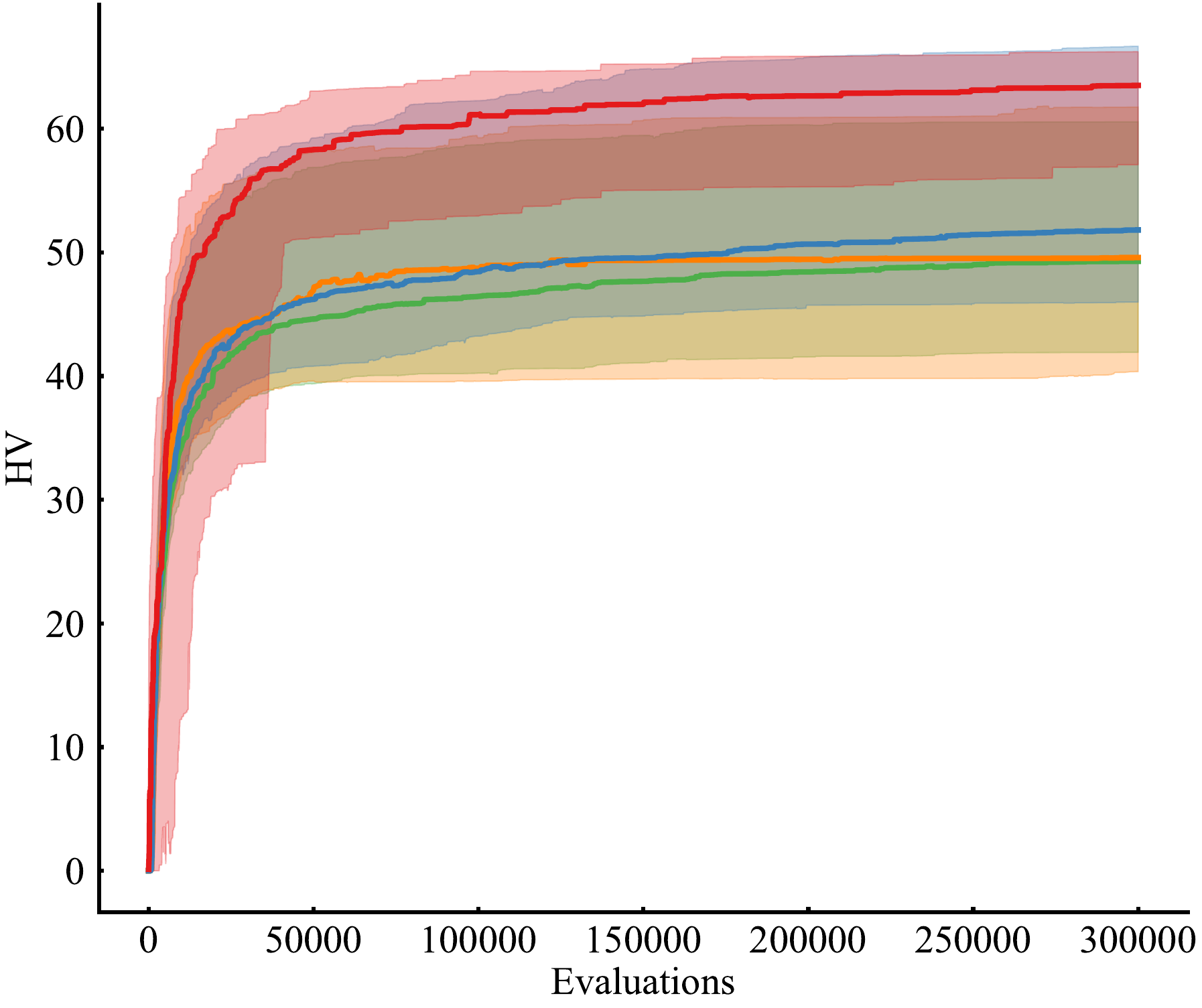}
        \begin{picture}(0,0)
            \put(0,195){b.} %
        \end{picture}
    \end{subfigure}

    \centering
    \vspace*{-1em} %
    {
    \begin{subfigure}[b]{0.6\textwidth}
        \includegraphics[width=\textwidth]{figures/appendix/alternative_targets/cs1/local_best_legend_only.pdf}
    \end{subfigure}
    }

    \caption{ EAF (a) and HV progression (b) of different algorithms, including $5^{th}$ and $95^{th}$ percentiles. Target point: $(0.39, 138.)$.}
    \label{fig:cs1_best_p3}
\end{figure}

\begin{figure}[H]
    \centering
    \begin{subfigure}[b]{0.49\textwidth}
        \includegraphics[width=\textwidth]{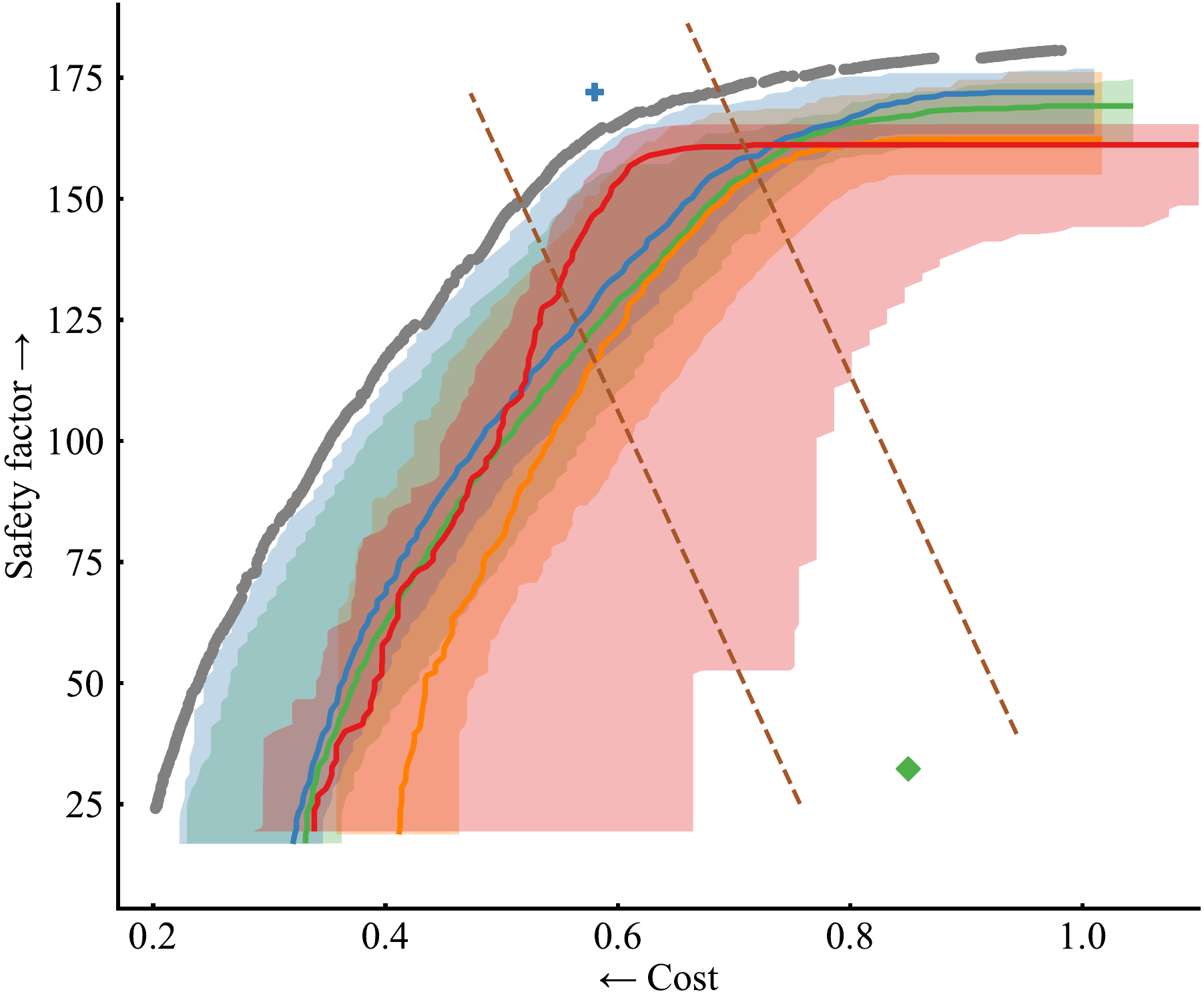}
        \begin{picture}(0,0)
            \put(0,195){a.} %
        \end{picture}

    \end{subfigure}
    \hfill
    \begin{subfigure}[b]{0.49\textwidth}
        \includegraphics[width=\textwidth]{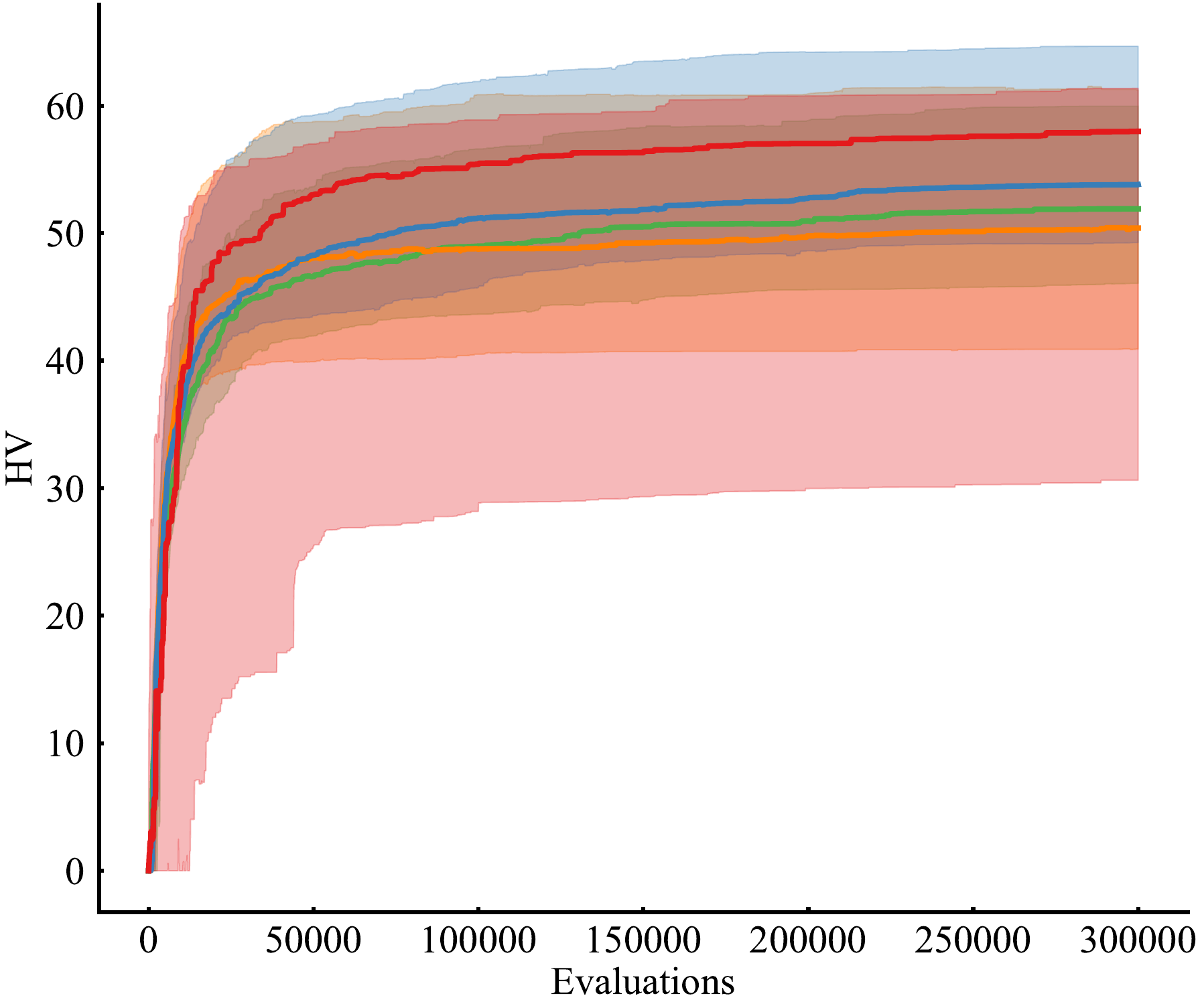}
        \begin{picture}(0,0)
            \put(0,195){b.} %
        \end{picture}
    \end{subfigure}

    \centering
    \vspace*{-1em} %
    {
    \begin{subfigure}[b]{0.6\textwidth}
        \includegraphics[width=\textwidth]{figures/appendix/alternative_targets/cs1/local_best_legend_only.pdf}
    \end{subfigure}
    }

    \caption{ EAF (a) and HV progression (b) of different algorithms, including $5^{th}$ and $95^{th}$ percentiles. Target point: $(0.58, 172.)$.}
    \label{fig:cs1_best_p4}
\end{figure}

\begin{figure}[H]
    \centering
    \begin{subfigure}[b]{0.49\textwidth}
        \includegraphics[width=\textwidth]{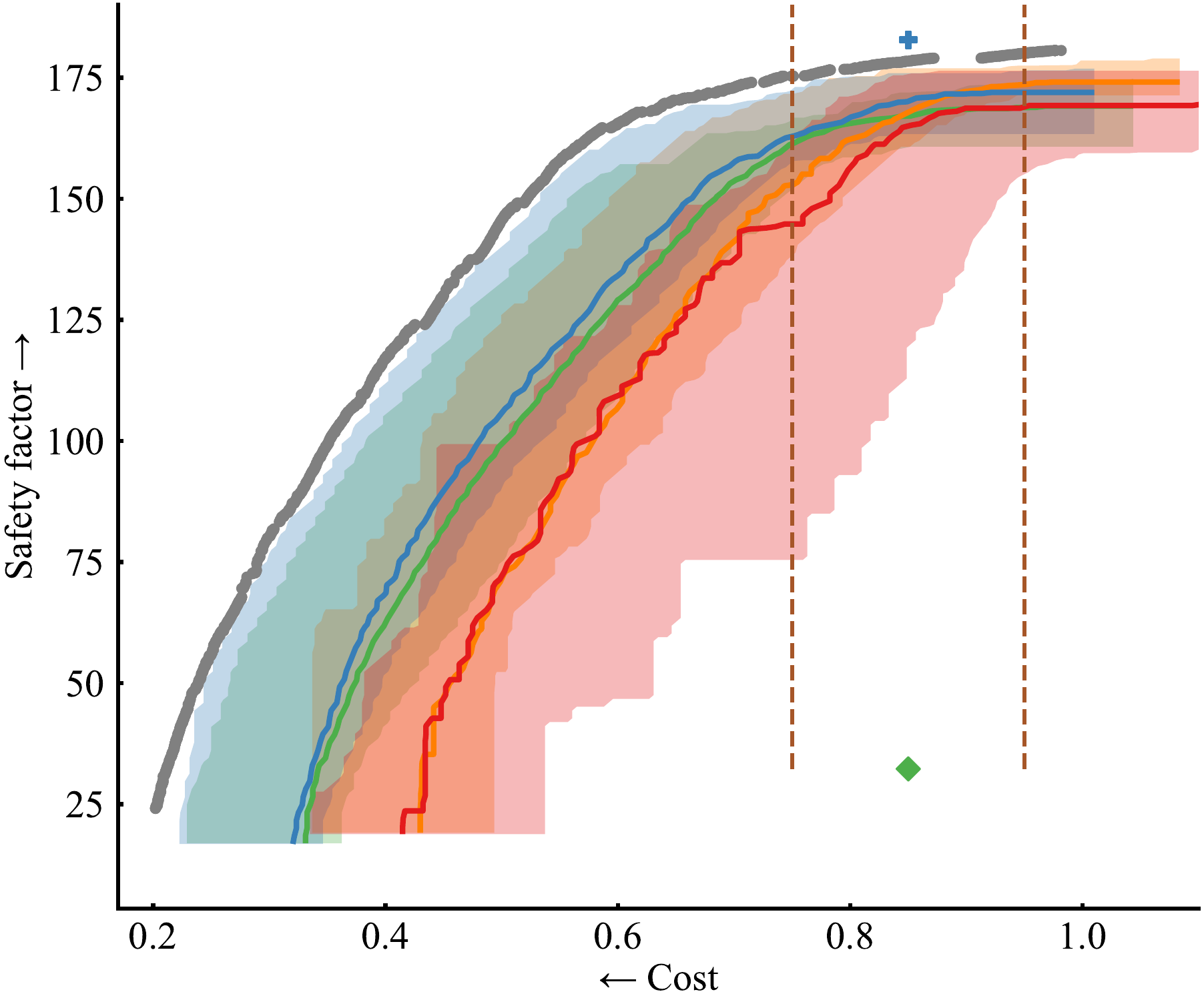}
        \begin{picture}(0,0)
            \put(0,195){a.} %
        \end{picture}

    \end{subfigure}
    \hfill
    \begin{subfigure}[b]{0.49\textwidth}
        \includegraphics[width=\textwidth]{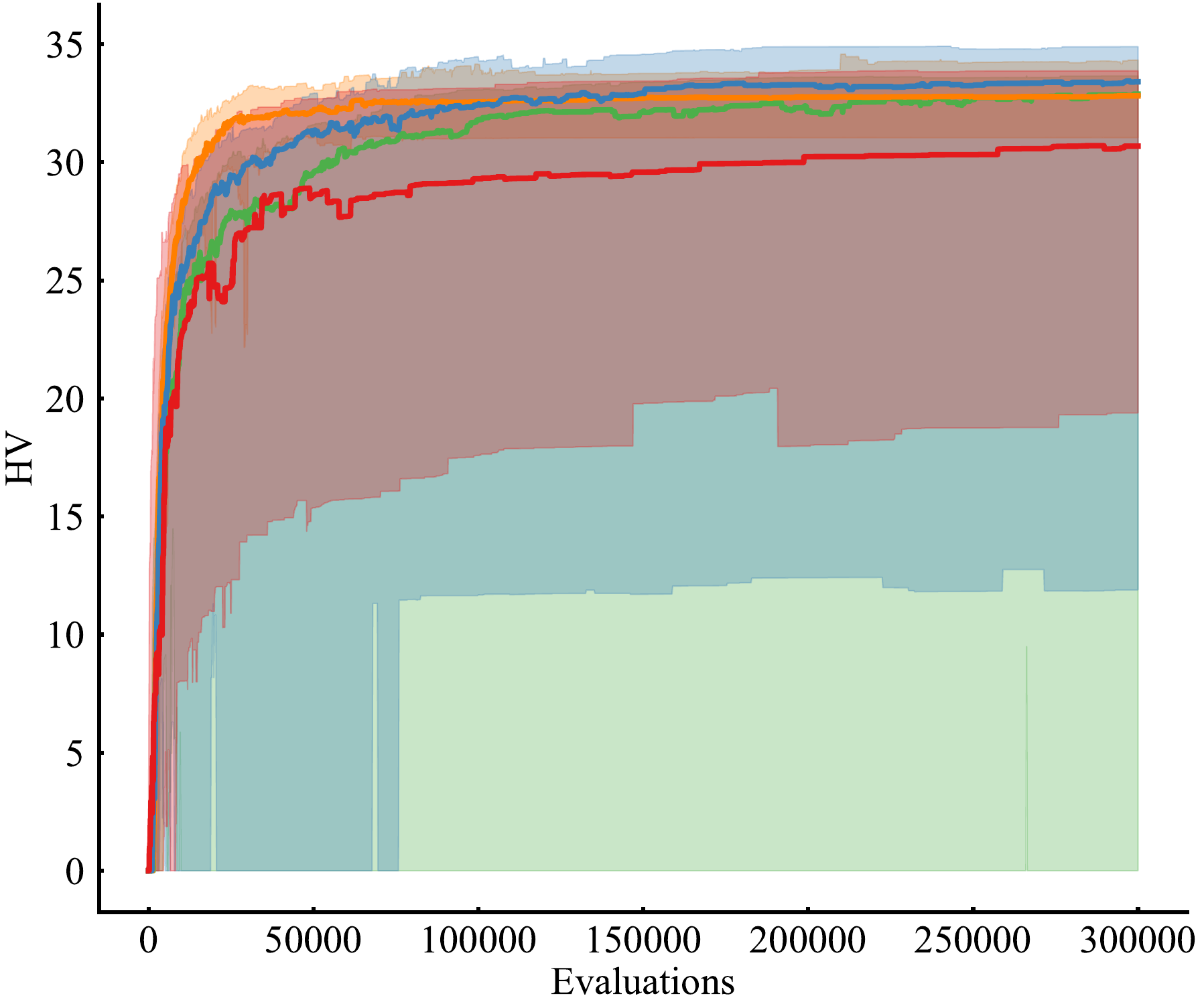}
        \begin{picture}(0,0)
            \put(0,195){b.} %
        \end{picture}
    \end{subfigure}

    \centering
    \vspace*{-1em} %
    {
    \begin{subfigure}[b]{0.6\textwidth}
        \includegraphics[width=\textwidth]{figures/appendix/alternative_targets/cs1/local_best_legend_only.pdf}
    \end{subfigure}
    }

    \caption{ EAF (a) and HV progression (b) of different algorithms, including $5^{th}$ and $95^{th}$ percentiles. Target point: $(0.85, 182.8)$.}
    \label{fig:cs1_best_p5}
\end{figure}

Figures \ref{fig:cs1_best_p2}, \ref{fig:cs1_best_p3}, \ref{fig:cs1_best_p4} and \ref{fig:cs1_best_p5} show the EAF and hypervolume progressions for experiments focused towards alternative target points, for ``CS1'' problem. The target points are: $(0.22, 51.2) $, $(0.39, 138.)$, $(0.58, 172.)$ and $(0.85, 182.8)$ respectively.

\clearpage

\clearpage
\newpage
\section{MODact ``CT1'' additional results}
\label{app:ct1-additional-results}

\begin{figure}[H]
    \centering
    \begin{subfigure}[b]{0.49\textwidth}
        \includegraphics[width=\textwidth]{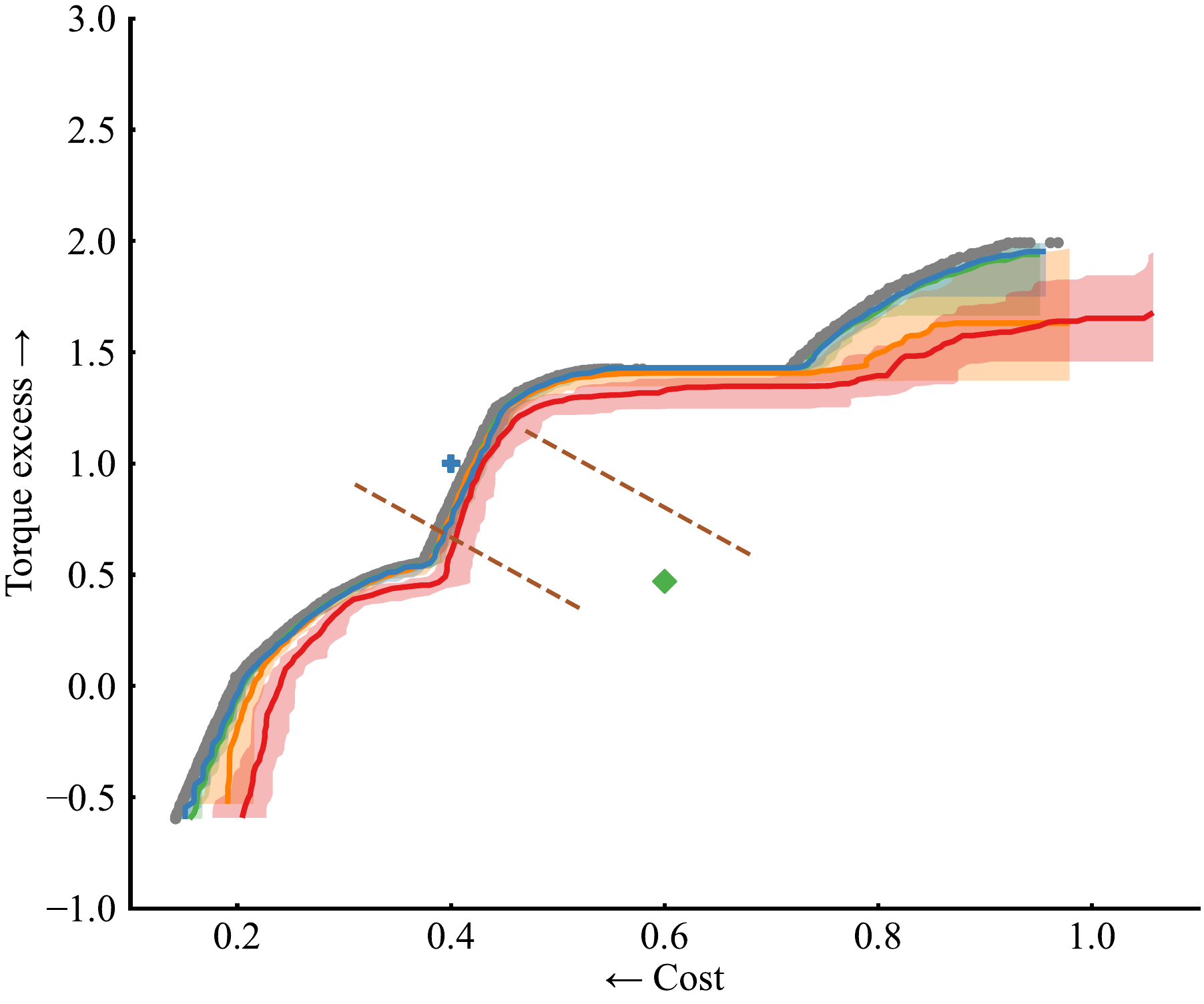}
        \begin{picture}(0,0)
            \put(0,195){a.} %
        \end{picture}

    \end{subfigure}
    \hfill
    \begin{subfigure}[b]{0.49\textwidth}
        \includegraphics[width=\textwidth]{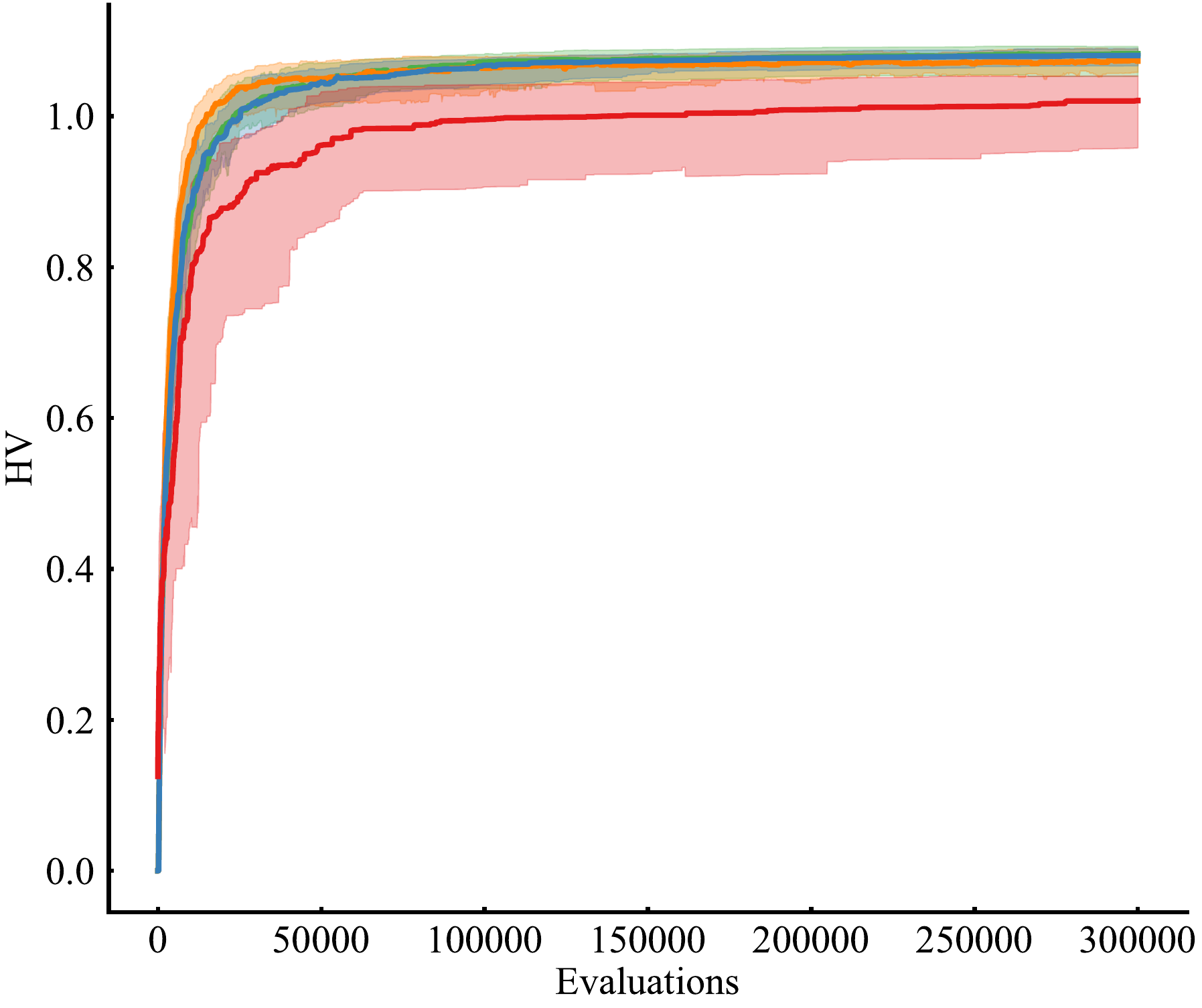}
        \begin{picture}(0,0)
            \put(0,195){b.} %
        \end{picture}
    \end{subfigure}

    \centering
    \vspace*{-1em} %
    {
    \begin{subfigure}[b]{0.6\textwidth}
        \includegraphics[width=\textwidth]{figures/appendix/alternative_targets/cs1/local_best_legend_only.pdf}
    \end{subfigure}
    }

    \caption{ EAF (a) and HV progression (b) of different algorithms, including $5^{th}$ and $95^{th}$ percentiles. Target point: $(0.4, 1)$}
    \label{fig:ct1_best_p1}
\end{figure}

\begin{figure}[H]
    \centering
    \begin{subfigure}[b]{0.49\textwidth}
        \includegraphics[width=\textwidth]{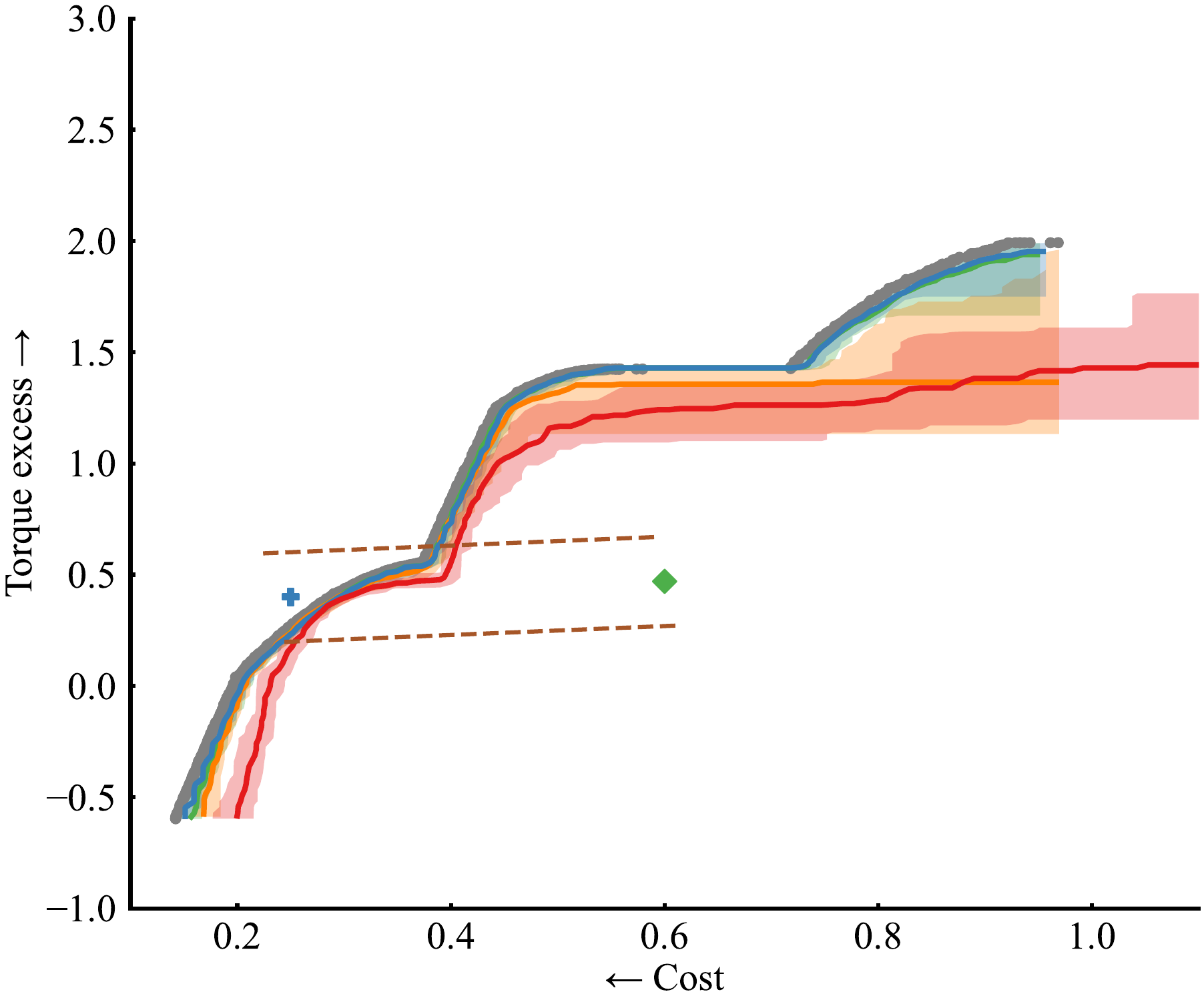}
        \begin{picture}(0,0)
            \put(0,195){a.} %
        \end{picture}

    \end{subfigure}
    \hfill
    \begin{subfigure}[b]{0.49\textwidth}
        \includegraphics[width=\textwidth]{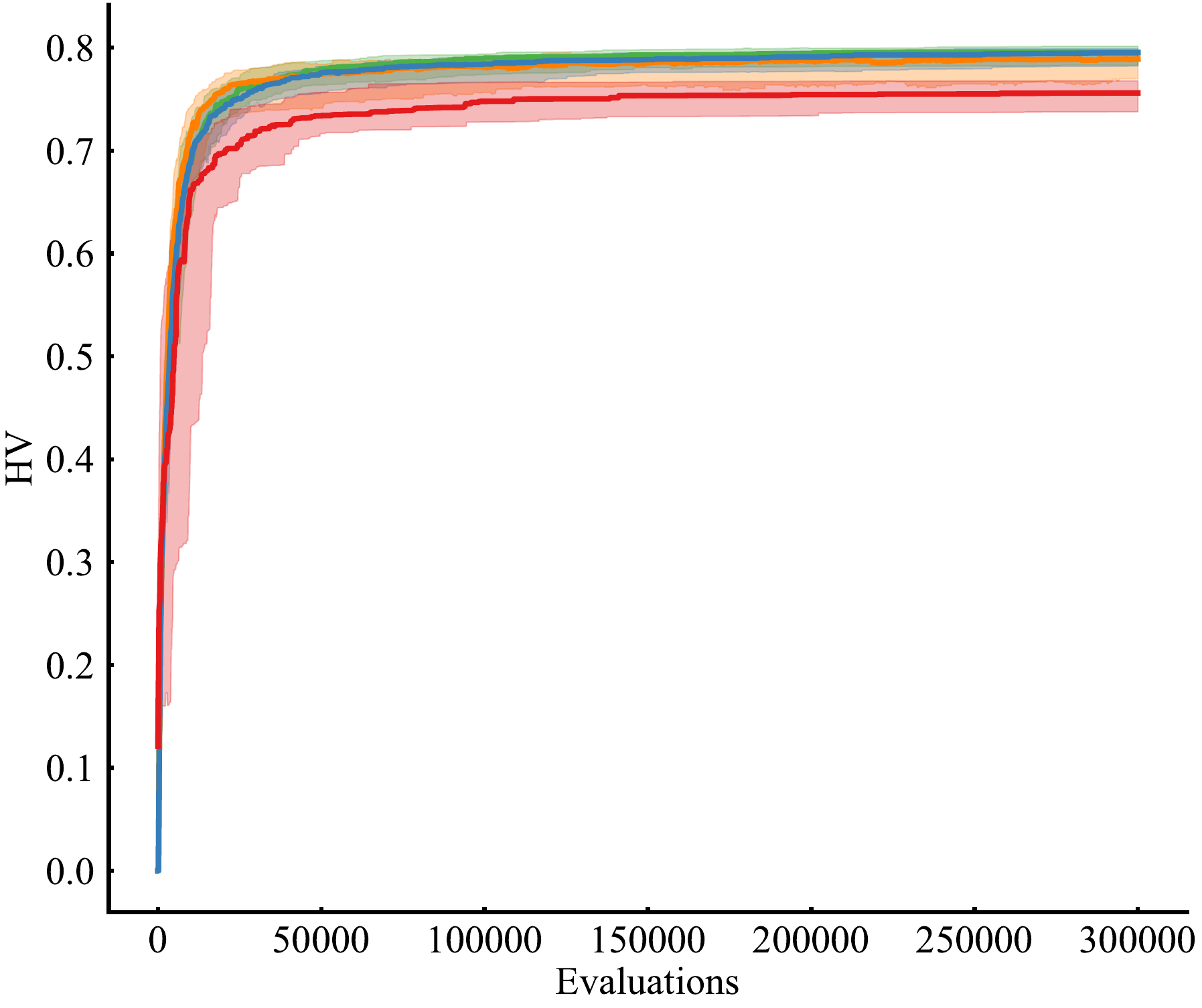}
        \begin{picture}(0,0)
            \put(0,195){b.} %
        \end{picture}
    \end{subfigure}

    \centering
    \vspace*{-1em} %
    {
    \begin{subfigure}[b]{0.6\textwidth}
        \includegraphics[width=\textwidth]{figures/appendix/alternative_targets/cs1/local_best_legend_only.pdf}
    \end{subfigure}
    }

    \caption{ EAF (a) and HV progression (b) of different algorithms, including $5^{th}$ and $95^{th}$ percentiles. Target point: $(0.25, 0.4)$.}
    \label{fig:ct1_best_p3}
\end{figure}

\begin{figure}[H]
    \centering
    \begin{subfigure}[b]{0.49\textwidth}
        \includegraphics[width=\textwidth]{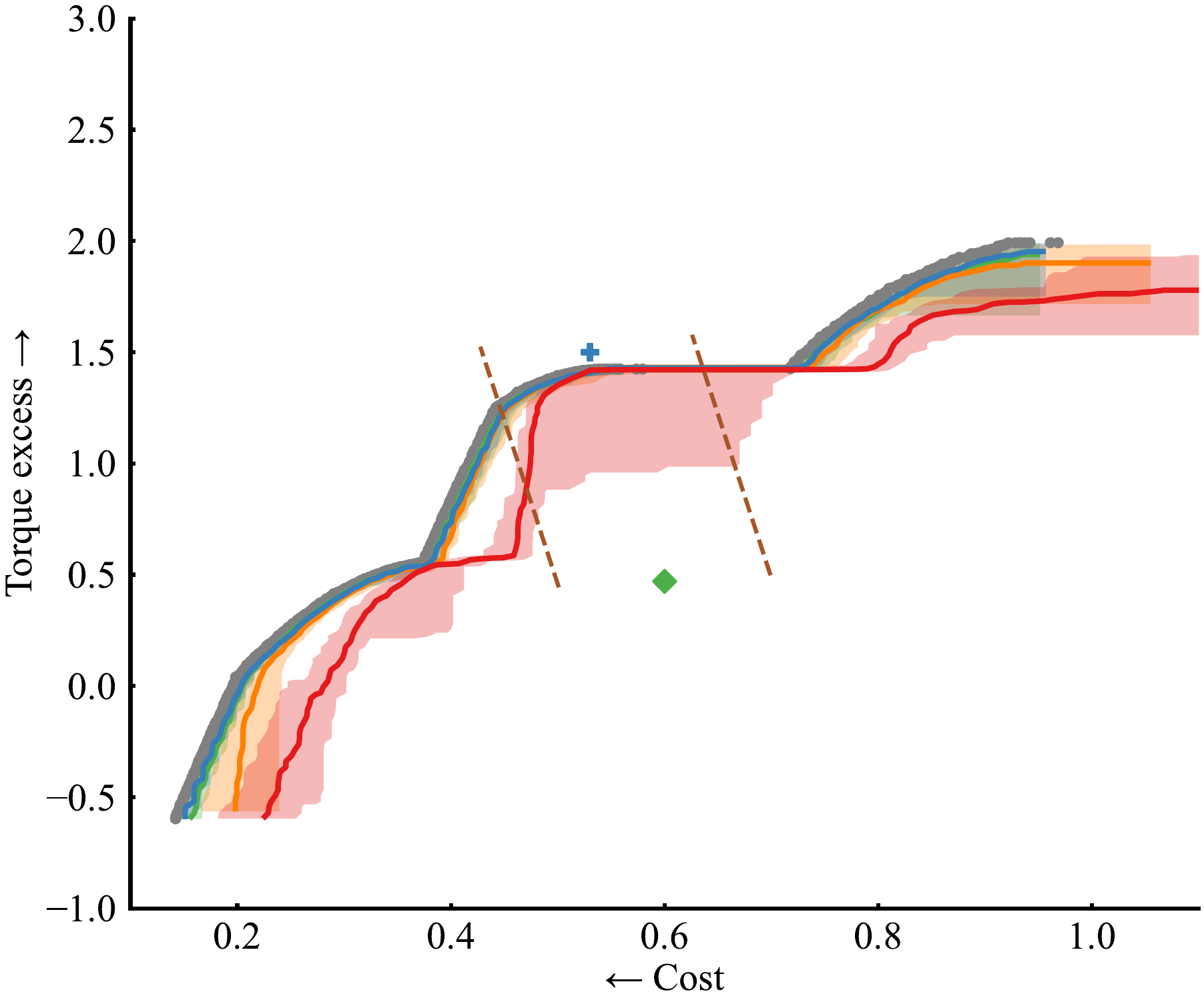}
        \begin{picture}(0,0)
            \put(0,195){a.} %
        \end{picture}

    \end{subfigure}
    \hfill
    \begin{subfigure}[b]{0.49\textwidth}
        \includegraphics[width=\textwidth]{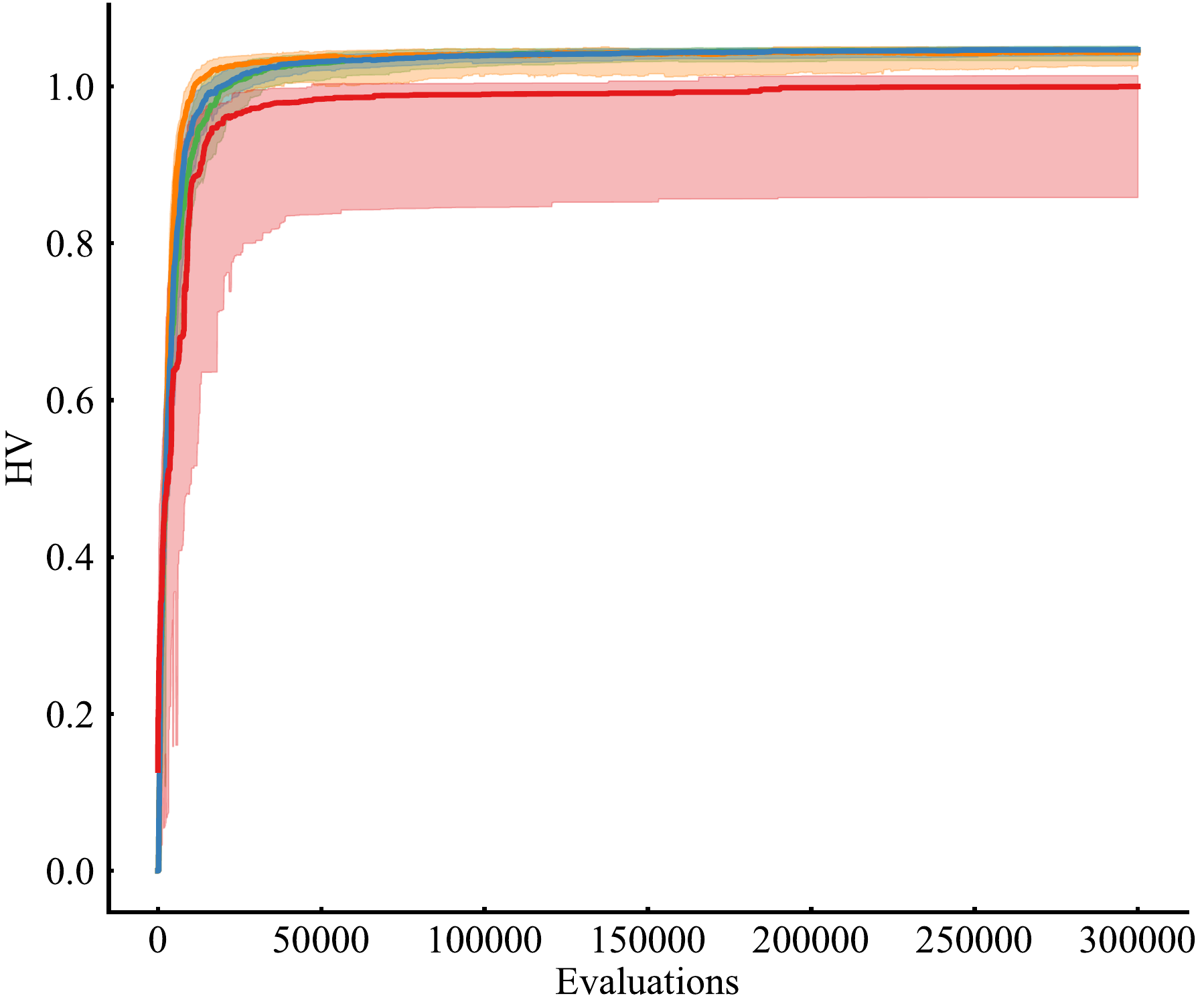}
        \begin{picture}(0,0)
            \put(0,195){b.} %
        \end{picture}
    \end{subfigure}

    \centering
    \vspace*{-1em} %
    {
    \begin{subfigure}[b]{0.6\textwidth}
        \includegraphics[width=\textwidth]{figures/appendix/alternative_targets/cs1/local_best_legend_only.pdf}
    \end{subfigure}
    }

    \caption{ EAF (a) and HV progression (b) of different algorithms, including $5^{th}$ and $95^{th}$ percentiles. Target point: $(0.53, 1.5)$.}
    \label{fig:ct1_best_p4}
\end{figure}

\begin{figure}[H]
    \centering
    \begin{subfigure}[b]{0.49\textwidth}
        \includegraphics[width=\textwidth]{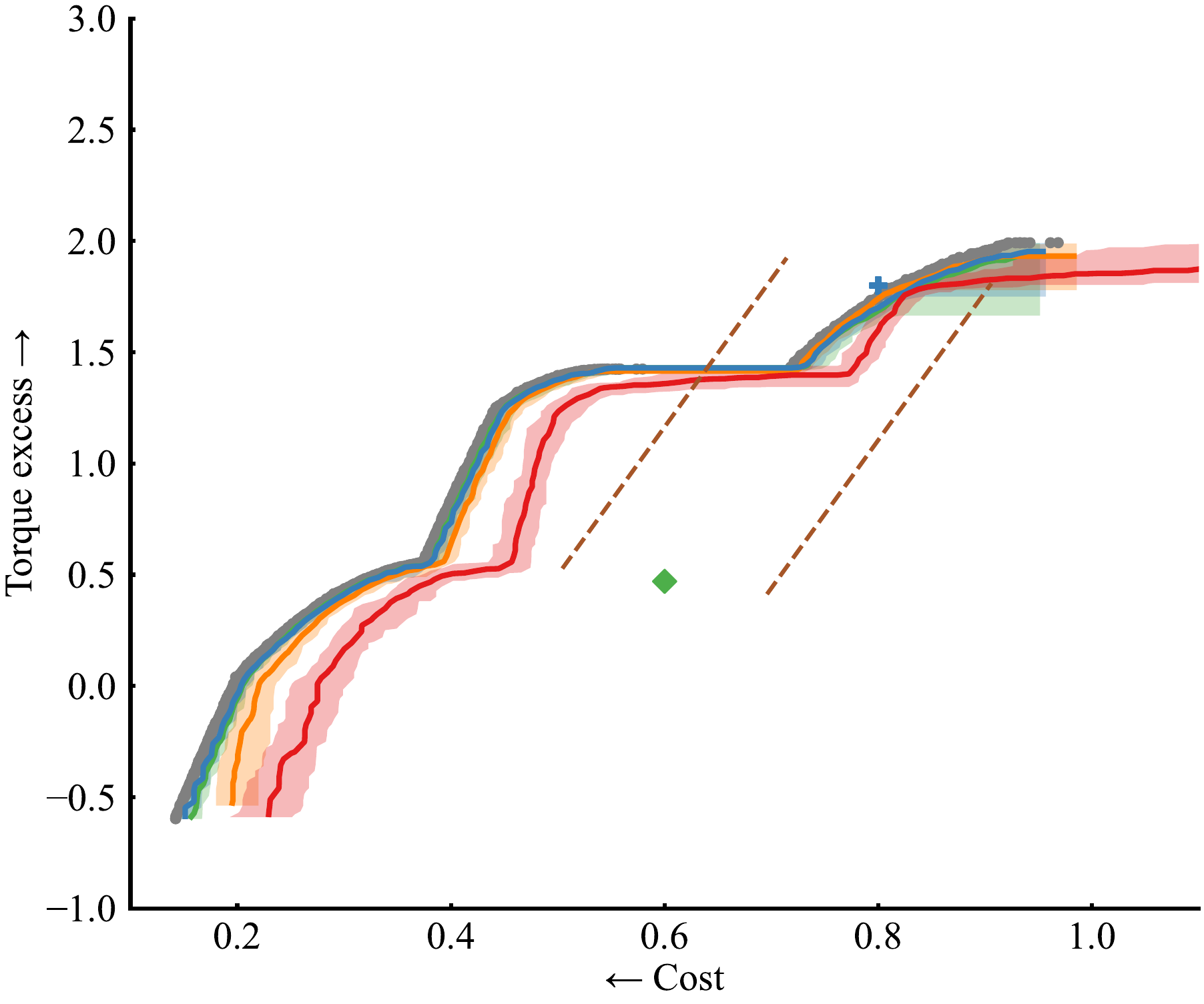}
        \begin{picture}(0,0)
            \put(0,195){a.} %
        \end{picture}

    \end{subfigure}
    \hfill
    \begin{subfigure}[b]{0.49\textwidth}
        \includegraphics[width=\textwidth]{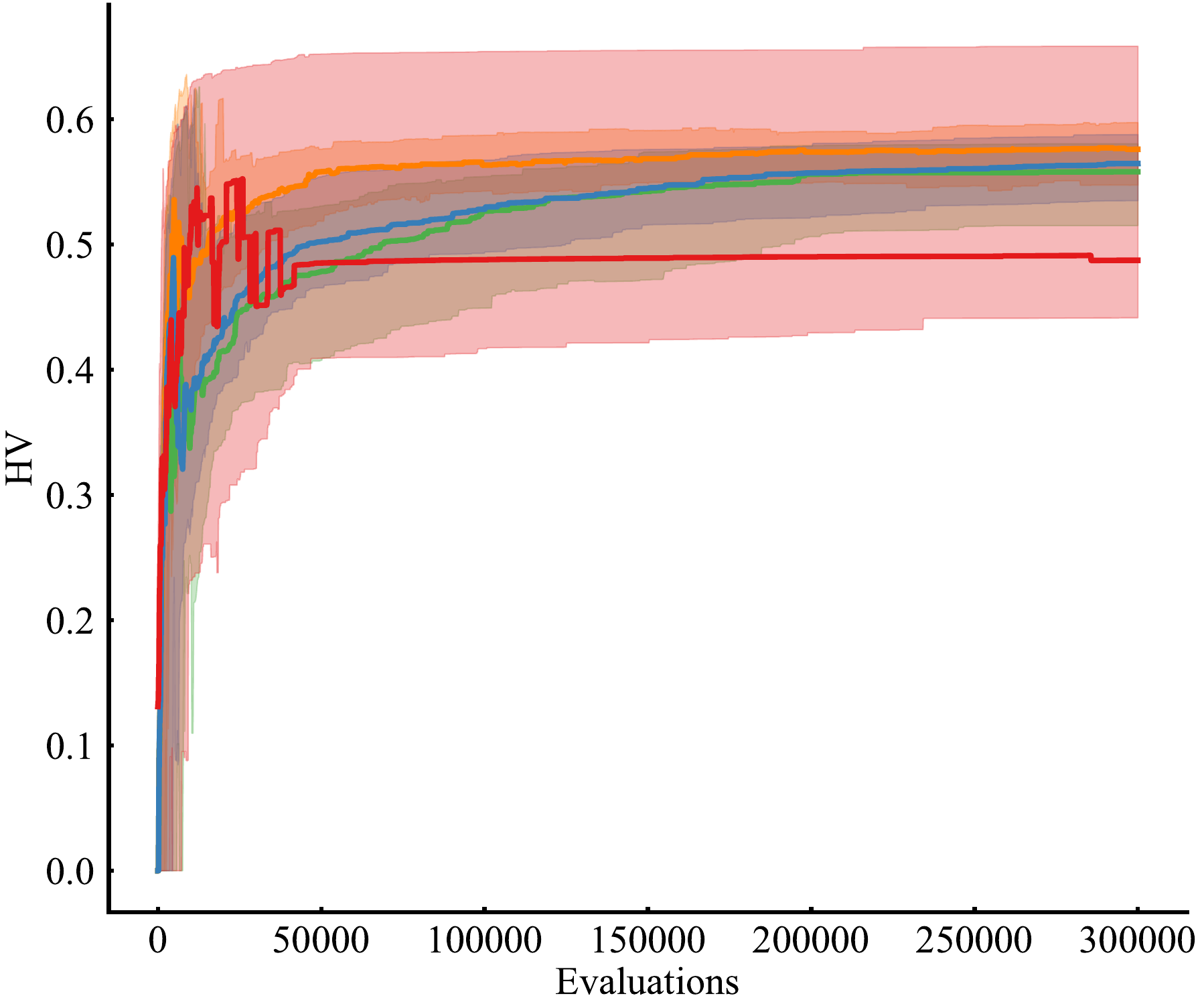}
        \begin{picture}(0,0)
            \put(0,195){b.} %
        \end{picture}
    \end{subfigure}

    \centering
    \vspace*{-1em} %
    {
    \begin{subfigure}[b]{0.6\textwidth}
        \includegraphics[width=\textwidth]{figures/appendix/alternative_targets/cs1/local_best_legend_only.pdf}
    \end{subfigure}
    }

    \caption{ EAF (a) and HV progression (b) of different algorithms, including $5^{th}$ and $95^{th}$ percentiles. Target point: $(0.8, 1.8)$}
    \label{fig:ct1_best_p5}
\end{figure}

Figures \ref{fig:ct1_best_p1}, \ref{fig:ct1_best_p3}, \ref{fig:ct1_best_p4} and \ref{fig:ct1_best_p5} show the EAF and hypervolume progressions for experiments focused towards alternative target points, for ``CT1'' problem. The target points are: $(0.4, 1)$, $(0.25, 0.4)$, $(0.53, 1.5)$ and $(0.8, 1.8)$ respectively.

\clearpage

\clearpage
\newpage
\section{MODact ``CTS1'' additional results}
\label{app:cts1-additional-results}

\begin{figure}[H]
    \centering
    \begin{subfigure}[b]{0.49\textwidth}
        \includegraphics[width=\textwidth]{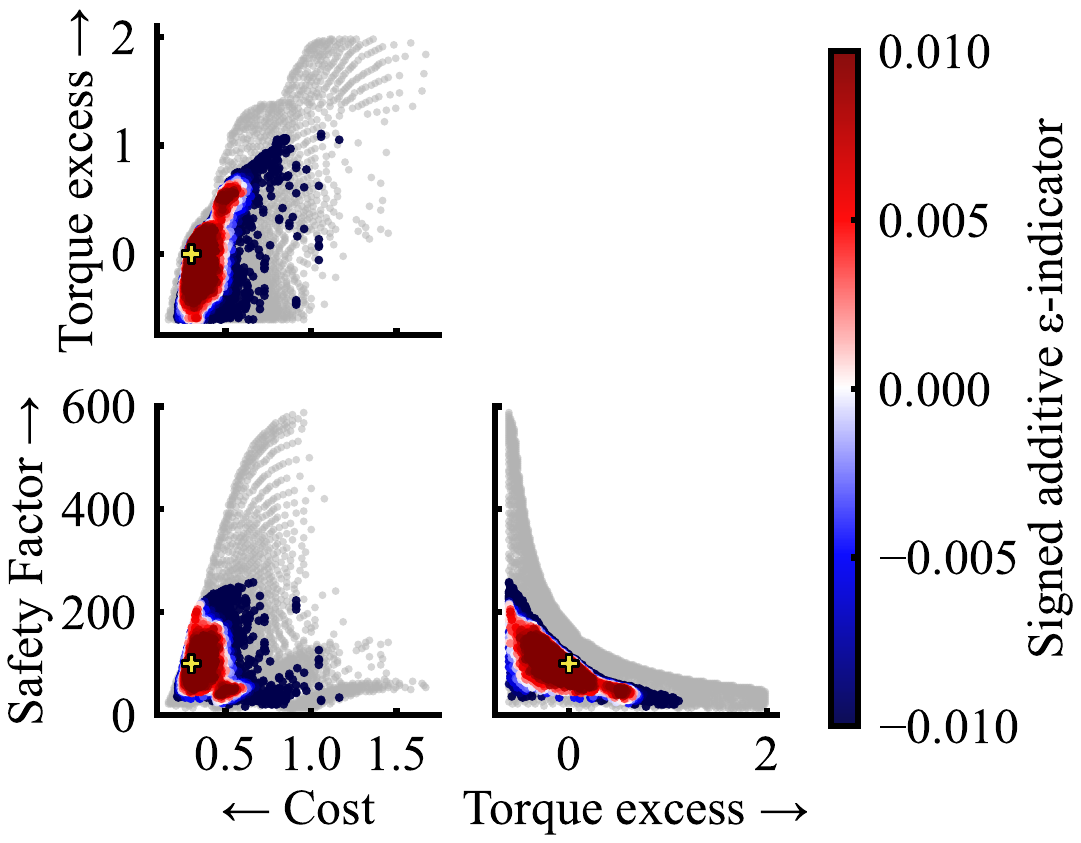}
        \begin{picture}(0,0)
            \put(0,180){a.} %
        \end{picture}

    \end{subfigure}
    \hfill
    \begin{subfigure}[b]{0.49\textwidth}
        \includegraphics[width=\textwidth]{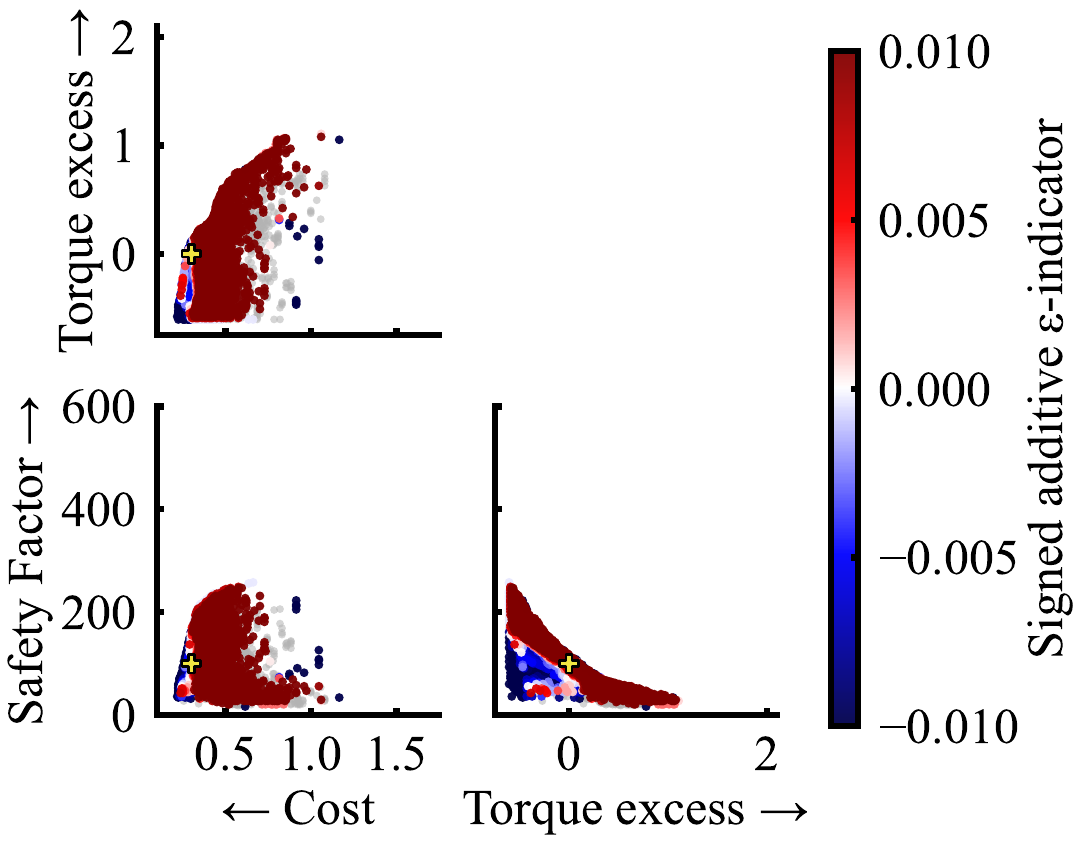}
        \begin{picture}(0,0)
            \put(0,180){b.} %
        \end{picture}
    \end{subfigure}

    \centering
    \vspace*{-1em} %

    \centering
    \begin{subfigure}[b]{0.49\textwidth}
        \includegraphics[width=\textwidth]{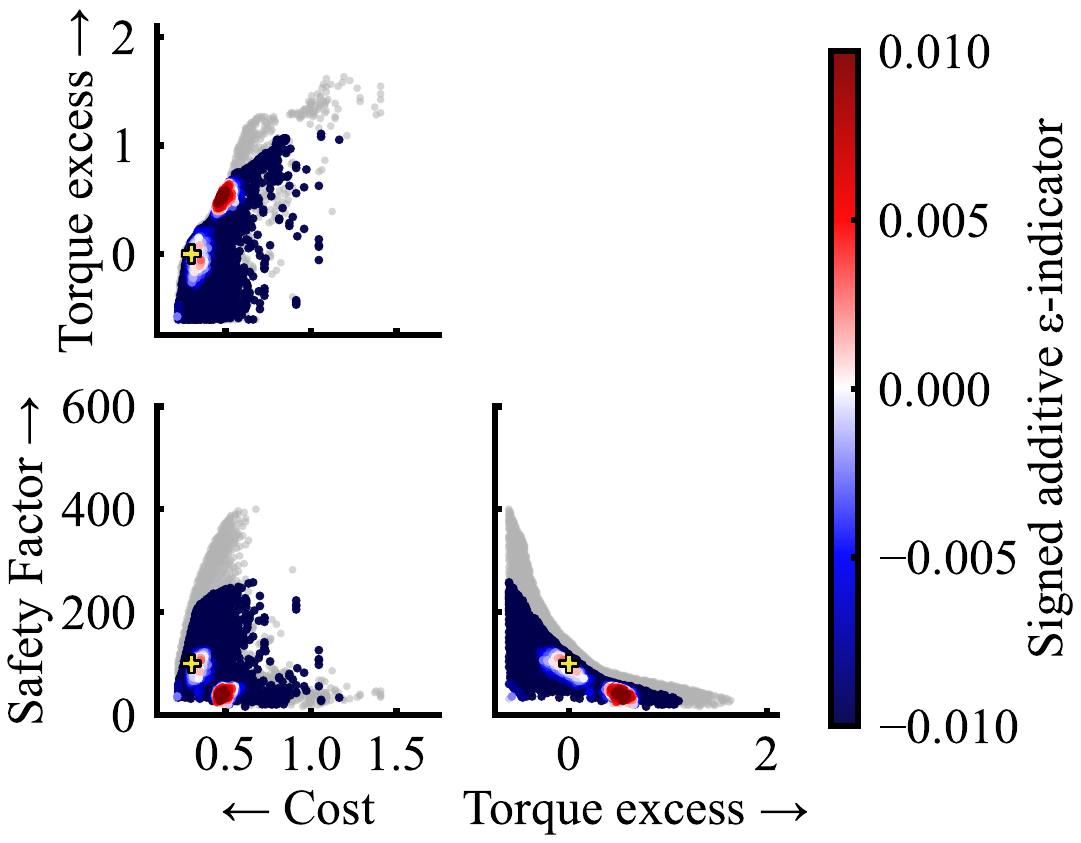}
        \begin{picture}(0,0)
            \put(0,180){c.} %
        \end{picture}

    \end{subfigure}
    \hfill
    \begin{subfigure}[b]{0.49\textwidth}
        \includegraphics[width=\textwidth]{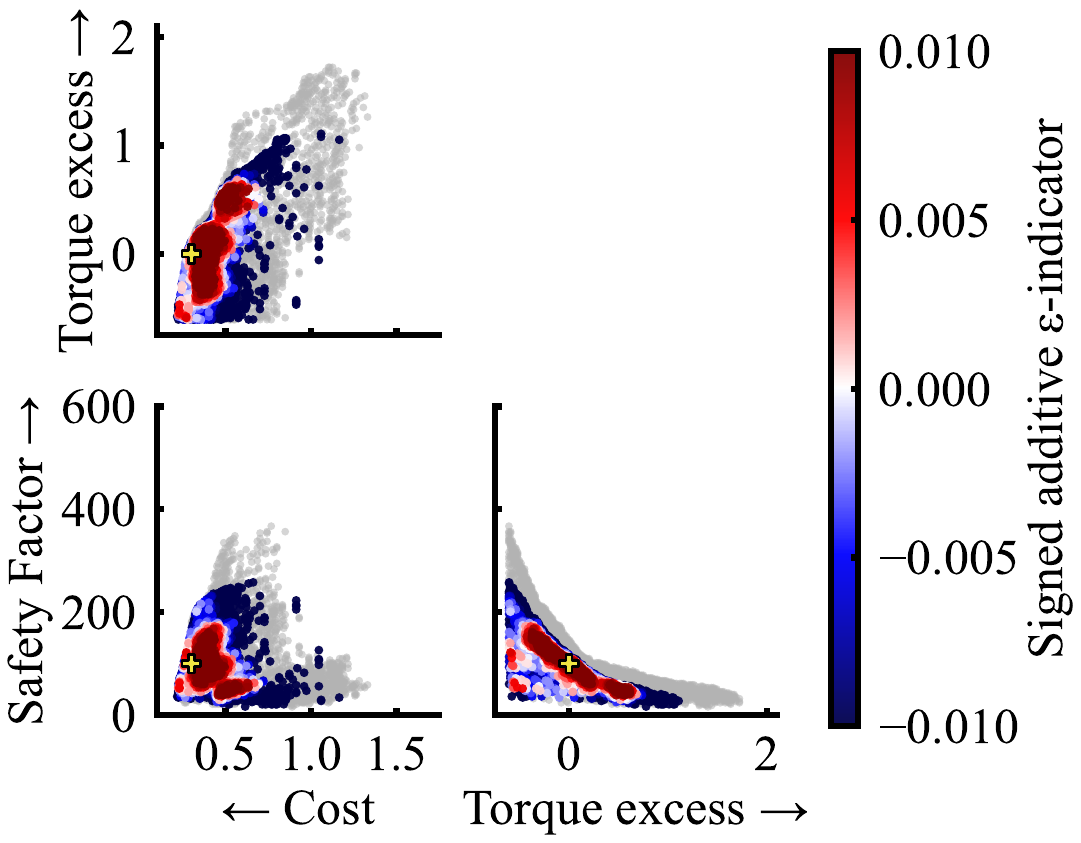}
        \begin{picture}(0,0)
            \put(0,180){d.} %
        \end{picture}
    \end{subfigure}

    \centering

    \caption{ Pairwise projections of median EAF for different algorithms: (a) NSGA-II (pop. $200$), (b) R-NSGA-II ($\varepsilon = 0.01$), (c) NSGA-III (200 ref. dir.) and (d) R-NSGA-III ($\mu = 0.01$). Target point is shown as yellow cross. Target point: $(0.3, 0.0, 100.)$.}
    \label{fig:cts1_p2}
\end{figure}

\begin{figure}[H]
    \centering
    \begin{subfigure}[b]{0.49\textwidth}
        \includegraphics[width=\textwidth]{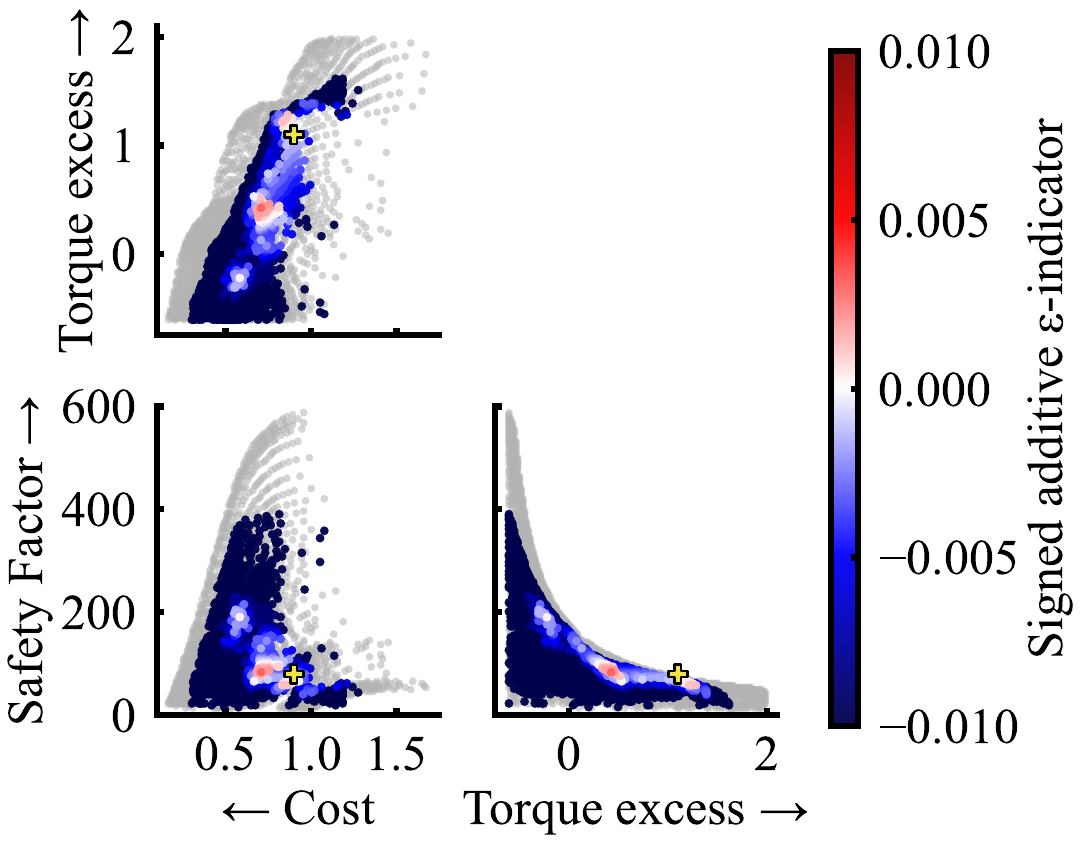}
        \begin{picture}(0,0)
            \put(0,180){a.} %
        \end{picture}

    \end{subfigure}
    \hfill
    \begin{subfigure}[b]{0.49\textwidth}
        \includegraphics[width=\textwidth]{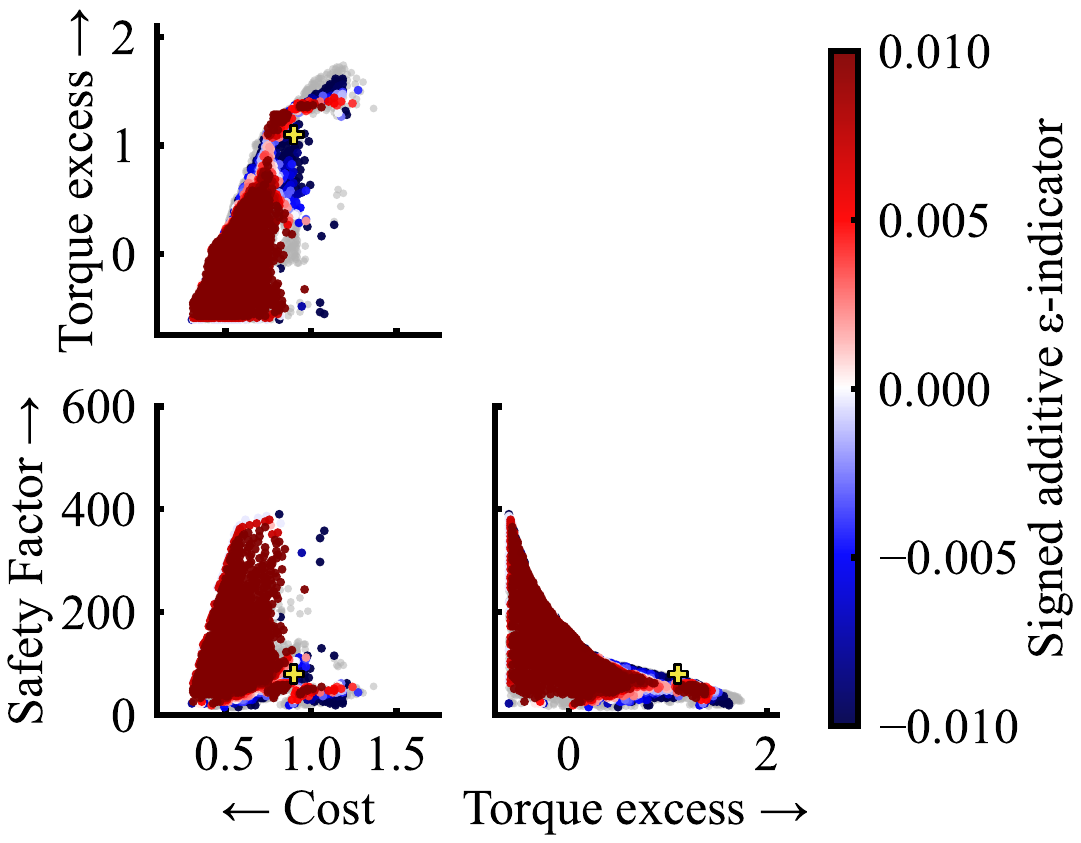}
        \begin{picture}(0,0)
            \put(0,180){b.} %
        \end{picture}
    \end{subfigure}

    \centering
    \vspace*{-1em} %

    \centering
    \begin{subfigure}[b]{0.49\textwidth}
        \includegraphics[width=\textwidth]{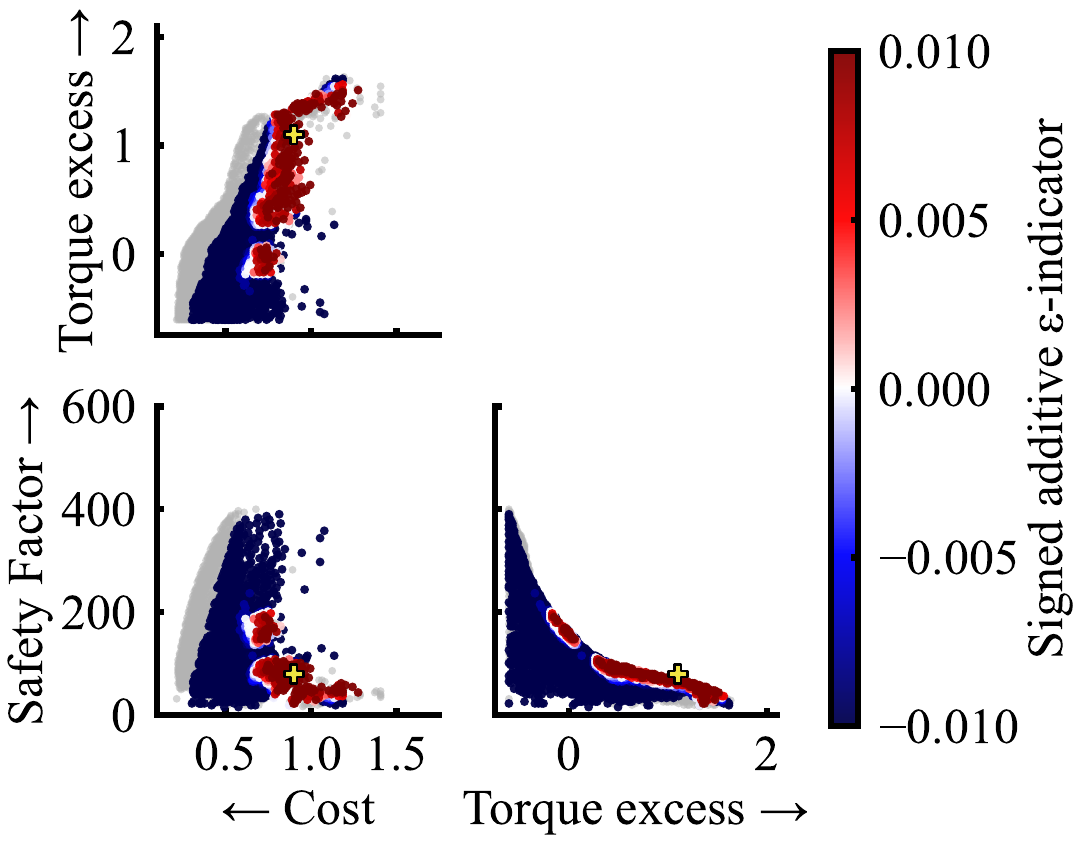}
        \begin{picture}(0,0)
            \put(0,180){c.} %
        \end{picture}

    \end{subfigure}
    \hfill
    \begin{subfigure}[b]{0.49\textwidth}
        \includegraphics[width=\textwidth]{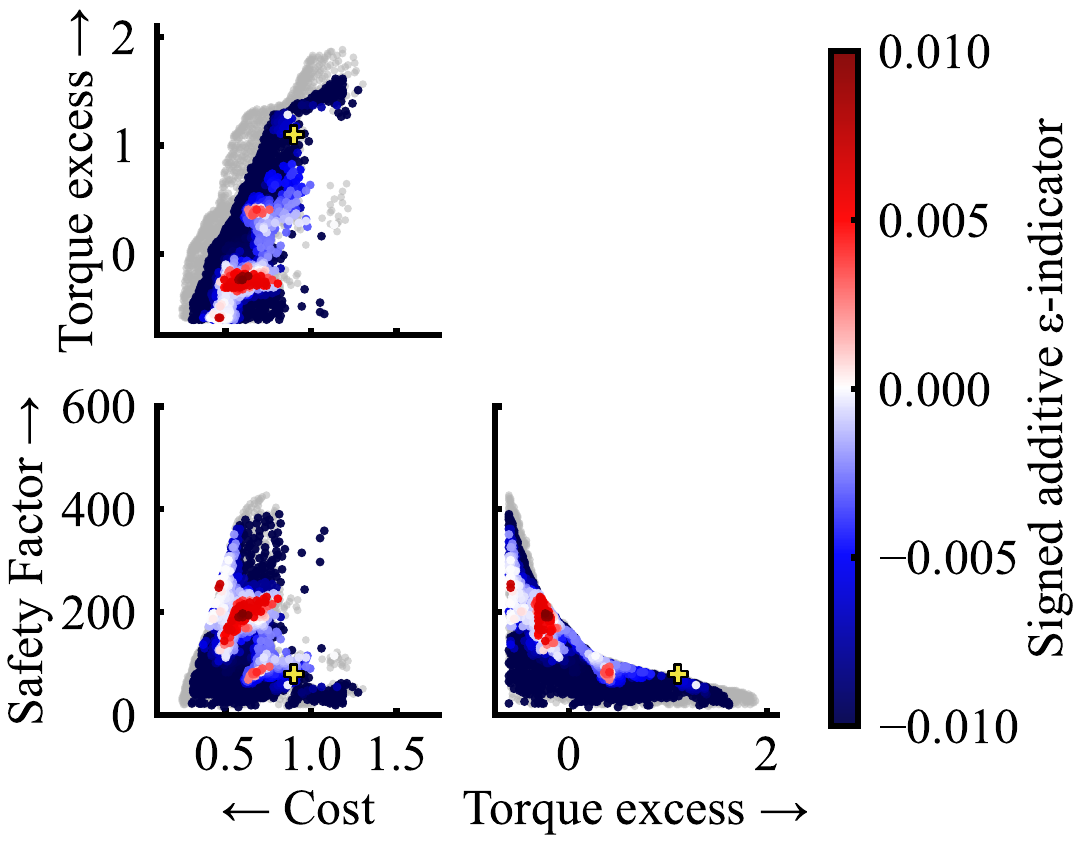}
        \begin{picture}(0,0)
            \put(0,180){d.} %
        \end{picture}
    \end{subfigure}

    \centering

    \caption{ Pairwise projections of median EAF for different algorithms: (a) NSGA-II (pop. $200$), (b) R-NSGA-II ($\varepsilon = 0.01$), (c) NSGA-III (200 ref. dir.) and (d) R-NSGA-III ($\mu = 0.01$). Target point is shown as yellow cross. Target point: $(0.9, 1.1, 80.)$.}
    \label{fig:cts1_p3}
\end{figure}

\begin{figure}[H]
    \centering
    \begin{subfigure}[b]{0.49\textwidth}
        \includegraphics[width=\textwidth]{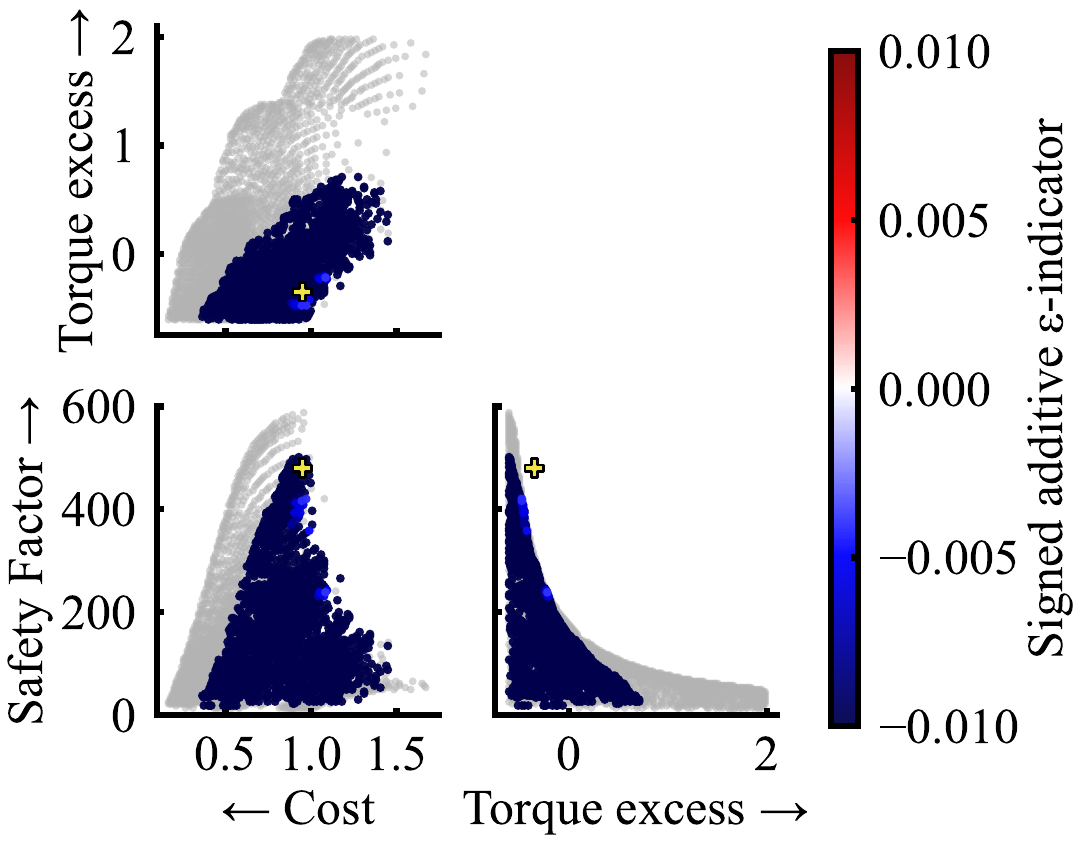}
        \begin{picture}(0,0)
            \put(0,180){a.} %
        \end{picture}

    \end{subfigure}
    \hfill
    \begin{subfigure}[b]{0.49\textwidth}
        \includegraphics[width=\textwidth]{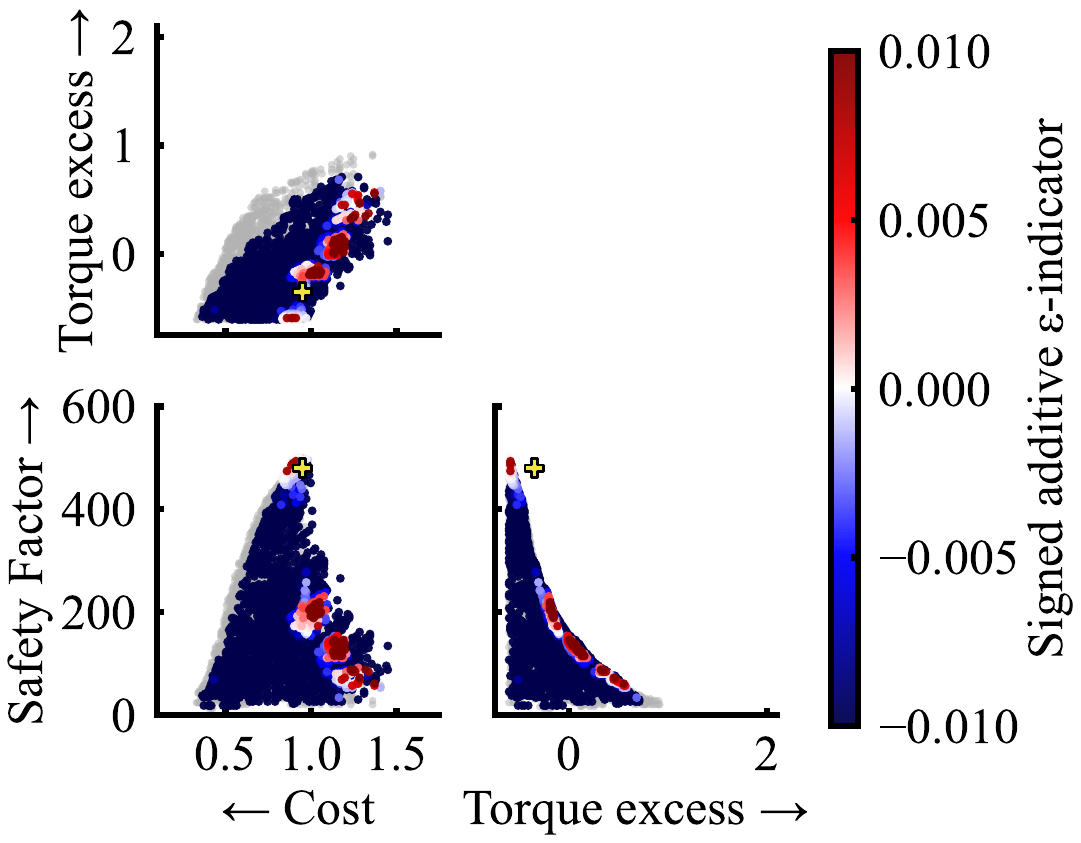}
        \begin{picture}(0,0)
            \put(0,180){b.} %
        \end{picture}
    \end{subfigure}

    \centering
    \vspace*{-1em} %

    \centering
    \begin{subfigure}[b]{0.49\textwidth}
        \includegraphics[width=\textwidth]{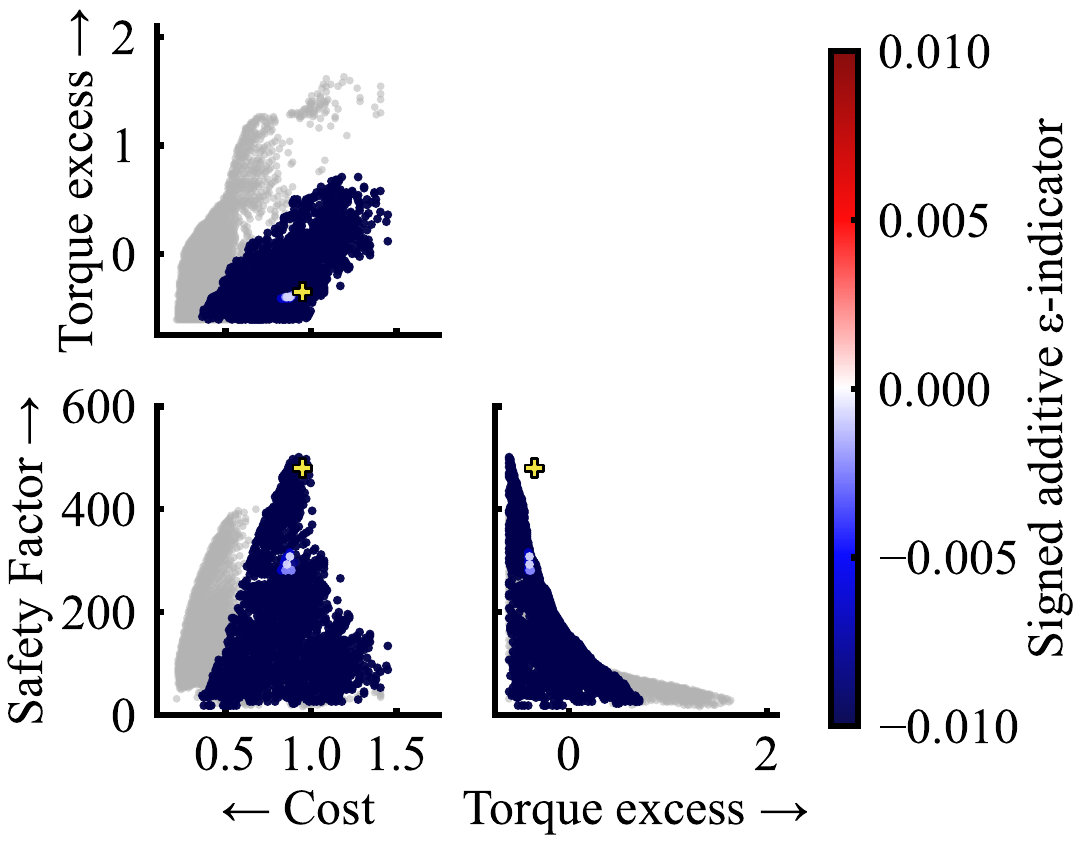}
        \begin{picture}(0,0)
            \put(0,180){c.} %
        \end{picture}

    \end{subfigure}
    \hfill
    \begin{subfigure}[b]{0.49\textwidth}
        \includegraphics[width=\textwidth]{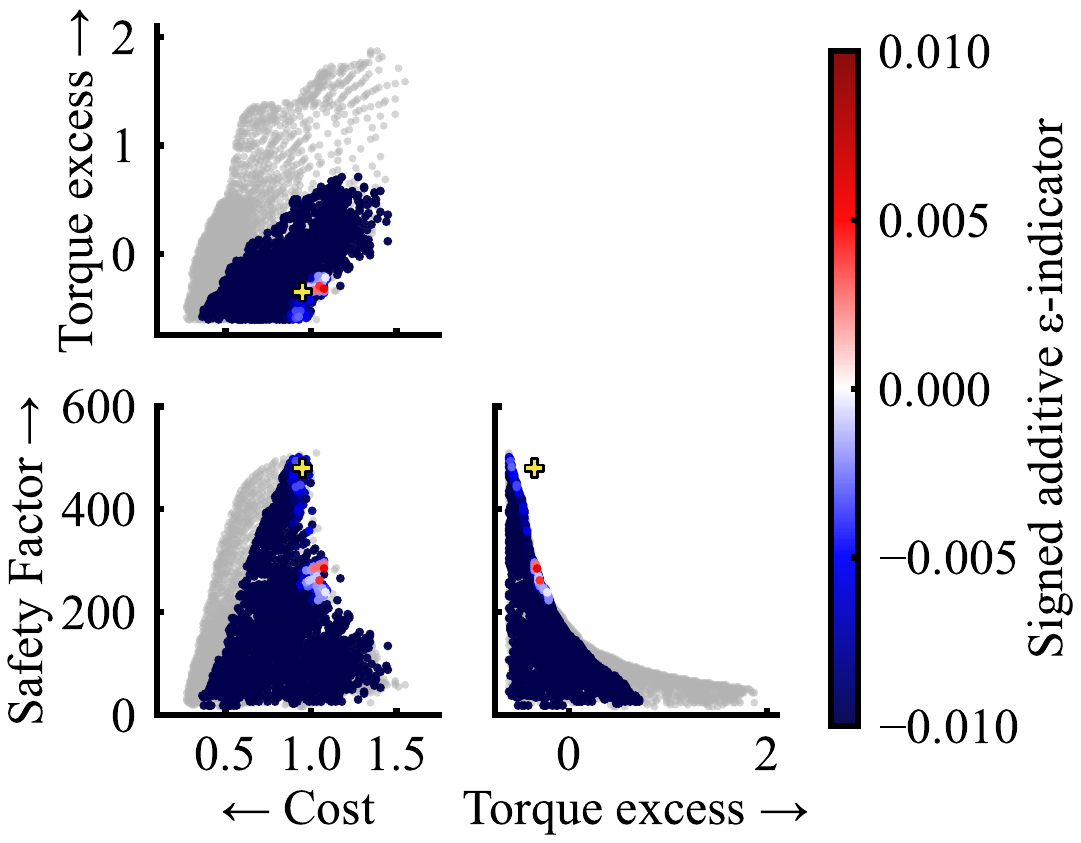}
        \begin{picture}(0,0)
            \put(0,180){d.} %
        \end{picture}
    \end{subfigure}

    \centering

    \caption{ Pairwise projections of median EAF for different algorithms: (a) NSGA-II (pop. $200$), (b) R-NSGA-II ($\varepsilon = 0.01$), (c) NSGA-III (200 ref. dir.) and (d) R-NSGA-III ($\mu = 0.01$). Target point is shown as yellow cross. Target point: $(0.95, -0.35, 480.)$.}
    \label{fig:cts1_p4}
\end{figure}

\begin{figure}[H]
    \centering
    \begin{subfigure}[b]{0.49\textwidth}
        \includegraphics[width=\textwidth]{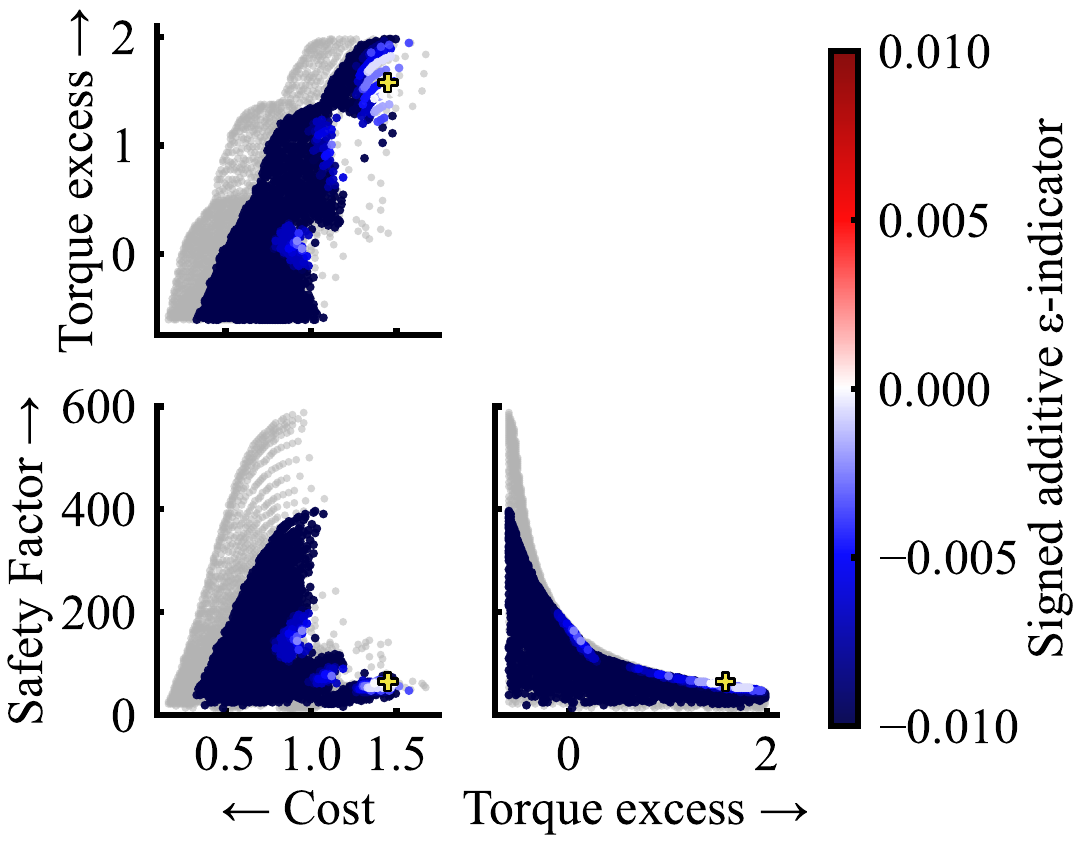}
        \begin{picture}(0,0)
            \put(0,180){a.} %
        \end{picture}

    \end{subfigure}
    \hfill
    \begin{subfigure}[b]{0.49\textwidth}
        \includegraphics[width=\textwidth]{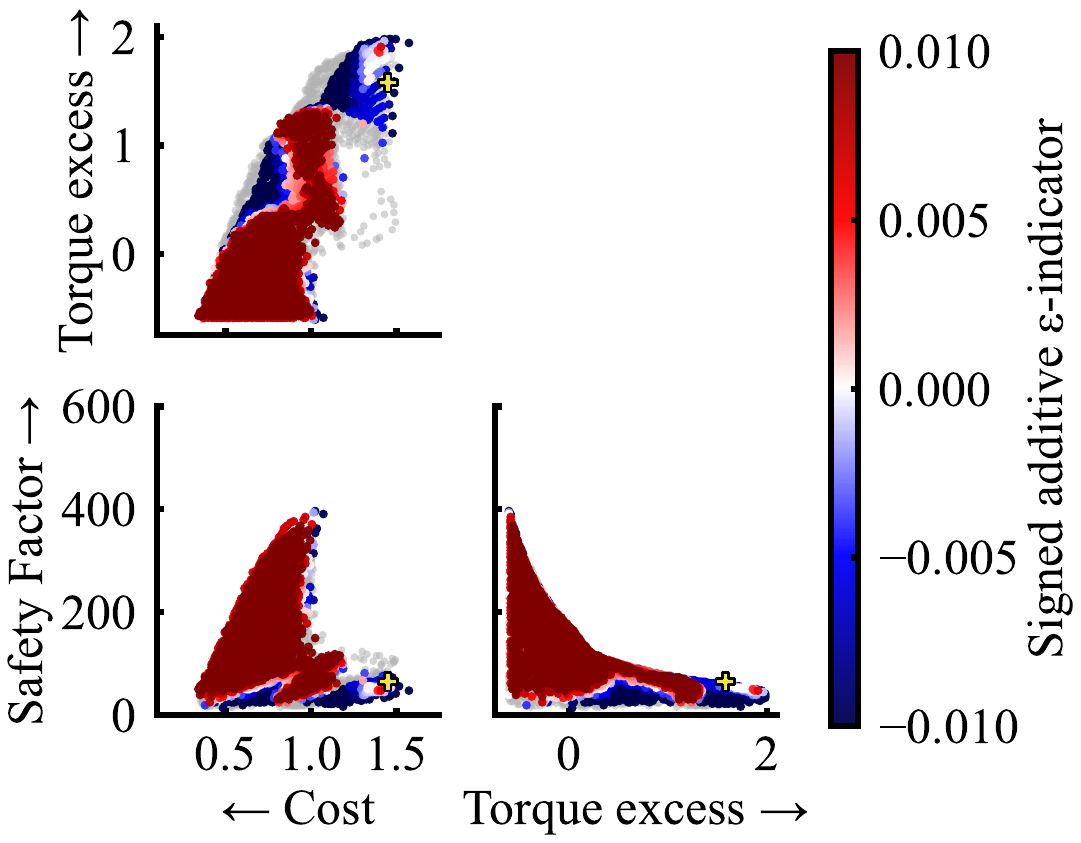}
        \begin{picture}(0,0)
            \put(0,180){b.} %
        \end{picture}
    \end{subfigure}

    \centering
    \vspace*{-1em} %

    \centering
    \begin{subfigure}[b]{0.49\textwidth}
        \includegraphics[width=\textwidth]{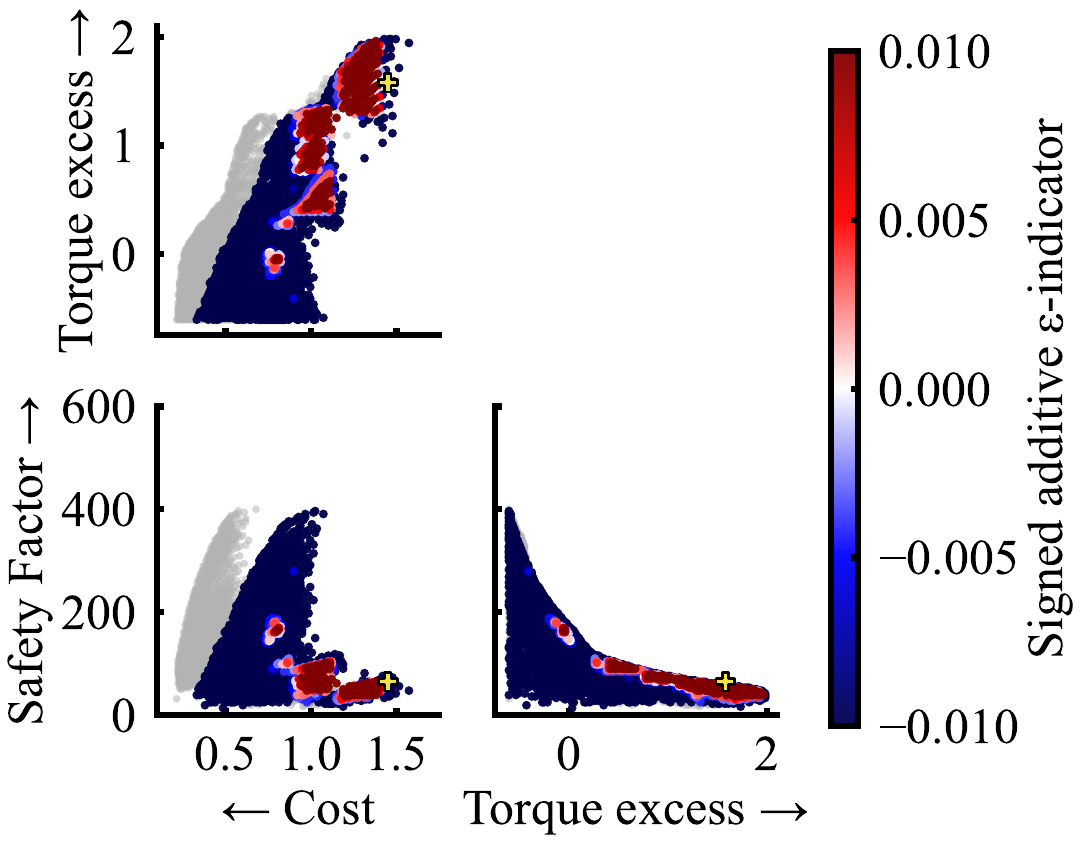}
        \begin{picture}(0,0)
            \put(0,180){c.} %
        \end{picture}

    \end{subfigure}
    \hfill
    \begin{subfigure}[b]{0.49\textwidth}
        \includegraphics[width=\textwidth]{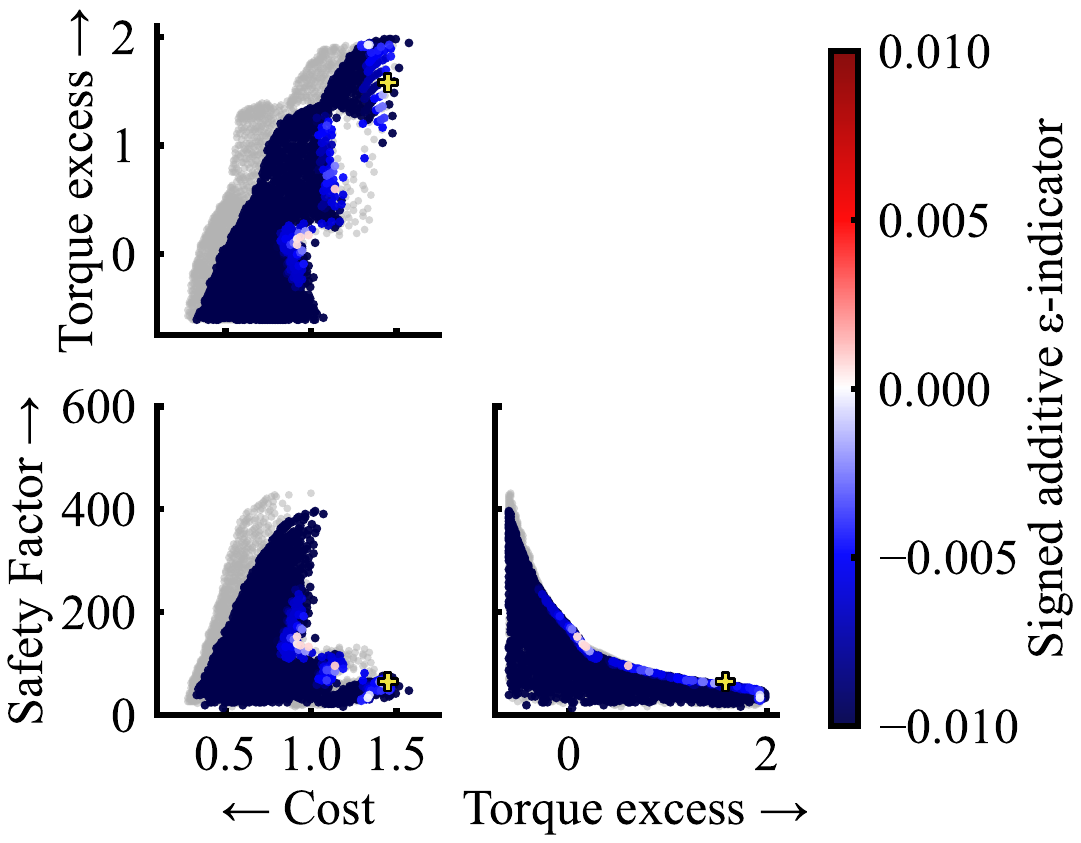}
        \begin{picture}(0,0)
            \put(0,180){d.} %
        \end{picture}
    \end{subfigure}

    \centering

    \caption{ Pairwise projections of median EAF for different algorithms: (a) NSGA-II (pop. $200$), (b) R-NSGA-II ($\varepsilon = 0.01$), (c) NSGA-III (200 ref. dir.) and (d) R-NSGA-III ($\mu = 0.01$). Target point is shown as yellow cross. Target point: $(1.45, 1.58, 65.)$}
    \label{fig:cts1_p5}
\end{figure}

Figures \ref{fig:cts1_p2}, \ref{fig:cts1_p3}, \ref{fig:cts1_p4} and \ref{fig:cts1_p5} show the EAF and hypervolume progressions for experiments focused towards alternative target points, for ``CS1'' problem. The target points are: $(0.3, 0.0, 100.)$, $(0.9, 1.1, 80.)$, $(0.95, -0.35, 480.)$, $(1.45, 1.58, 65.)$ respectively.

\clearpage

\clearpage
\newpage
\section{Ablations, additional results}
\label{app:ablation-additional}

\begin{figure}[H]
    \centering
    \begin{subfigure}[b]{0.49\textwidth}
        \includegraphics[width=\textwidth]{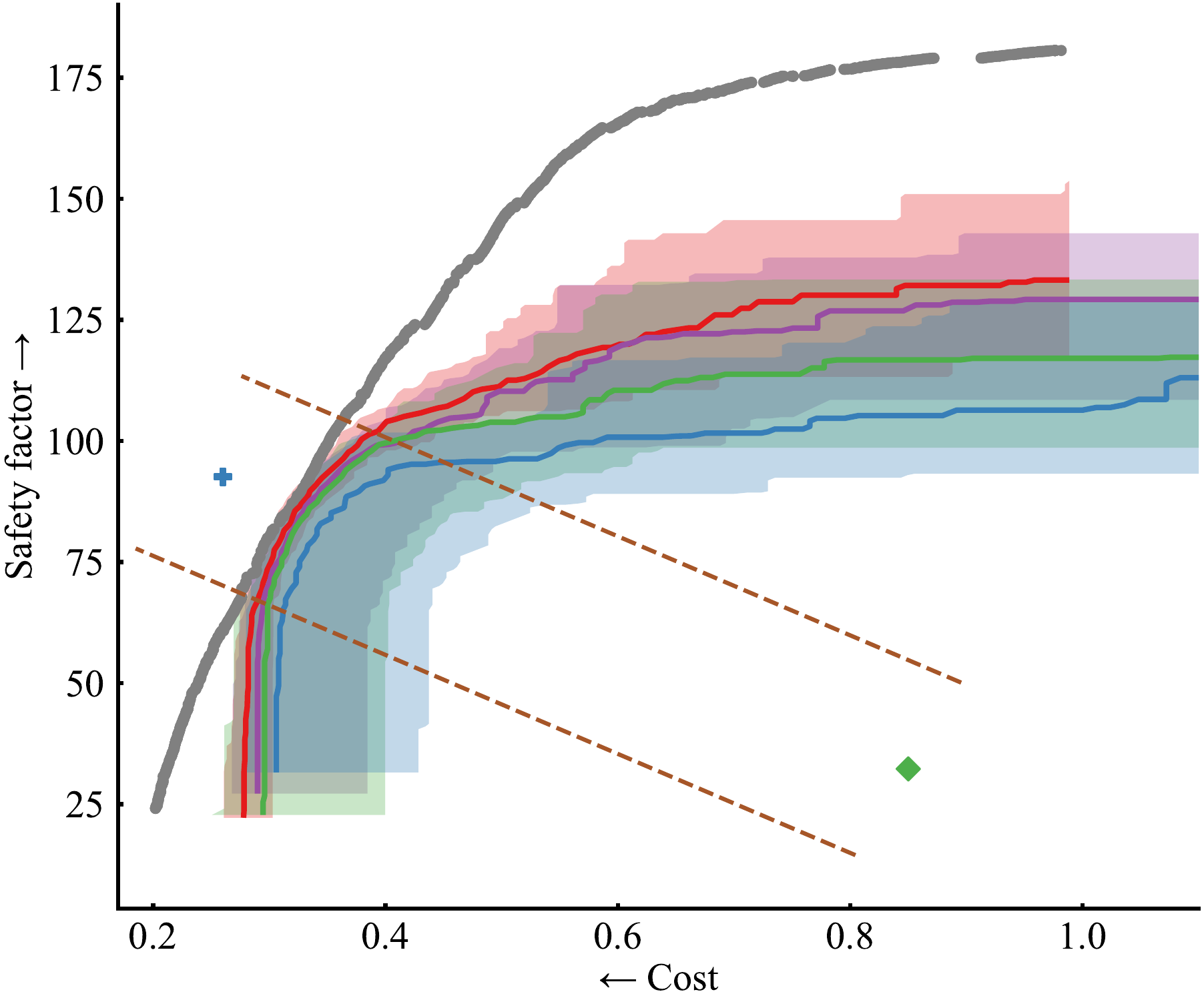}
        \begin{picture}(0,0)
        \end{picture}

    \end{subfigure}
    \vspace*{-1em} %
    {
    \begin{subfigure}[b]{0.8\textwidth}
        \includegraphics[width=\textwidth]{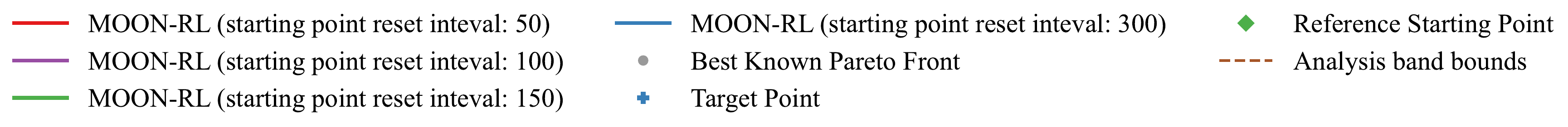}
    \end{subfigure}
    }
    \vspace*{1em} %

    \caption{Effect of starting point reset intervals}
    \label{fig:ablation-reset-intevals}
\end{figure}

\begin{figure}[H]
    \centering
    \begin{subfigure}[b]{0.49\textwidth}
        \includegraphics[width=\textwidth]{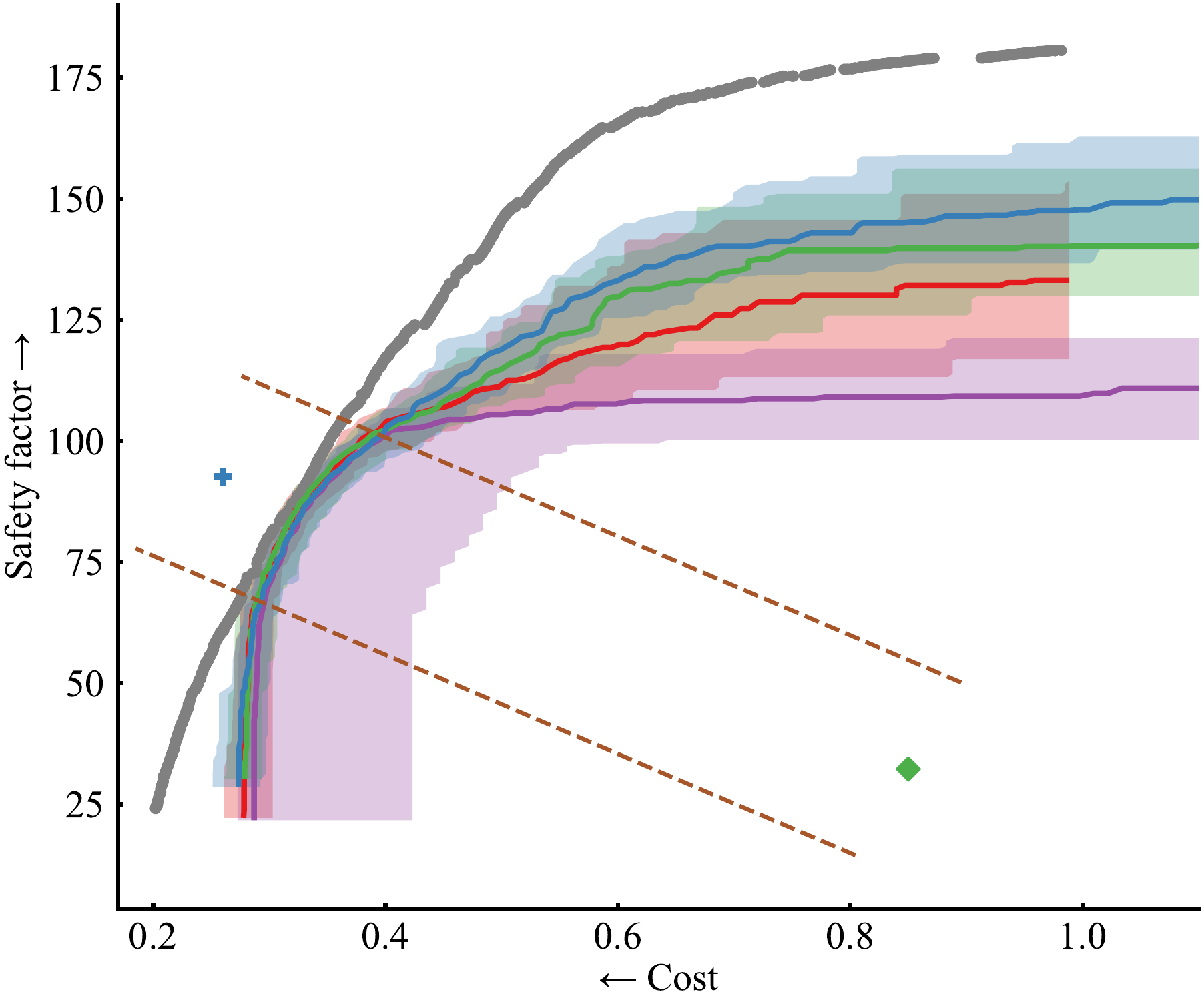}
        \begin{picture}(0,0)
        \end{picture}

    \end{subfigure}
    \vspace*{-1em} %
    {
    \begin{subfigure}[b]{0.8\textwidth}
        \includegraphics[width=\textwidth]{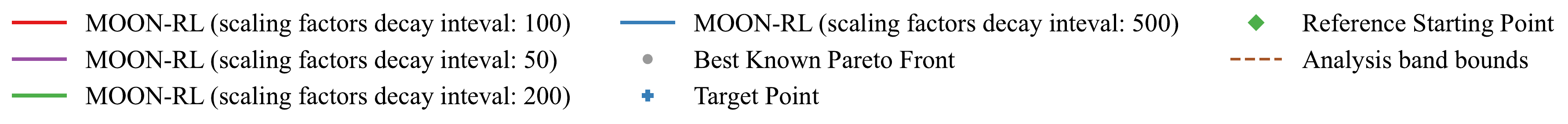}
    \end{subfigure}
    }
    \vspace*{1em} %

    \caption{Effect of scaling factors decay intervals}
    \label{fig:ablation-sf-decay-intevals}
\end{figure}

\begin{figure}[H]
    \centering
    \begin{subfigure}[b]{0.49\textwidth}
        \includegraphics[width=\textwidth]{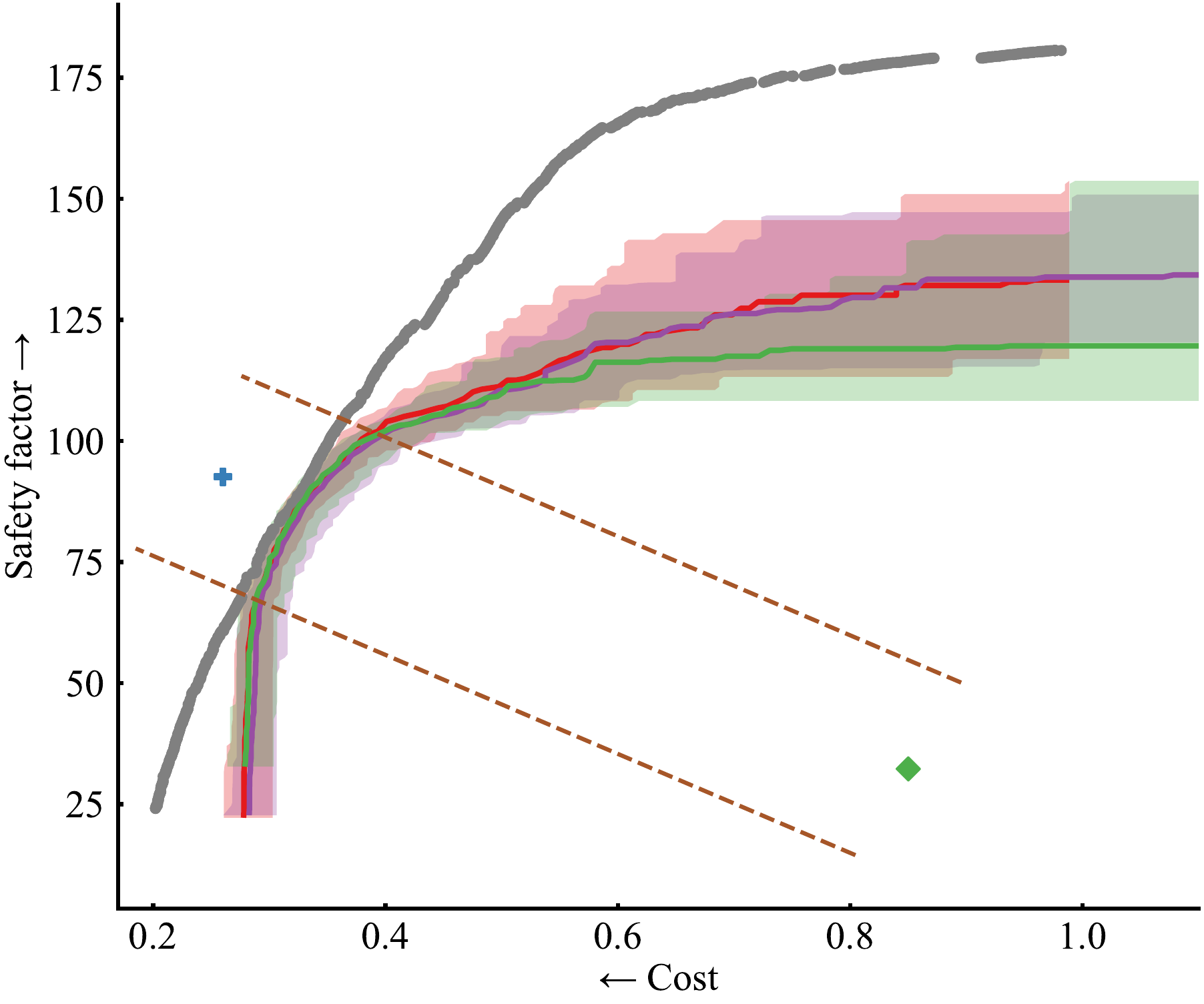}
        \begin{picture}(0,0)
        \end{picture}

    \end{subfigure}
    \vspace*{-1em} %
    {
    \begin{subfigure}[b]{0.8\textwidth}
        \includegraphics[width=\textwidth]{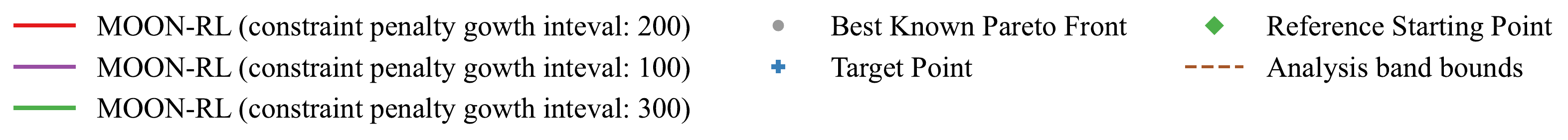}
    \end{subfigure}
    }
    \vspace*{1em} %

    \caption{ Effect of constraint penalty growth intervals}
    \label{fig:ablation-cv_intervals-intevals}
\end{figure}

Figures~\ref{fig:ablation-reset-intevals}, \ref{fig:ablation-sf-decay-intevals}, and \ref{fig:ablation-cv_intervals-intevals} show empirical attainment functions for different values of the starting point reset interval, the decay intervals of the scaling factors, and the constraint penalty growth intervals\textcolor{black}{, respectively}.

\noindent\textbf{Starting-point sampling temperature.} At every meta-algorithm reset, a new starting design is drawn from the feasible non-dominated archive using a rank-based softmax, \(p_i \propto \exp(-i/\tau)\), where rank \(i=0\) denotes the design closest to the target. We repeat the focused-search experiment on ``CS1'' with \(\tau \in \{2,4,8,16\}\), retaining the same settings, evaluation budget, and 30 random seeds.

\begin{figure}[h]
    \centering
    \begin{subfigure}[b]{0.49\textwidth}
        \includegraphics[width=\textwidth]{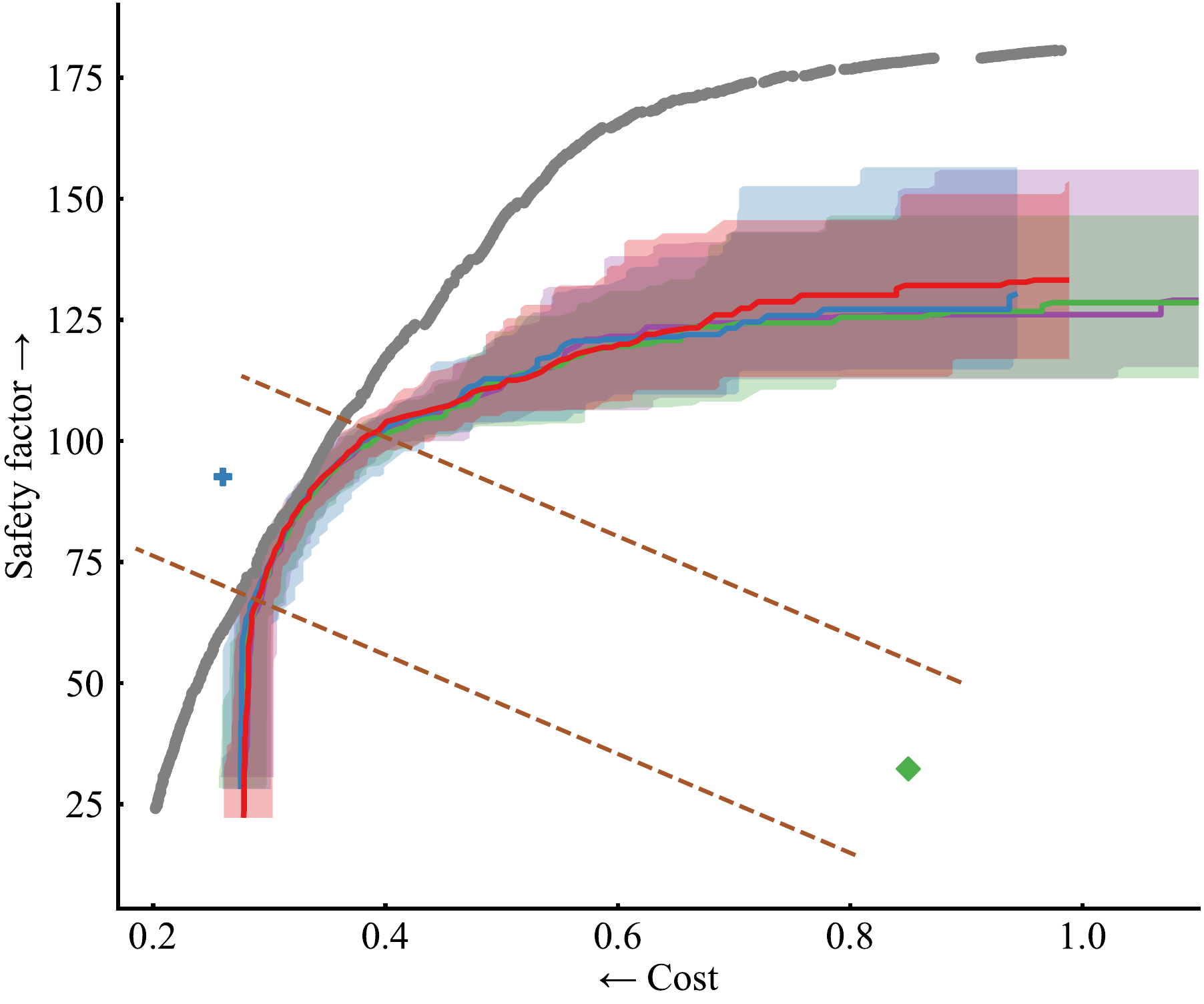}
        \begin{picture}(0,0)
            \put(0,195){a.}
        \end{picture}
        \label{fig:temperature_sweep_eaf}
    \end{subfigure}
    \hfill
    \begin{subfigure}[b]{0.49\textwidth}
        \includegraphics[width=\textwidth]{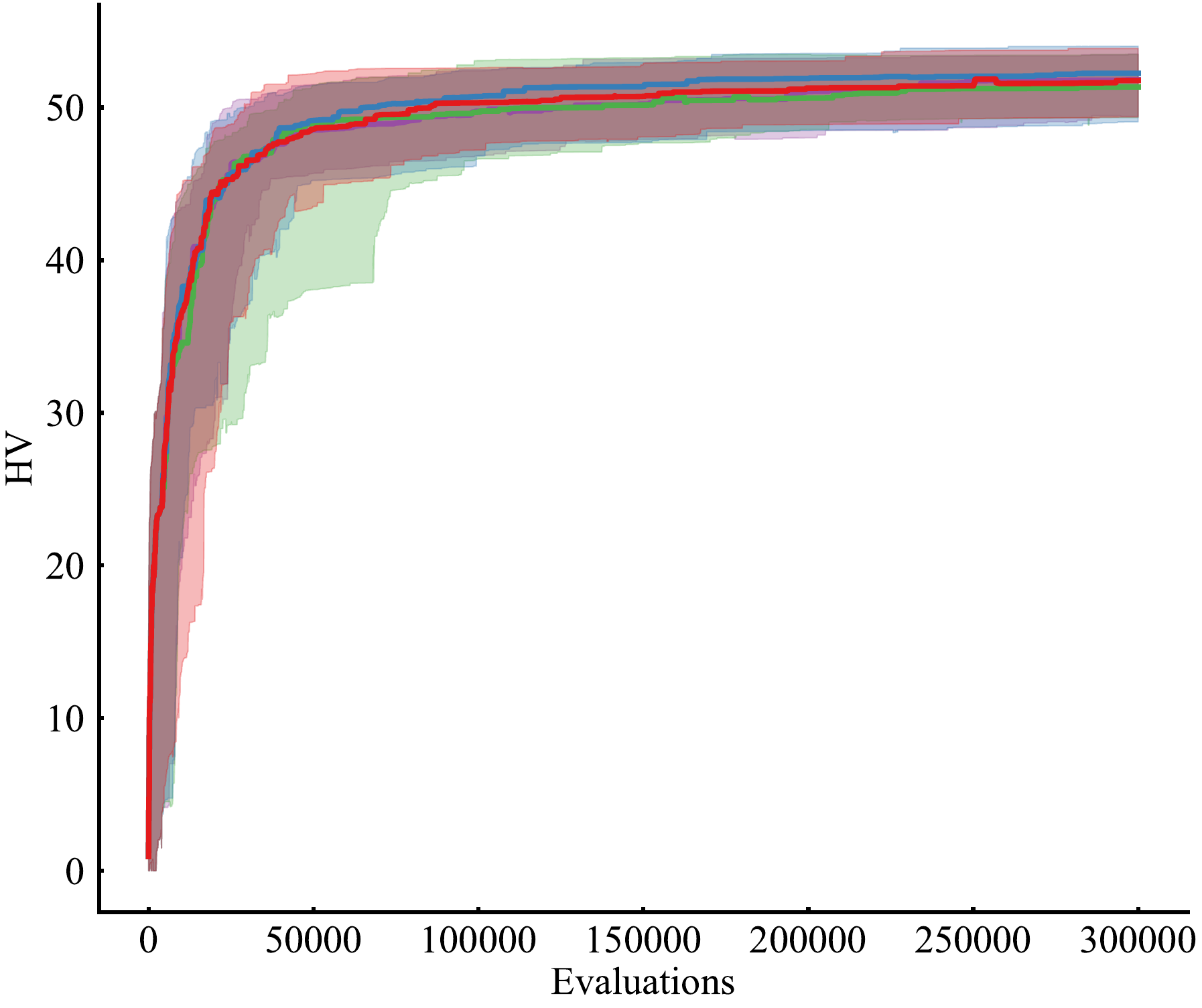}
        \begin{picture}(0,0)
            \put(0,195){b.}
        \end{picture}
        \label{fig:temperature_sweep_hv}
    \end{subfigure}

    \centering
    \vspace*{-1em}
    {
    \begin{subfigure}[b]{0.6\textwidth}
        \includegraphics[width=\textwidth]{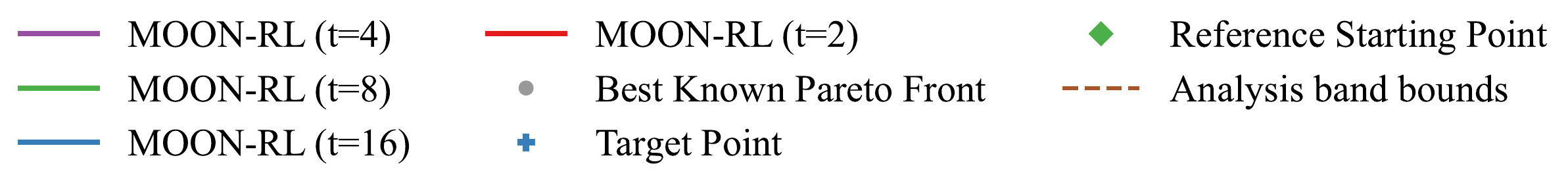}
    \end{subfigure}
    }

    \caption{Sensitivity of MOON-RL to the starting-point sampling temperature \(\tau\) on the focused-search ``CS1'' experiment. Panel (a) shows the empirical attainment functions and panel (b) shows hypervolume progression. Solid lines show medians and shaded regions show the 5th--95th percentiles across 30 random seeds.}
    \label{fig:temperature_sweep}
\end{figure}

Figure~\ref{fig:temperature_sweep} shows that the EAFs and hypervolume progressions remain similar across the examined values, with closely aligned median curves and broadly overlapping inter-percentile bands. No systematic improvement or deterioration is observed as the temperature increases. Performance on ``CS1'' is therefore insensitive to this parameter over the tested range, and the default value \(\tau=2\) is retained. A plausible explanation is that the feasible non-dominated archive is already concentrated in the region reached by the directional search. Changing the temperature then redistributes the restart probability among designs from the same focused region without materially changing the subsequent search behaviour.

\clearpage

\clearpage
\newpage

\section{Observation Ablation: Including Design Variables}
\label{app:observation_ablation}

To assess the effect of including the design vector in the policy observation, we performed an additional ablation on the ``CS1'' benchmark. The original MOON-RL observation, used in the main experiments, contains objective-space quantities and constraint-related information, but excludes the current design vector $\boldsymbol{x_t}$. In the ablation, we augment the observation with $x_t$ while keeping the rest of the algorithm unchanged.

\begin{figure}[H]
    \centering
    \begin{subfigure}[b]{0.6\textwidth}
        \includegraphics[width=\textwidth]{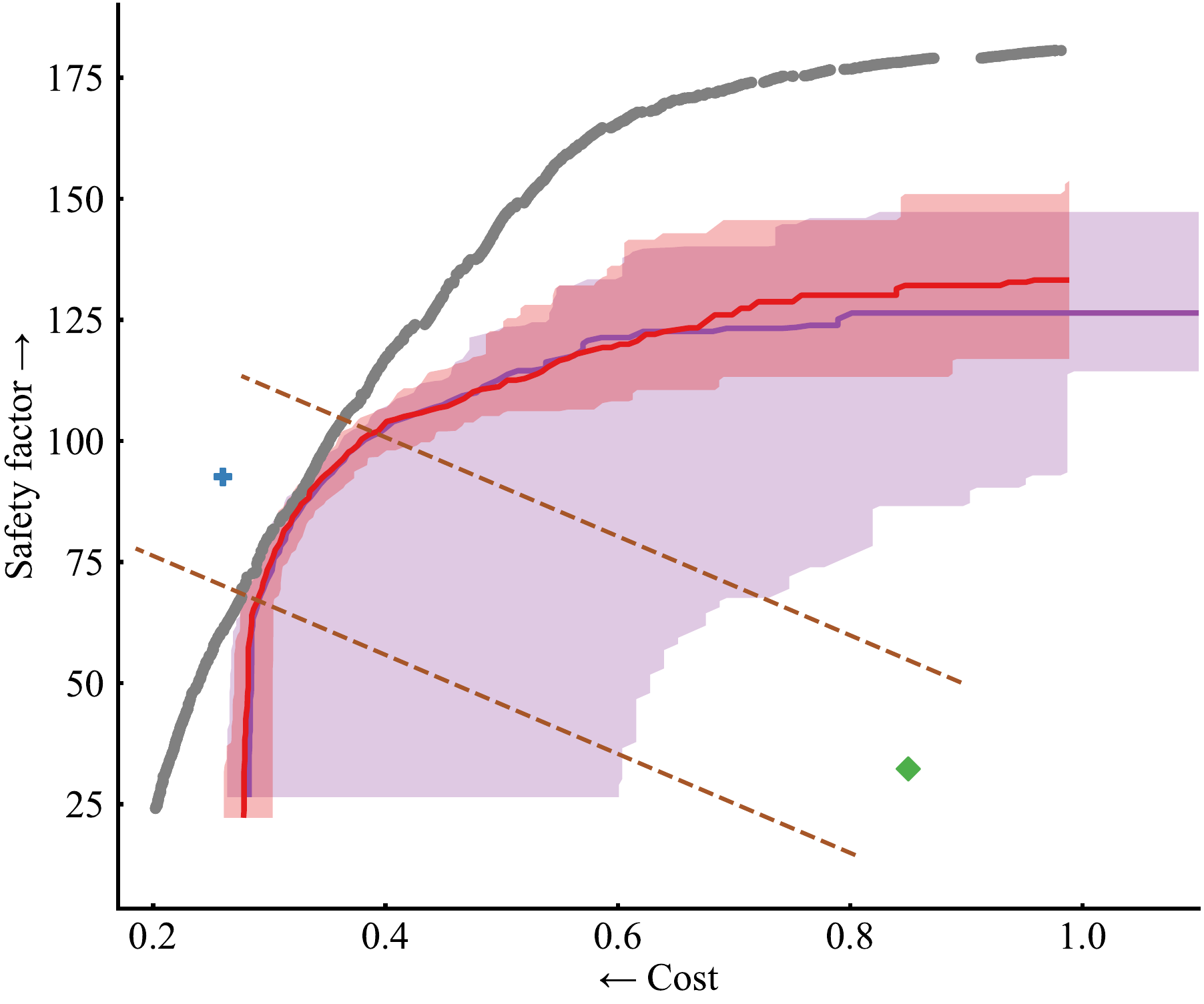}
        \begin{picture}(0,0)
        \end{picture}

    \end{subfigure}
    \vspace*{-1em} %
    {
    \begin{subfigure}[b]{0.8\textwidth}
        \includegraphics[width=\textwidth]{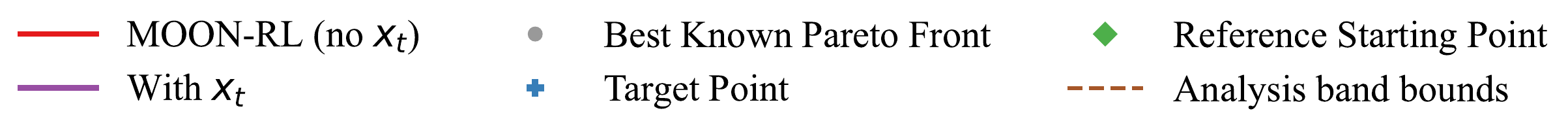}
    \end{subfigure}
    }
    \vspace*{1em} %

    \caption{Effect of including the design vector $\boldsymbol{x_t}$ in the observation on the ``CS1'' benchmark.}
    \label{fig:ablation-including_x_t}
\end{figure}

The comparison in Figure~\ref{fig:ablation-including_x_t} supports the observation choice used in the main paper. Since MOON-RL is formulated as an objective-space navigation method, the observation focuses on quantities directly related to the current objective-space position, the reference direction, and feasibility. Including $\boldsymbol{x_t}$ increases the observation dimension, but in this ablation it does not lead to improved navigation performance.

\clearpage

\clearpage
\newpage

\section{Effect of the Constraint Penalty Schedule}
\label{app:lambda_magnitude_ablation}

To assess the influence of the constraint-penalty schedule, we repeated the experiment with different initial-to-final ranges of the coefficient $\lambda$, while keeping the scheduling rate fixed. The tested schedules are $\lambda: 0 \to 4$, $\lambda: 1 \to 5$, $\lambda: 2 \to 6$, $\lambda: 4 \to 8$, and $\lambda: 6 \to 10$, where $\lambda: 2 \to 6$ corresponds to the setting used in the main experiments.

\begin{figure}[H]
    \centering
    \begin{subfigure}[b]{0.6\textwidth}
        \includegraphics[width=\textwidth]{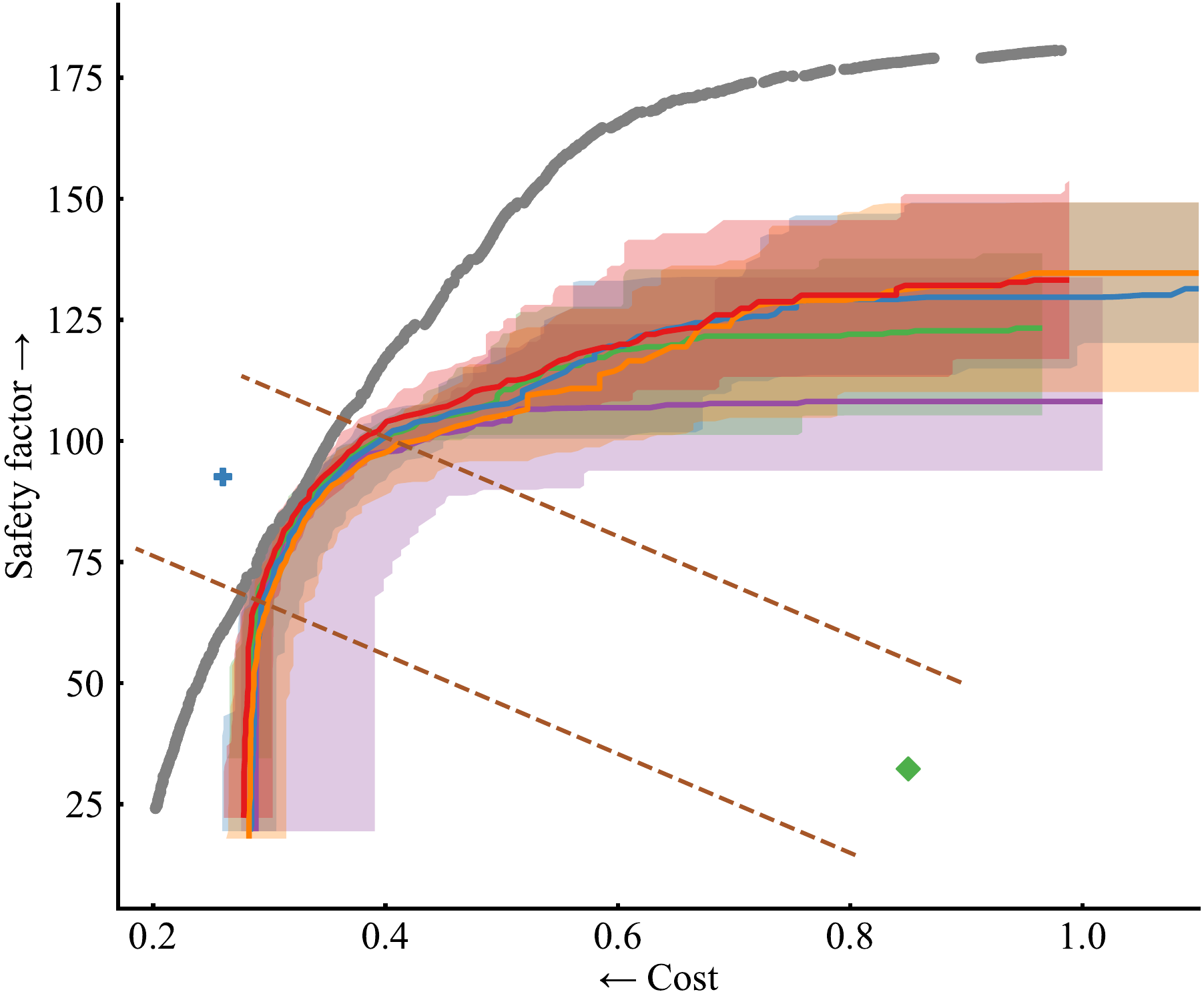}
        \begin{picture}(0,0)
        \end{picture}

    \end{subfigure}
    \vspace*{-1em} %
    {
    \begin{subfigure}[b]{0.8\textwidth}
        \includegraphics[width=\textwidth]{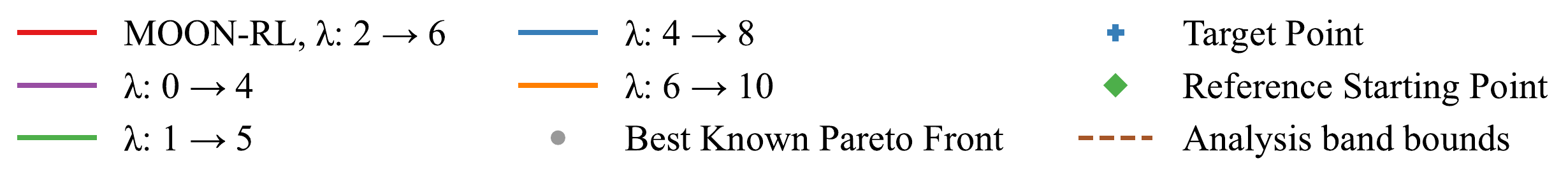}
    \end{subfigure}
    }
    \vspace*{1em} %

    \caption{Effect of the constraint penalty magnitude.}
    \label{fig:ablation-lambda_magnitude}
\end{figure}

Figure~\ref{fig:ablation-lambda_magnitude} shows that the empirical attainment functions remain broadly similar across the tested schedules. The percentile bands largely overlap, and no schedule produces a clearly distinct attained region on CS1. Within the tested $\lambda$ range on CS1, these results do not indicate strong sensitivity to the chosen magnitude range of the constraint-penalty schedule.

\clearpage

\clearpage
\newpage

\section{PPO learning diagnostics}
\label{app:ppo-diagnostics}

The diagnostic runs are performed in an isolated PPO setting, without the MOON-RL meta-algorithm active. In particular, the starting point is not reset during the run, and the policy is trained for a long uninterrupted sequence of episodes from different initial starting points. This setup is intentionally different from the full MOON-RL procedure, where in the default configuration the starting point is reset every 50 episodes. The purpose of this diagnostic experiment is to isolate the PPO learning dynamics, rather than to evaluate the full archive-management and reset mechanism.

\begin{figure}[h]
    \centering
    \begin{subfigure}[b]{0.49\textwidth}
        \includegraphics[width=\textwidth]{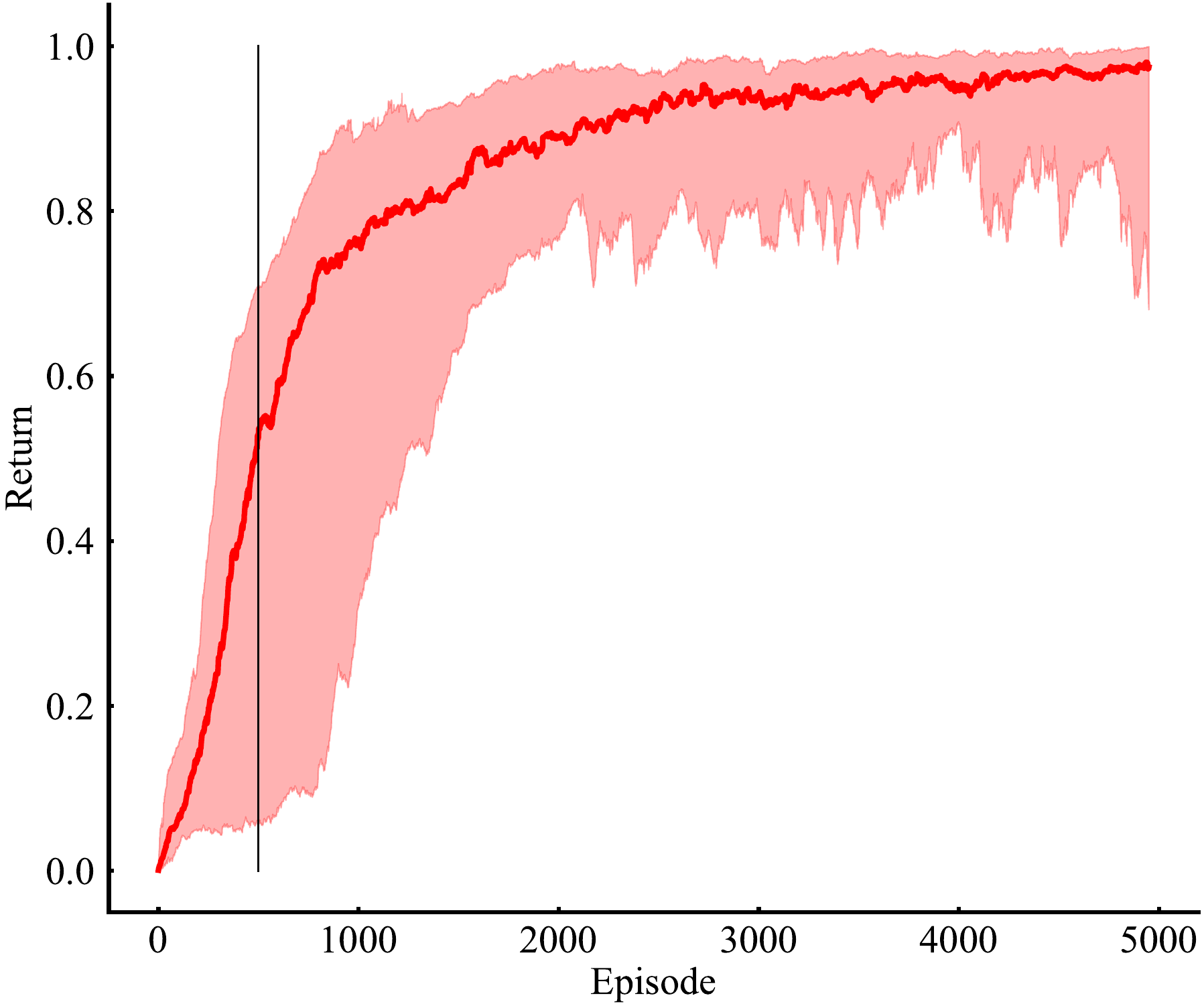}
        \begin{picture}(0,0)
            \put(0,195){a.} %
        \end{picture}

    \end{subfigure}
    \hfill
    \begin{subfigure}[b]{0.49\textwidth}
        \includegraphics[width=\textwidth]{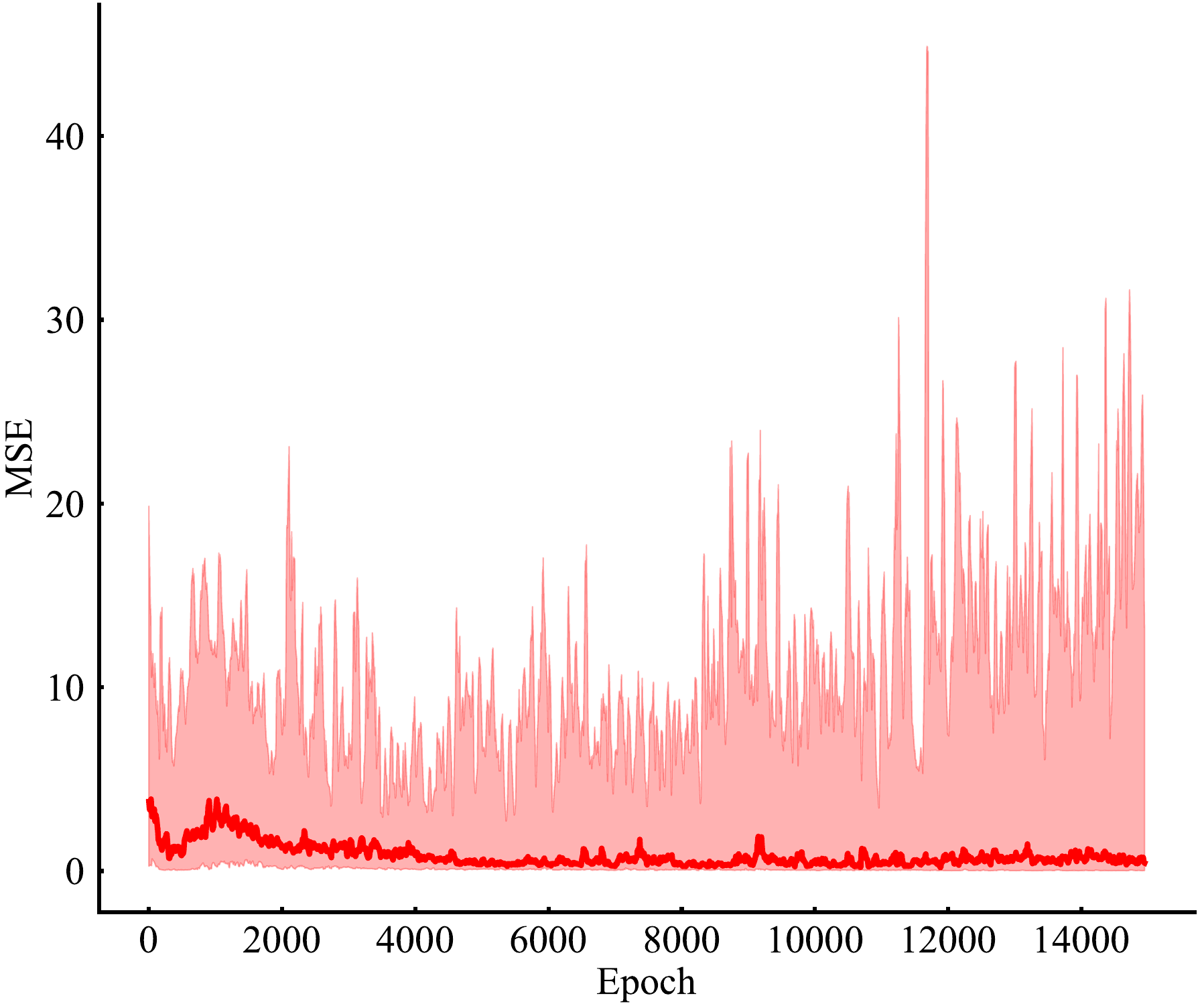}
        \begin{picture}(0,0)
            \put(0,195){b.} %
        \end{picture}
    \end{subfigure}

    \centering

    \caption{Rolling averages with a window size of 50 for min-max-normalized episodic return (a) and critic value loss (b) in isolated PPO runs without the MOON-RL meta-algorithm active. The critic value loss is computed as the mean-squared error between predicted values and return targets. The vertical dashed line indicates 500 individual trajectories, equivalent to 50 episodes and therefore to the starting-point reset interval used in the default MOON-RL configuration.}
    \label{fig:return}
\end{figure}

Figure \ref{fig:return} shows the rolling average, with a window size of 50, of the episodic return and critic value loss. For visualization, the plotted episodic returns are min-max normalized. The episodic return increases and then approaches a plateau, indicating that the actor learns to select actions that obtain higher reward under the proposed objective-space navigation reward. The critic value loss is computed as the mean-squared error between the critic prediction and the return target. This curve is noisier and should not be interpreted independently from the return curve. The critic loss measures value-function fitting error, whereas the return curve indicates whether the actor improves its reward-seeking behavior.

The noisiness of the critic loss is expected in this setting. PPO is updated online, so the sampled trajectories and return targets change as the policy changes. In addition, as the actor reaches higher-reward trajectories, the magnitude of the return targets can increase, which can increase the scale of the squared value-prediction error. This behavior is also related to the PPO settings used in MOON-RL, which were chosen to favor fast local adaptation from limited rollout data rather than slow conservative value-function convergence.

In the full MOON-RL procedure, the starting point is reset before this plateau is typically reached. Under a fixed evaluation budget, this is beneficial because frequent resets expose the policy to new local search regions, whereas longer uninterrupted training around the same starting point reduces overall performance, as shown by the reset-interval ablation in Section \ref{subsec:ablations} and Appendix \ref{app:ablation-additional}

\clearpage

\clearpage
\section{Additional DAS-CMOP Experiments}
\label{app:additional_dascmop}

We report additional EAF comparisons on DAS-CMOP1 and DAS-CMOP3 under four difficulty settings: 9, 2, 11, and 4. These settings correspond to high diversity difficulty, low feasibility difficulty, high convergence difficulty, and low combined difficulty, respectively. The aim of these additional experiments is to broaden the DAS-CMOP evaluation beyond the single DAS-CMOP3-11 case reported in the main text.

The feasibility-related settings are kept at the lowest feasibility level because the current MOON-RL implementation requires an initial feasible design. As discussed in the limitations of the method, finding such an initial feasible point by random sampling can become difficult when the feasible region is very small. The additional cases therefore test different DAS-CMOP difficulty types while remaining compatible with the current feasible-initialization procedure.

\begin{figure}[h]
    \centering
    \begin{subfigure}[b]{0.6\textwidth}
        \includegraphics[width=\textwidth]{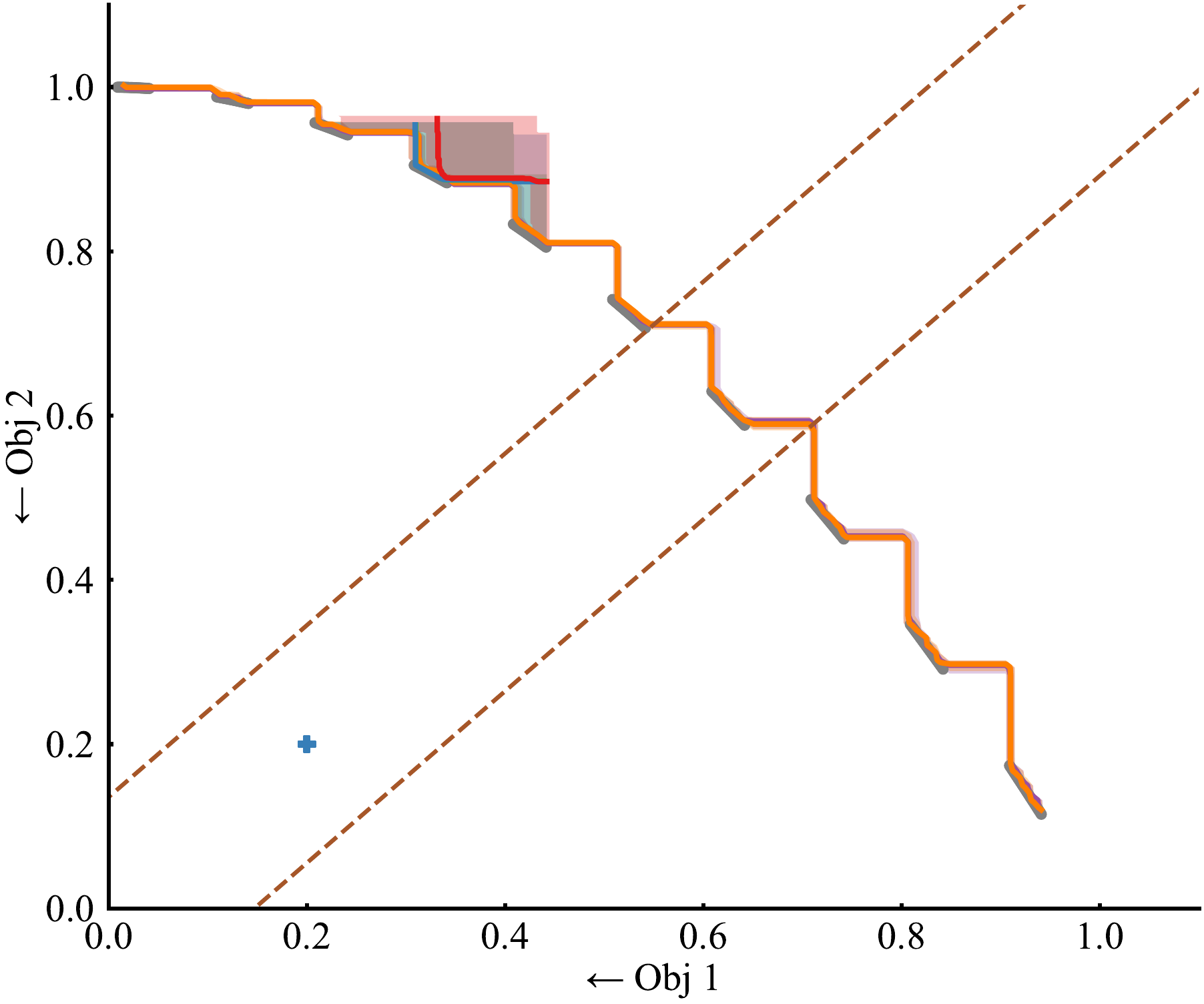}
        \begin{picture}(0,0)
        \end{picture}

    \end{subfigure}
    \vspace*{-1em} %
    {
    \begin{subfigure}[b]{0.8\textwidth}
        \includegraphics[width=\textwidth]{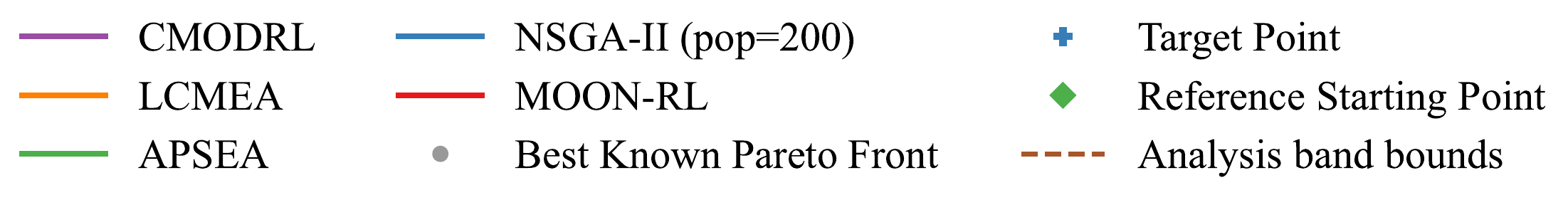}
    \end{subfigure}
    }
    \vspace*{1em} %

    \caption{EAF comparison on DAS-CMOP1-9, corresponding to high diversity difficulty.}
    \label{fig:das_1_9}
\end{figure}

\begin{figure}[h]
    \centering
    \begin{subfigure}[b]{0.6\textwidth}
        \includegraphics[width=\textwidth]{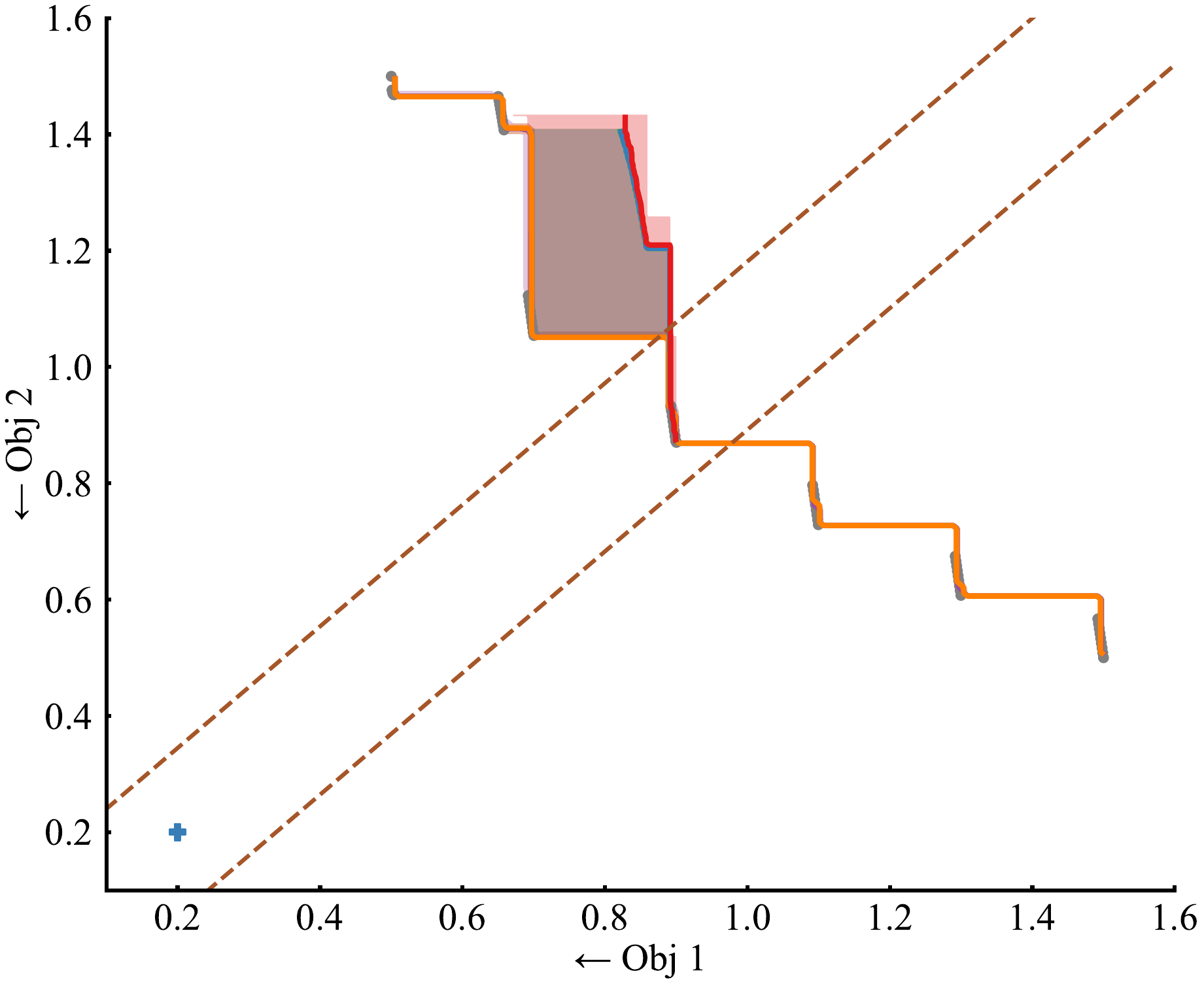}
        \begin{picture}(0,0)
        \end{picture}

    \end{subfigure}
    \vspace*{-1em} %
    {
    \begin{subfigure}[b]{0.8\textwidth}
        \includegraphics[width=\textwidth]{figures/appendix/additional_dascmop/legend.pdf}
    \end{subfigure}
    }
    \vspace*{1em} %

    \caption{EAF comparison on DAS-CMOP3-4, corresponding to low combined difficulty.}
    \label{fig:das_3_4}
\end{figure}

Figures~\ref{fig:das_1_9}--\ref{fig:das_3_4} show the resulting EAFs. To keep the appendix compact, MOON-RL is compared against a representative subset of baselines: APSEA, CMODRL, LCMEA, and NSGA-II. This subset includes recent constrained MOEAs together with a classical evolutionary baseline. The dashed lines indicate the bounds of the analysis band used to evaluate the reference-directed region.

The results show that MOON-RL reaches the prescribed analysis band on the additional DAS-CMOP cases. However, the strongest constrained MOEAs, especially LCMEA and often NSGA-II, generally achieve better convergence and broader coverage of the best-known Pareto front. This gap is most visible on DAS-CMOP1-11 and several DAS-CMOP3 settings. On DAS-CMOP1-9 and DAS-CMOP3-9, MOON-RL is closer to the compared baselines in the targeted region.

These results are consistent with the intended scope of MOON-RL. The method is not designed as a universal replacement for specialized constrained MOEAs for global Pareto-front approximation. Instead, it provides a sequential mechanism for objective-space navigation along a prescribed reference direction. The additional DAS-CMOP experiments therefore provide a broader empirical view of where this formulation remains competitive and where mature constrained MOEAs remain stronger.

\begin{figure}[h]
    \centering
    \begin{subfigure}[b]{0.6\textwidth}
        \includegraphics[width=\textwidth]{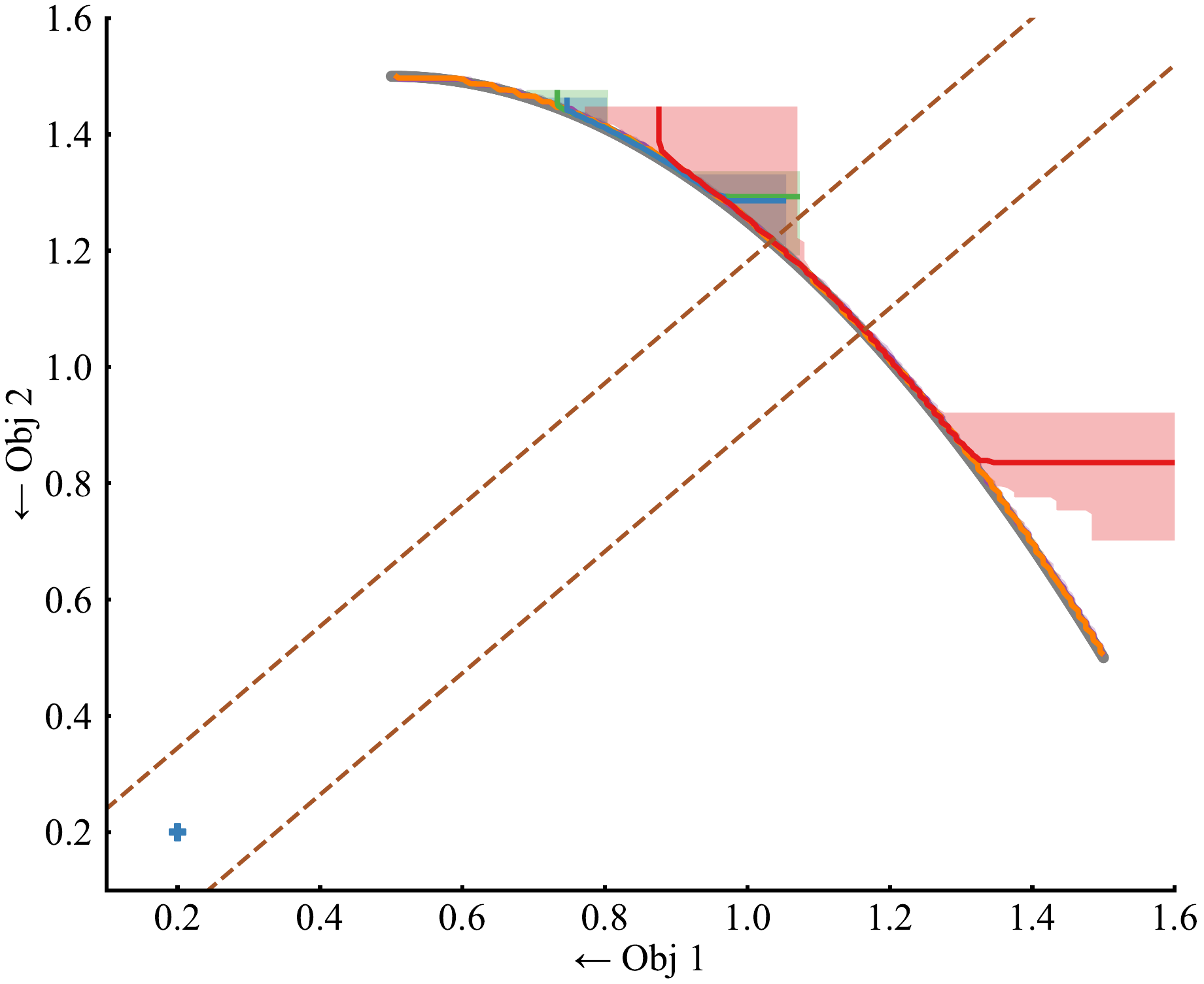}
        \begin{picture}(0,0)
        \end{picture}

    \end{subfigure}
    \vspace*{-1em} %
    {
    \begin{subfigure}[b]{0.8\textwidth}
        \includegraphics[width=\textwidth]{figures/appendix/additional_dascmop/legend.pdf}
    \end{subfigure}
    }
    \vspace*{1em} %

    \caption{EAF comparison on DAS-CMOP1-2, corresponding to low feasibility difficulty.}
\end{figure}
\begin{figure}[h]
    \centering
    \begin{subfigure}[b]{0.6\textwidth}
        \includegraphics[width=\textwidth]{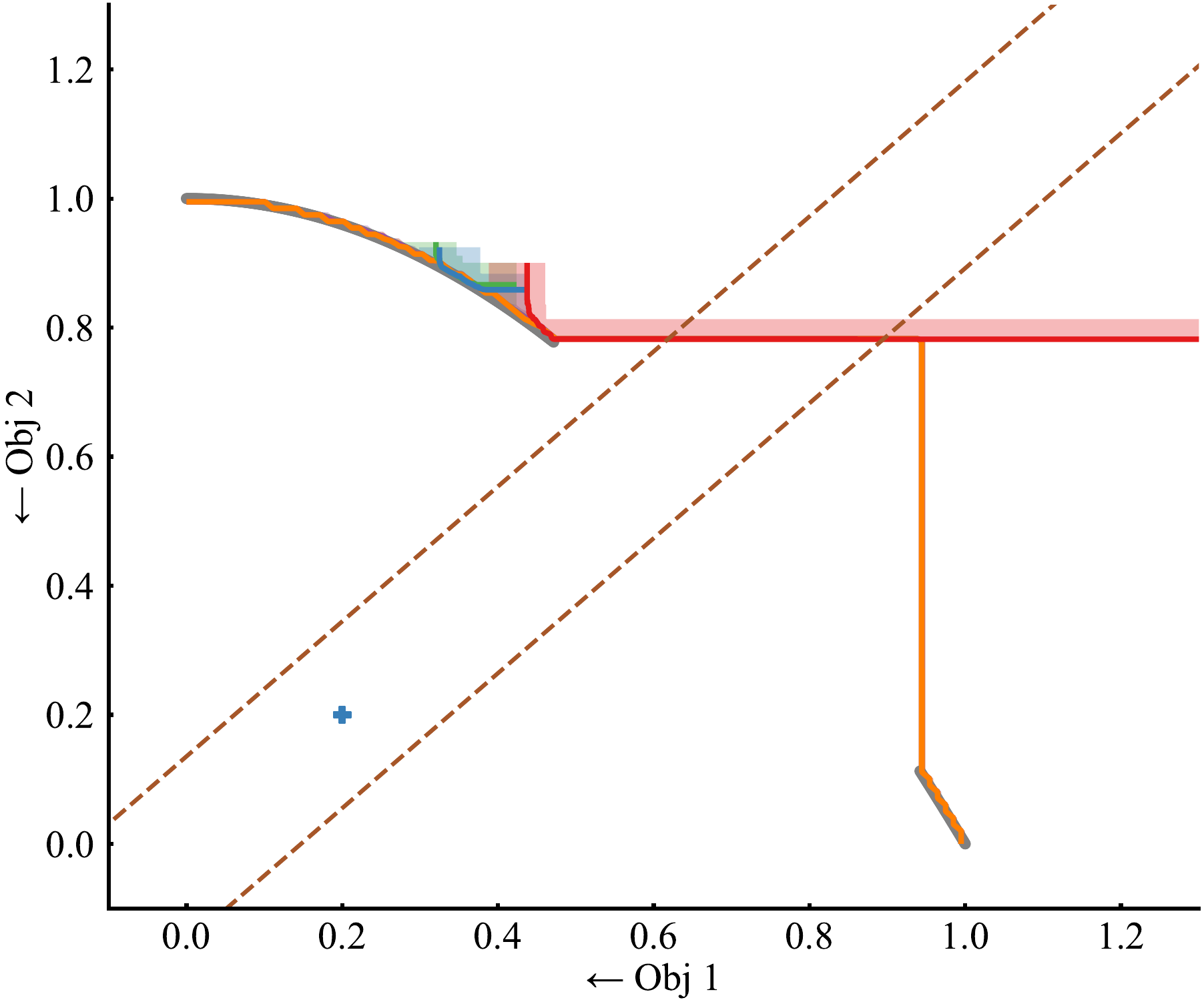}
        \begin{picture}(0,0)
        \end{picture}

    \end{subfigure}
    \vspace*{-1em} %
    {
    \begin{subfigure}[b]{0.8\textwidth}
        \includegraphics[width=\textwidth]{figures/appendix/additional_dascmop/legend.pdf}
    \end{subfigure}
    }
    \vspace*{1em} %

    \caption{EAF comparison on DAS-CMOP1-11, corresponding to high convergence difficulty.}
\end{figure}
\begin{figure}[h]
    \centering
    \begin{subfigure}[b]{0.6\textwidth}
        \includegraphics[width=\textwidth]{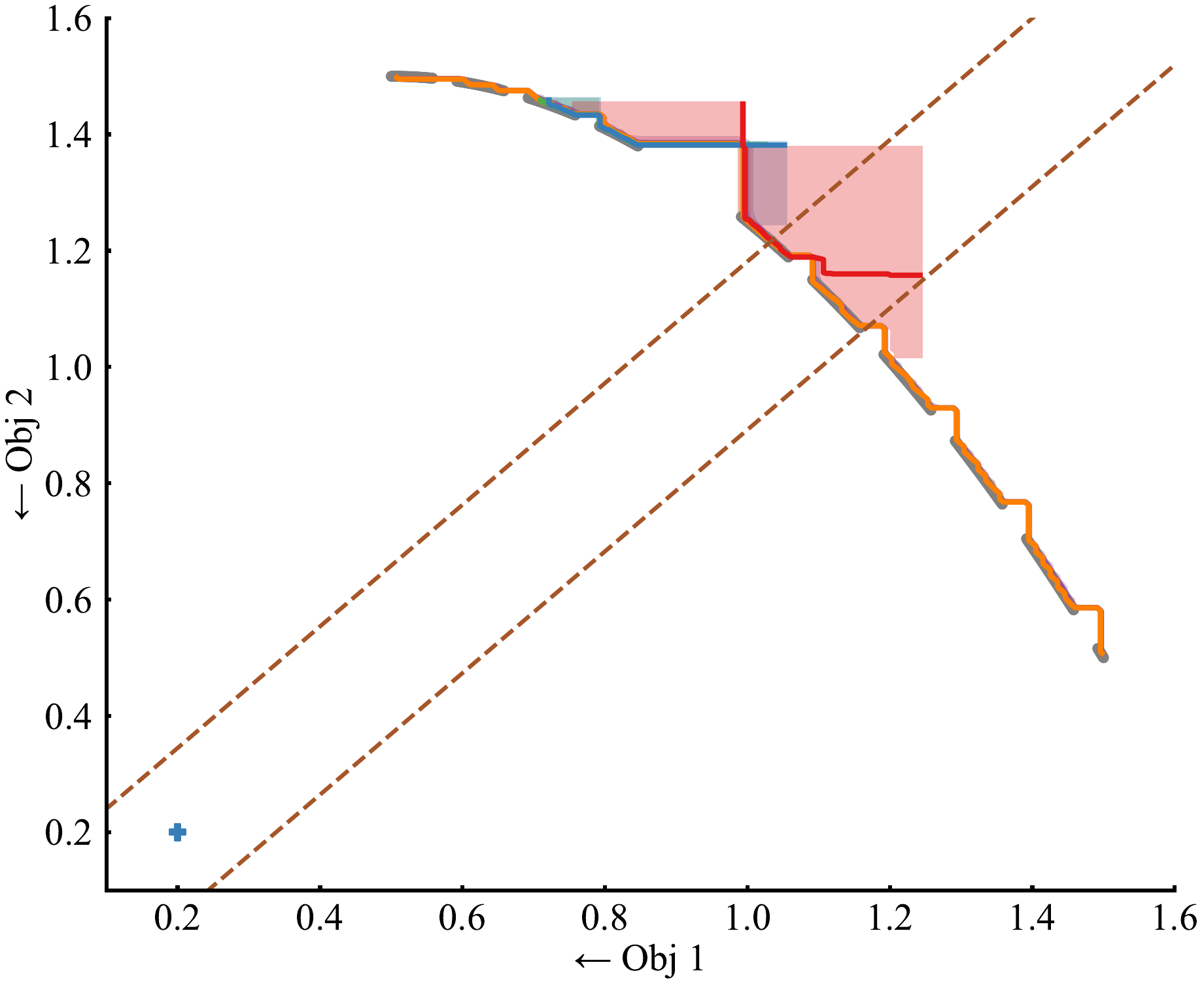}
        \begin{picture}(0,0)
        \end{picture}

    \end{subfigure}
    \vspace*{-1em} %
    {
    \begin{subfigure}[b]{0.8\textwidth}
        \includegraphics[width=\textwidth]{figures/appendix/additional_dascmop/legend.pdf}
    \end{subfigure}
    }
    \vspace*{1em} %

    \caption{EAF comparison on DAS-CMOP1-4, corresponding to low combined difficulty.}
\end{figure}

\begin{figure}[h]
    \centering
    \begin{subfigure}[b]{0.6\textwidth}
        \includegraphics[width=\textwidth]{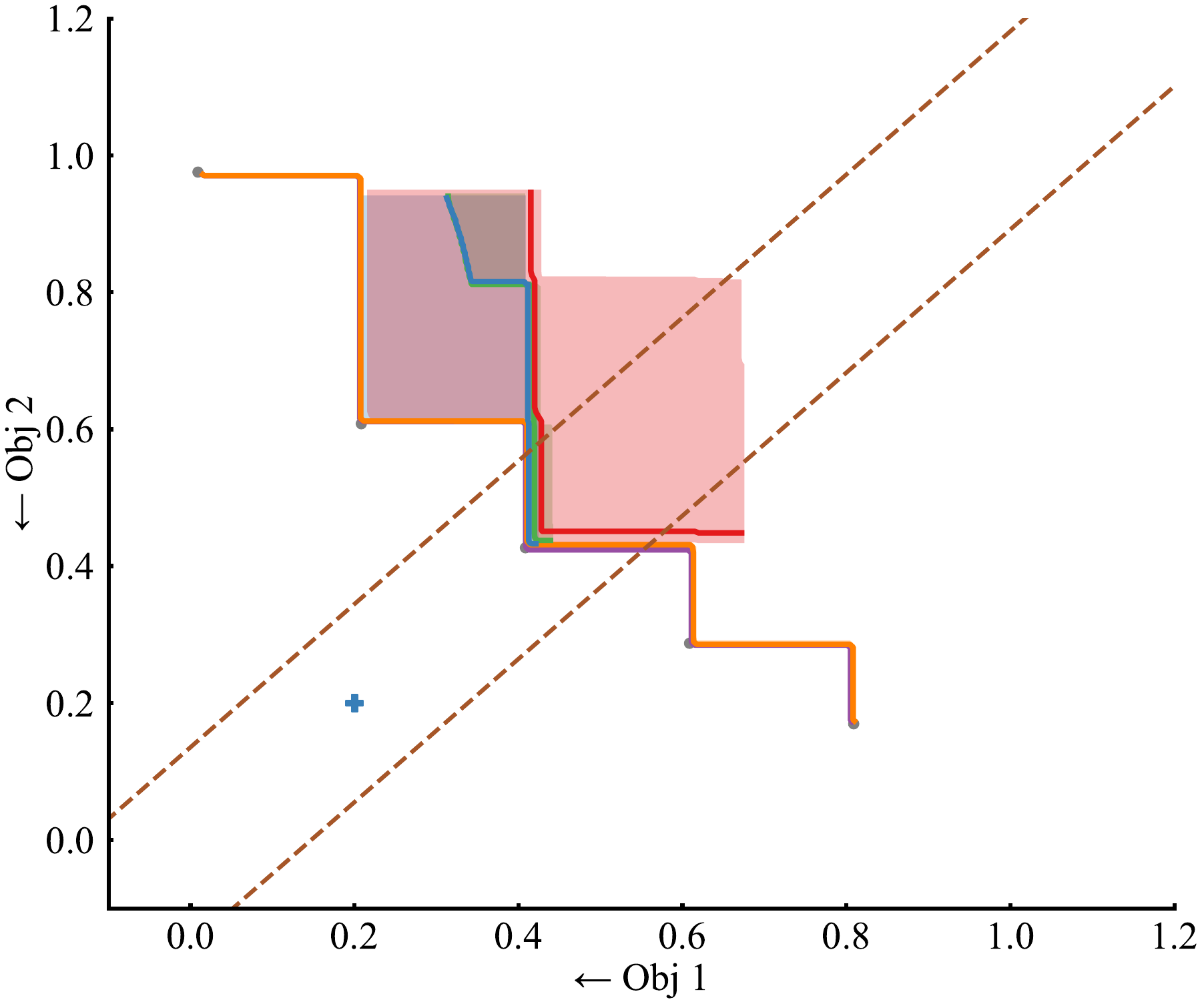}
        \begin{picture}(0,0)
        \end{picture}

    \end{subfigure}
    \vspace*{-1em} %
    {
    \begin{subfigure}[b]{0.8\textwidth}
        \includegraphics[width=\textwidth]{figures/appendix/additional_dascmop/legend.pdf}
    \end{subfigure}
    }
    \vspace*{1em} %

    \caption{EAF comparison on DAS-CMOP3-9, corresponding to high diversity difficulty.}
\end{figure}
\begin{figure}[h]
    \centering
    \begin{subfigure}[b]{0.6\textwidth}
        \includegraphics[width=\textwidth]{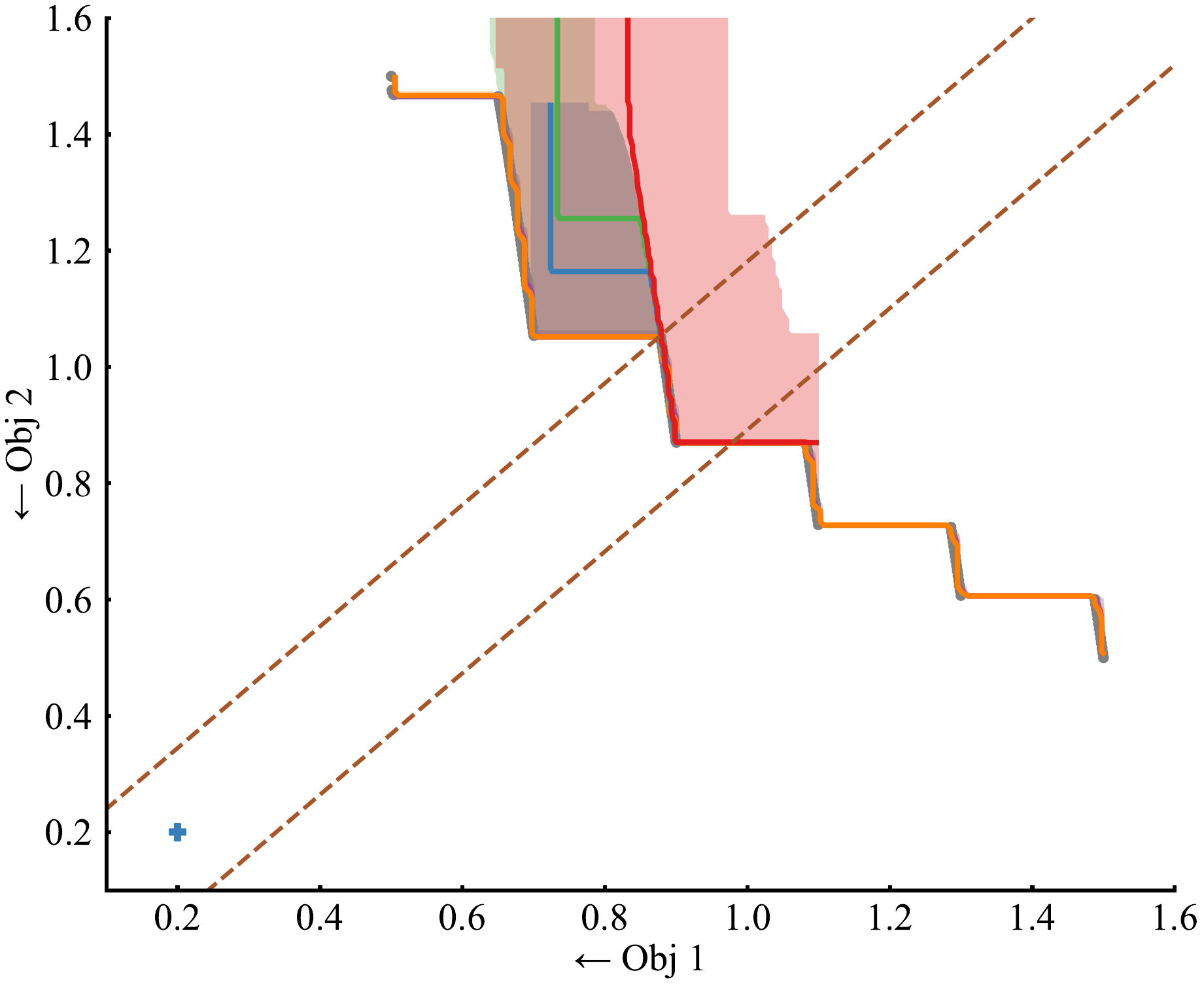}
        \begin{picture}(0,0)
        \end{picture}

    \end{subfigure}
    \vspace*{-1em} %
    {
    \begin{subfigure}[b]{0.8\textwidth}
        \includegraphics[width=\textwidth]{figures/appendix/additional_dascmop/legend.pdf}
    \end{subfigure}
    }
    \vspace*{1em} %

    \caption{EAF comparison on DAS-CMOP3-2, corresponding to low feasibility difficulty.}
\end{figure}

\clearpage

\clearpage
\newpage

\section{Effect of policy training and meta-algorithm on additional benchmarks}
\label{app:additional_meta_no_meta}

\begin{figure}[h]
    \centering
    \begin{subfigure}[b]{0.49\textwidth}
        \includegraphics[width=\textwidth]{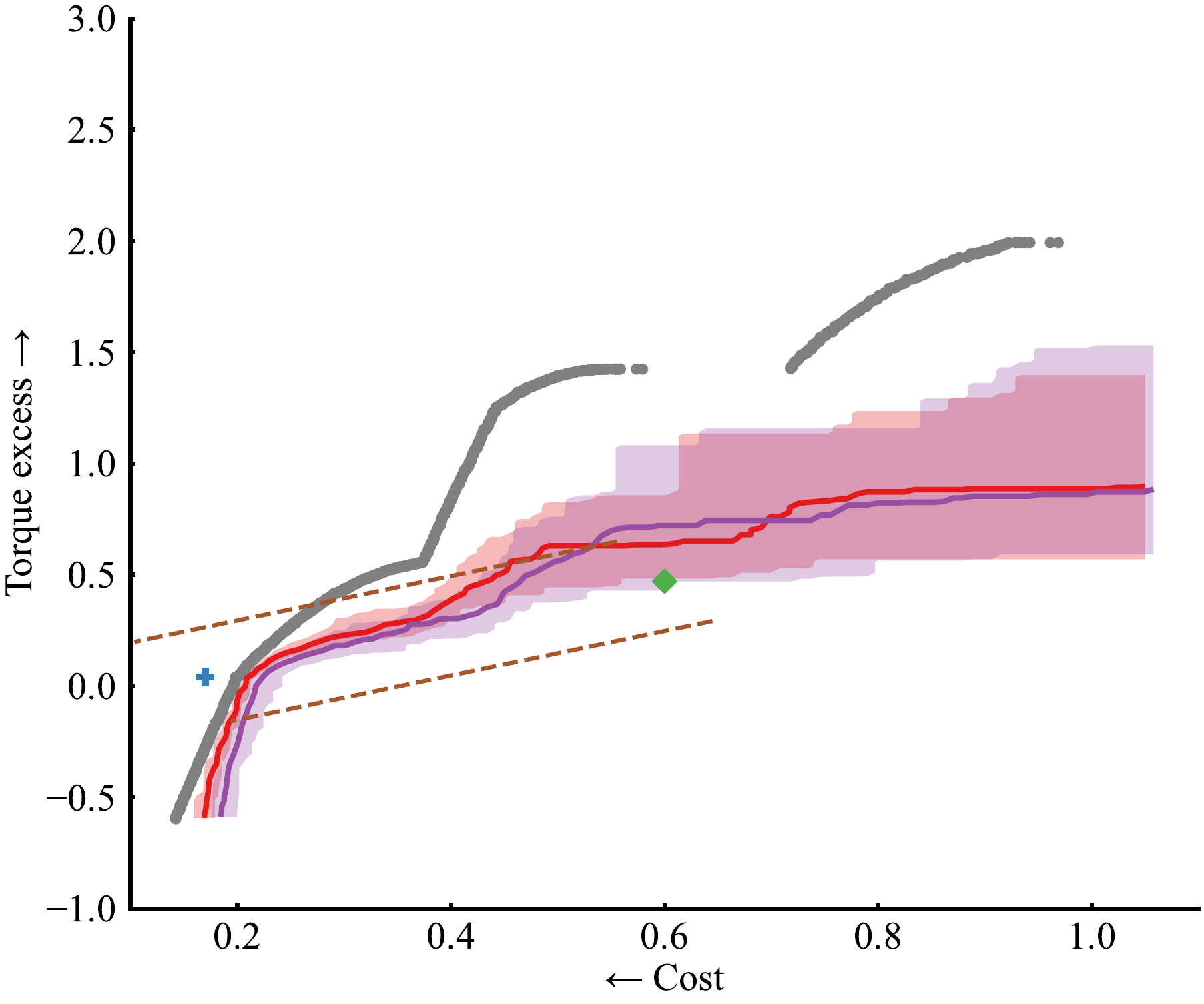}
        \begin{picture}(0,0)
            \put(0,195){a.} %
        \end{picture}

    \end{subfigure}
    \hfill
    \begin{subfigure}[b]{0.49\textwidth}
        \includegraphics[width=\textwidth]{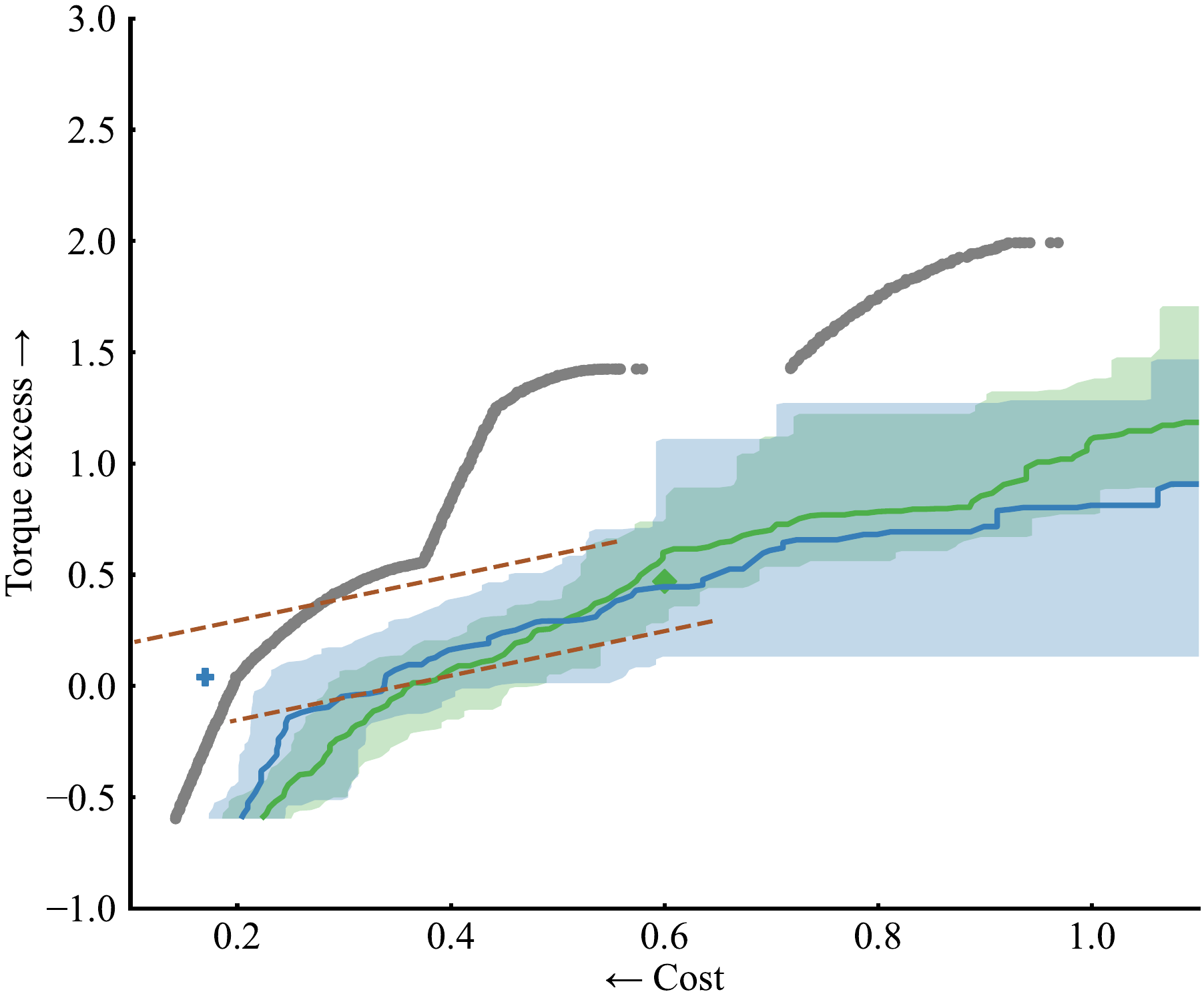}
        \begin{picture}(0,0)
            \put(0,195){b.} %
        \end{picture}
    \end{subfigure}

    \centering
    \vspace*{-1em} %
    {
    \begin{subfigure}[b]{0.6\textwidth}
        \includegraphics[width=\textwidth]{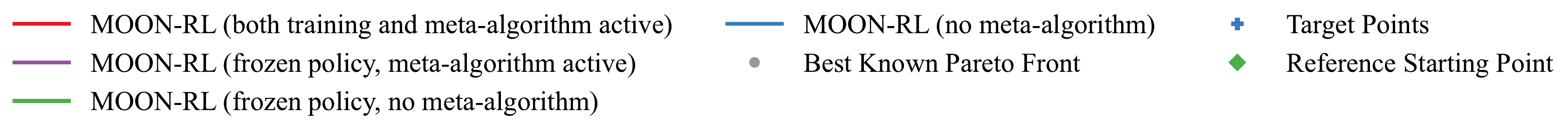}
    \end{subfigure}
    }

    \caption{EAF comparison of MOON-RL variants with and without the meta-algorithm and policy-network training. Panel (a) shows runs with the meta-algorithm enabled, and panel (b) shows runs without the meta-algorithm. In each panel, EAFs compare the effect of enabling or disabling policy-network training. ''CT1`` problem.}
    \label{fig:ct1_meta_no_meta}

\end{figure}

\begin{figure}[h]
    \centering
    \begin{subfigure}[b]{0.49\textwidth}
        \includegraphics[width=\textwidth]{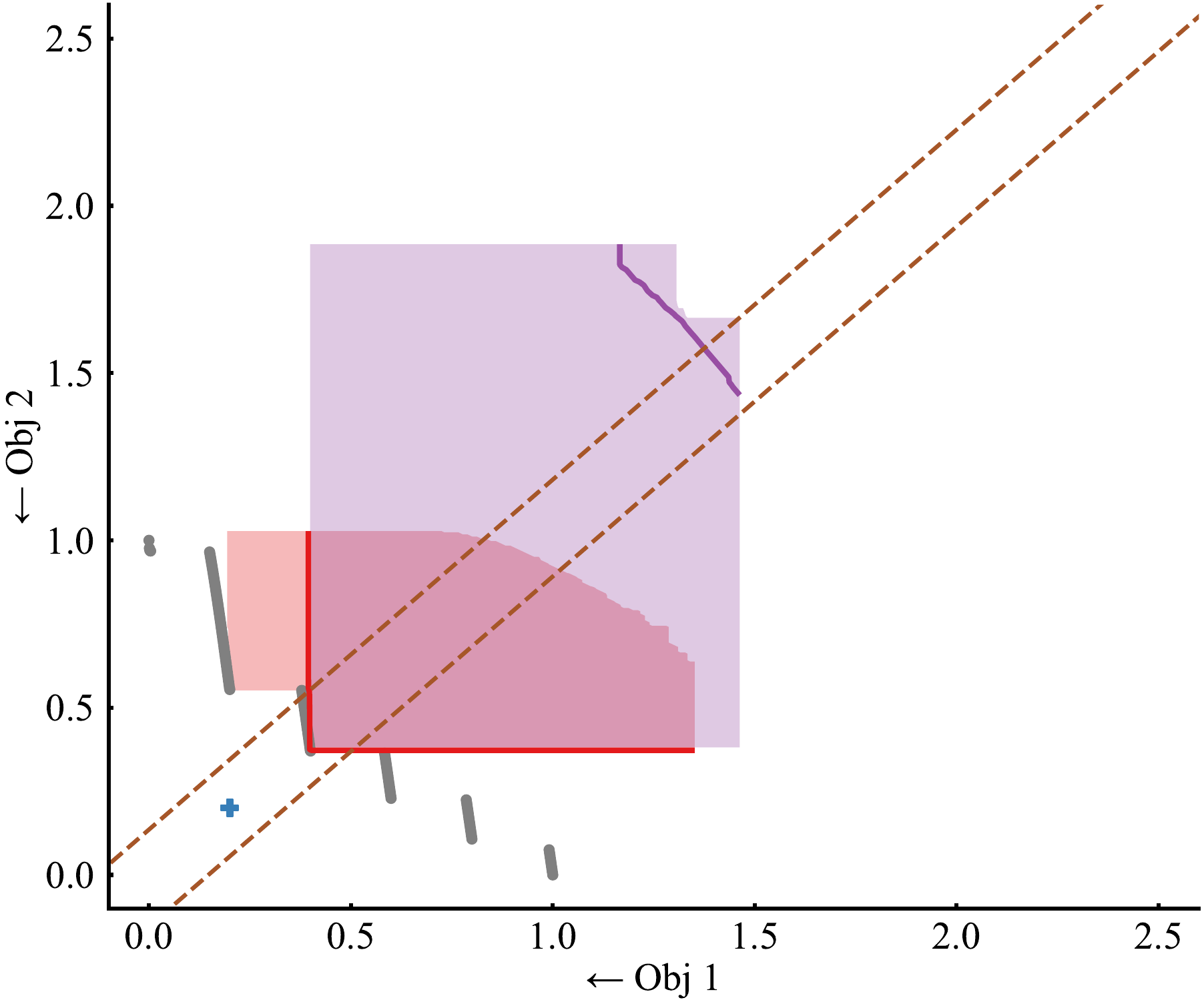}
        \begin{picture}(0,0)
            \put(0,195){a.} %
        \end{picture}

    \end{subfigure}
    \hfill
    \begin{subfigure}[b]{0.49\textwidth}
        \includegraphics[width=\textwidth]{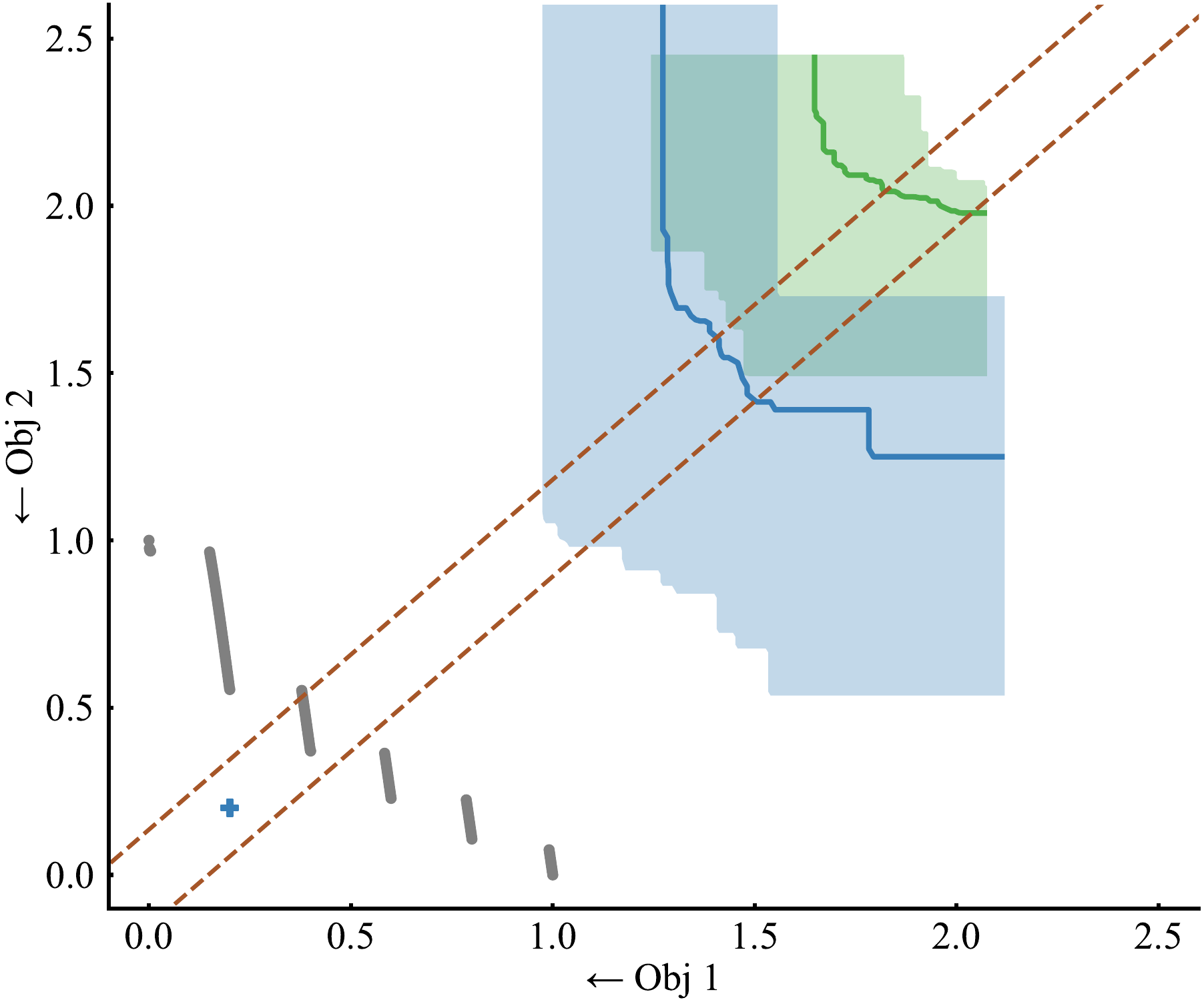}
        \begin{picture}(0,0)
            \put(0,195){b.} %
        \end{picture}
    \end{subfigure}

    \centering
    \vspace*{-1em} %
    {
    \begin{subfigure}[b]{0.6\textwidth}
        \includegraphics[width=\textwidth]{figures/appendix/additional_meta_algorithm_ablation/legend.pdf}
    \end{subfigure}
    }

    \caption{EAF comparison of MOON-RL variants with and without the meta-algorithm and policy-network training. Panel (a) shows runs with the meta-algorithm enabled, and panel (b) shows runs without the meta-algorithm. In each panel, EAFs compare the effect of enabling or disabling policy-network training. DAS-CMOP3-11 problem}
    \label{fig:dascmop3_11_meta_no_meta}

\end{figure}

\begin{figure}[h]
    \centering
    \begin{subfigure}[b]{0.49\textwidth}
        \includegraphics[width=\textwidth]{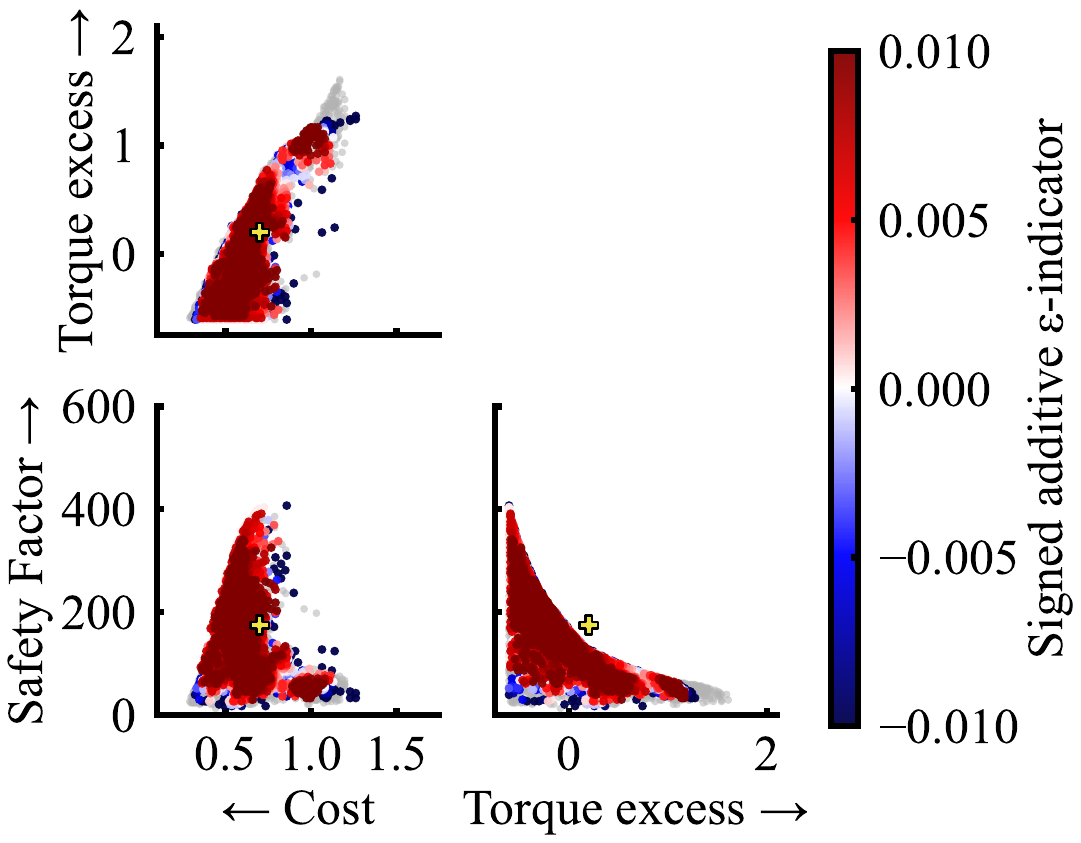}
        \begin{picture}(0,0)
            \put(0,195){a.} %
        \end{picture}

    \end{subfigure}
    \hfill
    \begin{subfigure}[b]{0.49\textwidth}
        \includegraphics[width=\textwidth]{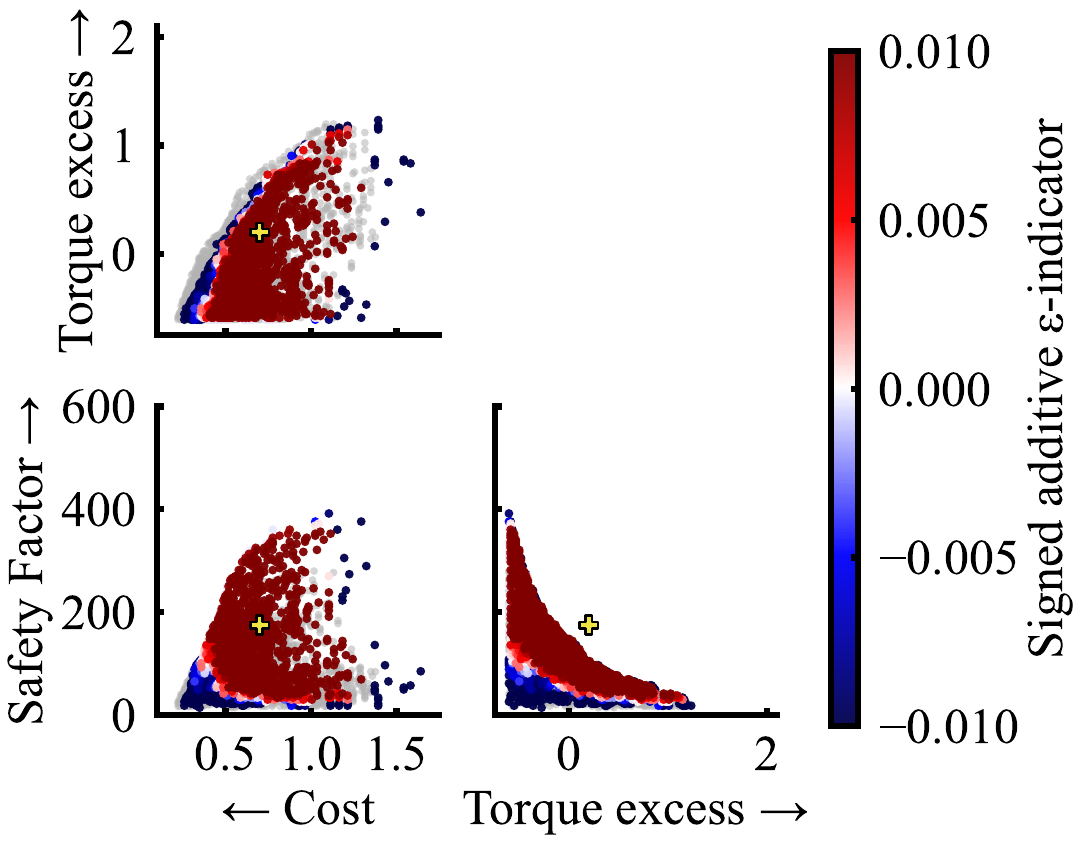}
        \begin{picture}(0,0)
            \put(0,195){b.} %
        \end{picture}
    \end{subfigure}

    \centering

    \caption{Pairwise projections of the median EAF for the CTS1 problem. Each panel compares MOON-RL with policy training enabled against MOON-RL with a frozen policy: (a) with the meta-algorithm enabled and (b) with the meta-algorithm disabled. ''CTS1`` problem.}
    \label{fig:cts1_meta_no_meta}

\end{figure}

To verify that the ablation results in Section \ref{subsec:ablations} are not specific to CS1, we repeat the same comparison on CT1, CTS1, and DAS-CMOP. We compare four variants: full MOON-RL, MOON-RL with a frozen policy, MOON-RL without the meta-algorithm, and MOON-RL with both the policy frozen and the meta-algorithm disabled. Figures~\ref{fig:ct1_meta_no_meta}--\ref{fig:cts1_meta_no_meta} show the corresponding EAF comparisons for CT1, DAS-CMOP3-11, and CTS1, respectively.

The results confirm the interpretation from Section \ref{subsec:ablations}. The frozen-policy variants still benefit from the surrounding MOON-RL infrastructure, including local design perturbations, scaling schedules, the reference direction, the feasible non-dominated archive, and archive-based restarts. However, across all tested problems, the variants with RL training enabled outperform the corresponding frozen-policy variants. The effect of RL training is most visible on DAS-CMOP, where the trained-policy variant reaches a clearly better attained region than the frozen-policy variant. This supports the view that the meta-algorithm and policy training play complementary roles: the meta-algorithm manages the search through restarts and scheduled parameters, while RL training improves the local search behavior between restarts.


\clearpage

\clearpage
\newpage

\section{Cross-Entropy (CEM) sampler implementation details}
\label{app:cem_implementation}

The cross-entropy sampler \cite{de2005tutorial} maintains a diagonal Gaussian distribution over design-variable
perturbations, with mean $\boldsymbol{\mu}$ and per-coordinate standard deviation $\boldsymbol{\sigma}$, both of
dimension $n$ equal to the number of design variables. The distribution does not depend on the
objective-space observation.

Two configurations of the cross-entropy sampler are reported. They differ in whether the last step of Algorithm \ref{alg:sampler_rollout} is applied, and in the elite count and variance estimator used in Algorithm \ref{alg:cem_update}. When it is, every design is advanced to
the highest-rewarded candidate of the step, so the $N$ concurrent evaluations act as candidates
around one moving design and an episode traces a single path through the design space. When it is
omitted, each design retains the candidate it produced, so the $N$ trajectories diverge over the
episode, as they do for the policy in Algorithm~\ref{algo:core-algo}. A single distribution
$\mathcal{D}$ is shared across the $N$ evaluations in both cases and is refitted once per step
from all $N$ perturbations.

\begin{algorithm}
\caption{Episode rollout for the non-learning samplers}
\label{alg:sampler_rollout}
\begin{algorithmic}[1]
\Require starting design $\boldsymbol{x}_0$, scaling factors $\boldsymbol{\kappa}$, bounds $\boldsymbol{\ell},\boldsymbol{u}$,
         episode length $T$, concurrent evaluations $N$, sampling distribution $\mathcal{D}$
\State $\boldsymbol{x}_i \gets \boldsymbol{x}_0$ \textbf{ for } $i = 1,\dots,N$
\For{$t = 1$ \textbf{to} $T$}
  \State draw $\boldsymbol{a}_1,\dots,\boldsymbol{a}_N \sim \mathcal{D}$
  \For{$i = 1$ \textbf{to} $N$}
    \State $\boldsymbol{m}_i \gets \boldsymbol{\kappa} \odot \boldsymbol{a}_i$; round the discrete components of $\boldsymbol{m}_i$
    \State $\boldsymbol{x}_i \gets \mathrm{clip}(\boldsymbol{x}_i + \boldsymbol{m}_i, \boldsymbol{\ell}, \boldsymbol{u})$
    \State evaluate $\boldsymbol{x}_i$ and compute $r_i$ \Comment{reward Eq.~\ref{eq:reward}}
    \State update archive $\mathcal{P}$ with $\boldsymbol{x}_i$ if feasible and non-dominated
  \EndFor
  \State update $\mathcal{D}$ from $\{(\boldsymbol{a}_i, r_i)\}_{i=1}^{N}$
        \Comment{Algorithm~\ref{alg:cem_update} for the cross-entropy sampler}
  \State $\boldsymbol{x}_i \gets \boldsymbol{x}_{\arg\max_j r_j}$ \textbf{ for all } $i$
        \Comment{omitted in the independent-trajectory configuration}
\EndFor
\end{algorithmic}
\end{algorithm}

\begin{algorithm}
\caption{Cross-entropy distribution update}
\label{alg:cem_update}
\begin{algorithmic}[1]
\Require perturbations $\boldsymbol{a}_1,\dots,\boldsymbol{a}_N$ with rewards $r_1,\dots,r_N$,
         current $\boldsymbol{\mu},\boldsymbol{\sigma}$, elite count $k$, smoothing $\alpha$, floor $\sigma_{\min}$
\State $\mathcal{E} \gets$ indices of the $k$ largest $r_i$
\State $\boldsymbol{\mu} \gets \alpha\,\boldsymbol{\mu} + (1-\alpha)\,\dfrac{1}{k}\sum_{i \in \mathcal{E}} \boldsymbol{a}_i$
\State $\boldsymbol{s} \gets$ coordinate-wise standard deviation of $\{\boldsymbol{a}_i\}_{i \in \mathcal{E}}$
\State $\boldsymbol{\sigma} \gets \max\!\big(\alpha\,\boldsymbol{\sigma} + (1-\alpha)\,\boldsymbol{s},\; \sigma_{\min}\big)$
        \Comment{maximum taken element-wise}
\State \Return $\boldsymbol{\mu}, \boldsymbol{\sigma}$
\end{algorithmic}
\end{algorithm}

The elites are ranked by the full reward, including the constraint penalty, so the ranking matches
the signal available to the policy. The distribution persists across steps and episodes and is
reinitialized only when the meta-algorithm resets the agent, at the same episodes as the policy
reset. On initialization the mean is set to zero and the standard deviation to the width at which
a freshly initialized policy network samples, so all arms begin from the same sampling
distribution.

Both configurations use $\alpha = 0.7$ and $\sigma_{\min} = 0.1$, with $N = 10$. The
sequential-trajectory configuration uses $k = 3$ and computes $\boldsymbol{s}$ as the population standard
deviation; the independent-trajectory configuration uses $k = 5$ and the sample standard
deviation.

\clearpage

\clearpage
\newpage

\section{CMA-ES sampler implementation details}
\label{app:cma_implementation}

The CMA-ES sampler uses the reference implementation \texttt{pycma}, version 3.2.2
\cite{hansen2019pycma}. The strategy is defined over design-variable perturbations, with its mean
initialized at the origin so that a freshly initialized strategy proposes zero-mean perturbations,
and its initial step size set to the width at which a freshly initialized policy network samples.

The rollout follows Algorithm~\ref{alg:sampler_rollout}, with the strategy supplying the sampling
distribution $\mathcal{D}$: one generation corresponds to one step, and the population size is
fixed to $N = 10$, equal to the number of concurrent evaluations, so that a generation and a step
coincide and the evaluation budget matches the other arms. Perturbations are scaled, rounded and
clipped as in Algorithm~\ref{alg:sampler_rollout}, and the resulting designs are evaluated, rewarded
and archived through the same code path as every other arm. The rewards are negated before being
returned to the strategy, which minimizes. The strategy persists across steps and episodes and is
rebuilt when the meta-algorithm resets the agent. Its random seed is derived from the run seed,
since the library otherwise seeds itself non-deterministically.

CMA-ES is used here as a perturbation sampler inside the navigation loop, not as a standalone
optimizer, which entails three deviations from its standard use. First, the design from which
perturbations are applied is advanced by the rollout, to the highest-rewarded candidate of each
step, and not by the strategy's own recombination; its step-size adaptation therefore operates on
an incomplete account of the movement of the search. Second, candidate designs are clipped to the
design-variable bounds outside the strategy, so that bound handling is identical across all arms.
Third, the population size is fixed at 10, where the default for this problem dimension would be
13, so that generations align with steps.

\clearpage

\clearpage
\newpage

\section{Additional Results for Alternative Adaptive Samplers}
\label{app:adaptive_sampler_results}

\begin{figure}[h]
        \centering
        \begin{subfigure}[b]{0.49\textwidth}
            \includegraphics[width=\textwidth]{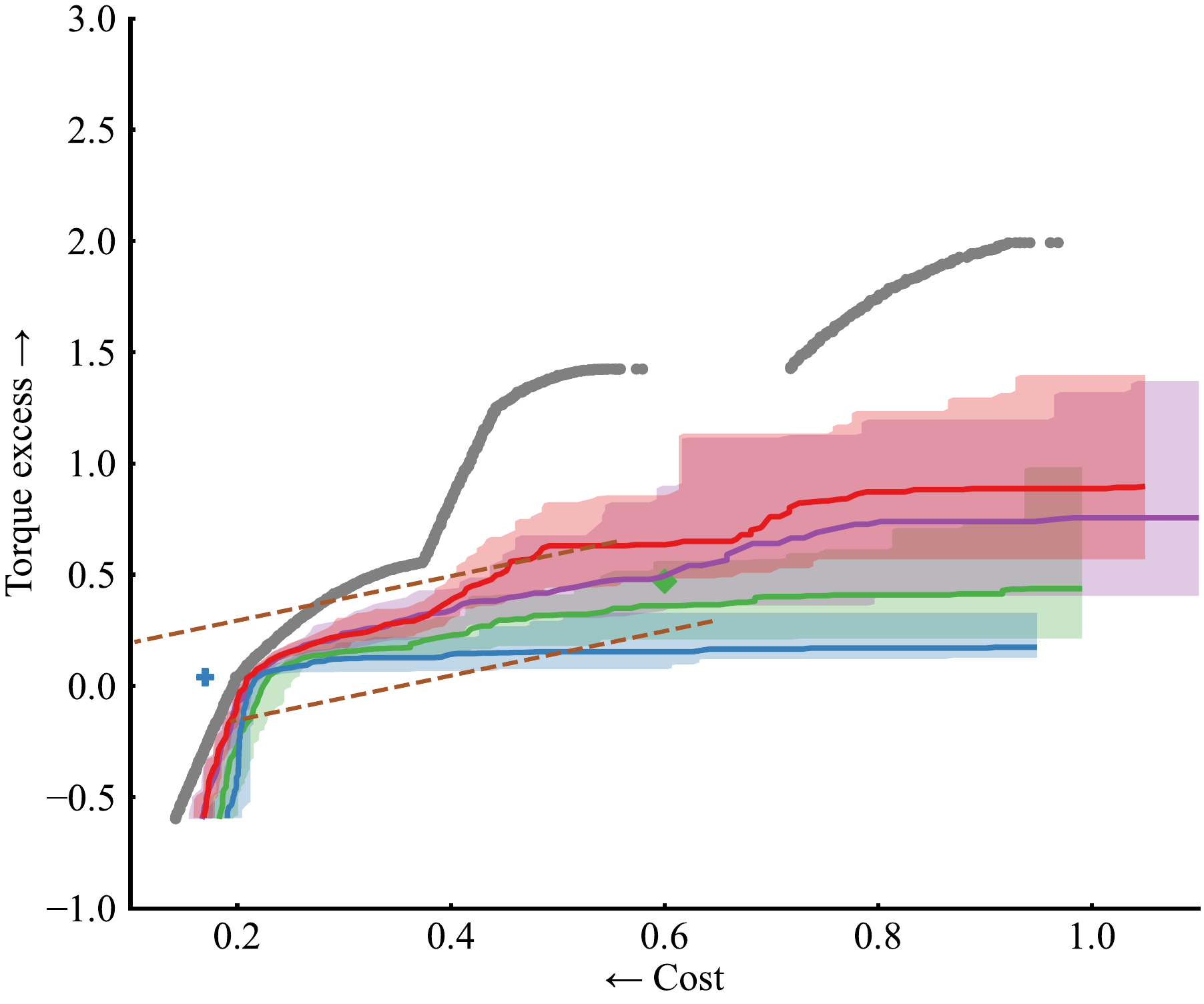}
            \begin{picture}(0,0)
                \put(0,195){a.} %
            \end{picture}

        \end{subfigure}
        \hfill
        \begin{subfigure}[b]{0.49\textwidth}
            \includegraphics[width=\textwidth]{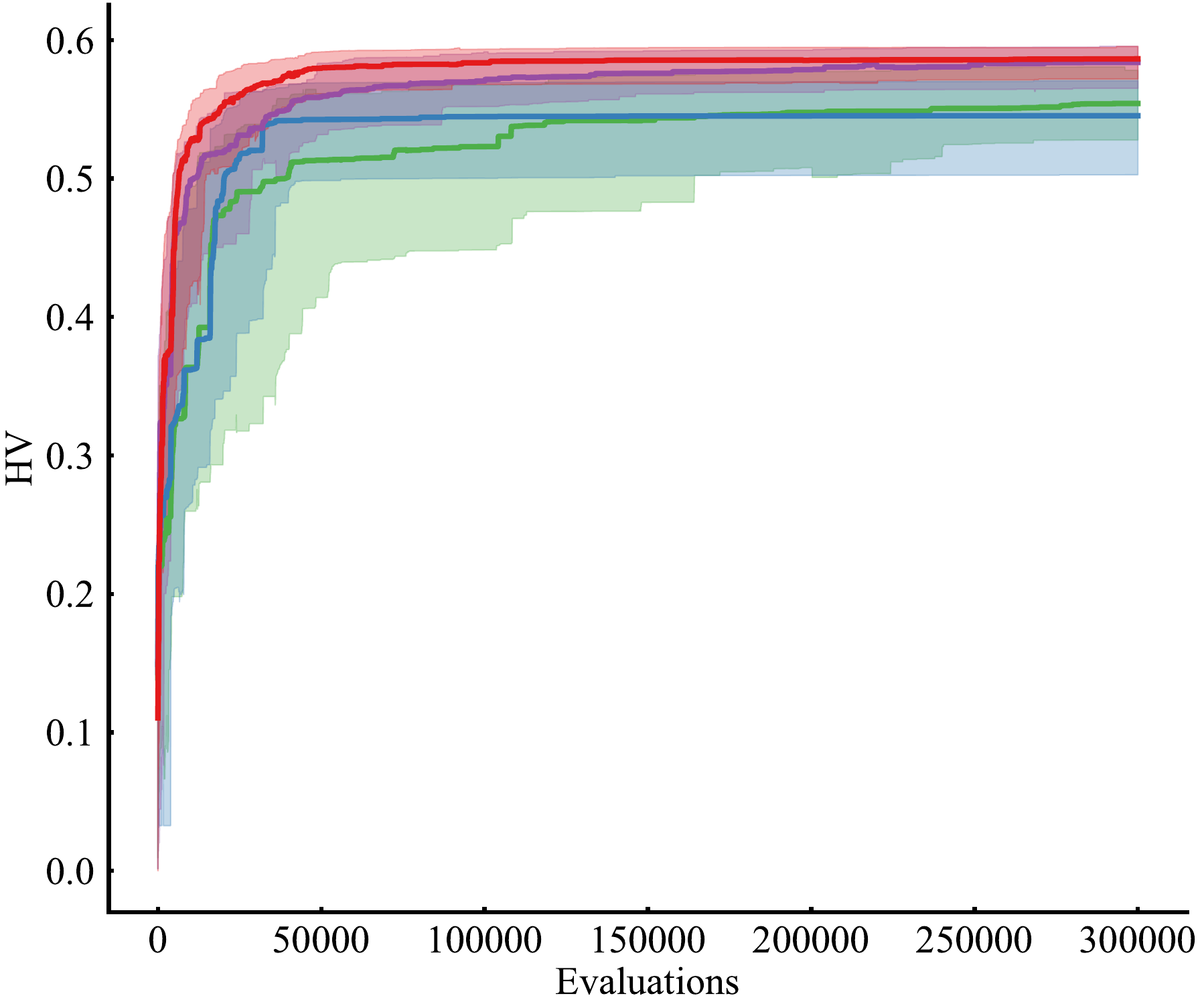}
            \begin{picture}(0,0)
                \put(0,195){b.} %
            \end{picture}
        \end{subfigure}

        \centering
        \vspace*{-1em} %
        {
        \begin{subfigure}[b]{0.6\textwidth}
            \includegraphics[width=\textwidth]{figures/adaptive_samplers/legend.pdf}
        \end{subfigure}
        }

        \caption{``CT1'' problem: EAF (a) and HV progression (b) of MOON-RL with different adaptive samplers, including $5^{th}$ and $95^{th}$ percentiles.}
        \label{fig:ct1_non_rl}
    \end{figure}

\begin{figure}[h]
        \centering
        \begin{subfigure}[b]{0.49\textwidth}
            \includegraphics[width=\textwidth]{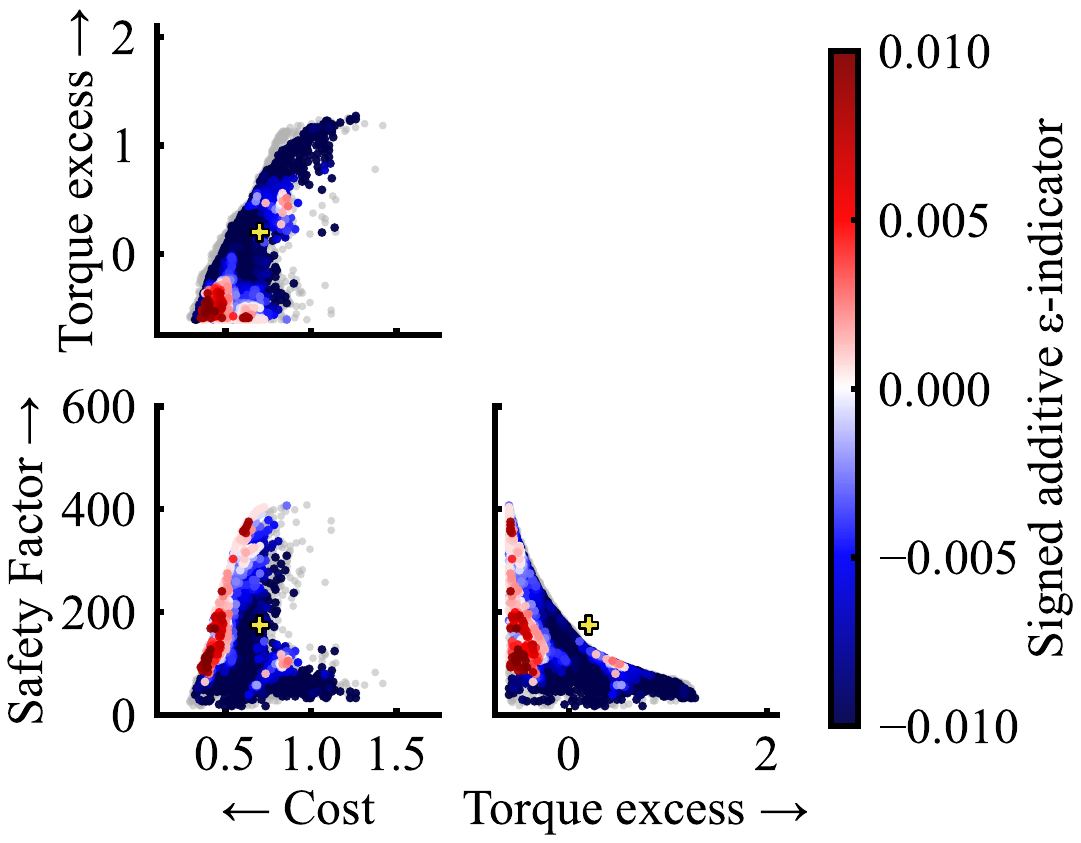}
            \begin{picture}(0,0)
                \put(0,180){a.} %
            \end{picture}
        \end{subfigure}
        \hfill
        \begin{subfigure}[b]{0.49\textwidth}
            \includegraphics[width=\textwidth]{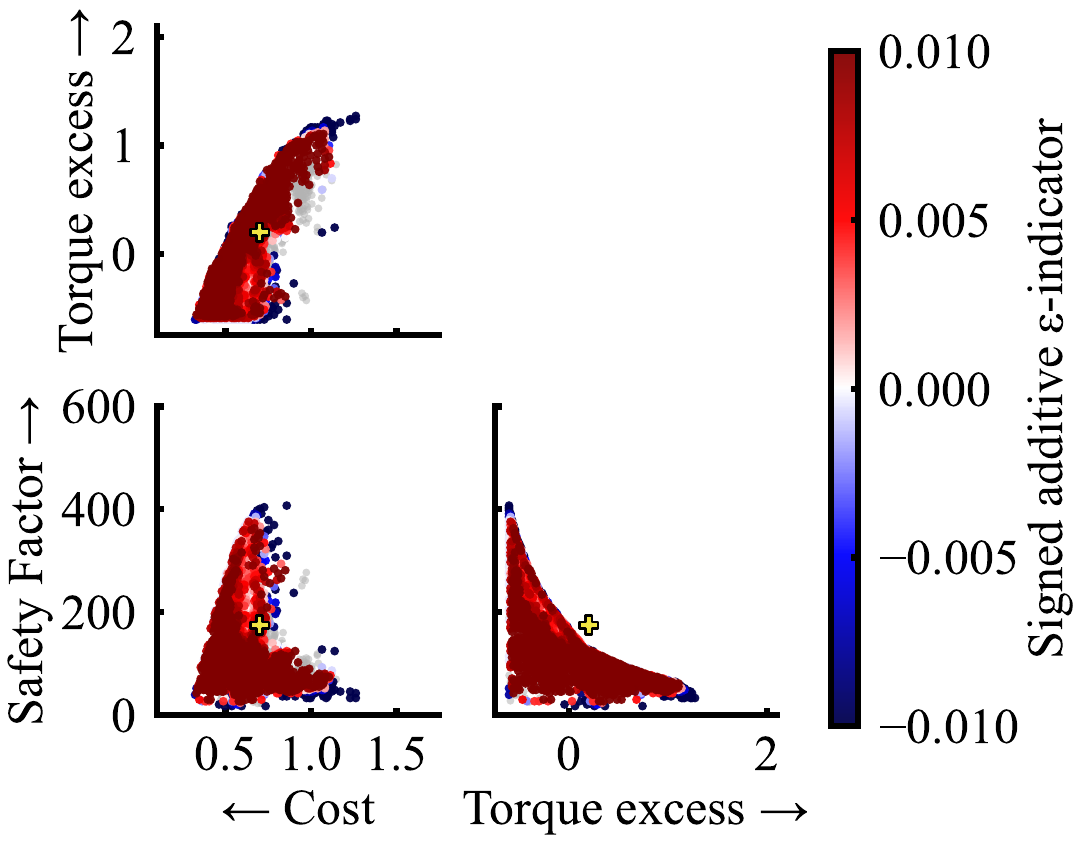}
            \begin{picture}(0,0)
                \put(0,180){b.} %
            \end{picture}
        \end{subfigure}

        \centering
        \vspace*{-1em} %

        \centering
        \begin{subfigure}[b]{0.49\textwidth}
            \includegraphics[width=\textwidth]{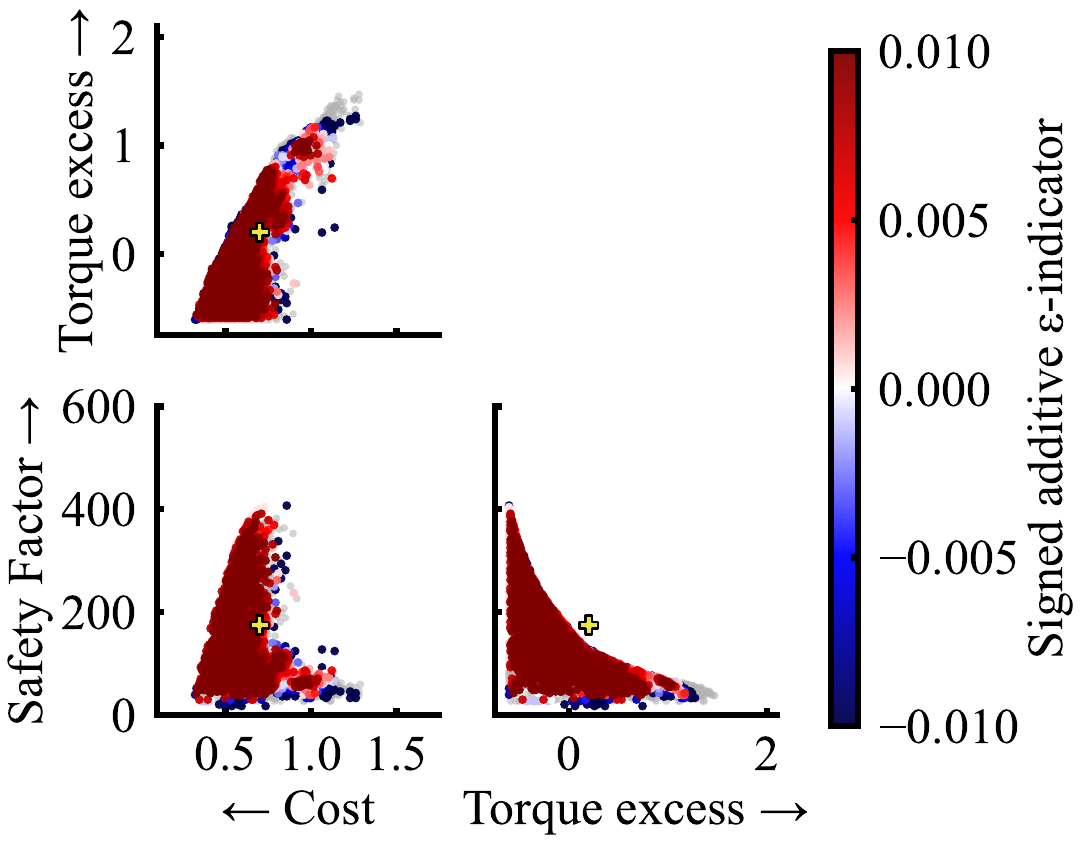}
            \begin{picture}(0,0)
                \put(0,180){c.} %
            \end{picture}
        \end{subfigure}
        \hfill
        \begin{subfigure}[b]{0.49\textwidth}
            \includegraphics[width=\textwidth]{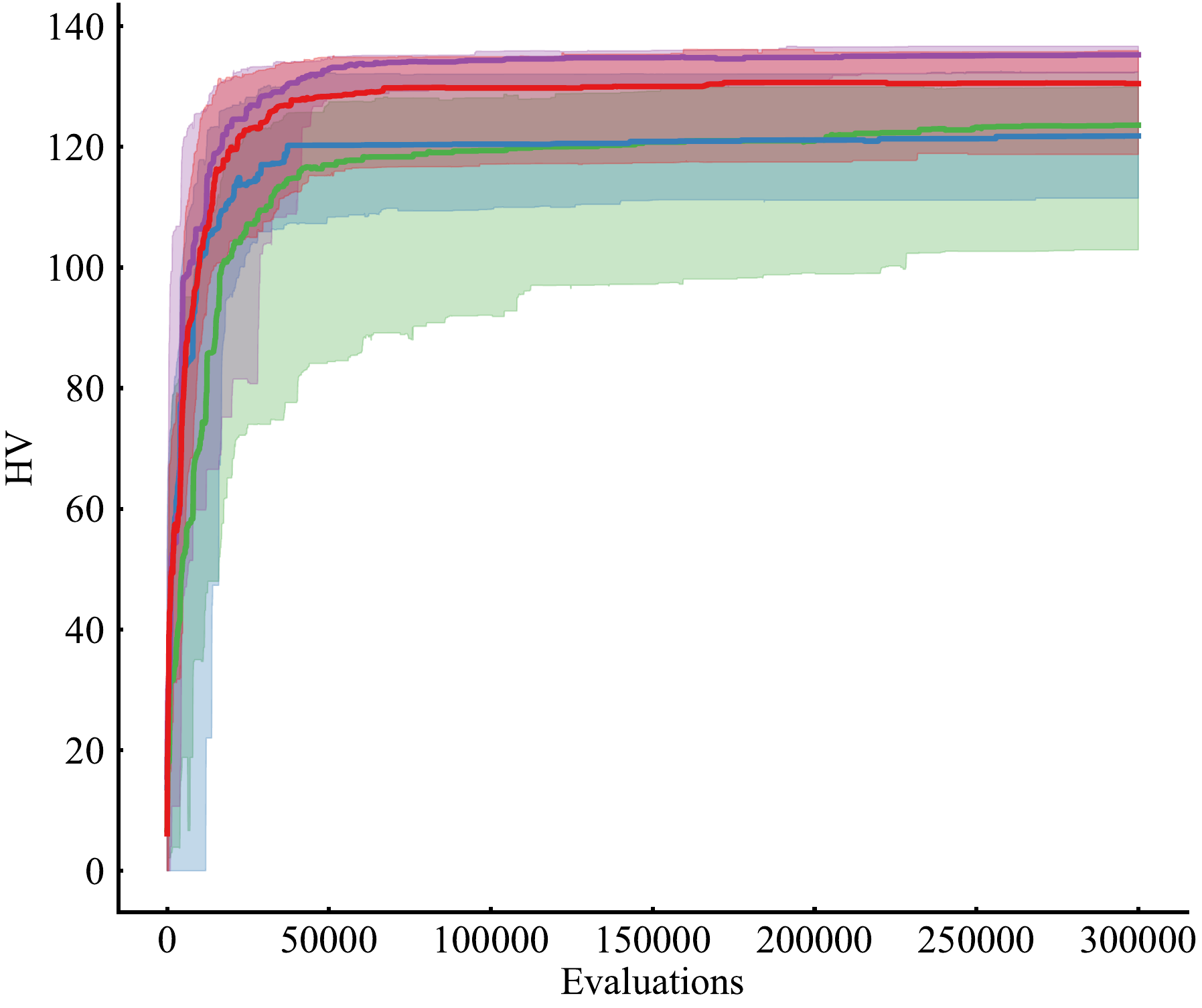}
            \begin{picture}(0,0)
                \put(0,180){d.} %
            \end{picture}
        \end{subfigure}

        \centering
        \vspace*{-1em} %

        \centering
        \begin{subfigure}[b]{0.49\textwidth}
            \includegraphics[width=\textwidth]{figures/adaptive_samplers/legend.pdf}
        \end{subfigure}

        \centering

        \caption{``CTS1'' problem: Comparison of median EAFs for MOON-RL (PPO) against MOON-RL (CMA-ES) (a), MOON-RL (CEM) (b) and MOON-RL (CEM, independent trajectories) (c). The target point is shown as a yellow cross. Grey points depict the baseline algorithm EAF, while colored points correspond to the MOON-RL EAF. The color encodes the signed additive $\varepsilon$-indicator: red regions indicate that MOON-RL outperforms the baseline, blue regions that the baseline outperforms MOON-RL, and white denotes parity. Pannel (d) shows the HV progression including $5^{th}$ and $95^{th}$ percentiles.}
        \label{fig:cts1_non_rl}
    \end{figure}

Figures~\ref{fig:ct1_non_rl} and~\ref{fig:cts1_non_rl} show the additional adaptive-sampler results for CT1 and CTS1, respectively. They compare MOON-RL using PPO, CMA-ES, CEM, and CEM with independent trajectories under the same experimental setting described in Section~\ref{subsec:alternative-samplers}.

\clearpage

\newpage